\documentclass[PhD]{UCLAthesis}

\graphicspath{ {Images/} }

\usepackage{bbm}
\usepackage{mleftright}
\usepackage{xparse}
\DeclareMathOperator{\divergence}{div}
\DeclareMathOperator*{\argmax}{arg\,max}

\defcitealias{feature_weights}{Nguyen et al., 2019}
\setcitestyle{numbers,square}

\title{Deep Learning of Unified Region, Edge, and Contour Models\\ for Automated Image Segmentation}

\author         {Ali Hatamizadeh}
\department     {Computer Science}
\degreeyear     {2020}

\chair          {Demetri Terzopoulos}
\member         {Quanquan Gu}
\member         {Fabien Scalzo}
\member         {Song-Chun Zhu}

\dedication {
To my beloved mother, brother, and Jinseo.
}

\acknowledgments { 
I would like to thank my academic advisor, Professor Demetri Terzopoulos, sincerely for his incredible mentorship, support, encouragement, and patience, and for all the opportunities that he made possible for me during my PhD journey. Demetri, who is a pioneer of computer vision and graphics with many significant contributions, taught me to be humble but hungry to make my mark. I consider myself to be lucky to have joined the elite group of people who have worked with Demetri on a topic related to his seminal active contours invention, and be able to contribute an extension of this powerful technique that has transformed the field of computer vision and image segmentation for the past three decades. I will miss our meetings and the three-hour midnight phone conversations to discuss new ideas and navigate my work in new directions. Indeed, Demetri has played a vital role in nurturing and maturing me as an academic and I will forever be indebted to him for his compassion, genuineness, encouragement, and mentorship.

I would like to thank the members of my PhD Committee, Professors Song-Chun Zhu, Quanquan Gu, and Fabien Scalzo, who dedicated their precious time to reviewing my thesis and providing invaluable suggestions.  

I would like to thank my industry collaborators, Dr.~Daguang Xu and Dr.~Andriy Myronenko, who provided an amazing opportunity for me to work with them on several projects at Nvidia Corporation and become involved in the BraTS19 and KiTS19 competitions. I would also like to thank Dr.~Xiaowei Ding and VoxelCloud, Inc., whose unrestricted corporate gift to UCLA funded a research assistantship that supported me in the first year of my PhD program and for providing an internship.

I would like to thank my collaborators at the UCLA David Geffen School of Medicine, Stein Eye Institute, Dr.~Steven D.~Schwartz and Dr.~Hamid Hosseini, who worked with me during my PhD program on projects to harness the power of AI in ophthalmic clinical workflow. 

I would like to thank Prof.~Daniel Rubin, Dr.~Hoogi, and other members of the Quantitative Imaging and Artificial Intelligence Laboratory in the Department of Biomedical Data Science of Stanford University for their efforts in our collaborative work and for providing the annotated lesion segmentation dataset, which has been invaluable to that work.

I would like to thank my co-author colleagues for their contributions and assistance with the various collaborative efforts in which I have participated during my PhD program.

I have enjoyed my time immensely in the UCLA Computer Science Department and the UCLA Computer Graphics and Vision Laboratory. Specifically, I would like to thank my former and current labmates, Ms.~Debleena Sengupta, Dr.~Masaki Nakada, Dr.~Tomer Weiss, Dr.~Garett Ridge, Mr.~Alan Litteneker, and Mr.~Abdullah Imran.

Lastly, I would like to thank my family for their amazing support: my mother, Ms.~Elham Pirhadi, who has loved and supported me through all the stages of my life and taught me how to stay strong and resilient even when the odds were against me; my brother, Mr.~Arshia Hatamizadeh, who has always been a true friend and someone on whom I can rely unconditionally; and my beloved, Ms.~Jinseo Choi, who has never stopped believing in me throughout my MSc and PhD studies, and is a major source of inspiration for me.
}

\vitaitem   {2017--2018}
{Machine Learning Intern \\
VoxelCloud Inc. \\
Los Angeles, CA.}

\vitaitem   {2017--2020}
{Research Assistant \\
Computer Science Department \\
University of California, Los Angeles\\
Los Angeles, CA.}

\vitaitem   {2018--2020}
{Teaching Assistant \\
Computer Science Department \\
University of California, Los Angeles\\
Los Angeles, CA.}

\vitaitem   {2019}
{Research Intern \\
NVIDIA Inc. \\
Santa Clara, CA.}

\vitaitem   {2019--2020}
{M.S Computer Science \\
University of California, Los Angeles\\
Los Angeles, CA.}

\publication{Hatamizadeh, A., Sengupta, D., Xu, D., and  Terzopoulos, D. (2020). Few-Shot Semantic Segmentation Using Aligned Variational Autoencoders. Under review by \emph{Advances in Neural Information Processing Systems (NeurIPS)}.}

\publication{Hatamizadeh, A., Sengupta, D., and Terzopoulos, D. (2020). Deep End-to-End Trainable Active Contours for Building Footprint Delineation. Under review by the \emph{European Conference on Computer Vision (ECCV)}.}

\publication{Hatamizadeh, A., Hoogi, A., Sengupta, D., Lu, W., Wilcox, B., Rubin, D., and Terzopoulos, D. (2019a). Deep Active Lesion Segmentation. In the \emph{International Workshop on Machine Learning in Medical Imaging (MLMI)}, pages 98–-105. Springer.}

\publication{Hatamizadeh, A., Terzopoulos, D., and Myronenko, A. (2019). End-to-End Boundary
Aware Networks for Medical Image Segmentation. In the  \emph{International Workshop on Machine
Learning in Medical Imaging (MLMI)}, pages 187–194. Springer.}

\publication{Myronenko, A., and Hatamizadeh, A. (2019). Robust Semantic Segmentation of Brain
Tumor Regions From 3D MRIs. In the \emph{MICCAI International  Brain Lesion Workshop}, pages
82–89. Springer.}

\publication{Myronenko, A., and Hatamizadeh, A. (2019). 3D Kidneys and Kidney Tumor Semantic Segmentation using Boundary-Aware Networks. In the \emph{2019 Kidney Tumor Segmentation Challenge: KiTS19}. University of Minnesota Digital Conservancy.}

\publication{Imran, A.-A.-Z., Hatamizadeh, A., Ananth, S. P., Ding, X., Tajbakhsh, N., and Terzopoulos, D. (2019). Fast and Automatic Segmentation of Pulmonary Lobes From Chest CT using a Progressive Dense B-Network. \emph{Computer Methods in Biomechanics and Biomedical Engineering: Imaging and Visualization}, November, pages 1–10.}

\publication{Hatamizadeh, A., Hosseini, H., Liu, Z., Schwartz, S. D., and Terzopoulos, D. (2019). Deep Dilated Convolutional Nets for the Automatic Segmentation of Retinal Vessels. In \emph{Proceedings of the 15th International Conference on Machine Learning and Data Mining (MLDM’19)}, pages 39–48.}

\publication{Imran, A.-A.-Z., Hatamizadeh, A., Ananth, S. P., Ding, X., Terzopoulos, D., and Tajbakhsh, N. (2018). Automatic Segmentation of Pulmonary Lobes Using a Progressive
Dense V-Network. In \emph{In Deep Learning in Medical Image Analysis (DLMIA)}, volume 11045 of \emph{Lecture Notes in Computer Science}, pages 282–290. Springer.\\
\textbf{NVIDIA Best Paper Award}.}

\abstract { 
Image segmentation is a fundamental and challenging problem in computer vision with applications spanning multiple areas, such as medical imaging, remote sensing, and autonomous vehicles. Recently, convolutional neural networks (CNNs) have gained traction in the design of automated segmentation pipelines. Although CNN-based models are adept at learning abstract features from raw image data, their performance is dependent on the availability and size of suitable training datasets. Additionally, these models are often unable to capture the details of object boundaries and generalize poorly to unseen classes. In this thesis, we devise novel methodologies that address these issues and establish robust representation learning frameworks for fully-automatic semantic segmentation in medical imaging and mainstream computer vision. In particular, our contributions include (1) state-of-the-art 2D and 3D image segmentation networks for computer vision and medical image analysis, (2) an end-to-end trainable image segmentation framework that unifies CNNs and active contour models with learnable parameters for fast and robust object delineation, (3) a novel approach for disentangling edge and texture processing in segmentation networks, and (4) a novel few-shot learning model in both supervised settings and semi-supervised settings where synergies between latent and image spaces are leveraged to learn to segment images given limited training data.
}

\begin{document}
\makeintropages

\chapter{Introduction}

Image segmentation has been considered a fundamental problem of
computer vision since the early days of the field
\citep[][Chapter~8]{rosenfeld1976digital} \citep[][Chapters~6 and
7]{sonka2014image}. Generally speaking, it refers to the task of
segmenting the image into parts, which may be objects or regions of
interest. So-called ``semantic segmentation'' further attempts to
classify each pixel in the image as belonging to some particular
object or region, thus elucidating the global semantics of the imaged
scene. In broad application areas, such as remote sensing, medical
image analysis, and autonomous vehicles, image segmentation is the
first step in building a fully automated perception system.

This thesis introduces methodologies that yield novel, reliable,
fully-automated image segmentation algorithms. Such algorithms have
numerous applications in all manner of quantitative image analysis.
For instance, to detect an aggressive cancerous lesion, measure its
clinically significant properties, and track its evolution over a
period of time, it is important to be able to localize and segment the
lesion in medical images and run further quantitative post-processing
operations. As another example, remote sensing systems benefit from
the rapid localization and delineation of buildings in aerial images
vital to applications such as urban planning and disaster relief
response.

Traditionally, model-based methods, such as Active Contour Models
(ACMs) \citep{kass1988snakes} have been a popular choice for
high-quality image segmentation, and they have evolved into
widely-used interactive tools such as the ``Lassos'' of GIMP and Adobe
PhotoShop. In recent years, however, machine learning approaches,
especially Deep Neural Networks (DNNs) have become popular due to
their data-driven nature and impressive performance. Various deep
Convolutional Neural Network (CNN) models have been successfully
applied in computer vision, including to automatic image segmentation
\citep{minaee2020image}. In particular, Fully Convolutional Networks
(FCNs) \citep{long2015fully} have gained traction for automated
semantic image segmentation. A good example of this is our own work,
reproduced in Appendix~\ref{cha:appendix}.

Despite some exceptions (e.g., \citep{mcinerney2002deformable}), the
dependence of ACMs on user interaction in the form of contour
initialization and parameter adjustment, has made it difficult to
deploy these models in large-scale image analysis tasks in which full
automation is needed. By contrast, although CNNs and FCNs have played
a major role in advancing automated image segmentation methodologies
and demonstrating state-of-the-art performance on benchmark datasets,
they typically rely on copious quantities training data and their
performance is often far from optimal in many applications where exact
segmentation predictions are needed, especially near object and region
boundaries.

\begin{figure}
\centering
\includegraphics[width=\linewidth]{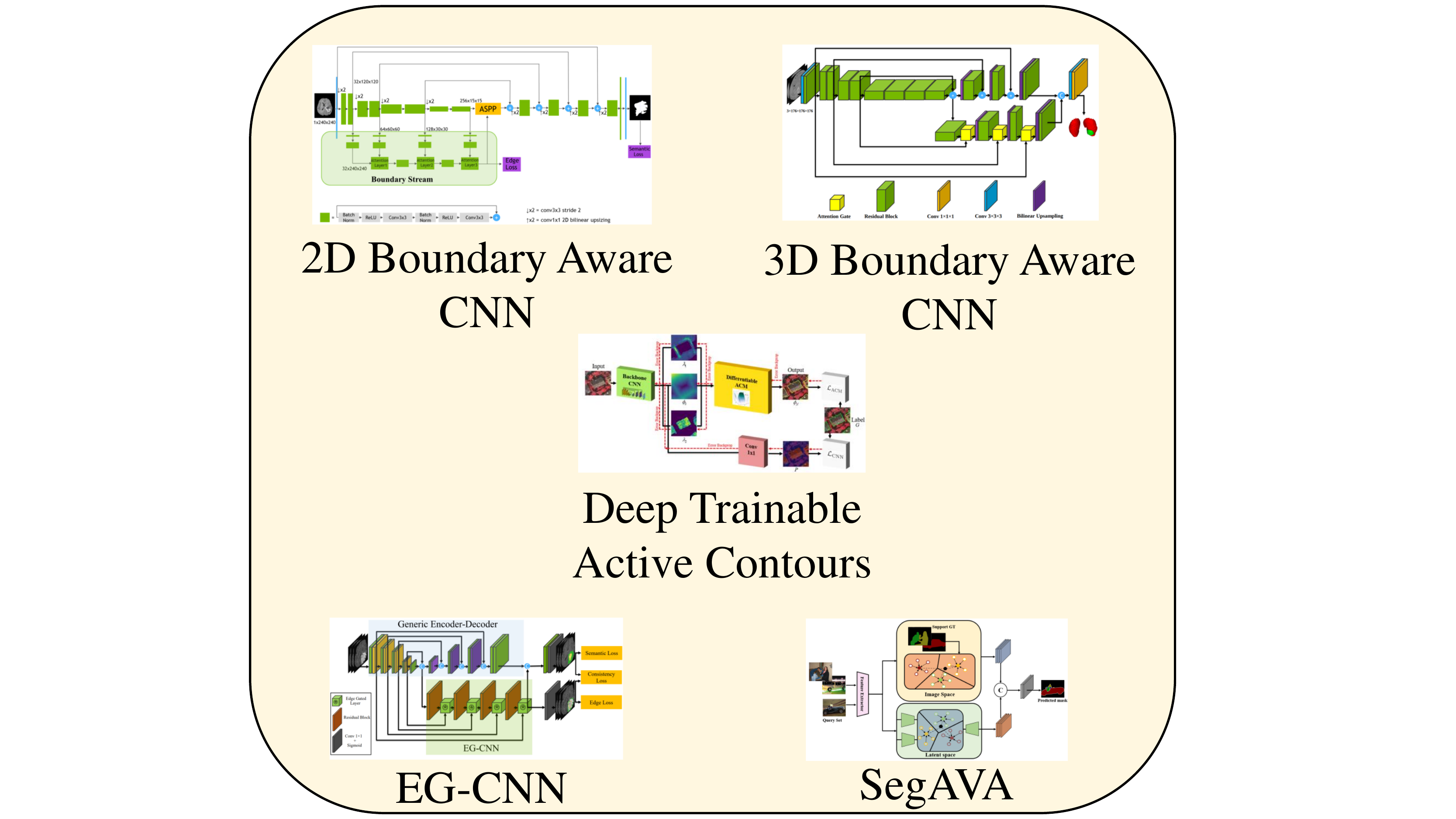}
\caption[Overview of methodologies proposed in this thesis] {In this
thesis, we introduce 2D and 3D edge-aware CNNs (EG-CNN) for
segmentation, Deep Trainable Active Contours (DTAC) for
segmentation with highly accurate boundaries, and segmentation with
aligned variational autoencoder (SegAVA) that can learn from small
datasets.}
\label{fig:framework_intro_overview}
\end{figure}

The goals of this thesis include devising novel deep learning models,
which are illustrated in Figure~\ref{fig:framework_intro_overview},
capable of learning powerful image representations, even with small
datasets, that can be leveraged to yield highly accurate segmentation
predictions and precisely delineate object and region boundaries. In
the remainder of this chapter, we discuss these issues in greater
detail and preview our solutions to some of these problems, which
comprise the technical contributions of this thesis.

\section{Edge-Aware Segmentation Networks}

Intensity edges and textures contribute different information to image
understanding. Edges (and boundaries) encode shape information, while
textures determine the appearance of regions. FCNs have proven to be
effective at representing and classifying textural information, thus
transforming image intensity into output class masks that achieve
semantic segmentation. In particular, the seminal U-Net architecture
\citep{Ronneberger15} demonstrated the effectiveness of down-sampling
and up-sampling paths for multi-scale feature representation learning,
and many encoder-decoder CNNs have since been introduced based on the
same principles.

\begin{figure}
\centering
\includegraphics[width=\linewidth]{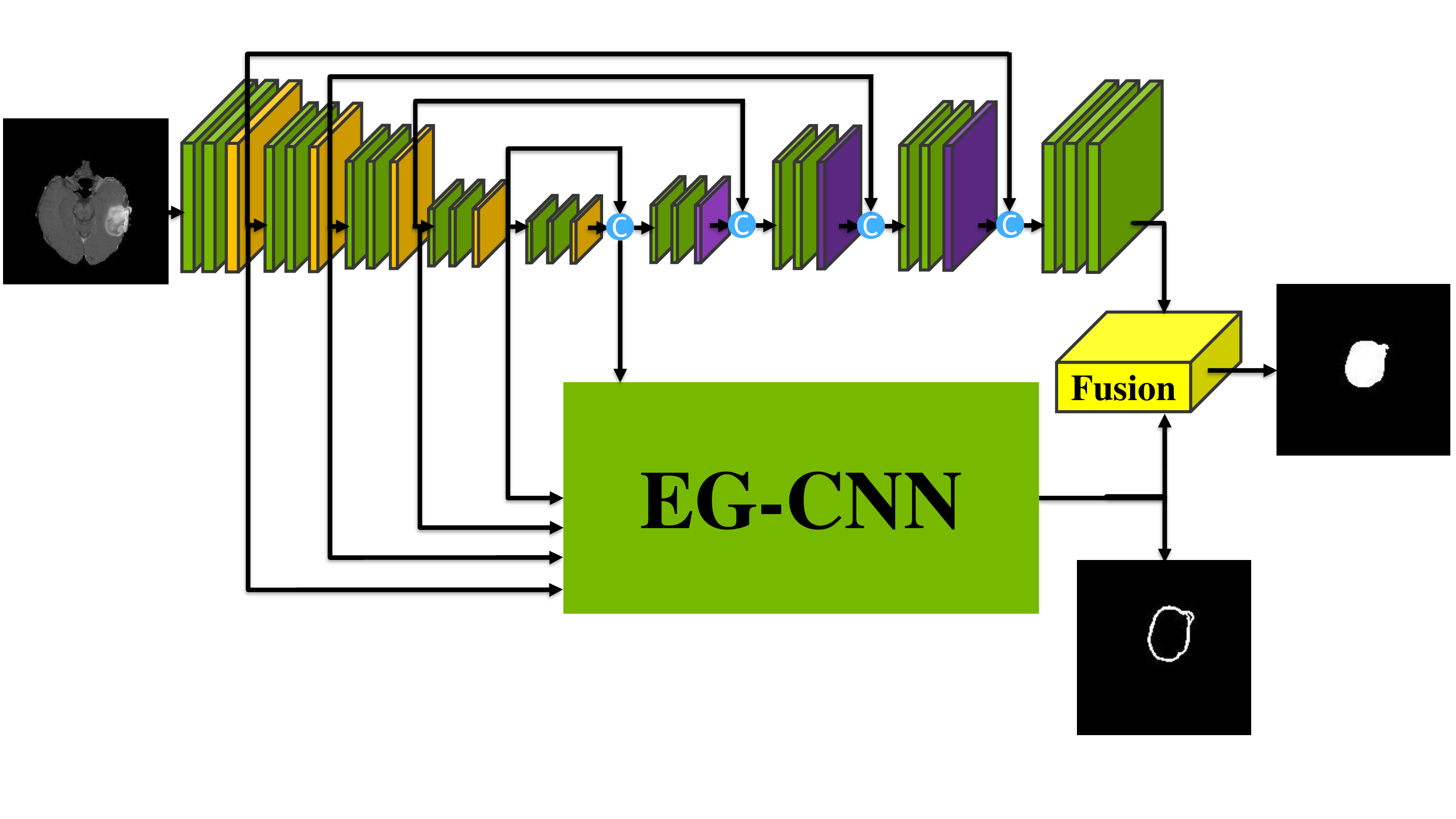}
\caption[Overview of the EG-CNN module] {We propose a plug-and-play
EG-CNN module that can be employed with any existing encoder-decoder
backbone to increase the segmentation accuracy by supervising the
edges.}
\label{fig:framework_intro2}
\end{figure}

\begin{figure} \centering
\def\x{0.242}
\includegraphics[width=\x\linewidth,height=\x\linewidth]{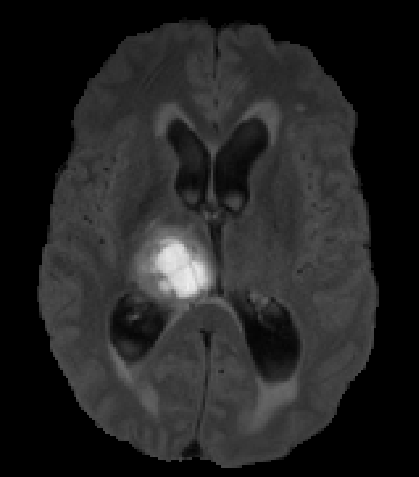}
\includegraphics[width=\x\linewidth,height=\x\linewidth]{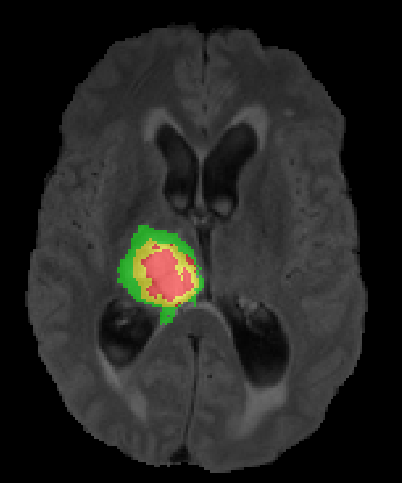}
\includegraphics[width=\x\linewidth,height=\x\linewidth]{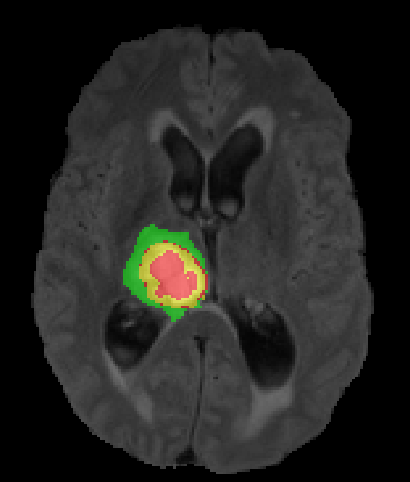}
\includegraphics[width=\x\linewidth,height=\x\linewidth]{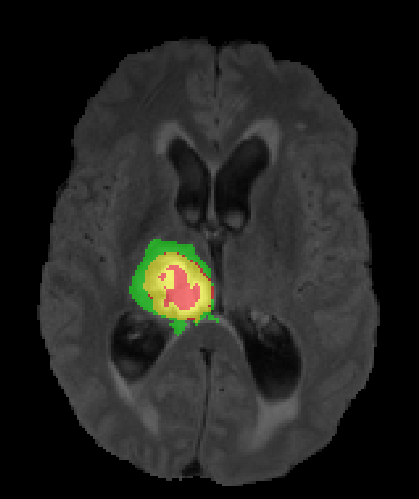}\\[3pt]
\includegraphics[width=\x\linewidth,height=\x\linewidth]{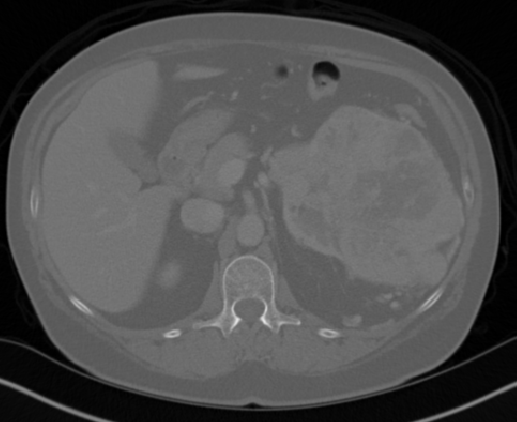}
\includegraphics[width=\x\linewidth,height=\x\linewidth]{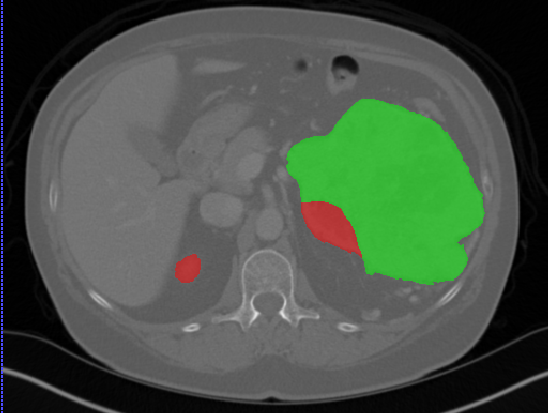}
\includegraphics[width=\x\linewidth,height=\x\linewidth]{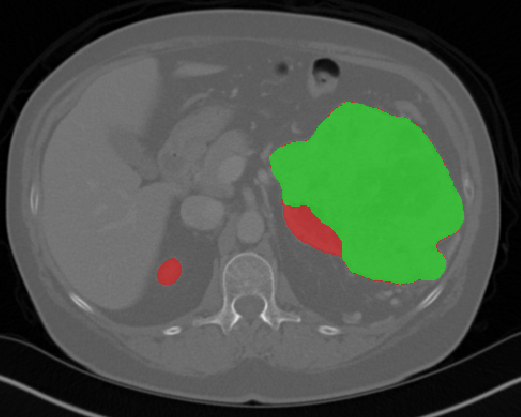}
\includegraphics[width=\x\linewidth,height=\x\linewidth]{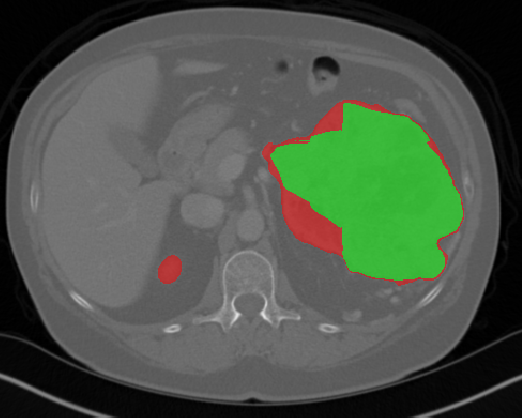}
\vspace{0.75pt}
\makebox[\x\linewidth]{(a) Input images} 
\makebox[\x\linewidth]{(b) Semantic Labels}
\makebox[\x\linewidth]{(c) Seg-Net+EG-CNN}
\makebox[\x\linewidth]{(d) Seg-Net}
\caption[Visualization of segmentation outputs in BraTS and KiTS] {Visualization of segmentation outputs when our proposed EG-CNN is employed along with Seg-Net for the tasks of brain and kidney tumor segmentation in BraTS and KiTS datasets. EG-CNN improves the segmentation accuracy by effectively accounting for edge representations.}
\label{fig:teas10}
\end{figure}

\citet{geirhos2018} empirically demonstrated that common CNN
architectures are biased towards recognizing textures in the image,
not object shape representations. This is in contrast to how humans
normally segment images. In medical imaging for instance, expert
manual segmentation often relies on the boundaries of anatomical
structures; for example, to manually segment a liver, a medical
practitioner usually identifies intensity edges first and subsequently
fills the interior region in the segmentation mask. CNNs, which
predominantly learn texture abstractions, often yield imprecise
boundary delineations. Thus, CNN predictions often need to be
post-processed to compensate for the shape details that the model
fails to learn during training.

We argue that the sub-optimal paradigm of processing different
abstractions within a single CNN pipeline can be remedied through the
effective processing of information in a structured manner.
Consequently, we devise strategies for disentangling the edge and
texture information within a single training pipeline.
Figure~\ref{fig:framework_intro2} illustrates how our proposed module,
dubbed EG-CNN, can be paired with any existing CNN encoder-decoder to
improve segmentation quality near intensity edges. We have applied our EG-CNN to the tasks of brain and liver tumor segmentation in medical images (Figure~\ref{fig:teas10}).

\section{End-to-End Trainable ACMs}

\begin{figure} \centering
\def\x{0.19}
\centerline{\makebox[\x\linewidth]{(1) Brain MR} \makebox[\x\linewidth]{(2) Liver MR} \makebox[\x\linewidth]{(3) Liver CT} \makebox[\x\linewidth]{(4) Lung CT}}
~\\[-5pt]
\subcaptionbox{Expert Manual}{%
\includegraphics[width=\x\linewidth,height=\x\linewidth]{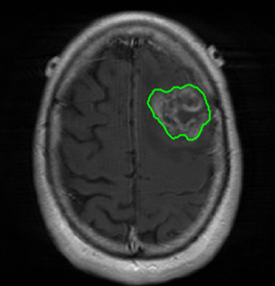} \hfill
\includegraphics[width=\x\linewidth,height=\x\linewidth]{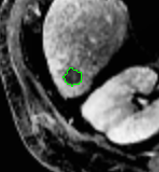} \hfill
\includegraphics[width=\x\linewidth,height=\x\linewidth,trim={270 80 230 100},clip]{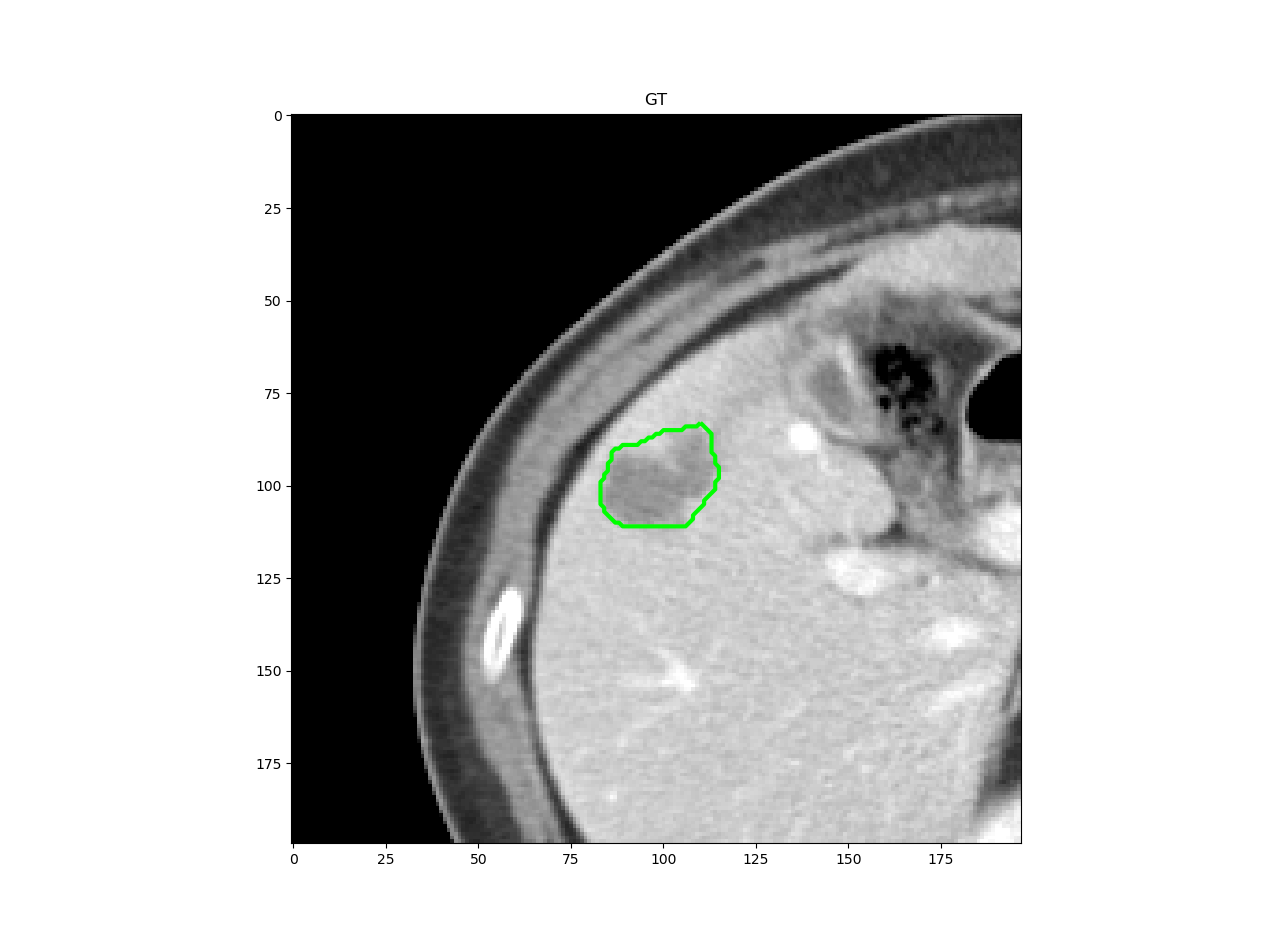} \hfill
\includegraphics[width=\x\linewidth,height=\x\linewidth,trim={270 80 230 100},clip]{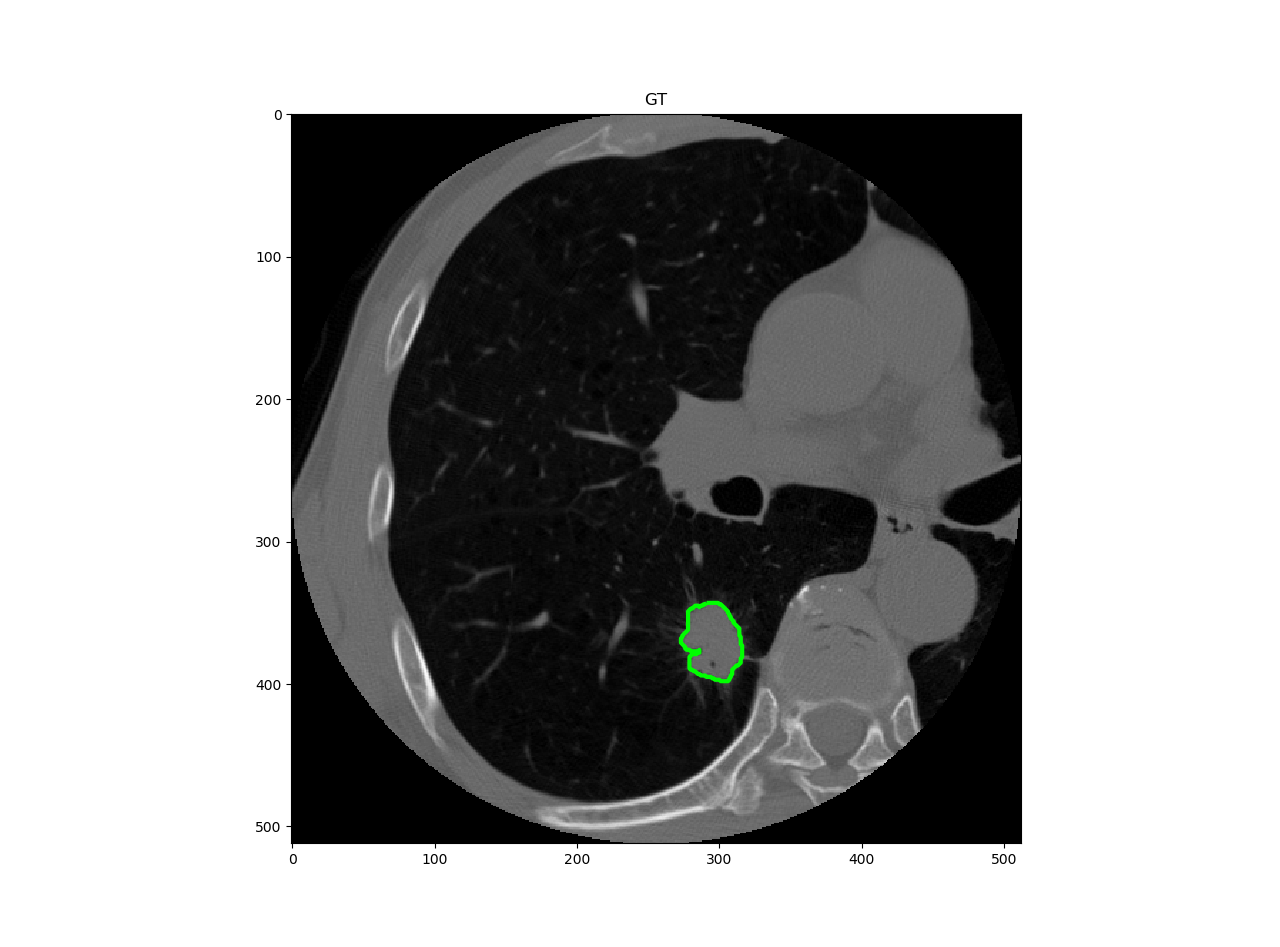}}\\[8pt]
\subcaptionbox{DALS Output}{%
\includegraphics[width=\x\linewidth,height=\x\linewidth]{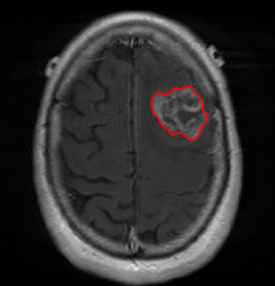} \hfill
\includegraphics[width=\x\linewidth,height=\x\linewidth]{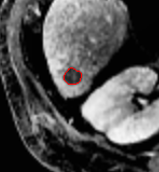} \hfill
\includegraphics[width=\x\linewidth,height=\x\linewidth,trim={270 80 230 100},clip]{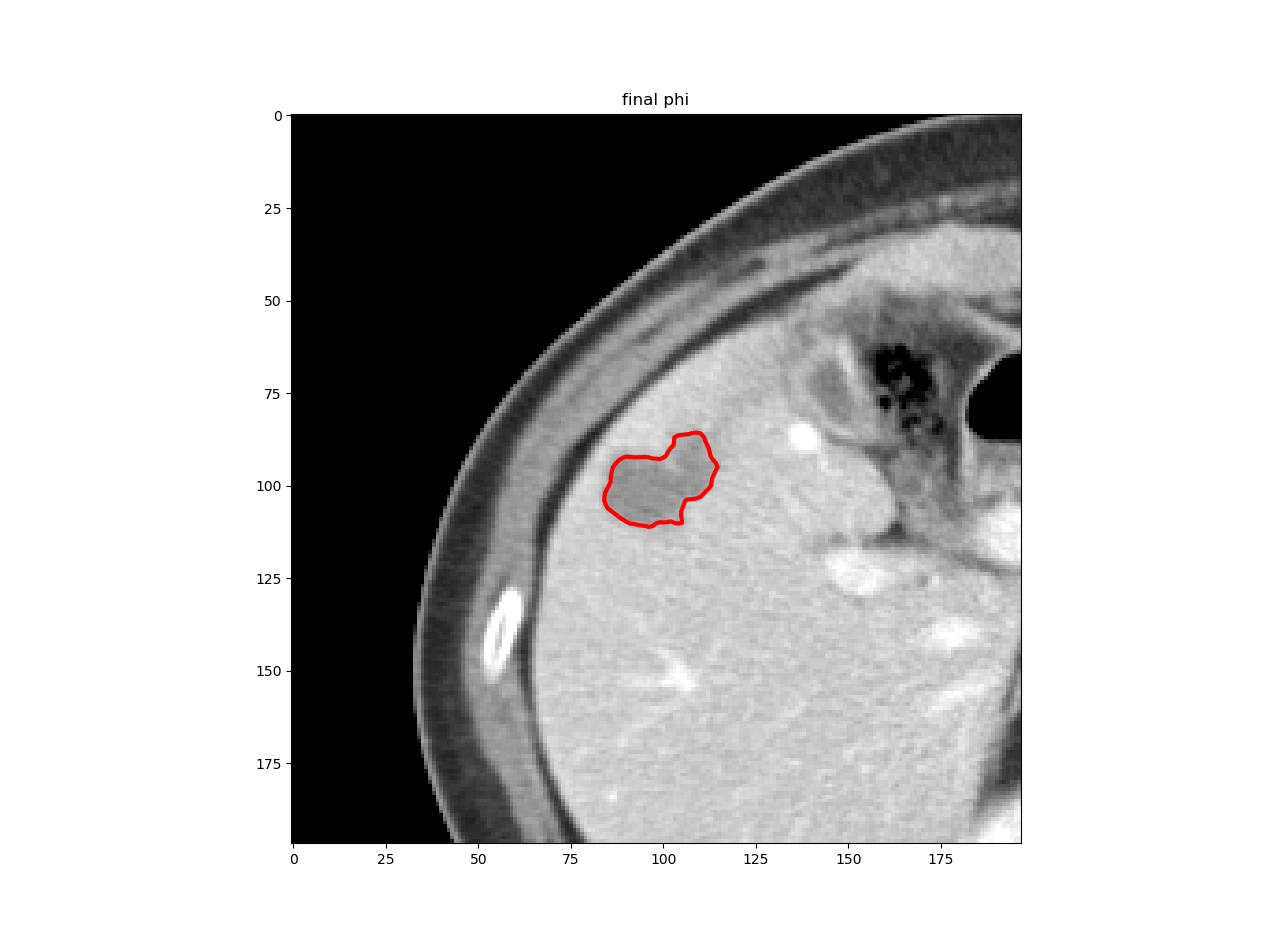} \hfill
\includegraphics[width=\x\linewidth,height=\x\linewidth,trim={270 80 230 100},clip]{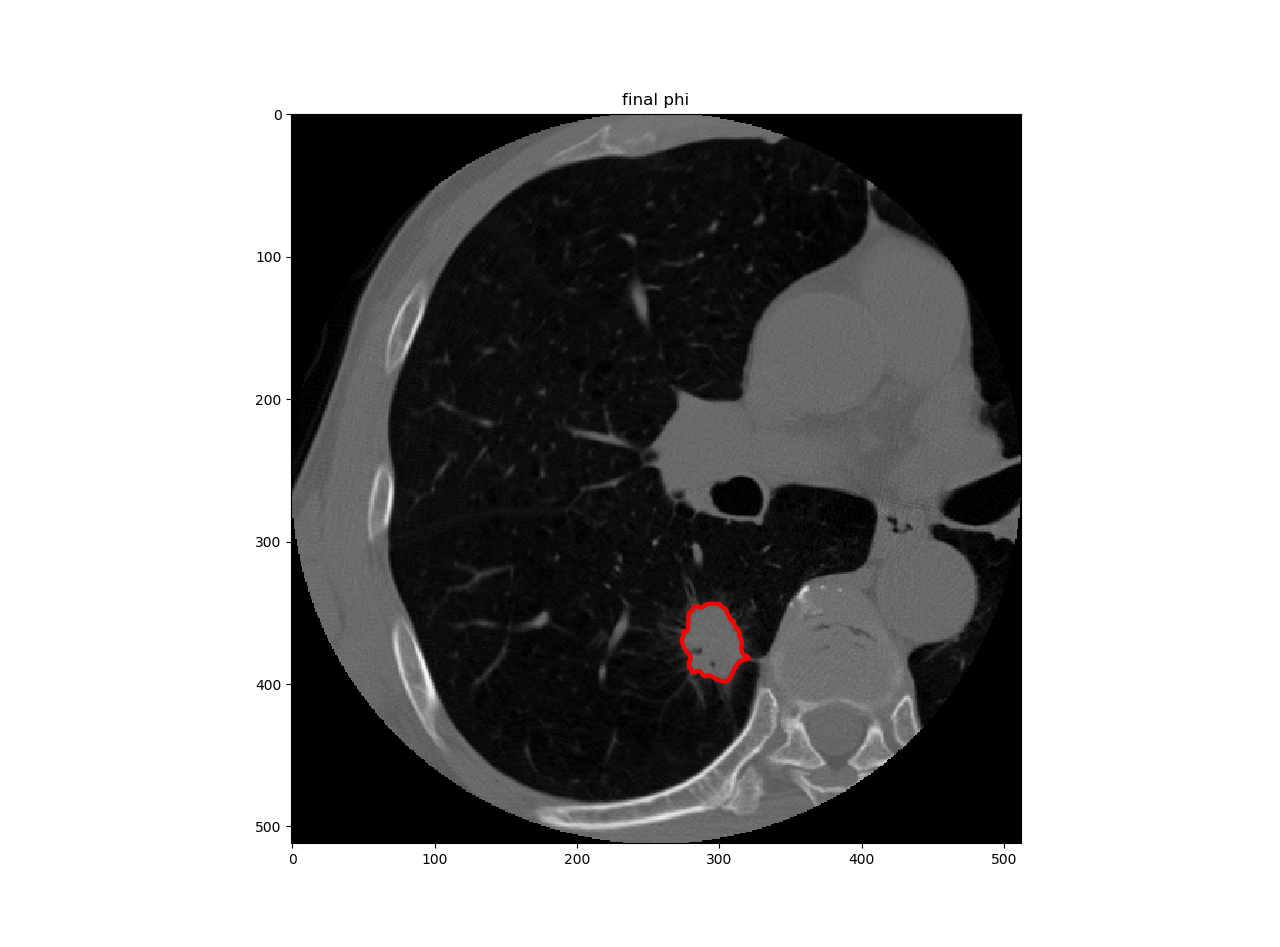}}\\[8pt]
\subcaptionbox{U-Net Output}{%
\includegraphics[width=\x\linewidth,height=\x\linewidth]{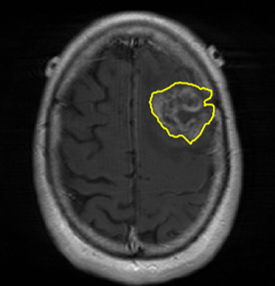} \hfill
\includegraphics[width=\x\linewidth,height=\x\linewidth]{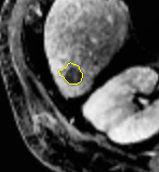} \hfill
\includegraphics[width=\x\linewidth,height=\x\linewidth,trim={270 80 230 100},clip]{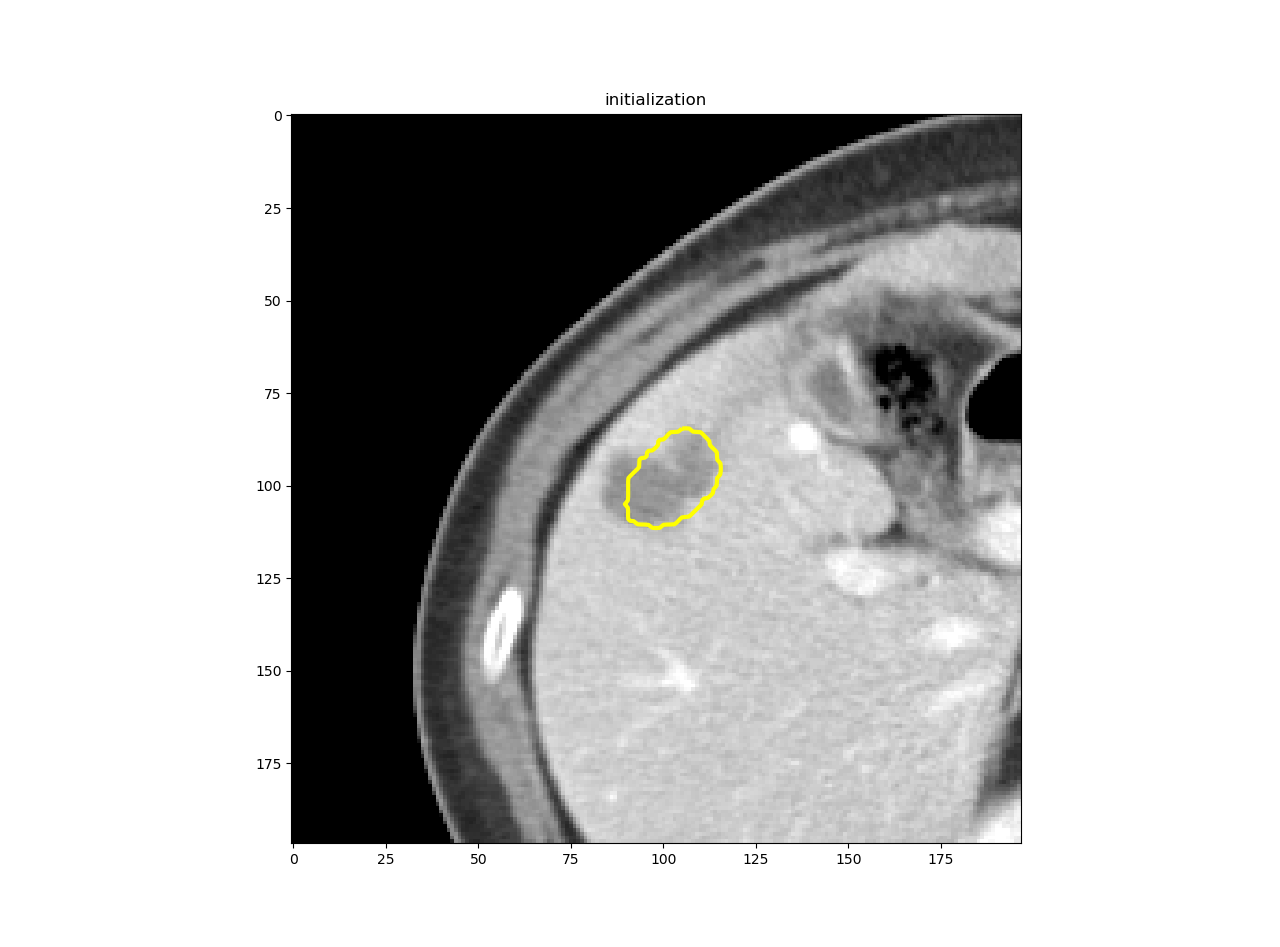} \hfill
\includegraphics[width=\x\linewidth,height=\x\linewidth,trim={270 80 230 100},clip]{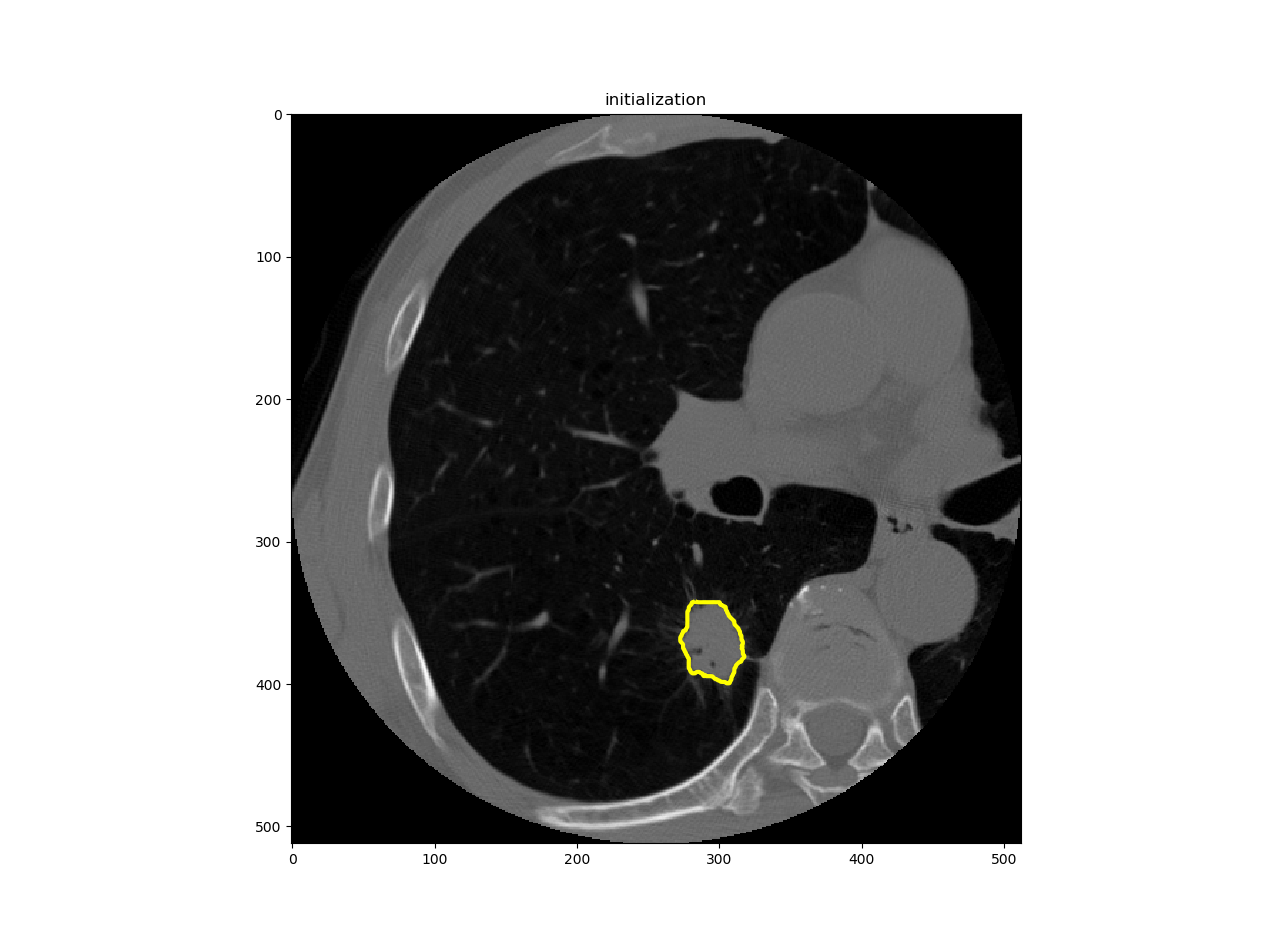}}
\caption[Segmentation comparison of DALS and competing methods]
        {Segmentation comparison of (a) medical expert manual with (b)
        our DALS and (c) U-Net \citep{Ronneberger15}, in (1) Brain
        MR, (2) Liver MR, (3) Liver CT, and (4) Lung CT images.}
\label{fig:teas1}
\end{figure}

Despite attempts to disentangle texture and edge information within a
single pipeline, accurately delineating object boundaries remains a
challenging task even for the most promising CNN architectures
\citep{chen2017deeplab, he2017mask, zhao2017pyramid} that have
achieved state-of-the-art performance on benchmark datasets (see also
Appendix~\ref{cha:appendix}). The recently proposed Deeplabv3+
\citep{chen2018encoder} mitigates this problem to some extent by
leveraging dilated convolutions, but such improvements were made
possible by extensive pre-training consuming vast computational
resources.

Unlike CNNs that rely on large annotated datasets, massive
computation, and hours of training, conventional ACMs are
non-learning-based segmentation models that rely mainly on the content
of the input image itself. ACMs have been successfully employed in
various image analysis tasks, including object segmentation and
tracking. In most ACM variants, the deformable curve(s) of interest
dynamically evolves by an iterative process that minimizes an
associated energy functional. However, the classic ACM
\citep{kass1988snakes} relies on some degree of user interaction to
specify the initial contour and tune the parameters of the energy
functional, which undermines its applicability to the automated
analysis of large quantities of images.

\begin{figure}
\centering
\includegraphics[width=0.5\linewidth,trim={0 150 0 100},clip]{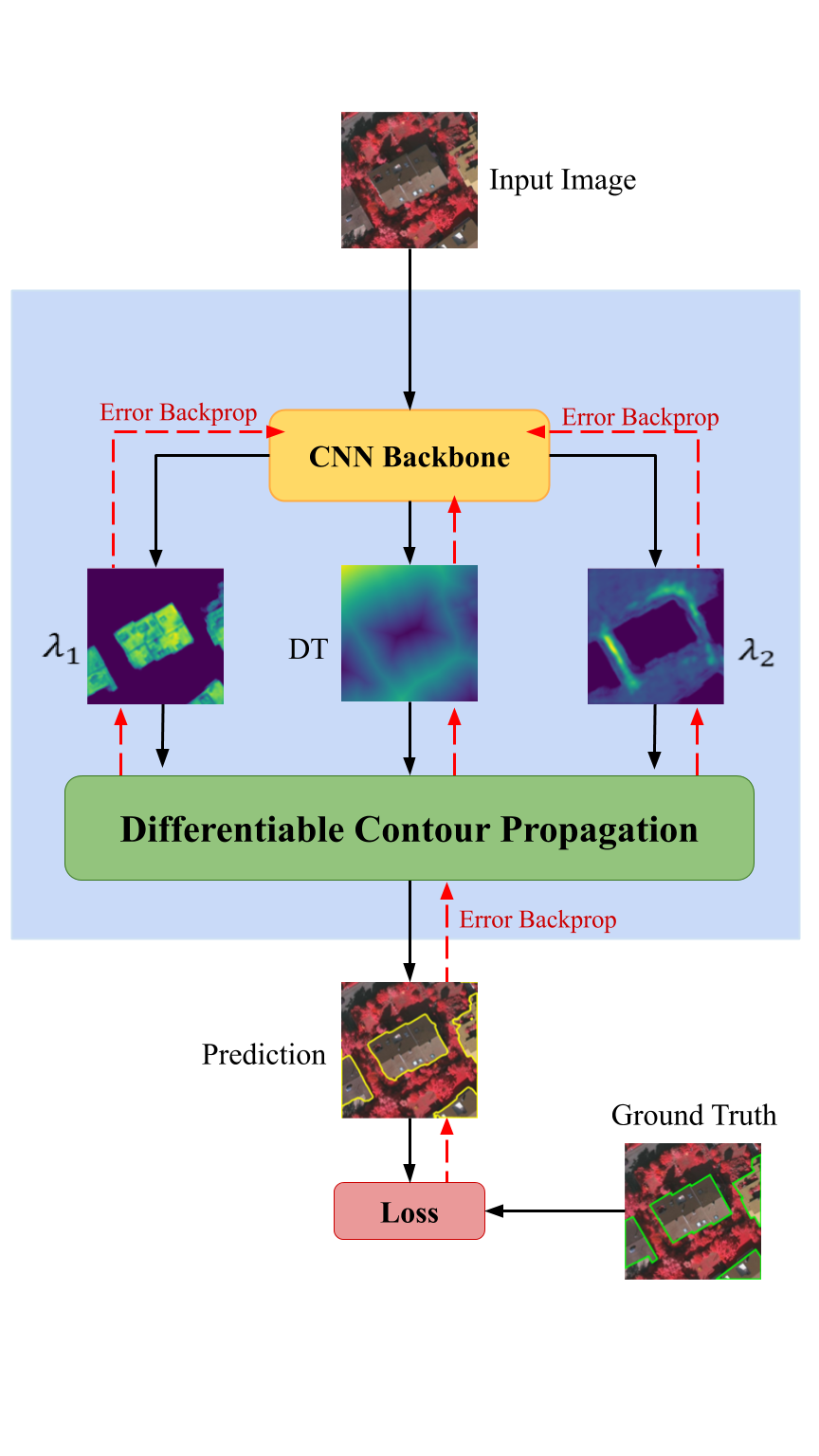}
\caption[Overview of End-to-End Trainable ACMs] {We propose a unified
ACM-CNN framework that is automatically differentiable, hence
end-to-end trainable without user supervision. The backbone CNN learns
to initialize the ACM, via a generalized distance transform, and tune
the per-pixel parameter maps in the ACM's energy functional.}
\label{fig:framework_intro_acm}
\end{figure}

\begin{figure}
\centering
\def\x{0.24}
\includegraphics[width=\x\linewidth,height=\x\linewidth]{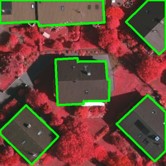}
\includegraphics[width=\x\linewidth,height=\x\linewidth]{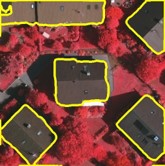}
\includegraphics[width=\x\linewidth,height=\x\linewidth,trim={80 35 20 20},clip]{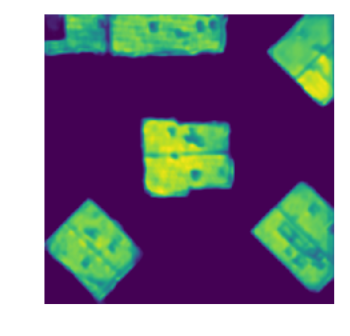}
\includegraphics[width=\x\linewidth,height=\x\linewidth,trim={80 35 20 20},clip]{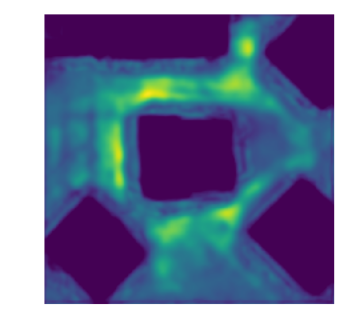}\\[3pt]
\includegraphics[width=\x\linewidth,height=\x\linewidth]{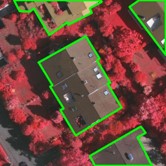}
\includegraphics[width=\x\linewidth,height=\x\linewidth]{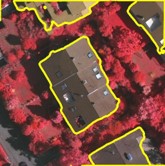}
\includegraphics[width=\x\linewidth,height=\x\linewidth,trim={80 35 20 20},clip]{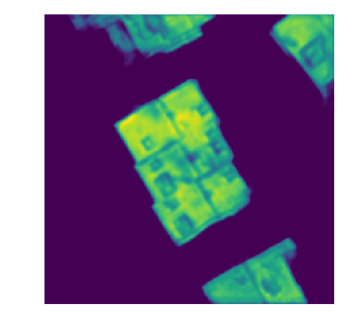}
\includegraphics[width=\x\linewidth,height=\x\linewidth,trim={80 35 20 20},clip]{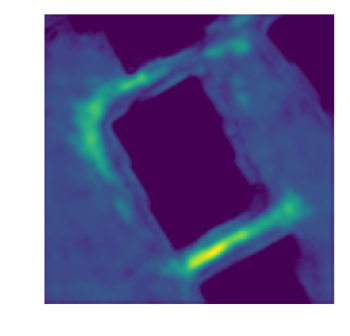}\\[3pt]
\includegraphics[width=\x\linewidth,height=\x\linewidth]{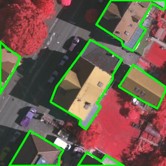}
\includegraphics[width=\x\linewidth,height=\x\linewidth]{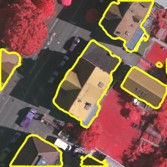}
\includegraphics[width=\x\linewidth,height=\x\linewidth,trim={80 35 20 20},clip]{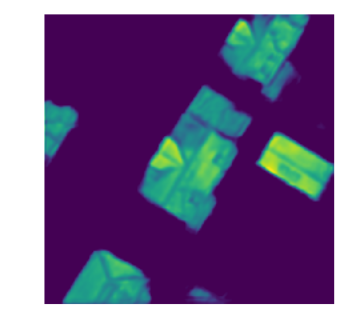}
\includegraphics[width=\x\linewidth,height=\x\linewidth,trim={80 35 20 20},clip]{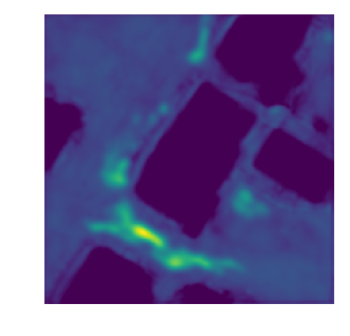}\\[3pt]
\makebox[\x\linewidth]{(a) Input image} \makebox[\x\linewidth]{(b) DTAC Output}
\makebox[\x\linewidth]{(c) $\lambda_{1}(x,y)$} \makebox[\x\linewidth]{(d) $\lambda_{2}(x,y)$}
\caption[Visualization of DTAC outputs] {Visualization of DTAC segmentation outputs and learned feature maps $\lambda_{1}(x,y)$ and $\lambda_{2}(x,y)$ used in DTAC's energy functional.}
\label{fig:teas5}
\end{figure}

We first introduce a method for connecting the output of a CNN to an ACM, yielding a model for the precise delineation of lesions, to which we refer as Deep Active Lesion Segmentation (DALS) (Figure~\ref{fig:teas1}). We then go further to introduce a truly unified framework
(Figure~\ref{fig:framework_intro_acm}) that bridges the gap between ACMs
and CNNs by leveraging a novel, automatically differentiable level-set
ACM with trainable parameters that allows for back-propagation of
gradients and can be end-to-end trained along with a backbone CNN from
scratch, without any CNN pre-training. The ACM is initialized directly
by the CNN and utilizes an energy functional that is locally-tunable
by the backbone CNN, through 2D feature maps. Thus, our work overcomes
the big hurdle of fully automating the powerful ACM approach to image
segmentation. We have applied our proposed framework to the task of building segmentation in aerial images (Figure~\ref{fig:teas5}). 

\section{Few-Shot Learning for Segmentation}

In essence, CNNs and FCNs are hierarchical filter learning models in
which the weights of the network are usually tuned by using a
stochastic back-propagation error gradient decent optimization scheme.
Since CNN architectures often include millions of trainable
parameters, the training process is relies the sheer size of the
dataset. Moreover, although fully-supervised models generally tend to
perform better when given more training samples, they can still
generalize poorly to unseen/novel classes not present in the
training set.

For the task of semantic segmentation, establishing large-scale
datasets with pixel-level annotations (that are not synthetic
\citep{DBLP:journals/ijcv/JiangQZHLYTZ18}) is time-consuming and
prohibitively costly, and it may not be possible to include all
possible classes in the training set. Although semi-supervised
approaches aim to relax the level of supervision to bounding boxes and
image-level tags, these models still require copious training samples
and are prone to sub-optimal performance on unseen classes.

\begin{figure}
\includegraphics[width=\linewidth]{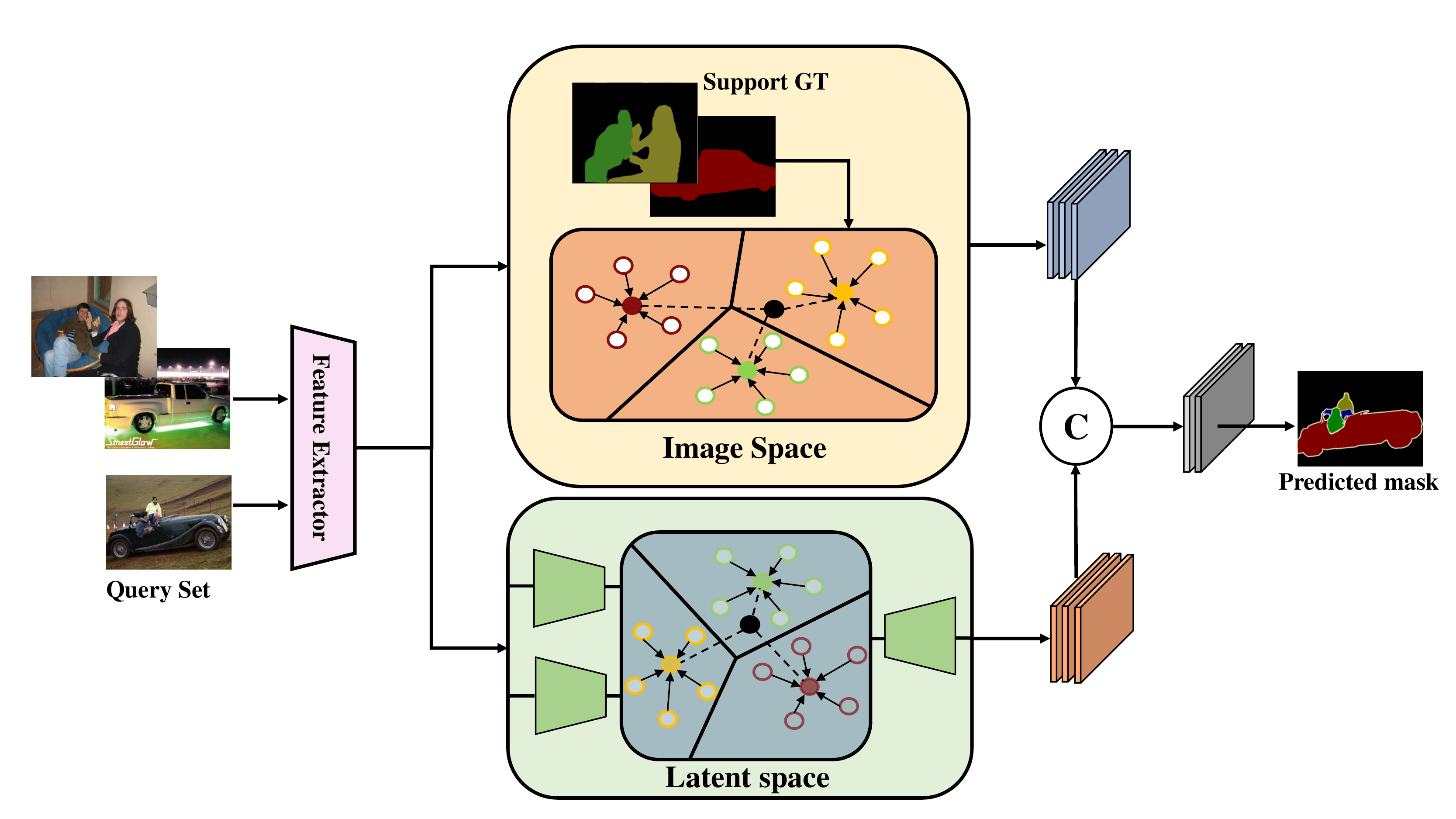}
\caption[Overview of the SegAVA's architecture] {Overview of the
SegAVA architecture for few-shot semantic segmentation.}
\label{fig:pipeline_highlevel_intro}
\end{figure}

\begin{figure}
\begin{subfigure}{\linewidth}
\centering
\includegraphics[width=\linewidth]{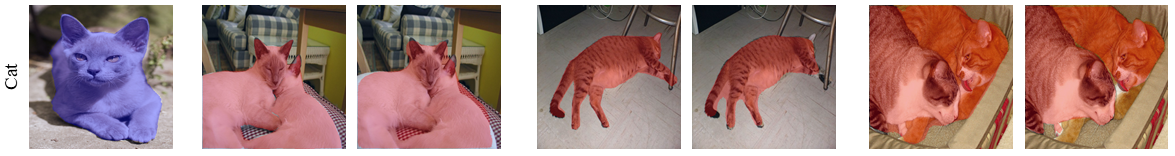}  
\label{fig:sub-first1teas2}
\end{subfigure}
\begin{subfigure}{\linewidth}
\centering
\includegraphics[width=\linewidth]{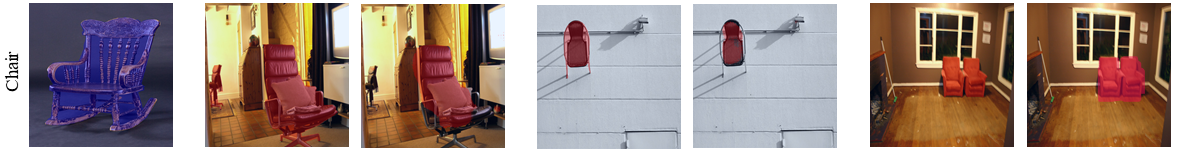}  
\label{fig:sub-first3teas2}
\end{subfigure}
\begin{subfigure}{\linewidth}
\centering
\includegraphics[width=\linewidth]{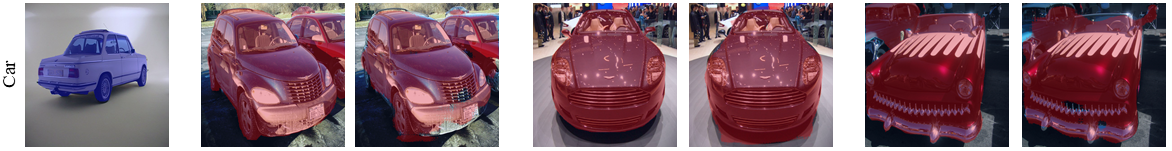}  
\label{fig:sub-first4teas2}
\end{subfigure}
\caption[SegAVA's outputs in 1-way, 1-shot task on the PASCAL-5i
dataset] {Example results from evaluating SegAVA in 1-way, 1-shot
segmentation on the PASCAL-5i dataset.}
\label{fig:teas2}
\end{figure}

By contrast, the few-shot learning \citep{lake2015human} paradigm
attempts to utilize a few annotated samples, referred to as ``support
samples'', to learn novel representations that belong to unseen
classes, denoted as ``query samples''.
The few-shot learning paradigm was initially focused on image
classification and later expanded to image segmentation
\citep{shaban2017one, dong2018few}. We propose a novel framework for
few-shot image segmentation
(Figure~\ref{fig:pipeline_highlevel_intro}), which we call
Segmentation with Aligned Variational Autoencoders (SegAVA), that
explores the latent and image spaces of support and query sets to find
the most common class-specific embeddings, and fuses them to produce
the final semantic segmentation. We have applied SegAVA to the task of semantic segmentation of natural images (Figure~\ref{fig:teas2}). 

\section{Contributions}

The specific contributions of this thesis are as follows:
\begin{enumerate}

\item \textbf{Edge-Aware 2D Image Segmentation Networks\\
\citep{hatamizadeh2019endbound,hatamizadeh2019boundary}:} Fully convolutional neural networks
(CNNs) have proven to be effective at representing and classifying
textural information, thus transforming image intensity into output
class masks that achieve semantic image segmentation. In medical image
analysis, however, expert manual segmentation often relies on the
boundaries of anatomical structures of interest. We propose 2D
edge-aware CNNs for medical image segmentation. Our networks are
designed to account for organ boundary information, both by providing
a special network edge branch and edge-aware loss terms, and they are
trainable end-to-end. We validate their effectiveness on the task of
brain tumor segmentation using the BraTS 2018 dataset. Our experiments
reveal that our approach yields more accurate segmentation results,
which makes it promising for more extensive application to medical
image segmentation.

\item \textbf{Edge-Aware 3D Image Segmentation Networks\\
\citep{hatamizadeh2019kidney}:}
Automated segmentation of kidneys and kidney tumors is an important
step in quantifying the tumor's morphometrical details to monitor the
progression of the disease and accurately compare decisions regarding
the kidney tumor treatment. Manual delineation techniques are often
tedious, error-prone and require expert knowledge for creating
unambiguous representation of kidneys and kidney tumors segmentation.
We propose a 3D end-to-end edge-aware FCN for reliable kidney and
kidney tumor semantic segmentation from arterial phase abdominal 3D CT
scans. Our segmentation network consists of an encoder-decoder
architecture that specifically accounts for organ and tumor semantics.
We evaluate our model on the 2019 MICCAI KiTS Kidney Tumor
Segmentation Challenge dataset.

\item \textbf{Plug-and-Play Edge-gated 3D Image Segmentation
Networks\\ \citep{hatamizadeh2020edge}:} We propose a plug-and-play
module, dubbed Edge-Gated CNNs (EG-CNNs), that can be used with
existing encoder-decoder architectures to process both edge and
texture information. The EG-CNN learns to emphasize the edges in the
encoder, to predict crisp boundaries by an auxiliary edge supervision,
and to fuse its output with the original CNN output. We evaluate the
effectiveness of the EG-CNN against various mainstream CNNs on the
publicly available BraTS19 dataset for brain tumor semantic
segmentation, and demonstrate how the addition of EG-CNN consistently
improves segmentation accuracy and generalization performance.

\item \textbf{Deep Active Lesion Segmentation\\
\citep{hatamizadeh2019deeplesion}:} Lesion segmentation is an
important problem in computer-assisted diagnosis that remains
challenging due to the prevalence of low contrast, irregular
boundaries that are unamenable to shape priors. We introduce Deep
Active Lesion Segmentation (DALS), a fully automated segmentation
framework that leverages the powerful nonlinear feature extraction
abilities of FCNs and the precise boundary delineation abilities of
ACMs. Our DALS framework benefits from an improved level-set ACM
formulation with a per-pixel-parameterized energy functional and a
novel multiscale encoder-decoder CNN that learns an initialization
probability map along with parameter maps for the ACM. We evaluate our
lesion segmentation model on a new Multiorgan Lesion Segmentation
(MLS) dataset that contains images of various organs, including brain,
liver, and lung, across different imaging modalities---MR and CT. Our
results demonstrate favorable performance compared to competing
methods, especially for small training datasets.

\item \textbf{End-to-End Trainable Deep Active Contour Models\\
\citep{hatamizadehdcac2019}:} The automated segmentation of buildings
in aerial images is an important task in many applications, which
requires the accurate delineation of multiple building instances of
interest over a typically large area of pixel space. Manual methods
are often laborious and current deep learning approaches typically
suffer from inaccurate delineation of segmented instances. We
introduce Deep Trainable Active Contours (DTAC), an end-to-end
trainable image segmentation framework that unifies a CNN and a
differentiable localized ACM with learnable parameters for fast and
robust delineation of buildings in satellite imagery. The ACM's
Eulerian energy functional includes per-pixel parameter maps predicted
by the backbone CNN, which also initializes the ACM. Importantly, both
the CNN and ACM components are fully implemented in TensorFlow, and
the entire DTAC architecture is end-to-end automatically
differentiable and backpropagation trainable without user
intervention. Unlike earlier efforts employing Lagrangian ACMs for
building segmentation, our DTAC enables the fast and fully automated
simultaneous delineation of arbitrarily many instances of buildings.
We validate our model on two publicly available aerial image datasets
for building segmentation (Vaihingen and Bing Huts), and our results
demonstrate that DTAC establishes a new state-of-the-art performance.

\item \textbf{Few-Shot Semantic Segmentation:} We address the
challenging problem of few-shot image segmentation by feature
alignment in the image and latent spaces of support and query samples.
Our model, which is dubbed SegAVA, leverages a latent stream as well
as an encoder-decoder stream to extract the most essential
discriminative semantic embeddings and learn similarities in both
spaces. The latent stream consists of two variational autoencoders,
conditioned on support and query sets, that jointly learn to generate
the input images and discriminatively identify the most common
class-specific representations using a Wasserstein-2 metric. These
embedding are then decoded to the image space and concatenated into a
common representation found by comparing support and query extracted
features using our fully convolutional decoder. We train and test our
SegAVA model using the PASCAL-5i dataset, and our results demonstrate
new state-of-the-art performance in 1-shot and 5-shot scenarios. We
also validate the SegAVA model in a semi-supervised setting where only
bounding boxes are provided, and the results demonstrate the
effectiveness of our approach.

\end{enumerate}

\section{Overview}

The remainder of the thesis is organized as follows:

In Chapter~\ref{cha:related-work}, we review the relevant literature
in the area of edge-aware CNN networks that utilize edge and texture
information in a specialized manner, hybrid frameworks that leverage
ACMs and CNNs within a single segmentation pipeline, and few-shot
learning with an emphasis on semantic image segmentation.

In Chapter~\ref{cha:bound}, we propose an end-to-end edge-aware
network that processes texture and edge information in dedicated
branches, the latter supervised with edge-aware loss functions.
Additionally, we propose EG-CNN, which is a plug-and-play, volumetric
(3D) segmentation module that can be paired with any existing
volumetric CNN architecture so as to disentangle texture and edge
processing and improve the segmentation accuracy near intensity edges.

In Chapter~\ref{cha:dtac}, we propose DTAC, an end-to-end trainable
image segmentation framework that unifies ACMs and CNNs, resulting in
a differentiable ACMs with learnable parameters for fast and robust
segmentation and delineation.

In Chapter~\ref{cha:few}, we propose SegAVA, an end-to-end, few-shot
segmentation framework that leverages a latent stream as well as an
encoder-decoder stream to extract the most essential discriminative
semantic embeddings and learn similarities in both spaces and
efficiently segment images, given only a handful of labeled examples.

In Chapter~\ref{cha:experiments}, we describe our experiments with the
models developed in the previous chapters and benchmark our results.

Chapter~\ref{cha:conclusions} presents the conclusions of the thesis
and suggests promising future research directions.

Appendix~\ref{cha:appendix} presents a novel deep learning-based
methodology for 3D human lung lobe segmentation.

\chapter{Related Work}
\label{cha:related-work}

In this chapter, we first review the relevant research focusing on
image segmentation using FCNs. We then review efforts at designing
networks that are more aware of boundaries, as well efforts to combine
ACMs and CNNs. Finally, we review relevant work in few-shot learning
and, in particular, few-shot image segmentation.

\section{Fully Convolutional Networks for Image Segmentation}

\subsection{Natural Image Segmentation}

\begin{figure}
\subcaptionbox{\label{fig:fcn}}{\includegraphics[width=0.47\linewidth]{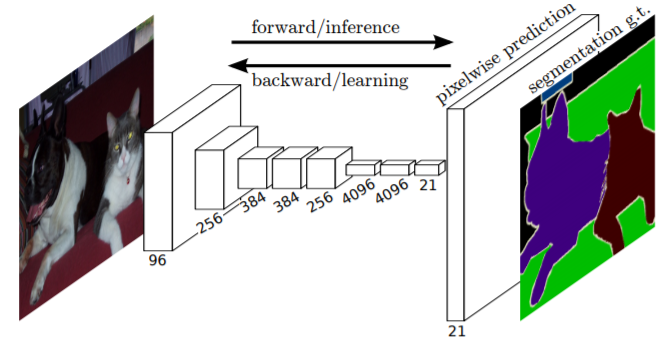}}
\hfill
\subcaptionbox{\label{fig:parsnet}}{\includegraphics[width=0.47\linewidth]{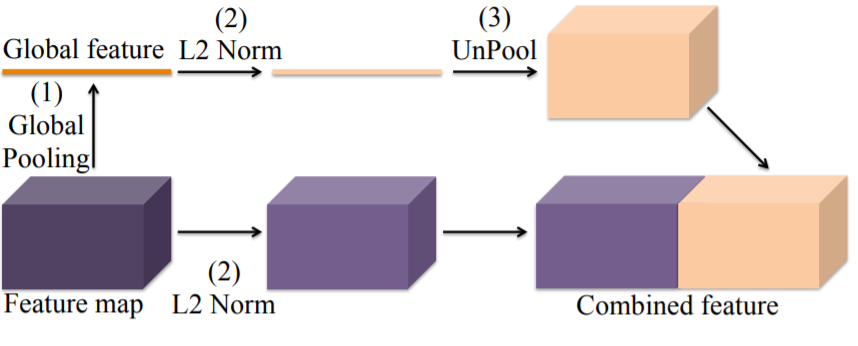}}
\caption[Visualization of the FCN and ParsNet architectures] {(a)
Architecture of FCNs for image segmentation. (b) Architecture of the
ParsNet context module. Images from \citep{chen2014semantic},
\citep{noh2015learning}, and \citep{badrinarayanan2017segnet}.}
\label{fig:convpars}
\end{figure}

\citet{long2015fully} introduced fully convolutional neural networks
(Figure~\ref{fig:fcn}) for semantic segmentation, interleaving
convolutional and pooling layers to learn the combined semantic and
appearance information, eventually generating per-pixel prediction
maps wherein boundaries were often blurred due to the reduction of
resolution. \citet{liu2015parsenet} proposed a global context module
(Figure~\ref{fig:parsnet}) that alleviated the issue of local
confusion.

\begin{figure}
\subcaptionbox{\label{fig:deeplab}}{\includegraphics[width=\linewidth]{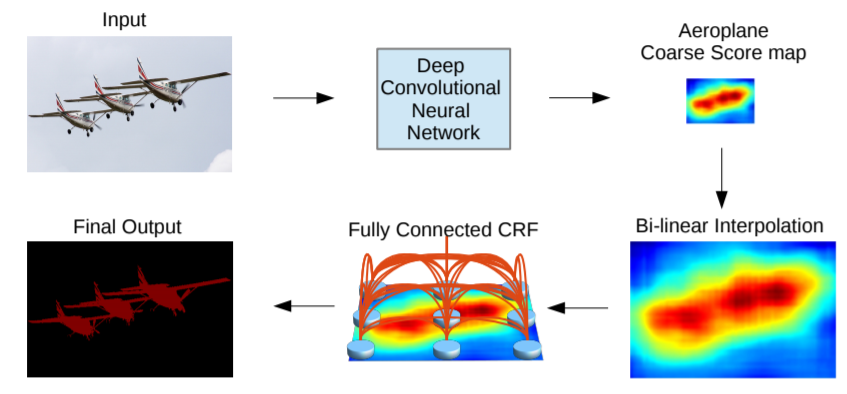}}\\
\subcaptionbox{\label{fig:bggnet}}{\includegraphics[width=0.48\linewidth]{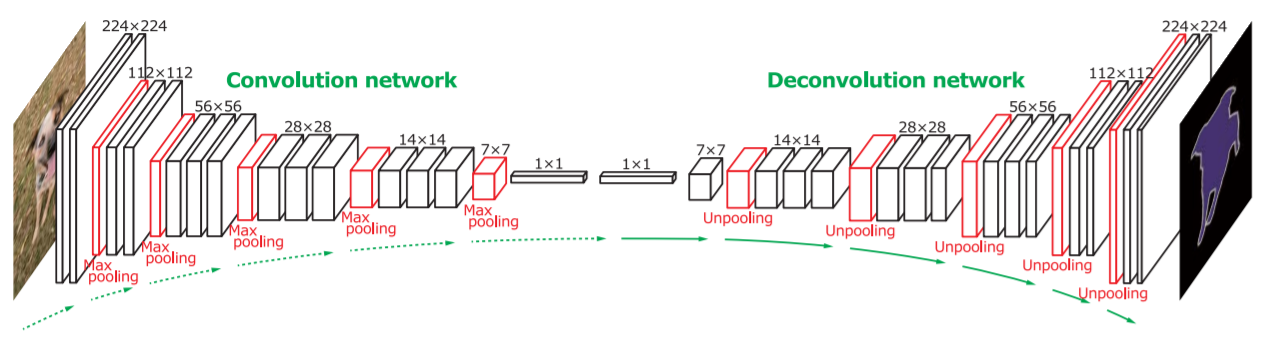}}
\hfill
\subcaptionbox{\label{fig:segnet}}{\includegraphics[width=0.48\linewidth]{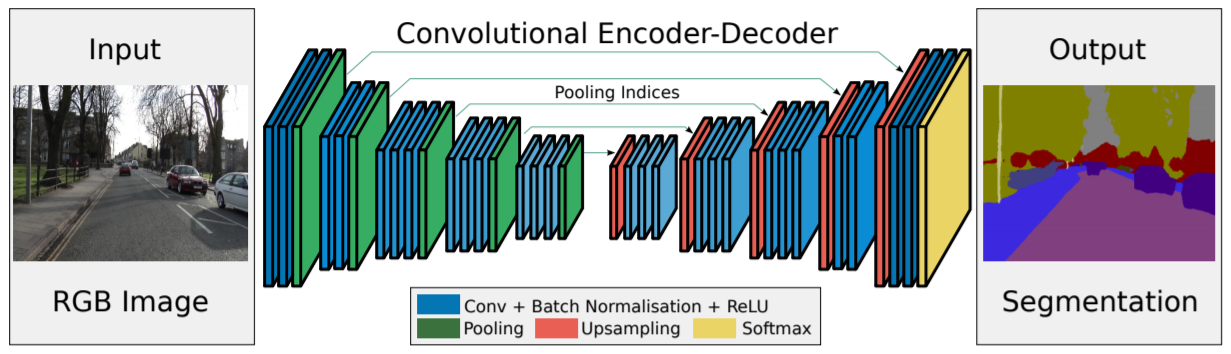}}
\caption[Architectures of DeepLab, DeconvNet, and SegNet] {(a)
Overview of DeepLab. (b) Architecture of deconvolutional network. (C)
Architecture of SegNet. Images from \citep{long2015fully} and
\citep{liu2015parsenet}.}
\label{fig:deeplab_segnet}
\end{figure}

Furthermore, \citet{chen2014semantic} proposed to combine the output
of the last layer of a CNN with a fully connected Conditional Random
Field (CRF) in order to overcome the poor localization property of
CNNs. Their model, which they called DeepLab
(Figure~\ref{fig:deeplab}), achieved significantly better segmentation
predictions near edges due to the ability of CRFs to fully delineate
mis-segmented regions. One of the early efforts that utilized an
encoder-decoder-like architecture for semantic segmentation is by
\citet{noh2015learning}, where a decoding network consisting of
deconvolutional and unpooling layers was added to a VGG16 backbone
\citep{simonyan2014very} for predicting pixel-wise outputs
(Figure~\ref{fig:bggnet}). Following this work,
\citet{badrinarayanan2017segnet} proposed to use an encoder-decoder
architecture (Figure~\ref{fig:segnet}), without the VGG16 backbone,
where the low-resolution, encoded feature maps are decoded back up to
the original input image resolution.

A follow-up effort by \citet{chen2017deeplab}, called DeepLabv2,
extended this DeepLab framework by leveraging the power of dilated
convolutional layers to explicitly control the resolution of the
feature responses and enlarge the field of view of filters without
additional free parameters. In addition, this work introduced a novel
module, dubbed Dilated Spatial Pyramid Pooling (DSPP), which enabled
accurate segmentation at multiple resolutions.

\begin{figure}
\subcaptionbox{\label{fig:pspnet}}{\includegraphics[width=0.47\linewidth]{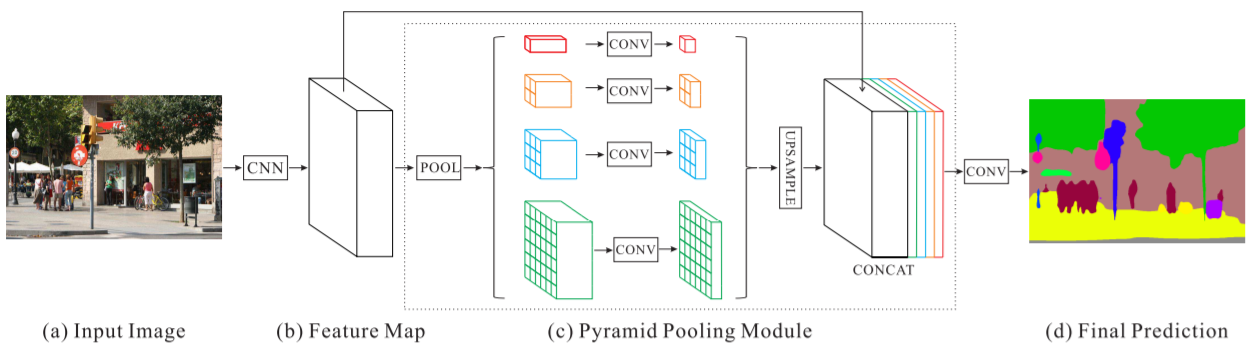}}
\hfill
\subcaptionbox{\label{fig:deeplabv3plus}}{\includegraphics[width=0.47\linewidth]{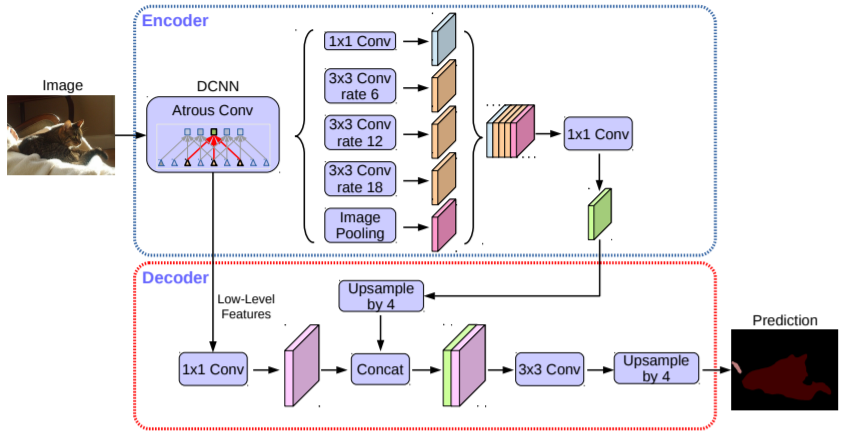}}
\caption[Visualization of the PSPNet and DeepLabv3+ architectures]
        {(a) Architecture of the PSPNet. (b) Architecture of
        DeepLabv3+. Images from \citep{zhao2017pyramid} and
        \citep{chen2018encoder}.}
\end{figure}

The use of multi-scale information for semantic segmentation has also
been explored by various researchers and shown to be effective.
\citet{yu2015multi} proposed an architecture that uses dilated
convolutions in order to increase the receptive fields in an efficient
manner while aggregating multi-scale semantic information.
\citet{zhao2017pyramid} introduced the pyramid scene parsing network
(PSPNet) (Figure~\ref{fig:pspnet}), which extracted and aggregated
global context information and improved the quality of segmentation
without employing computationally expensive post processing methods
like the CRF used in \citep{chen2014semantic}.

DeepLabv3 \citep{chen2017rethinking} attempted to capture multi-scale
context by using multiple dilation rates in cascaded and DSPP modules
that leveraged dilated convolutions. Furthermore, DeepLabv3+
\citep{chen2018encoder} (Figure~\ref{fig:deeplabv3plus}) employed an
architecture similar to DeepLabv3 \citep{chen2017rethinking}, but
proposed the use of an decoder network to improve segmentation
accuracy around edges. In DeepLabv3+, depthwise separable
convolutional layers were used in both the DSPP module and decoder
network and reportedly improved the computational performance.

\subsection{Medical Image Segmentation}

A seminal paper in deep learning applied to medical image segmentation
is that by \citet{Ronneberger15}, which introduces a 2D FCN comprising
an encoder and decoder that are connected by skip connections at
different resolutions. This work was later extended
\citep{cciccek20163d} to 3D segmentation. \citet{Milletari2016}
proposed an encoder-decoder architecture with residual blocks, denoted
as V-Net, for volumetric medical image segmentation.
\citet{gibson2018automatic} expanded the V-Net work by introducing
dense feature blocks in the encoder network. \citet{Myronenko18}
applied an asymmetric encoder-decoder architecture with residual
blocks to 3D brain tumor segmentation.

Variants of the U-Net encoder-decoder architecture have been proposed
for various applications. \citet{li2018h} introduced a hybrid
architecture consisting of 2D and 3D U-Nets with dense blocks for the
task of liver segmentation. \citet{jin2019dunet} proposed a 2D U-Net
architecture with deformable convolutions for the task of retinal
vessel segmentation. For this segmentation task,
\citep{hatamizadeh2019deepvessel, hatamizadeh2020artificial} proposed
an encoder-decoder architecture that leverages dilated spatial pyramid
pooling with multiple dilation rates to recover the lost content in
the encoder and add multiscale contextual information to the decoder.

\section{Edge-Aware Networks for Image Segmentation}

This section separately reviews relevant literature on natural image
segmentation and additional literature on medical image segmentation.

\subsection{Natural Image Segmentation}

Since the advent of deep learning, several efforts have been dedicated
in particular to edge prediction and enhancing the quality of
boundaries in the segmented areas. \citet{ZhidingYu17} proposed a
multi-label semantic boundary detection network to improve a wide
variety of vision tasks by predicting edges directly. They included a
new skip-layer architecture in which category-wise edge activations at
the top convolution layer share and are fused with the same set of
bottom layer features, along with a multi-label loss function to
supervise the fused activations.

\citet{yu2017casenet} proposed a category-aware semantic edge
detection framework in which direct predictions of edges improved a
wide variety of vision tasks. Their method includes a skip-layer CNN
architecture in which category-wise edge activations of the top and
bottom convolution layers are shared and fused together. In addition,
\citet{Yu_2018_ECCV} demonstrated the vulnerability of CNNs to
misaligned edge labels and proposed a framework for the simultaneous
alignment and learning of the edges.

For the task of portrait image segmentation, \citet{chen2019boundary}
proposed a lightweight 2D encoder-decoder architecture with an added
branch, consisting of boundary feature mining for selectively
extracting detailed information of boundaries from the output
segmentation of the CNN. Aiming to learn semantic boundaries,
\citet{HuPanoptic2019} presented a framework that aggregates different
tasks of object detection, semantic segmentation, and instance edge
detection into a single holistic network with multiple branches,
demonstrating significant improvements over conventional approaches
through end-to-end training.

\citet{Acuna_2019_CVPR} predicted object edges by identifying pixels
that belong to class boundaries, proposing a new layer and a loss that
enforces the detector to predict a maximum response along the normal
direction at an edge, while also regularizing its direction.
\citet{takikawa2019gated} proposed a framework for semantic instance
segmentation of objects in the Cityscapes dataset
\citep{cordts2016cityscapes} in which such gates are employed to
remove the noise from higher-level activations and process the
relevant boundary-related information separately.

\subsection{Medical Image Segmentation}

An early model for medical image segmentation with an emphasis on edge
learning is DCAN \citep{chen2016dcan}, in which the output of the decoder is also branched to learn the edges. However, DCAN does not prioritize
such a learning scheme in a dedicated path and fusion simply amounts
to the concatenation of the learned feature maps to the output of the
main CNN. Consequently, this approach does not generalize well to more
sophisticated segmentation tasks with irregular shapes. Subsequently,
the CIA-Net \citep{zhou2019cia} was introduced to address some of
these issues by incorporating a more sophisticated fusion module.

For the application of 2D brain tumor segmentation,
\citet{shen2017boundary} proposed the use of separate decoders for
learning the edges and tumor regions and concatenated the probability
outputs of each before feeding them into two consecutive convolutional
layers and a final softmax function. However, no specialized loss
functions were designated for the edge predictions and utilizing
replicated decoders with no effective connections is inefficient.

\citet{murugesan2019psi} introduced a edge-aware joint multi-task
framework for medical image segmentation that utilizes parallel
decoders, along with the main encoder-decoder stream, to perform
contour prediction and distance map estimation. The proposed effort
uses the same encoder for three parallel decoder streams, but does not
utilize the predicted contour and distance map in making the final
prediction.

\citet{zhang2019net} use a 2D edge attention guidance network to learn
the edge attention representation in the earlier stages of the
encoding process and transfer them to multi-scale decoding layers
where they are fused with the main encoder-decoder prediction using a
weighted aggregation module.

\subsubsection{Kidney and Kidney Tumor Segmentation}

Kidney cancer accounted for nearly 175,000 deaths worldwide in 2018
\citep{bray2018global}, and it is projected that 14,770 deaths will
occur due to the disease in 2019 in the US \citep{siegel2019cancer}.
Current kidney tumor treatment planning includes Radical Nephrectomy
(RN) and Partial Nephrectomy (PN). In RN, both the tumor and the
affected kidney are removed whereas in PN the tumor is removed but
kidneys are saved \citep{sun2012treatment}. Although RNs were
historically prevalent as a standard treatment procedure for kidney
tumors, new capabilities for earlier detection of the tumors as well
as advancements in surgery has made PNs a viable treatment approach
\citep{heller2019kits19}.

Traditionally, various techniques such as deformable models
\citep{mcinerney1996deformable}, GrabCuts, region growing and
atlas-based methods have been applied to the problem of kidney
segmentation. In recent years, researchers have attempted to leverage
the power of deep learning and CNNs to build segmentation frameworks
that are more automated and less dependant on incorporation of prior
shape statistics. \citet{thong2018convolutional} proposed a 2D
patch-based approach for kidney segmentation in contrast-enhanced CT
scans by leveraging a modified ConvNet.

\citet{jackson2018deep} developed a framework for detection and
segmentation and of kidneys in non-contrast CT images by utilizing a
3D U-Net. \citet{yang2018automatic} proposed a method for kidney and
renal tumor segmentation in CT angiography images by a modified
residual FCN that is equipped with a pyramid pooling module.
Furthermore, \citet{yin2019deep} employed a cascaded approach for
segmentation of kidneys with renal cell carcinoma by training a CNN
that predicts a bounding box around the kidney and a subsequent CNN
that segments the kidneys. Recently, \citet{xia2019deep} proposed a
two-stage approach for the segmentation of kidney and space-occupying
lesion areas by using SCNN and ResNet for image retrieval and
SIFT-flow and MRF for smoothing and pixel matching.

\section{End-to-End Trainable Deep Active Contours}

In this section, we first present relevant work on ACMs with an
emphasis on level-set ACMs. We then present a review of notable FCNs
for 2D image segmentation including approaches used for building image
segmentation. Finally, we review efforts that have attempted to
combine ACMs and CNNs within a segmentation pipeline.

\subsection{Level-Set ACMs}

Eulerian active contours evolve the segmentation curve by dynamically
propagating the zero level set of an implicit function so as to
minimize a corresponding functional \citep{osher2001level}. Level-set
ACM segmentation requires determining suitable parameter values for
the associated Partial Differential Equation (PDE), usually in a
tedious trial and error process where each parameter value is tested
over a series of images and remains the same for the entire image set.
New images with different statistics typically require re-tuning of
the parameters. Moreover, for images with diverse spatial statistics,
a fixed set of parameters may result in suboptimal segmentation
performance over all the images. Spatially adaptive parameters are
better suited to accurate segmentation.

Most notable approaches that utilize this formulation are active
contours without edges \citep{chan2001active} and geodesic active
contours \citep{caselles1997geodesic}. The Caselles-Kimmel-Sapiro
model is mainly dependent on the location of the level-set, whereas
the Chan-Vese model mainly relies on the content difference between
the interior and exterior of the level-set. In addition,
\citet{lankton2008localizing} reformulate the Chan-Vese model such
that the energy functional incorporates image properties in local
regions around the level-set, and it was shown to more accurately
segment objects with heterogeneous features.

\citet{oliveira2009liver} present a solution for liver segmentation
based on a deformable model in which the parameters are adjusted via a
genetic algorithm, but all the segmentations in their test set were
obtained by using the same set of parameters. They and
\citet{baillard} define the problem of parameter tuning as a
classification of each point along the contour, performed by
maximizing the posterior segmentation probability---if a point belongs
to the object, then the implicit surface should locally extend,
otherwise it should contract. However, only the direction of the curve
evolution is considered, not its magnitude, which is critical
especially in heterogeneous regions wherein convergence to local
minima should be prevented.

\citet{marquez2013morphological} proposed a morphological approach
that approximates the numerical solution of the PDE by successive
application of morphological operators defined on the equivalent
binary level set. \citet{hoogi2017adaptive} presented an alternative,
fully automatic model for the adaptive tuning of parameters, based on
estimating the zero level set contour location relative to the lesion
using the location probabilities, and showed significantly improved
segmentations.

\subsection{FCNs for Building Segmentation}

An early effort in leveraging CNN-based models for building
segmentation is by \citet{audebert2016semantic} who used SegNet
\citep{badrinarayanan2017segnet} with multi-kernel convolutional
layers at three different resolutions. Subsequently,
\citet{wang2017torontocity} proposed using ResNet \citep{he2016deep},
first to identify the instances, followed by an MRF to refine the
predicted masks. \citet{wu2018automatic} employed a U-Net
encoder-decoder architecture with loss layers at different scales to
progressively refine the segmentation masks. \citet{xu2018building}
proposed a cascaded approach in which pre-processed hand-crafted
features are fed into a Residual U-Net to extract the building
locations and a guided filter to refine the results.

Furthermore, \citet{bischke2019multi} proposed a cascaded multi-task
loss function to simultaneously predict the semantic masks and
distance classes in an effort to address the problem of poor boundary
predictions by CNN models. Recently, \citet{rudner2019multi3net}
proposed a method to segment flooded buildings using multiple streams
of encoder-decoder architectures that extract spatiotemporal
information from medium-resolution images and spatial information from
high-resolution images along with a context aggregation module to
effectively combine the learned feature map.

\subsection{Deep Learning Assisted Active Contours}

\citet{hu2017deep} proposed a model in which the network learns a
level-set function for salient objects; however, the authors
predefined a fixed weighting parameter $\lambda$, which will not be
optimal for all cases in the analyzed set of images. In medical image
analysis, the challenges are much more complex---variability between
images is high, there are many low-contrast images, and noise is very
common. \citet{ngo2017combining} proposed to combine deep belief
networks with implicit ACMs for cardiac left ventricle segmentation;
However, their approach requires additional prepossessing steps such
as edge detection and needs user intervention for setting the ACM's
parameters.

\citet{le2018reformulating} proposed a framework in which level-set
ACMs are implemented as RNNs for the task of semantic segmentation of
natural images. There are 3 key differences between that effort and
our proposed model: (1) our model does not reformulate ACMs as RNNs,
which makes it more computationally efficient. (2) our model benefits
from a novel locally-penalized energy functional, as opposed to
constant weighted parameters. (3) our model has an entirely different
pipeline---we employ a single CNN that is trained from scratch along
with the ACM, as opposed to requiring two \emph{pre-trained} CNN
backbones.

\citet{marcos2018learning} proposed Deep Structured Active Contours
(DSAC), an integration of ACMs with CNNs in a structured prediction
framework for building instance segmentation in aerial images. There
are 3 key differences between that work and our work: (1) our model is
fully automated and runs without any external supervision, as opposed
to depending heavily on the manual initialization of contours. (2) our
model leverages the Eulerian ACM, which naturally segments multiple
building instances simultaneously, as opposed to a parametric
formulation that can handle only a single building at a time. (3) our
approach fully automates the direct back-propagation of gradients
through the entire DTAC framework due to its automatically
differentiable ACM.

\citet{cheng2019darnet} proposed the Deep Active Ray Network (DarNet)
that uses polar coordinates instead of Euclidean coordinates, and rays
to prevent the problem of self-intersection, and employs a
computationally expensive multiple initialization scheme to improve
the performance of the proposed model. Like DSAC, DarNet can handle
only single instances of buildings due to its explicit formulation.
Our approach is inherently different from DarNet, as (1) it uses an
implicit ACM formulation that handles multiple building instances and
(2) leverages a CNN to automatically and precisely initialize the
implicit ACM.

\citet{gur2019end} used an explicit ACM, represented by a neural renderer, along with a backbone encoder-decoder that predicts a shift map to efficiently evolve the contour via edge displacement.  

Some efforts have also focused on deriving new loss functions that are
inspired by ACM principles. Inspired by the global energy formulation
of \citet{chan2001active}, \citet{chen2019learning} proposed a
supervised loss layer that incorporated area and size information of
the predicted masks during training of a CNN and tackled the problem
of ventricle segmentation in cardiac MRI. Similarly,
\citet{gur2019unsupervised} presented an unsupervised loss function
based on morphological active contours without edges
\citep{marquez2013morphological} for microvascular image segmentation.

\section{Few-Shot Learning}

\subsection{Few-Shot Classification}

In few-shot classification, the goal is to learn unseen classes given
a few labeled training examples for each class. Among different
approaches that have been proposed for this problem, metric-based
methodologies \citep{koch2015siamese, snell2017prototypical,
lifchitz2019dense} have grained the most traction. In such a paradigm,
a metric function compares the similarity between the extracted
features of labeled and unlabeled samples. \citet{vinyals2016matching}
introduced Matching Networks, which consisted of a recurrent neural
network and a cosine similarity metric function for one-shot
classification tasks. Similarly, \citet{snell2017prototypical}
presented a prototypical learning framework that used a Euclidean
distance function as the learning metric.

In contrast to these approaches that utilize fixed-distance metrics,
\citet{sung2018learning} used a convolutional neural network, denoted
as Relation Network, to learn to learn a deep distance metric in an
end-to-end manner. \citet{garcia2017few} expanded this idea and used a
graph convolutional neural network to learn the distance metric. Other
approaches have also sought to utilize the latent space for learning
the semantic embeddings. \citet{kim2019variational} introduced a
variational prototype encoder in which a generalizable embedding
latent space is learned for identifying novel categories.
\citet{schonfeld2019generalized} proposed to use a shared latent space
to identify important multi-domain information for unseen categories.

\subsection{Few-Shot Segmentation}

Few-shot semantic segmentation extends the idea of few-shot learning
to dense pixel-wise predictions. \citet{shaban2017one} were the first
to study the problem of 1-way semantic segmentation and used a
conditional branch to learn the important embedding in the support set
and combine it with query features in a separate branch to produce the
final segmentation. Furthermore, \citet{rakelly2018conditional}
introduced a network that was conditional on the support set and
performed inference on the query set via feature fusion.
\citet{Hu2019AttentionBasedMG} proposed an attention mechanism to
highlight multi-scale context features between support and query
features and used a Conv-LSTM to fuse learned features.

In contrast to the approaches that separately processed support and
query embeddings, \citet{zhang1810sg} used a
masked average pooling scheme to create guidance features from support
images and aggregated them with query features to obtain the final
segmentation using a unified pipeline. In this work, cosine similarity
was used to measure the distance between features in the support and
query sets. Following this single-branch strategy, \citet{AMP}
proposed a multi-resolution adaptive imprinting to identify the
similarities of extracted features.

\citet{feature_weights} computed a class feature vector as the average
of foreground areas in the extracted support features and used it to
compare against query features by a cosine similarity metric. In a
similar approach, \citet{wang2019panet} employed a prototypical
learning framework, PANet, in which support prototypes are extracted
by a masked average pooling and compared against query prototypes by
using the cosine similarity metric. Additionally, PANet uses a
prototype alignment regularization by using the predicted query masks
to further align the support and query embeddings.

Unlike earlier efforts, we utilize both latent and image spaces to
find the most common class-specific representations for the task of
few-shot semantic segmentation. Additionally, we introduce a fully
convolutional decoder to learn the similarities in the image space.
Our method achieves state-of-the-art results on the popular PASCAL-5i
dataset \citep{shaban2017one} and effectively segments images using
weaker levels of supervision, such as bounding boxes.

\chapter{Edge-Aware Semantic Segmentation Networks}
\label{cha:bound}

In this chapter, we first introduce a 2D encoder-decoder architecture
that leverages a special interconnected edge layer module that is
supervised by edge-aware losses in order to preserve boundary
information and emphasize it during training. By explicitly accounting
for the edges, we encourage the network to internalize edge importance
during training. Our method utilizes edge information only to assist
training for semantic segmentation, not for the main purpose of
predicting edges directly. This strategy enables a structured
regularization mechanism for our network during training and results
in more accurate and robust segmentation performance during inference.

Furthermore, we extend our methodology and propose 3D boundary-aware
FCNs for end-to-end and reliable semantic segmentation of kidneys and
kidney tumor by encoding the information of edges in a dedicated
stream that is supervised by edge-aware losses.

Lastly, we create a 3D plug-and-play module that we call the
Edge-Gated CNN (EG-CNN), which can be incorporated with any
encoder-decoder architecture to disentangle the learning of texture
and edge representations.
The contribution of the proposed EG-CNN is two-fold. First, EG-CNN
leverages an effective way to progressively learn to highlight the
edge semantics from multiple scales of feature maps in the main
encoder-decoder architecture by a novel and efficient layer denoted
the edge-gated layer. Second, instead of separately supervising the
edge and texture outputs, the EG-CNN uses a dual-task learning scheme,
in which these representations are jointly learned by a consistency
loss. Therefore, without increasing the cost of data annotation and by
exploiting the duality between edge and texture predictions, the
EG-CNN improves the overall segmentation performance with highly
detailed boundaries.

\section{2D Edge-Aware Encoder-Decoders}
\label{sec:2d-edge-aware}

\subsection{Architecture}

\begin{figure}
\includegraphics[width=\linewidth]{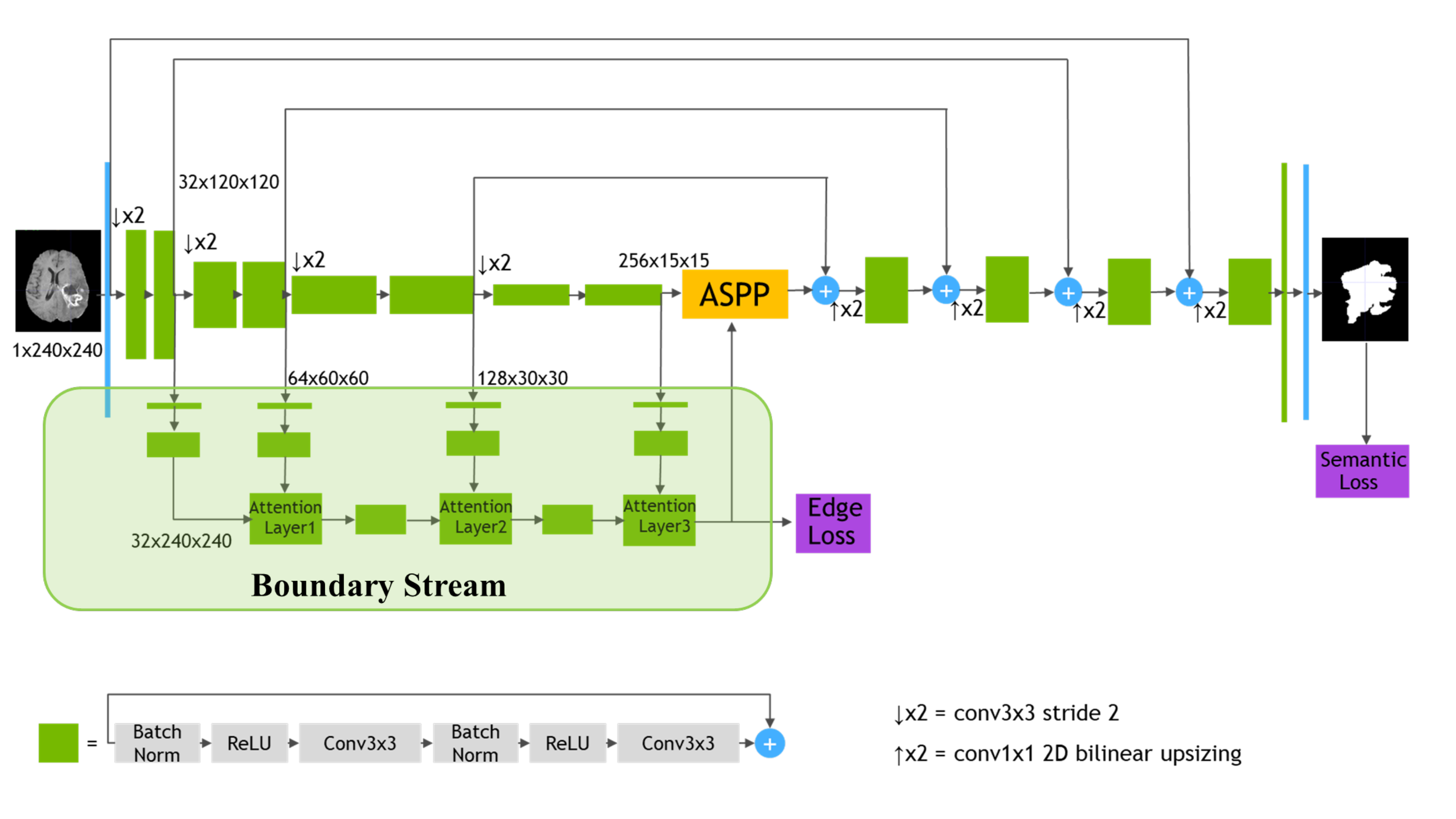}
\caption[2D fully convolutional edge-aware architecture] {2D fully
convolutional edge-aware architecture.}
\label{fig:pipeline_2d}
\end{figure}

Our network comprises a main encoder-decoder stream for semantic
segmentation as well as a shape stream that processes the feature maps
at the boundary level (Figure~\ref{fig:pipeline_2d}). In the encoder
portion of the main stream, every resolution level includes two
residual blocks whose outputs are fed to the corresponding resolution
of the shape stream. A $1\times1$ convolution is applied to each input
to the shape stream and the result is fed into an attention layer that
is discussed in the next section.

The outputs of the first two attention layers are fed into connection
residual blocks. The output of the last attention layer is
concatenated with the output of the encoder in the main stream and fed
into a dilated spatial pyramid pooling layer. Losses that contribute
to tuning the weights of the model come from the output of the shape
stream that is resized to the original image size, as well as the
output of the main stream.

\subsection{Attention Layer}

Each attention layer receives inputs from the previous attention layer
as well as the main stream at the corresponding resolution. Let $s_l$
and $m_l$ denote the attention layer and main stream layer inputs at
resolution $l$. First, $s_l$ and $m_l$ are concatenated and a
$1\times1$ convolution layer $C_{1 \times 1}$ is applied, followed by
a sigmoid function $\sigma$, to obtain an attention map:
\begin{equation}
\alpha_{l}=\sigma\big(C_{1 \times 1}(s_{l}\mathbin\Vert m_{l})\big).
\label{eq:att3}
\end{equation} 
An element-wise multiplication is then performed with the input to the
attention layer to obtain the output of the attention layer, denoted
as
\begin{equation}
o_{l}= s_{l} \odot \alpha_{l}.
\label{eq:att2}
\end{equation} 

\subsection{Edge-Aware Segmentation}

Our network jointly learns the semantics and boundaries by supervising
the output of the main stream as well as the edge stream. We use the
generalized Dice loss on predicted outputs of the main stream and the
shape stream. Additionally, we add a weighted binary cross entropy
loss to the shape stream loss in order to deal with the large
imbalance between the boundary and non-boundary pixels. The overall
loss function of our network is
\begin{equation}
L_\text{total}=\lambda_{1}
L_\text{Dice}(y_\text{pred},y_\text{true})+
\lambda_{2}L_\text{Dice}(s_\text{pred},s_\text{true})+
\lambda_{3} L_\text{Edge}(s_\text{pred},s_\text{true}),
\label{eq:finalloss}
\end{equation}
where $y_\text{pred}$ and $y_\text{true}$ denote the pixel-wise
semantic predictions of the main stream while $s_\text{pred}$ and
$s_\text{true}$ denote the boundary predictions of the shape stream;
$s_\text{true}$ can be obtained by computing the spatial gradient of
$y_\text{true}$.

The Dice loss \citep{Milletari2016} in (\ref{eq:finalloss}) is
\begin{equation}
L_\text{Dice}= 1- \frac{2 \sum y_\text{true} y_\text{pred} }{\sum
y_\text{true}^2 + \sum y_\text{pred}^2 + \epsilon},
\label{eq:dice}
\end{equation} 
where summation is carried over the total number of pixels and
$\epsilon$ is a small constant to prevent division by zero.

The edge loss in (\ref{eq:finalloss}) is
\begin{equation}
L_\text{Edge}= -\beta \sum_{j\in y_{+}} \log P(y_{\text{pred},j}=1\mid
x;\theta)-(1-\beta) \sum_{j\in y_{-}} \log P(y_{\text{pred},j}=0\mid
x;\theta),
\label{eq:bce}
\end{equation} 
where $x$, $\theta$, $y_{-}$, and $y_{+}$ denote the input image, CNN
parameters, and edge and non-edge pixel sets, respectively, $\beta$ is
the ratio of non-edge pixels over the entire number of pixels, and
$P(y_{\text{pred},j})$ denotes the probability of the predicated class
at pixel $j$.

\section{3D Edge-Aware Encoder-Decoders}
\label{sec:3d-edge-aware}

\begin{figure}
\includegraphics[width=\linewidth]{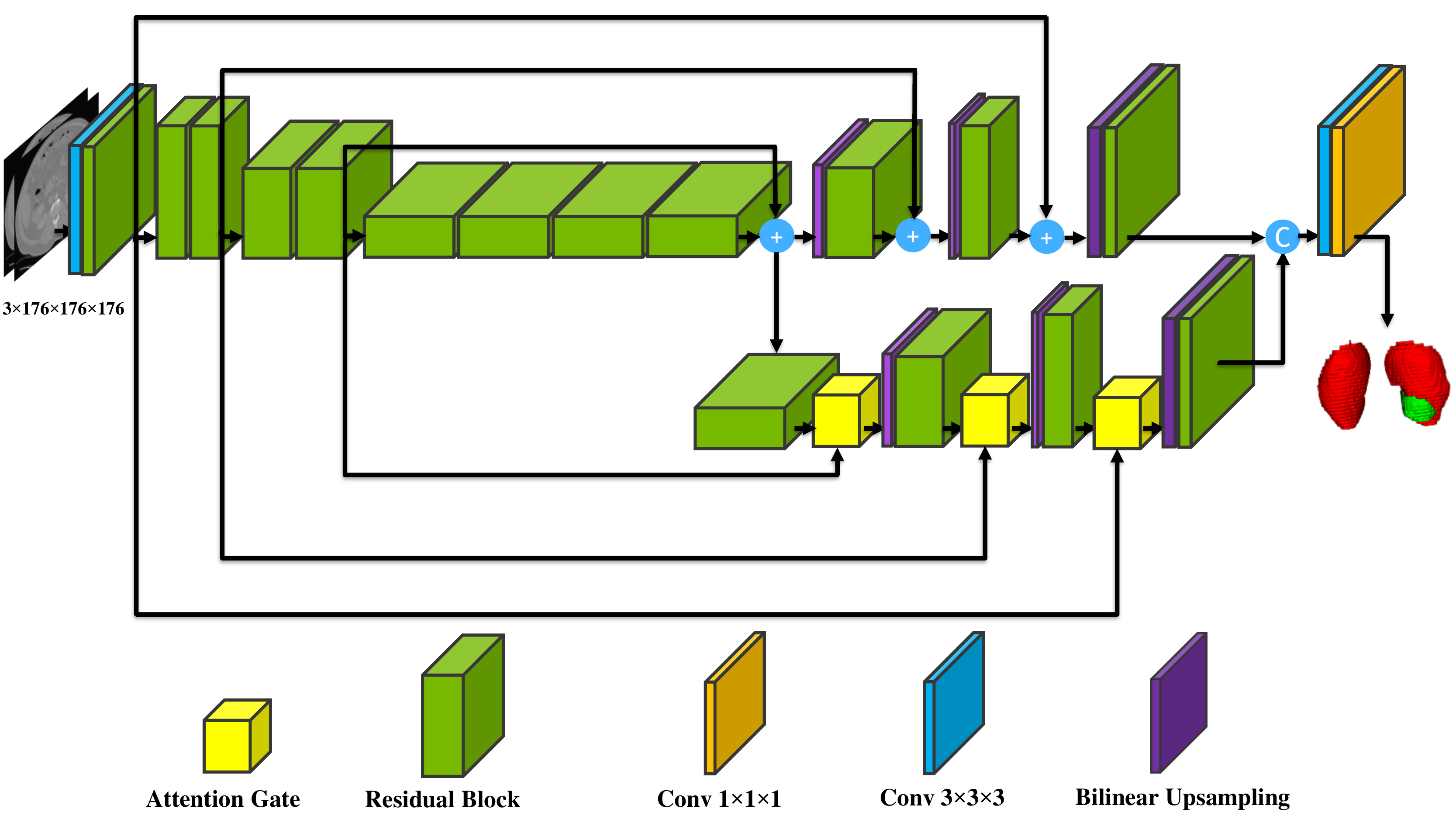}
\caption[Proposed volumentric edge-aware architecture] {Proposed
volumetric (3D) edge-aware architecture for kidney and kidney
tumor segmentation.}
\label{fig:pipeline}
\end{figure}

\subsection{Framework Architecture}

As is illustrated in Figure~\ref{fig:pipeline}, our network consists
of the main segmentation branch and the additional boundary stream
that processes the feature maps at the boundary level. The main
branch, following \citep{Myronenko18}, is an asymmetric
encoder-decoder structure. The input to the encoder is a
$176\times176\times176$ crop which is initially fed into a
$3\times3\times3$ convolution with 16 filters. Feature maps are then
extracted at each resolution by feeding them into a residual block
followed by a strided $3\times3\times3$ convolution (for downsizing
and doubling of the feature dimension).

The bottom of the encoder entails four consecutive residual blocks
that are connected to the decoder. The extracted feature maps in the
decoder are upsampled using bilinear interpolation and added with
feature maps from the encoder. The output of the decoder is
concatenated with the output of the boundary and fed into a
$1\times1\times1$ convolution with 2 channels where channel-wise
sigmoid activation $\sigma(X) =1/(1+e^{-X})$ determines the
probability of each voxel belonging to kidneys and tumor or only tumor
classes.

\subsection{Boundary Stream}

The purpose of the boundary stream is to highlight the edge
information of the feature maps extracted in the main encoder by
leveraging an additional attention-driven decoder. The attention gates
in every resolution of the boundary stream process the feature maps
that are learned in the main encoder as well as the output of the
previous attention gates.

For the first attention gate, we first concatenate the output of the
encoder with its previous resolution and feed it into a residual
block. In the attention gates, each input is first fed into a
$3\times3\times3$ convolutional layer with matching number of feature
maps and then fused together, followed by ReLU. The output of the ReLU
is fed into a $1\times1\times1$ convolution layer followed by sigmoid
function $\sigma$ to obtain the attention map. Consecutively, an
element-wise multiplication between the boundary stream feature maps
and the computed attention map results in the output of the attention
gates.

\subsection{Loss Functions}

We use a dice loss function on the predicted outputs of the main
stream as well as the boundary stream. The dice loss is as follows
\citep{Milletari2016}:
\begin{equation}
L_\text{Dice}= 1- \frac{2 \sum y_\text{true}\, y_\text{pred} }{\sum
y_\text{true}^2 + \sum y_\text{pred}^2 + \epsilon},
\label{eq:dice3}
\end{equation}   
where $y_\text{pred}$ and $y_\text{true}$ denote the voxel-wise
semantic predictions of the main stream and their corresponding
labels, $\epsilon$ is a small constant to avoid division by zero and
summation is carried over the total number of voxels.

Additionally, we add a weighted Binary Cross Entropy (BCE) loss to the
boundary stream loss in order to deal with the imbalanced number of
boundary and non-boundary voxels:
\begin{equation}
L_\text{BCE}= -\beta \sum_{j\in y_{+}} \log P(y_{\text{pred},j}=1\mid
x;\theta) -(1-\beta) \sum_{j\in y_{-}} \log P(y_{\text{pred},j}=0\mid
x;\theta),
\label{eq:finalloss2}
\end{equation}
where $x$, $\theta$, $y_{-}$, and $y_{+}$ denote the 3D input image,
CNN parameters, edge, and non-edge voxel sets, respectively, $\beta$
is the ratio of non-edge pixels over the entire number of voxels, and
$P(y_{\text{pred},j})$ denotes the probability of the predicated class
at voxel $j$.

The total loss function that is minimized during training is computed
by taking the average of losses for tumor-only and foreground class
predictions.

\section{Plug-and-Play Edge-Aware CNNs (EG-CNNs)}
\label{sec:EG-CNN}

We next present a plug-and-play edge-aware CNN, dubbed EG-CNN, and
introduce its architecture. The main stream, a generic CNN
encoder-decoder, learns feature representations that span multiple
resolutions. Our EG-CNN receives each of the feature maps in the main
stream and learns to highlight the edge representations. In
particular, the EG-CNN consists of a sequence of residual blocks
followed by tailored layers, as we denote the edge-gated layers, to
progressively extract the edge representations.

The output of the EG-CNN is then concatenated with the output of the
main stream in order to produce the final segmentation output.
Furthermore, the main stream and the EG-CNN are supervised by their
own dedicated loss layers as well as a consistent loss function that
jointly learns the output of both streams. The edge ground-truth is
generated online by applying a 3D Sobel filter to the original ground
truth masks.

Each edge-gated layer requires two inputs that originate from the main
stream and the EG-CNN stream. The intermediate feature maps from every
resolution of the main stream as well as the first up-sampled feature
maps in the decoder are fed to the EG-CNN as inputs.

\begin{figure}
\includegraphics[width=\linewidth]{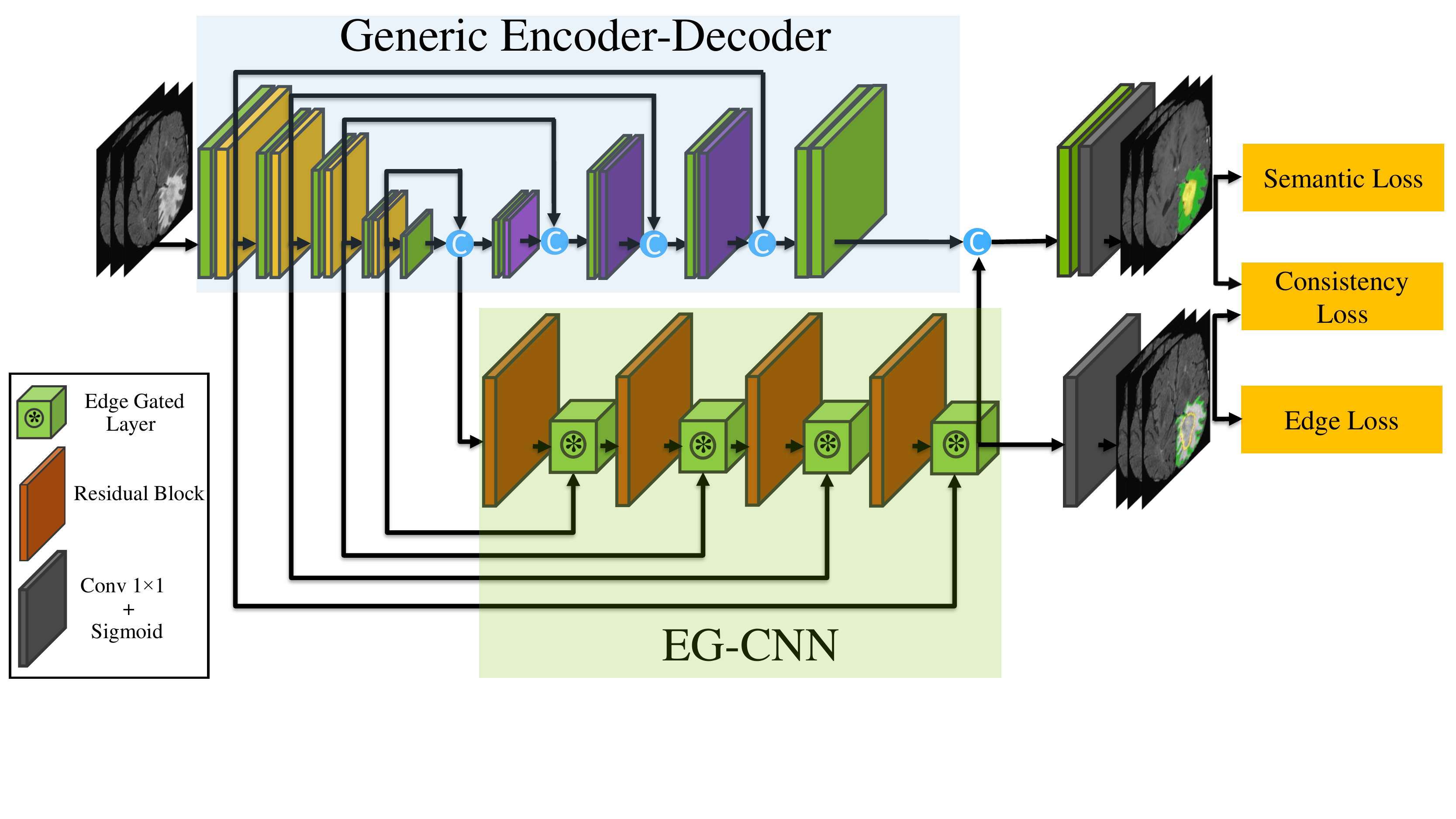}
\caption[EG-CNN architecture] {The EG-CNN module can be integrated
with any generic encoder-decoder architecture and highlight the edge
representations of the intermediate feature maps.}
\label{fig:pipeline_eg}
\end{figure}

The latter is first fed into a residual block followed by bilinear
upsampling before being fed into the edge-gated layer along with the
input from its previous resolution in the encoder. The output of each
edge-gated layer (except for the last one) is fed into another
residual block followed by bilinear upsampling before being fed to the
next edge-gated layer along with its corresponding input from the
encoder (Figure~\ref{fig:pipeline_eg}).

\subsection{Edge-Gated Layer}

Edge-gated layers highlight the edge features and connect the feature
maps learned in the main and edge streams. They receive inputs from
the previous edge-gated layers as well as the main stream at its
corresponding resolution. Let $e_{r,in}$ and $m_r$ denote the inputs
coming from edge and main streams, respectively, at resolution $r$.
First, an attention map, $\alpha_{r}$ is obtained by feeding each
input into a $1\times1\times1$ convolutional layer,
$C_{1\times1\times1}$, fusing the outputs and passing them into a
rectified linear unit (ReLU) $Re(X) = \max(0, X)$ according to
\begin{equation}
\alpha_{r}= \sigma\big(Re(C_{1\times1\times1}(e_{r,in}) +
C_{1\times1\times1}(m_{r}))).
\label{eq:att1}
\end{equation} 
The obtained attention map $\alpha_{r}$ is then pixel-wise multiplied
by $e_{r,in}$ and fed into a residual layer with kernel $w_{r}$.
Therefore, the output of each resolution in EG-CNN ,$e_{r,out}$, can
be represented as
\begin{equation}
e_{r,out}= e_{r,in} \odot \alpha_{r}+e_{r,in}.
\label{eq:att4}
\end{equation} 
The computed attention map highlights the edge semantics that are
embedded in the main stream feature maps. In general, there will be as
many edge-gated layers as the number of different resolutions in the
main encoder-decoder CNN architecture.

\subsection{Loss Functions}

The total loss of the EG-CNN is as follows:
\begin{equation}
L_\text{Tot}= L_\text{Semantic} + L_\text{Consistency}+ L_\text{Edge},
\label{eq:egcnnloss}
\end{equation}
where $L_\text{Semantic}$ represent standard loss functions used for
supervising the main stream in a semantic segmentation network,
$L_\text{Edge}$ represent tailored losses for learning the edge
representations, and $L_\text{Consistency}$ is a dual-task loss for
the joint learning of edge and texture and enforces the class
consistency of predictions.

\paragraph{Semantic Loss:}
Without loss of generality, we use the Dice loss \citep{Milletari2016}
for learning the semantic representations of texture according to
\begin{equation}
L_\text{Dice}= 1- \frac{2 \sum y_\text{true} y_\text{pred} }{\sum
y_\text{true}^2 + \sum y_\text{pred}^2 + \epsilon},
\label{eq:dice4}
\end{equation} 
where summation is carried over the total number of pixels,
$y_\text{pred}$ and $y_\text{true}$ denote the pixel-wise semantic
predictions of the main stream, and $\epsilon$ is a small constant to
prevent division by zero.

\paragraph{Edge Loss:}
The edge loss used in EG-CNN comprises of Dice loss
\citep{Milletari2016} and balanced cross entropy \citep{yu2017casenet},
as follows:
\begin{equation}
L_\text{Edge}= \lambda_{1} L_\text{Dice} + \lambda_{2}L_\text{BCE},
\label{eq:attloss2}
\end{equation}
where $\lambda_{1}$ and $\lambda_{2}$ are hyper-parameters. Let
$e_{\text{pred},j}$ and $e_{\text{true},j}$ denote the edge prediction
outputs of the EG-CNN and its corresponding groundtruth at voxel $j$,
respectively. Then the balanced cross entropy $L_\text{BCE}$ used in
(\ref{eq:attloss2}) can be defined as
\begin{equation}
L_\text{BCE}= -\beta \sum_{j\in e_{+}} \log P(e_{\text{pred},j}=1\mid
x;\theta)-(1-\beta) \sum_{j\in e_{-}} \log P(e_{\text{pred},j}=0\mid
x;\theta),
\label{eq:bce2}
\end{equation} 
where $x$, $\theta$, $e_{-}$, and $e_{+}$ denote the input image, CNN
parameters, edge, and non-edge voxel sets, respectively, $\beta$ is
the ratio of non-edge voxels to all voxels, and $P(e_{\text{pred},j})$
is the probability of the predicated class at voxel $j$. The cross
entropy loss follows (\ref{eq:bce2}) except for the fact that non-edge
voxels are not weighted.

\paragraph{Consistency Loss:}
We exploit the duality of edge and texture predictions and
simultaneously supervise the outputs of the edge and main stream by
the consistency loss. Inspired by \citep{takikawa2019gated}, the
semantic probability predictions of the main CNN architectures and the
ground truth masks are first converted into edge predictions by taking
the spatial derivative in a differentiable manner. Subsequently, we
penalize the mismatch between the boundary predictions of the semantic
masks and the corresponding ground truth by utilizing an $L_1$ loss.
Let $y_{\text{pred},j}$ denote the output of the main stream and $c$
represent the segmentation class. We propose a consistency loss
function
\begin{equation}
L_\text{Consistency}= \sum_{j\in e_{+}} \bigl(\|
{\nabla(\argmax(P(y_{\text{pred},j}=1\mid e;c))} \|)-\|
{\nabla(y_{\text{true},j}} \|\bigr).
\label{eq:consistency}
\end{equation} 
Due to the non-differentiability of the $\argmax$ function, we
leverage the Gumbel softmax trick \citep{jang2016categorical} to avoid
blocking the error-gradient. Thus, the gradient of the $\argmax$ can
be approximated according to
\begin{equation}
\frac{\partial \argmax_{t}P(y^{t})}{\partial \gamma_{j}}= \nabla_{j}
\frac{e^{(\log P(y_{t})+g_{t})/\tau}}{\sum_{i} e^{(\log
P(y_{i})+g_{i})/\tau}},
\label{eq:gumbeltrick}
\end{equation} 
where $\gamma$ is a differentiation dummy variable, $\tau$ is the
temperature, set as a hyper-parameter, and $g_{i}$ denotes the Gumbel
density function.

\chapter{End-to-End Trainable Deep Active Contour Models}
\label{cha:dtac}

ACMs \citep{kass1988snakes} have been extensively applied to computer
vision tasks such as image segmentation, especially for medical image
analysis \citep{mcinerney1996deformable}. ACMs leverage parametric
(``snake'') or implicit (level-set) formulations in which the contour
evolves by minimizing an associated energy functional, typically using
a gradient descent procedure. In the level-set formulation, this
amounts to solving a PDE to evolve object boundaries that are able to
handle large shape variations, topological changes, and intensity
inhomogeneities. Alternative approaches to image segmentation that are
based on deep learning have recently been gaining in popularity. CNNs
can perform well in segmenting images within datasets on which they
have been trained, but they may lack robustness when cross-validated
on other datasets. Moreover, in medical image segmentation, CNNs tend
to be less precise in boundary delineation than ACMs.

In this chapter, we establish a modeling framework that benefits from
data-driven non-linear feature extraction capabilities of CNNs and
versatility of ACMs. In essence, our goal is to employ a backbone CNN
for initializing and guiding the ACM in a fully automated manner and
without any user interaction.

First, we introduce a fully automatic framework for medical image
segmentation that combines the strengths of CNNs and level-set ACMs to
overcome their respective weaknesses. We apply our proposed Deep
Active Lesion Segmentation (DALS) framework to the challenging problem
of segmenting lesions in MR and CT medical images, dealing with lesions of substantially
different sizes within a single framework. In particular, our proposed
encoder-decoder architecture learns to localize the lesion and
generates an initial attention map along with associated parameter
maps, thus instantiating a level-set ACM in which every location on
the contour has local parameter values.

By automatically initializing and tuning the segmentation process of
the level-set ACM, our DALS yields significantly more accurate
boundaries in comparison to conventional CNNs and can reliably segment
lesions of various sizes.

Furthermore, we combine CNNs and ACMs in an end-to-end trainable
framework that leverages an automatically differentiable ACM with
trainable parameters. By enabling the backpropagation of gradients for
stochastic optimization, the ACM and a backbone CNN can be trained
together from scratch, without pre-training. Moreover, our ACM
utilizes a locally-penalized energy functional that is directly
predicted by its backbone CNN, through 2D feature maps, and it is
initialized directly by the CNN. Thus, our work alleviates the biggest
obstacle to exploiting the power of ACMs---eliminating the need for
any type of user supervision or intervention.

\section{Level-Set Active Contour Model With Parameter Functions}
\label{sec:dals}

\begin{figure}
\centering
\includegraphics[scale=0.50]{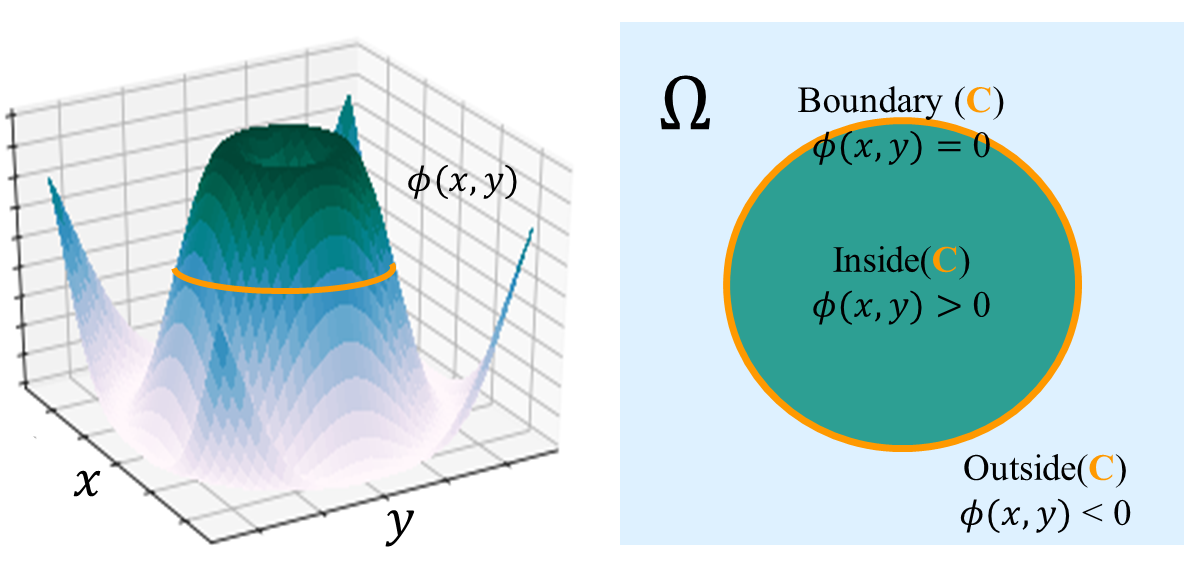}
\caption[Boundary represented as the zero level set of an implicit
function] {Boundary $C$ represented as the zero level set of implicit
function $\phi(x,y)$.}
\label{fig:level-set}
\end{figure}

First proposed by Osher and Sethian \citep{osher1988fronts} to evolve
wavefronts in CFD simulations, a level-set is an implicit
representation of a hypersurface that is dynamically evolved according
to the nonlinear Hamilton-Jacobi equation. Similarly, instead of
working with a parametric contour that encloses the desired area to be
segmented, we represent the contour as the zero level set of an
implicit function. Let $I$ represent an input image and $C= \big\{(x,
y) \mid \phi(x, y) = 0 \big\}$ be a closed contour in $\Omega\in R^2$
represented by the zero level set of the signed distance map
$\phi(x,y)$ (Figure~\ref{fig:level-set}). The interior and exterior of
$C$ are represented by $\phi(x,y)>0$ and $\phi(x,y)<0$, respectively.
Following \citep{chan2001active}, we use a smoothed Heaviside function
$H$ to represent the interior $(H(\phi))$ and exterior $(1-H(\phi))$
according to
\begin{equation}
H(\phi(x, y))=\frac{1}{2}+\frac{1}{\pi}\arctan\Big(
\frac{{\phi(x,y)}}{\epsilon}\Big).
\label{eq:heavi}
\end{equation}
The derivative of $H(\phi(x, y))$ is
\begin{equation}
\delta\phi(x, y)=\frac{\partial H(\phi(x, y))}{\partial
\phi(x,y)}=\frac{1}{\pi}\frac{\epsilon}{\epsilon^2+\phi(x,y)^2}.
\end{equation}

\subsection{Energy Functional}

In our formulation, we evolve $C$ to minimize an energy functional
according to
\begin{equation}
E({\phi})= E_\text{length}({\phi})+E_\text{image}({\phi}),
\label{eq:contourloss}
\end{equation}
where
\begin{equation}
E_\text{length}({\phi})= \int_\Omega
\mu\delta(\phi(x,y))|\nabla\phi(x,y)|\,dx\,dy
\label{eq:contourloss2}
\end{equation}
penalizes the length of the contour while
\begin{equation}
\label{eq:uniform_density}
\begin{split}
E_\text{image}(\phi) &= \int_\Omega\delta(\phi(x,y))
\biggl[H(\phi(x,y))(I(x,y)-m_1)^2+\\ &\qquad\qquad(1-H(\phi(x,y)))(I(x,y)-m_2)^2\biggr]\,dx\,dy
\end{split}
\end{equation}
takes into account the mean image intensities $m_1$ and $m_2$ of the
regions interior and exterior to the curve $C$ \citep{chan2001active}.

\begin{figure}
\def\x{0.48}
\subcaptionbox{}{\includegraphics[width=\x\linewidth]{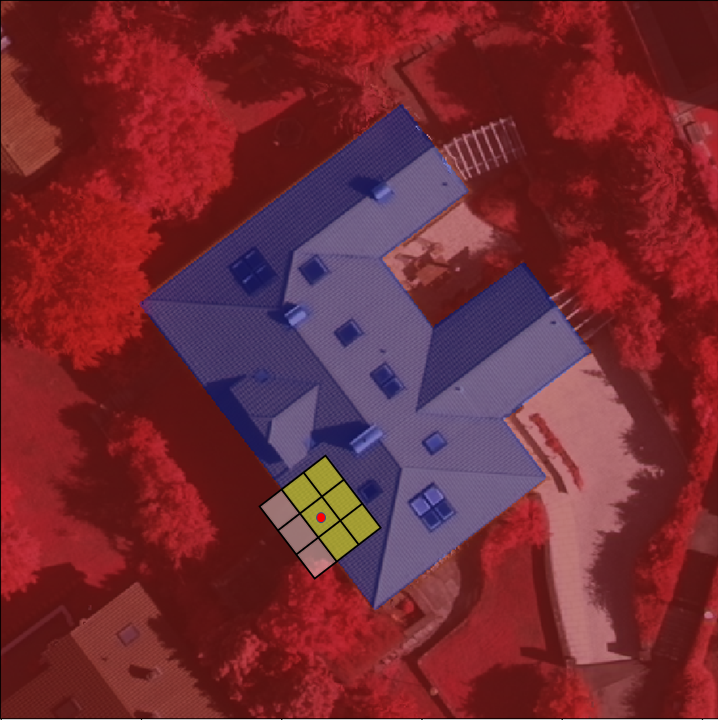}}
\hfill
\subcaptionbox{}{\includegraphics[width=\x\linewidth]{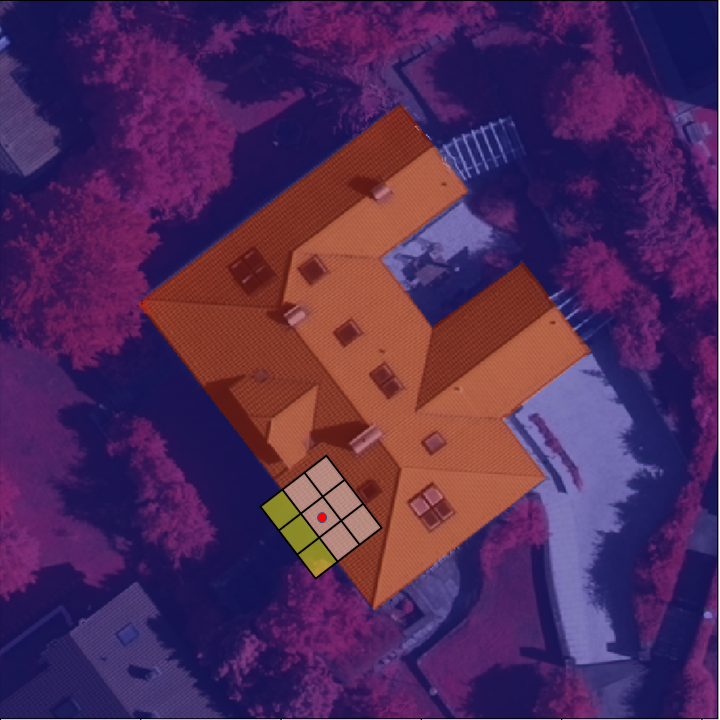}}
\caption[Visualization of the exterior and interior regions] {The
filter is divided by the contour into interior and exterior regions.
The point $x$ is represented by the red dot and the interior (a) and
exterior (b) regions are shaded in yellow.}
\label{fig:local_filter_maps}
\end{figure}

We compute these local statistics using a characteristic function
$W_{s}$ with local window (Figure~\ref{fig:local_filter_maps}) of size
$f_{s}$, as follows:
\begin{equation}
\label{eq:B}
W_{s} =
  \begin{cases}
    1 & \quad \text{if } x-f_{s}\leq u \leq x+f_{s}, \quad y-f_{s}\leq
    v \leq y+f_{s}; \\ 0 & \quad \text{otherwise}, \\
  \end{cases}
\end{equation} 
where $x,y$ and $u,v$ are the coordinates of two independent points.

We introduce feature maps $\lambda_1(x,y)$ and $\lambda_2(x,y)$ for
learning the foreground and background energies and allow them to be
functions over the image domain $\Omega$. Therefore, our energy
functional may be written as
\begin{equation}
\label{eq:f_pc}
E(\phi)= \int_\Omega \delta(\phi(x,y)) \biggl[\mu|\nabla\phi(x,y)| +
\int_\Omega W_s F(\phi(u,v)) \,du\,dv\biggr]\,dx\,dy,
\end{equation}
in which $F(\phi)$ is
\begin{equation}
\label{eq:density}
\begin{split}
F(\phi) = &\lambda_1(x,y) (I(u,v)-m_1(x,y))^2 (H(\phi(x,y)) +
\\ &\lambda_2(x,y) (I(u,v)-m_2(x,y))^2 (1-H(\phi(x,y)).
\end{split}
\end{equation}

It is important to note that our localized formulation enables us to
capture the fine-grained details of boundaries, and our use of
\emph{pixel-wise masks} $\lambda_1(x,y)$ and $\lambda_2(x,y)$ allows
them to be directly predicted by the backbone CNN along with an
initialization map $\phi_0(x,y)$. Thus, not only does the implicit ACM
propagation now become fully automated, but it can also be directly
controlled by a CNN through these learnable parameter functions.

\subsection{Euler-Lagrange Partial Differential Equation}

Following \citet{lankton2008localizing}, we now derive the
Euler-Lagrange PDE governing the evolution of the ACM.

Using the characteristic function $W_s$ that selects regions within a
square window of size $s$, the energy functional of contour $C$ in
terms of a generic internal energy density $F$ may be written as
\begin{equation}
\label{eq:f_pc_region_final}
E(\phi)= \int_{\Omega_{X_{1}}}
\delta(\phi(X_{1}))\int_{\Omega_{X_{2}}} W_s
F(\phi,X_{1},X_{2})\,dX_{2}\,dX_{1},
\end{equation}
where $X_{1}=(u,v)$ and $X_{2}=(x,y)$ are two independent spatial
variables, each of which represents a point in $\Omega$. To compute
the first variation of the energy functional, we add to $\phi$ a
perturbation function $\epsilon\psi$, where $\epsilon$ is a small
number; hence,
\begin{equation}
\label{eq:f_pc_region_purt}
E(\phi+\epsilon\psi)= \int_{\Omega_{X_{1}}}
\delta(\phi(X_{1})+\epsilon\psi)\int_{\Omega_{X_{2}}} W_s
F(\phi+\epsilon\psi,X_{1},X_{2})\,dX_{2}\,dX_{1}.
\end{equation}
Taking the partial derivative of (\ref{eq:f_pc_region_purt}) with
respect to $\epsilon$ and evaluating at $\epsilon=0$ yields, according
to the product rule,
\begin{equation}
\label{eq:f_pc_derv}
\begin{split}
\left.\dfrac{\partial E}{\partial\epsilon}\right|_{\epsilon=0} =
\int_{\Omega_{X_{1}}} \delta(\phi(X_{1}))\int_{\Omega_{X_{2}}}\psi W_s
\nabla_{\phi}F(\phi,X_{1},X_{2})\,dX_{2}\,dX_{1}
+ \\ \psi\int_{\Omega_{X_{1}}}\gamma\phi(X_{1})\int_{\Omega_{X_{2}}}
W_s F(\phi,X_{1},X_{2})\,dX_{2}\,dX_{1},
\end{split}
\end{equation}
where $\gamma\phi$ is the derivative of $\delta(\phi)$. Since
$\gamma\phi$ is zero on the zero level set, it does not affect the
movement of the curve. Thus the second term in (\ref{eq:f_pc_derv})
and can be ignored. Exchanging the order of integration, we obtain
\begin{equation}
\label{eq:f_pc_derv_v1}
\left.\dfrac{\partial E}{\partial\epsilon}\right|_{\epsilon=0} =
\int_{\Omega_{X_{2}}}\int_{\Omega_{X_{1}}} \psi\delta(\phi(X_{1})) W_s
\nabla_{\phi}F(\phi,X_{1},X_{2})\,dX_{1}\,dX_{2}.
\end{equation}
Invoking the Cauchy–Schwartz inequality yields
\begin{equation}
\label{eq:evol_region_v2}
\frac{\partial\phi}{\partial t} =
\int_{\Omega_{X_{2}}}\delta(\phi(X_{1})) W_s
\nabla_{\phi}F(\phi,X_{1},X_{2})\,dX_{2}.
\end{equation}
Adding the contribution of the curvature term and expressing the
spatial variables by their coordinates, we obtain the desired curve
evolution PDE:
\begin{equation}
\label{eq:ACWEalgorithm}
\frac{\partial\phi}{\partial t} = \delta(\phi) \biggl[\mu \divergence
\left(\frac{\nabla\phi}{|\nabla\phi|}\right) + \int_\Omega W_s
\nabla_\phi F(\phi)\,dx\,dy \biggr],
\end{equation}
where, assuming a uniform internal energy model and defining $m_1$ and
$m_2$ as the mean image intensities inside and outside $C$ and within
$W_s$, we have
\begin{equation}
\nabla_\phi F = \delta(\phi) \left(\lambda_1(u,v)[I(u,v)-m_1(x,y)]^2 -
\lambda_2(u,v)[I(u,v)-m_2(x,y)]^2\right).
\end{equation}

\subsection{DALS CNN Backbone}

Our encoder-decoder is an FCN architecture that is tailored and
trained to estimate a probability map from which the initial distance
function $\phi(x,y,0)$ of the level-set ACM and the functions
$\lambda_1(x,y)$ and $\lambda_2(x,y)$ are computed. In each dense
block of the encoder, a composite function of batch normalization,
convolution, and ReLU is applied to the concatenation of all the
feature maps $[x_0, x_1, \dots , x_{l-1}]$ from layers 0 to $l-1$ with
the feature maps produced by the current block. This concatenated
result is passed through a transition layer before being fed to
successive dense blocks. The last dense block in the encoder is fed
into a custom multiscale dilation block with 4 parallel convolutional
layers with dilation rates of 2, 4, 8, and 16. Before being passed to
the decoder, the output of the dilated convolutions are then
concatenated to create a multiscale representation of the input image
thanks to the enlarged receptive field of its dilated convolutions.
This, along with dense connectivity, assists in capturing local and
global context for highly accurate lesion localization.

\begin{figure}
\includegraphics[width=\linewidth]{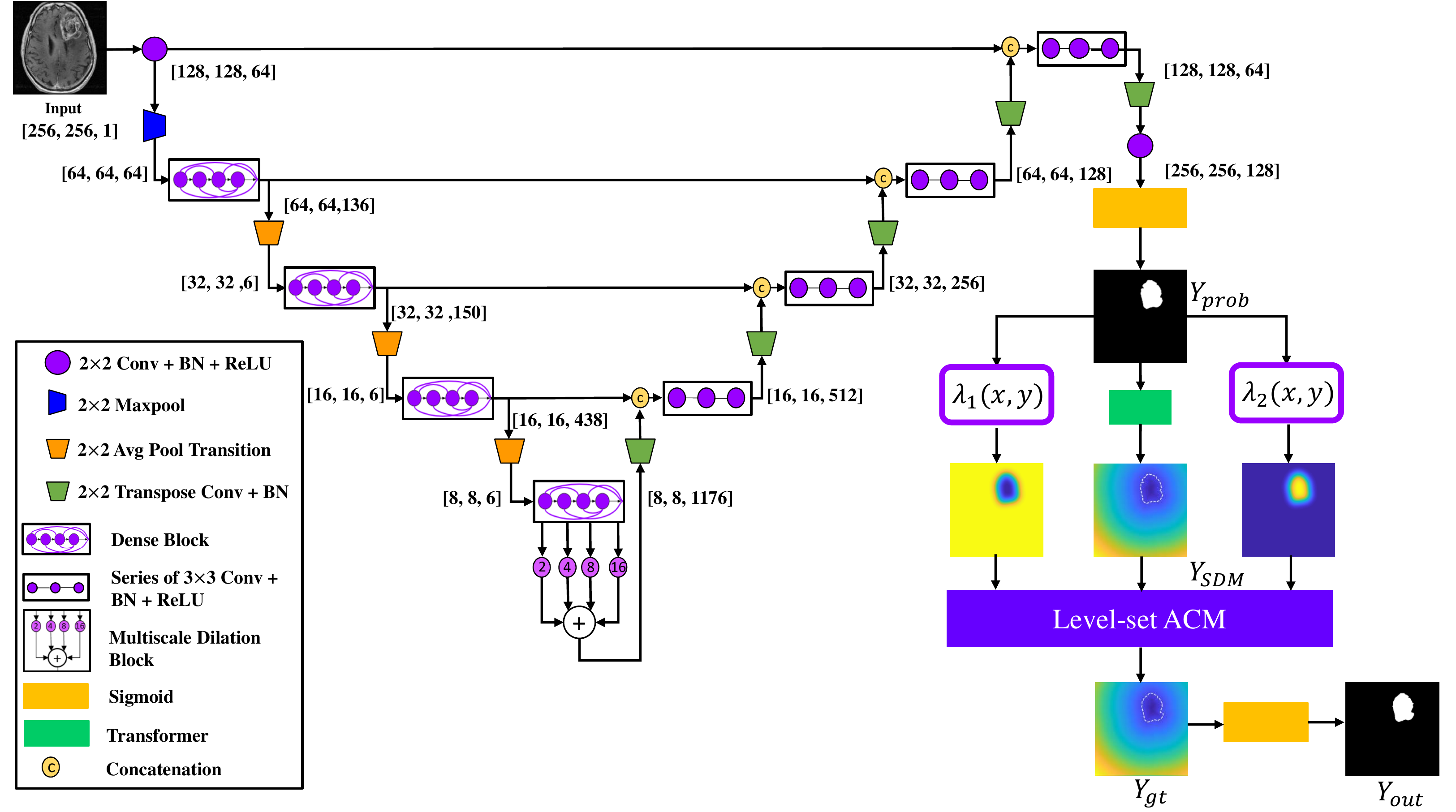}
\caption[The DALS architecture] {The DALS architecture. DALS is a
fully automatic segmentation framework. The CNN initializes and guides
the ACM by its learning local weighted parameters.}
\label{fig:dlac}
\end{figure}

\subsection{The DALS Framework}

Our DALS framework is illustrated in Figure~\ref{fig:dlac}. 
The boundaries of the segmentation map generated by the
encoder-decoder are fine-tuned by the level-set ACM that takes
advantage of information in the CNN maps to set the per-pixel
parameters and initialize the contour. The input image is fed into the
encoder-decoder, which localizes the lesion and, after $1\times1$
convolutional and sigmoid layers, produces the initial segmentation
probability map $Y_\text{prob}(x,y)$, which specifies the probability
that any point $(x,y)$ lies in the interior of the lesion. The
Transformer converts $Y_\text{prob}$ to a Signed Distance Map (SDM)
$\phi(x,y,0)$ that initializes the level-set ACM. Map $Y_\text{prob}$
is also utilized to estimate the parameter functions $\lambda_1(x,y)$
and $\lambda_2(x,y)$ in the energy functional (\ref{eq:f_pc}).
Extending the approach of \citet{hoogi2017adaptive}, the $\lambda$
functions in Figure~\ref{fig:dlac} are chosen as follows:
\begin{equation}
\label{eq:lambdas}
\lambda_1(x,y)=\exp\left(\frac{2-Y_\text{prob}(x,y)}{1+Y_\text{prob}(x,y)}\right);
\quad
\lambda_2(x,y)=\exp\left(\frac{1+Y_\text{prob}(x,y)}{2-Y_\text{prob}(x,y)}\right).
\end{equation}
The exponential amplifies the range of values that the functions can
take. These computations are performed for each point on the zero
level set contour $C$. During training, $Y_\text{prob}$ and the ground
truth map $Y_\text{gt}(x,y)$ are fed into a Dice loss function and the
error is back-propagated accordingly. During inference, a forward pass
through the encoder-decoder and level-set ACM results in a final SDM,
which is converted back into a probability map by a sigmoid layer,
thus producing the final segmentation map $Y_\text{out}(x,y)$.

\section{The DTAC Framework}
\label{sec:dtac}

We further propose a model, dubbed Deep Trainable Active
Contours (DTAC), that establishes a tight merger between our ACM with
any backbone CNN for segmenting images in a robust manner and capture
the fine-grained details of their boundaries.

\subsection{Differentiable Level Set}
\label{sec:diff_acm}

We dynamically evolve the contour according to
(\ref{eq:ACWEalgorithm}) in a differentiable manner using TensorFlow.
The first term,
$\divergence\bigl(\frac{\nabla\phi}{|\nabla\phi|}\bigr)$, necessitates
computing the surface curvature according to
\begin{equation}
\label{eq:f_pc3}
\divergence \left(\frac{\nabla\phi}{|\nabla\phi|}\right) =
\dfrac{\phi_{xx}\phi_{y}^{2}-2\phi_{xy}\phi_{x}\phi_{y} +
\phi_{yy}\phi_{x}^{2}}{(\phi_{x}^2+\phi_{y}^{2})^{3/2}},
\end{equation}
where the subscripts denote spatial derivatives of $\phi$, which we
compute using central finite differences. For the second term, we find
the regions in the image that correspond to the interior and exterior
of the curve and leverage average pooling layers to efficiently
compute $m_{1}$ and $m_{2}$ used in (\ref{eq:density}). Therefore we
can evaluate $\frac{\partial\phi}{\partial t}$ in
(\ref{eq:ACWEalgorithm}) and update the level-set according to
\begin{equation}
\phi^{t}= \phi^{t-1} +\Delta t \frac{\partial\phi^{t-1}}{\partial t},
\label{eq:attloss}
\end{equation}
where $\Delta t$ is the time step size.

\subsection{DTAC CNN Backbone}

\begin{figure}
\includegraphics[width=\linewidth]{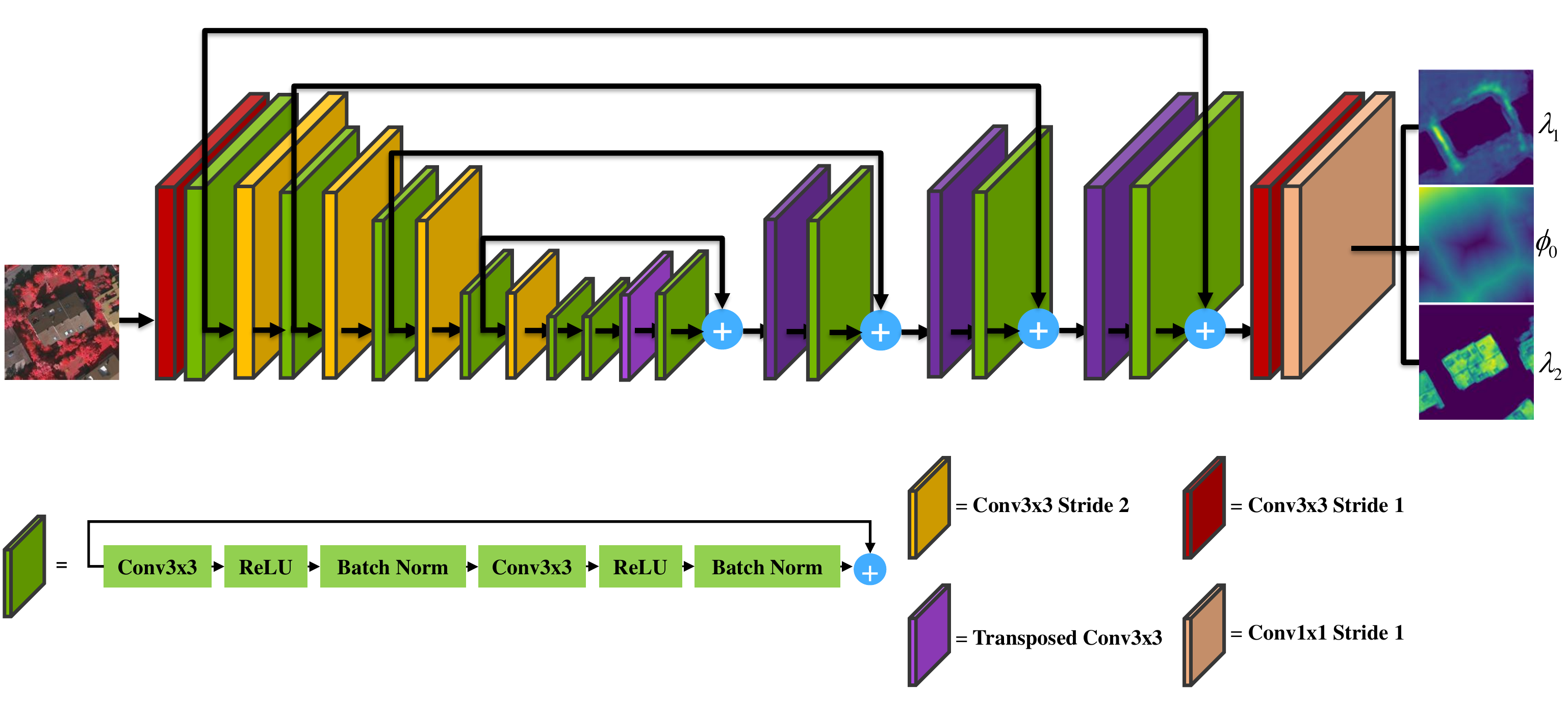}
\caption[CNN backbone architecture for DTAC] {DTAC's CNN backbone has
a standard encoder-decoder architecture.}
\label{fig:cnn}
\end{figure}

We use a standard encoder-decoder architecture with residual blocks
and skip connections between the encoder and decoder sub-networks.
Each residual block consists of two convolutions with batch
normalization, ReLU, and an additive identity skip connection. As is
illustrated in Figure~\ref{fig:cnn}, each stage of the encoder
comprises of residual blocks and convolutions with stride of two.
Similarly, each stage of the decoder has a residual block followed by
a transposed convolution. The encoder is connected to the decoder via
a residual block at the lowest resolution as well as skip connections
at every stage. The output of the decoder is connected to a
convolution with three output channels for predicting the
$\lambda_{1}(x,y)$ and $\lambda_{2}(x,y)$ feature maps as well as the
initialization map $\phi_0(x,y)$. Detailed information regarding the
encoder and decoder of DTAC is presented in Tables~\ref{tab:encoder}
and~\ref{tab:decoder}.

\begin{table} \centering
\begin{tabular}{ll}
\toprule
Operations  &Output size    \\ 
\midrule
Input &$512\times512\times1$   \\
Conv, ReLu, BN & $512\times512\times64$    \\
Conv, ReLU, BN, Conv, ReLU, BN, Add &$512\times512\times64$    \\
Conv stride 2 &$256\times256\times128$    \\
Conv, ReLU, BN, Conv, ReLU, BN, Add &$256\times256\times128$    \\
Conv stride 2 &$128\times128\times256$    \\
Conv, ReLU, BN, Conv, ReLU, BN, Add &$128\times128\times256$    \\
Conv stride 2&$64\times64\times512$    \\
Conv, ReLU, BN, Conv, ReLU, BN, Add &$64\times64\times512$    \\
Conv stride 2&$32\times32\times1024$    \\
Conv, ReLU, BN, Conv, ReLU, BN, Add &$32\times32\times1024$   \\
Conv, ReLU, BN, Conv, ReLU, BN, Add &$32\times32\times1024$   \\
\bottomrule
\end{tabular}
\caption[Detailed information about the DTAC encoder] {Detailed
information about the encoder of DTAC. BN and Add denote batch
normalization and additive identity skip connections. Conv denotes a
$3\times3$ convolutional layer.}
\label{tab:encoder}
\end{table}

\begin{table} \centering
\begin{tabular}{ll}
\toprule
Operations  & Output size    \\ 
\midrule
Input &$32\times32\times1024$   \\
TransConv stride 2 &$64\times64\times512$    \\
Conv, ReLU, BN, Conv, ReLU, BN, Add &$64\times64\times512$    \\
TransConv stride 2 &$128\times128\times256$    \\
Conv, ReLU, BN, Conv, ReLU, BN, Add &$128\times128\times256$    \\
TransConv stride 2 &$256\times256\times128$    \\
Conv, ReLU, BN, Conv, ReLU, BN, Add &$256\times256\times128$    \\
TransConv stride 2 &$512\times512\times64$    \\
Conv, ReLU, BN, Conv, ReLU, BN, Add &$512\times512\times64$    \\
Conv, ReLu, BN & $512\times512\times32$ \\
Conv1, Sigmoid &$512\times512\times3$    \\
\bottomrule
\end{tabular}
\caption[Detailed information about the DTAC decoder] {Detailed
information about the decoder of DTAC. BN and Add denote batch
normalization and additive identity skip connections. Conv and Conv1
denote $3\times3$ and $1\times1$ convolutional layers, respectively.
TransConv denotes a $3\times3$ transposed convolutional
layer with a kernel size of 2.}
\label{tab:decoder}
\end{table}

\subsection{The DTAC Architecture and Network Training}
\label{sec:loss}

\begin{figure}
\centering \includegraphics[width=\linewidth]{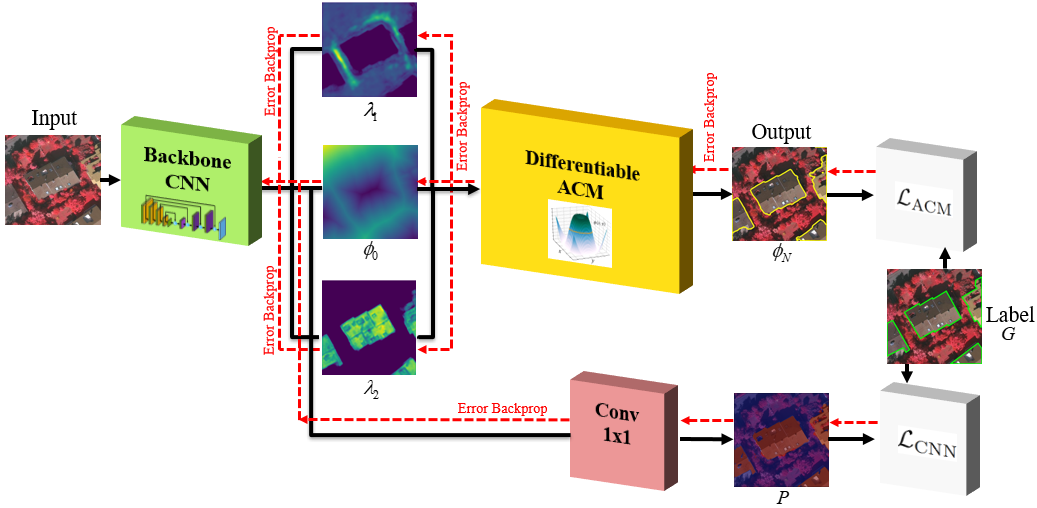}
\caption[The DTAC architecture] {DTAC is a fully-automated, end-to-end
automatically differentiable and backpropagation trainable ACM and
backbone CNN framework.}
\label{fig:dtac-framework}
\end{figure}

We simultaneously train the CNN and levelset components of DTAC in an
end-to-end manner with no human supervision. The CNN guides the ACM by
predicting the $\lambda_{1}(x,y)$ and $\lambda_{2}(x,y)$ feature maps
as well as an initialization map $\phi_0(x,y)$. The level set evolves
in a differentiable manner, thus allowing for directly backpropagating
the error. The initialization map output of the CNN is further passed
into another convolution layer followed by a sigmoid activation
function (Figure~\ref{fig:dtac-framework}). Therefore, the total loss
for training the DTAC is
\begin{equation}
\mathcal{L}= \mathcal{L}_\text{CNN} + \mathcal{L}_\text{ACM},
\label{eq:attloss3}
\end{equation}
where $\mathcal{L}_\text{CNN}$ and $\mathcal{L}_\text{ACM}$ denote the
losses computed over the output of backbone CNN and final iteration of
level-set ACM, respectively. $\mathcal{L}_\text{ACM}$ is computed
using a binary cross entropy loss function according to
\begin{equation}
\mathcal{L}_\text{ACM}= -\frac{1}{N}\sum_{j=1}^{N}\left[ G_{j}\log
H(\phi_{j})+(1-G_{j})\log (1-H(\phi_{j})) \right],
\label{eq:ce}
\end{equation} 
where $H$ is defined according to (\ref{eq:heavi}), $\phi_{j}$ and
$G_{j}$ denote the ACM output and ground truth at pixel $j$
respectively, and $N$ is the total number of pixels in the image.
$\mathcal{L}_\text{CNN}$ is calculated in a similar manner to
(\ref{eq:ce}) by replacing $H_{j}$ with the output prediction
probabilities of $P_{j}$ from the CNN.
Algorithm~\ref{alg:dtac-algorithm} presents the details of DTAC
training.

\begin{algorithm}[ht]
\caption{DTAC Training Algorithm}
\label{alg:dtac-algorithm}
\KwData{$X$, $G$: Paired image and label; $W$: CNN with parameters $\omega$; $g$: ACM energy function with parameters $\lambda_1,\lambda_2$; $\mathcal{L}$: Loss function; $N$: Number of ACM iterations; $\eta$: Learning rate; $\phi$: Levelset; $P$: CNN probability output}
\KwResult{Trained model}
\While{not converged}{
  $\lambda_1,\lambda_2,\phi_{0},P=W(X)$\\
 \For{$t=1$ \textbf{to} $N$}{
    $ \frac{\partial\phi_{t-1}}{\partial t}=g(\phi_{t-1};\lambda_1,\lambda_2, X)$\\
    $\phi^{t}= \phi^{t-1} +\Delta t \frac{\partial\phi^{t-1}}{\partial t}$

    }
    $\mathcal{L}=\mathcal{L}_\text{ACM}(\phi_{N},G)+\mathcal{L}_\text{CNN}(P,G)$\\
  Compute $\frac{\partial \mathcal{L}}{\partial \omega}$ and backpropagate the error\\
  Update the weights of $f$: $\omega \leftarrow \omega-\eta  \frac{\partial \mathcal{L}}{\partial \omega}$
}
\end{algorithm}

\chapter{Few-Shot Semantic Segmentation}
\label{cha:few}

\begin{figure}
\includegraphics[width=\linewidth]{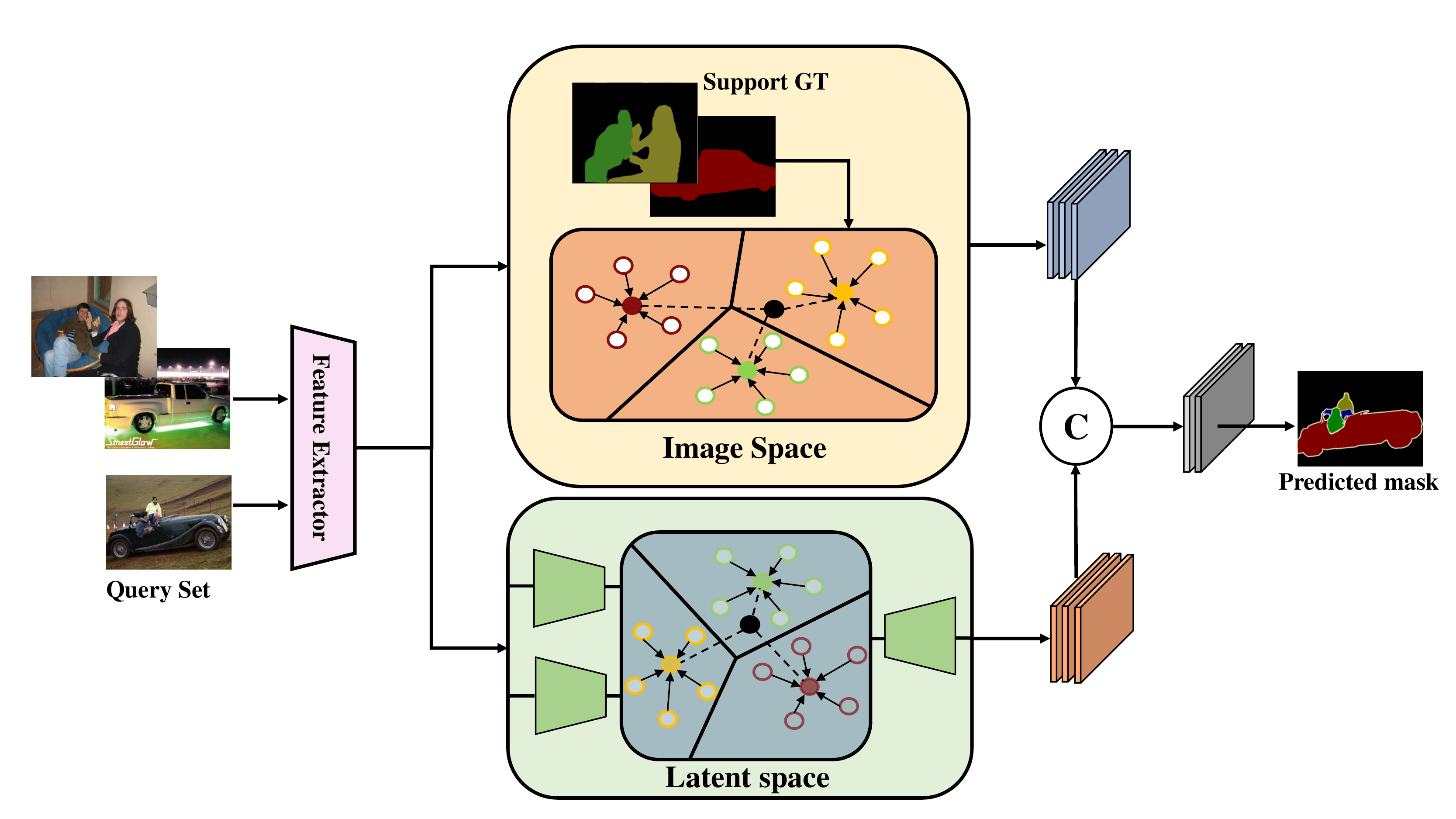}
\caption[The SegAVA architecture] {Overview of the SegAVA
architecture, showing the two parallel branches. In the image space
branch, SegAVA maps the support and query images to embedding
features, learning features for each class (represented by the red,
green, and yellow circles), matching query features to the nearest
embedded feature. In the latent space branch, SegAVA employs two
variational auto-encoders to learn the latent space of support and
query images and uses Wasserstein-2 metric to learn similarities
between embeddings. Results from the two branches are concatenated and
passed through convolutional layers, yielding the final segmentation
of the query image.}
\label{fig:pipeline_highlevel}
\end{figure}

In this chapter, we propose a novel metric-based framework for
few-shot image segmentation, which we call Segmentation with Aligned
Variational Auto-Encoders (SegAVA), that explores the latent and image
spaces of support and query sets to find the most common
class-specific embeddings and fuses them to produce the final semantic
segmentation (Figure~\ref{fig:pipeline_highlevel}). Specifically,
SegAVA features a latent stream consisting of two Variational
Auto-Encoders (VAEs) that generate support and query images and learn
the most essential discriminative information by aligning their
learned features in the latent space.

Additionally, SegAVA uses an encoder-decoder in the image space to
extract the most similar features of the support and query images and
concatenate them with the learned embeddings of the latent space to
produce the segmentation in an end-to-end manner without additional
post-processing.

We argue that the latent space of the support and query sets provides
rich semantics for identifying the most essential discriminative
features, and aggregation with image space embeddings leads to
improved segmentation accuracy. Our work can be regarded an extension
to that of \citet{deudon2018learning} who used the latent space for
learning semantic similarity in natural language processing, but
differs in that SegAVA is trained jointly for image generation and
semantic similarity extraction.

\section{Problem Setting}

In the $N$-way $k$-shot semantic segmentation problem, given a
training set of $K$ samples with $N$ classes, the goal is to learn to
segment new images with categories that belong to the $N$ classes. We
follow the same training and testing protocols in prior efforts
\citep{rakelly2018conditional, wang2019panet} and formulate our
problem as follows: Given, two sets of non-overlapping seen and unseen
categories, denoted as $\mathcal{C}_\text{unseen}$ and
$\mathcal{C}_\text{seen}$, we define two sets for training and testing
the model. The train set $\mathcal{D}_\text{train}$ =
\{${(\mathcal{S}_{i}, \mathcal{Q}_{i})}_{i=1}^{N_\text{train}}$\} and
test set $\mathcal{D}_\text{test}$ = \{${(\mathcal{S}_{i},
\mathcal{Q}_{i})}_{i=1}^{N_\text{test}}$\} are defined in a sequence
of episodes. Each episode, denoted by $i$, has a set of support
samples $\mathcal{S}_{i}$ and query samples $\mathcal{Q}_{i}$ with
total numbers $N_\text{train}$ and $N_\text{test}$ for the train and
test episodes, respectively.

In a $N$-way, $k$-shot setting, the episode $i$ comprises a support
set $\mathcal{S}_{i}$ = \{($\mathcal{I}_{c,k}$, $\mathcal{L}_{c,k}$)\}
in which for each class, there exist $K$ samples of image and label
pairs, and there are $N$ distinct semantic classes in total.
Furthermore, from the categories that are present in the support set,
there are $N_\text{query}$ samples of image and label pairs in the the
query set. In each training episode, the goal is to utilize the
support set $\mathcal{S}_{i}$, with images $\mathcal{I}$ and
corresponding pixel-wise annotations $\mathcal{L}$, to segment images
in the query set $\mathcal{Q}_{i}$. Eventually, the trained
segmentation model is employed to perform segmentation on the cases
from the test set $\mathcal{D}_\text{test}$ in each of its episodes.

\section{SegAVA Framework}

\begin{figure}
\includegraphics[width=\linewidth]{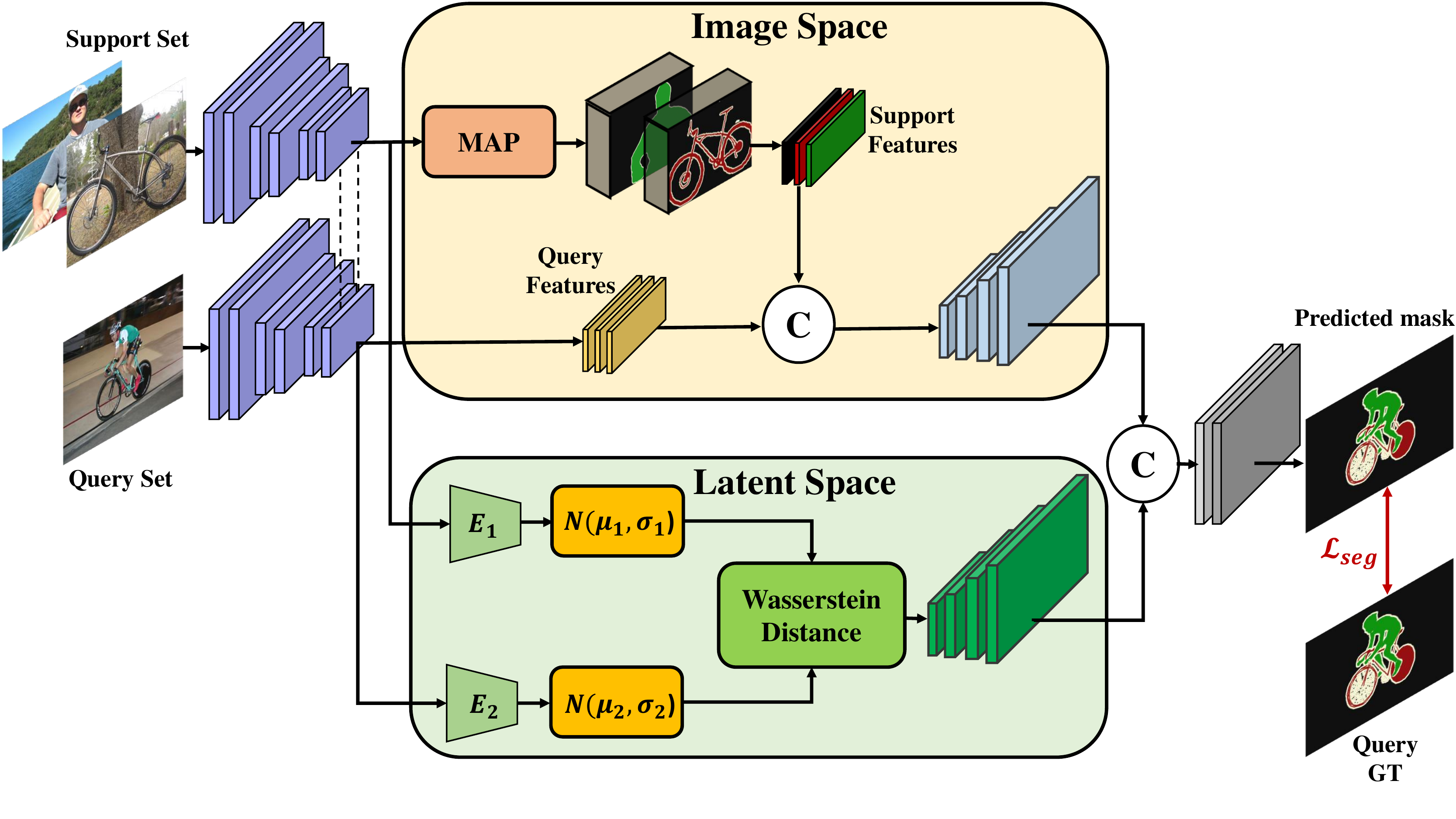}
\caption[Detailed diagram of the SegAVA architecture] {Detailed
diagram of the SegAVA architecture. MAP denotes the Masked Average
Pooling operation. $E_{1}$ and $E_{2}$ denote the encoders of support
and query features, respectively. The pretrained feature extractor of
support and query sets share the same weights.}
\label{fig:pipeline_lowlevel}
\end{figure}

As illustrated in Figure~\ref{fig:pipeline_lowlevel}, images in the
support and query sets are first fed into a pre-trained network for
initial feature extraction, and the extracted features are
subsequently aligned in the image space (upper stream) as well as the
latent space (lower stream). The aligned features in both latent and
image space are further concatenated and fed into a series of
convolutional layers that produce the final segmentation. We detail
the working principles of feature alignments in the next two sections.

\section{Latent Space Alignment}
\label{sec:latent}

In SegAVA, the building blocks of feature alignment in the latent
space are VAEs \citep{kingma2013auto}. Given a VAE with an encoder
$\phi$, decoder $\theta$, and input $x$, the goal of the encoder is to
parameterize $p_{\theta}(z \mid x)$ over the latent variable $z$.
Furthermore, the decoder parameterizes $p_{\theta}(z\mid x)$ over $x$,
given a random latent variable $z$. Using a variational lower bound
limit on the marginal likelihood of $p(x\mid \theta,\phi)$, the VAE
loss function can be expressed as
\begin{equation}
\label{eq:VAE_1}
\mathcal{L} = \mathbbm{E}_{q_{\phi}(z\mid
x)}[\text{log}p_{\theta}(x\mid z)] - \text{KL}(q_{\phi}(z\mid x)
\,||\, p(z)),
\end{equation}
where the first term represents the reconstruction error and the
second term is the Kullback-Leibler (KL) divergence between the prior
on the latent code $p(z)$ and a posterior distribution $q_{\phi}(z\mid
x)$. The decoder predicts the posterior, normally a Gaussian
distribution such that $q_{\phi}(z\mid x)= \mathcal{N}(\mu,\sigma).$
Consequently, the final loss function of SegAVA's VAEs, for the
support and query sets is
\begin{equation}
\label{eq:VAE_2}
\mathcal{L}_\text{VAE} = \sum_{i}^{M=2}\mathbbm{E}_{q_{\phi}(z\mid
x)}[\log p_{\theta}(x^{(i)}\mid z)] - \text{KL}[q_{\theta}(z\mid
x^{(i)}) \,||\, p_{\theta}(z)].
\end{equation}
Inspired by \citep{deudon2018learning}, we further utilize a
Wasserstein-2 metric between the latent multivariate Gaussian
distributions of the support and query sets for alignment in the
latent space, according to
\begin{equation}
\label{eq:VAE_5}
W_{2}^{2}(p_{1},p_{2}) = \sum_{i}^{}(\mu_{1}^{i}-\mu_{2}^{i})^{2} +
(\sigma_{1}^{i}-\sigma_{2}^{i})^{2},
\end{equation}
where $p_{1}=\mathcal{N}(\mu_{1},\sigma_{1})$ and
$p_{2}=\mathcal{N}(\mu_{2},\sigma_{2})$, the diagonal covariance
matrices of two Gaussians. It is important to note that we utilize
(\ref{eq:VAE_5}) in an element-wise manner and feed the result it to a
dense layer followed by a fully convolutional decoder to estimate the
similarity between the support and query embeddings.

\section{Image Space Alignment}
\label{sec:image}

In the image space, query and support images are first fed into a
pre-trained network to obtain feature embeddings that can be used to
estimate the similarities. Given a support set $\mathcal{S}_{i}$ =
\{($\mathcal{I}_{c,k}$, $\mathcal{L}_{c,k}$)\} in which $c$ denotes
the index corresponding to each semantic class and $k=1,2,\ldots,K$ is
the index for each sample in the support set, we use a masked average
pooling operation \citep{zhang1810sg},
\begin{equation}
\label{eq:PANet_1}
p_{c} = \frac{1}{K}\sum_{k}\frac{\sum_{x,y}F_{c,k}^{(x,y)}
\mathbbm{1}[\mathcal{L}_{c,k}^{(x,y)} =
c]}{\sum_{x,y}\mathbbm{1}[\mathcal{L}_{c,k}^{(x,y)} = c]},
\end{equation}
where $(x,y)$ are spatial location indexes and $F_{c,k}^{(x,y)}$ are
the extracted features for an input image $\mathcal{I}_{c,k}$ at
spatial location $(x,y)$. Subsequently, the masked features are fed
into a fully convolutional decoder each layer of which consists of a
$3\times3$ transposed convolution with stride of $2$ followed by a
batch normalization \citep{ioffe2015batch} operation and a ReLU
activation function.

Furthermore, the upsampled similarity features from the image space
are concatenated with decoded features from the latent space and fed
into a $3\times3$ convolution followed by a $1\times1$ convolution.
The output segmentation map is subsequently calculated according to
\begin{equation}
\label{eq:PANet_3}
\tilde{L}_{q;j}^{(x,y)} = \frac{\exp( F_{q}^{(x,y)})}{\sum_{p_{j} \in
\mathcal{P}}\exp(F_{q}^{(x,y)})},
\end{equation}
where $F_{q}^{(x,y)}$ is the pixel-wise output of the last
convolutional layer. Accordingly, the segmentation loss can be defined
as
\begin{equation}
\label{eq:PANet_5}
\mathcal{L}_{\text{seg}} =
-\frac{1}{N}\sum_{x,y}\sum_{p_{j}\in\mathcal{P}}
\mathbbm{1}[L_{q}^{(x,y)} = j]\log\tilde{L}_{q;j}^{(x,y)},
\end{equation}
where $L_{q}^{(x,y)}$ and $\tilde{L}_{q;j}^{(x,y)}$ denote the
ground-truth and predictions at spatial location $(x,y)$. To jointly
train the latent and image streams, we use the hybrid loss function
\begin{equation}
\label{eq:PANet_6}
\mathcal{L}_{\text{SegAVA}} =\mathcal{L}_{\text{seg}} + \gamma
\mathcal{L}_{\text{VAE}},
\end{equation}
where $\gamma$ is a hyper-parameter.

\section{Active Contour Assisted Few-Shot Segmentation}

SegAVA can additionally benefit from a post-processing module that can
refine the segmentation predictions. As such, we leveraged our DALS
framework to fully delineate the boundaries.

The probability predictions by SegAVA are used to initialize the
contour as well as the $\lambda_{1}(x,y)$ and $\lambda_{2}(x,y)$
feature maps. The contour $C$ is then evolved according to
\begin{equation}
\label{eq:ACWEalgorithm2}
\frac{\partial\phi}{\partial t} = \delta(\phi) \biggl[\mu \divergence
\left(\frac{\nabla\phi}{|\nabla\phi|}\right) + \int_\Omega W_s
\nabla_\phi F(\phi)\,dx\,dy \biggr],
\end{equation}
where $m_1$ and $m_2$ denote the mean image intensities inside and
outside $C$, and
\begin{equation}
\nabla_\phi F = \delta(\phi) \bigl(\lambda_1(u,v)(I(u,v)-m_1(x,y))^2 -
\lambda_2(u,v)(I(u,v)-m_2(x,y))^2\bigr).
\end{equation}

\chapter{Implementation Details, Data, Experiments, Results}
\label{cha:experiments}

This chapter presents our experiments with the models that we developed in Chapters~\ref{cha:bound}, \ref{cha:dtac}, and \ref{cha:few}, and it reports our results. We also provide information about the datasets that we use in our empirical studies and implementation details about the models themselves.  

\section{2D Edge-Aware Encoder-Decoders}

In this section, we empirically study the models developed in Section~\ref{sec:2d-edge-aware}.

\subsection{Dataset}

In our experiments, we used the BraTS 2018 \citep{brats2}, which
provides multimodal 3D brain MRIs and ground truth brain tumor
segmentations annotated by physicians, consisting of 4 MRI modalities
per case (T1, T1c, T2, and FLAIR). Annotations include 3 tumor
subregions---the enhancing tumor, the peritumoral edema, and the
necrotic and non-enhancing tumor core. The annotations were combined
into 3 nested subregions---whole tumor (WT), tumor core (TC), and
enhancing tumor (ET). The data were collected from 19 institutions,
using various MRI scanners.

For simplicity, we use only a single input MRI modality (T1c) and aim
to segment a single tumor region---TC, which includes the main tumor
components (nectrotic core, enhancing, and non-enhancing tumor
regions). Furthermore, even though the original data is 3D
($240\times240\times155$), we operate on 2D slices for simplicity. We
have extracted several axial slices centered around the tumor region
from each 3D volume, and combined them into a new 2D dataset.

\begin{figure}
\centering
\def\x{0.24}
\includegraphics[width=\x\linewidth,height=\x\linewidth]{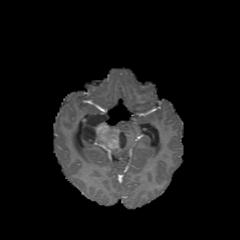}
\includegraphics[width=\x\linewidth,height=\x\linewidth]{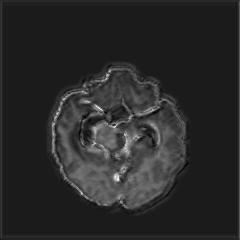}
\includegraphics[width=\x\linewidth,height=\x\linewidth]{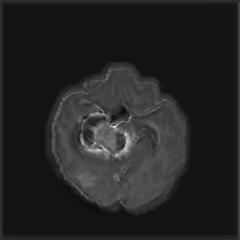}
\includegraphics[width=\x\linewidth,height=\x\linewidth]{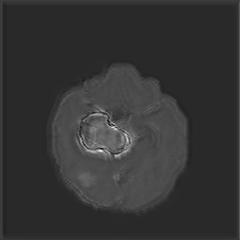}\\[3pt]
\includegraphics[width=\x\linewidth,height=\x\linewidth]{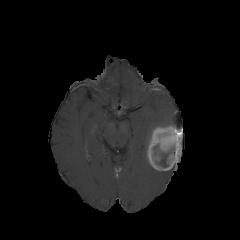}
\includegraphics[width=\x\linewidth,height=\x\linewidth]{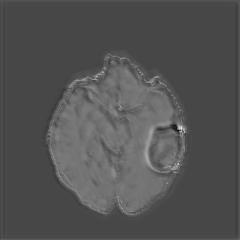}
\includegraphics[width=\x\linewidth,height=\x\linewidth]{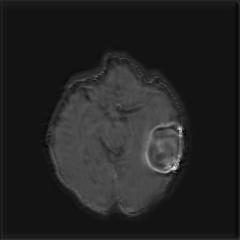}
\includegraphics[width=\x\linewidth,height=\x\linewidth]{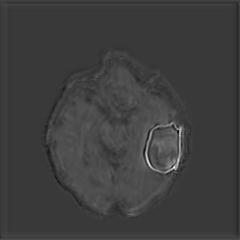}\\[3pt]
\includegraphics[width=\x\linewidth,height=\x\linewidth]{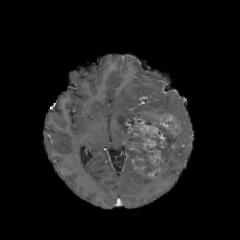}
\includegraphics[width=\x\linewidth,height=\x\linewidth]{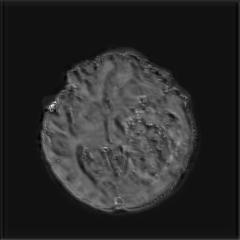}
\includegraphics[width=\x\linewidth,height=\x\linewidth]{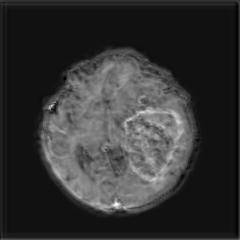}
\includegraphics[width=\x\linewidth,height=\x\linewidth]{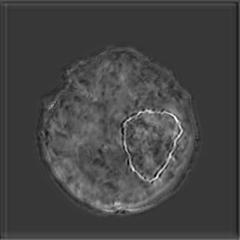}\\[3pt]
\makebox[\x\linewidth]{(a) Input image} \makebox[\x\linewidth]{(b) Att.~Layer~1}
\makebox[\x\linewidth]{(c) Att.~Layer~2} \makebox[\x\linewidth]{(d) Att.~Layer~3}
\caption[Visualization of learned feature maps in 2D edge-aware
network] {Visualization of learned feature maps in 2D edge-aware
network. (b--d) Outputs of the attention layers. The boundary emphasis
becomes more prominent with each attention layer.}
\label{fig:attention_maps}
\end{figure}

\subsection{Implementation Details}

We have implemented our model in Tensorflow. The brain input images
were resized to predefined sizes of $240\times240$ and normalized to
the intensity range $[0,1]$. The model was trained on NVIDIA Titan RTX
and an Intel® Core™ i7-7800X CPU @ 3.50GHz $\times$ 12 with a batch
size of 8 for all models. We used $\lambda_{1}=1.0$,
$\lambda_{2}=0.5$, and $\lambda_{3}=0.1$ in (\ref{eq:finalloss}). The
Adam optimization algorithm was used with initial learning rate of
$\alpha_{0} = 1.0^{-3}$ and further decreased according to
\begin{equation}
\label{eq:learningrate}
\alpha = \alpha_{0}\left(1-e/N_{e}\right)^{0.9},
\end{equation}
where $e$ denotes the current epoch and $N_{e}$ the total number of
epochs, following \citep{myronenko2019robust}. We have evaluated the
performance of our model by using the Dice score, Jaccard index, and
Hausdorff distance.

\subsection{Results}
\label{sec:results}

\paragraph{Boundary Stream:}
Figure~\ref{fig:attention_maps} demonstrates the output of each of the
attention layers in our dedicated boundary stream. In essence, each
attention layer progressively localizes the tumor and refines the
boundaries. The first attention layer has learned rough estimate of
the boundaries around the tumor and localized it, whereas the second
and third layers have learned more fine-grained details of the edges
and boundaries, refining the localization.

Moreover, since our architecture leverages a dilated spatial pyramid
pooling to merge the learned feature maps of the regular segmentation
stream and the boundary stream, multiscale regional and boundary
information have been preserved and fused properly, which has enabled
our network to capture the small structural details of the tumor.

\begin{table} \centering
\begin{tabular}{lccc}
\toprule
Model & Dice Score & Jaccard Index &  Hausdorff Distance \\
\midrule
U-Net & 0.731$\pm$0.230   & 0.805~$\pm$0.130  & 3.861$\pm$1.342     \\
V-Net & 0.769$\pm$0.270  &0.837$\pm$0.140  & 3.667$\pm$1.329 \\
Ours (no edge loss) & 0.768$\pm$0.236   & 0.832$\pm$0.136  &  3.443$\pm$1.218    \\
Ours  & \textbf{0.822$\pm$0.176} & \textbf{0.861$\pm$0.112} & \textbf{3.406$\pm$1.196} \\
\bottomrule
\end{tabular}
\caption[Quantitative comparison of 2D edge-aware network and others
on BraTS] {Performance evaluations of different models. We validate
the contribution of the edge loss by measuring performance with and
without this layer.}
\label{tab:res}
\end{table}

\paragraph{Edge-Aware Losses:}
To validate the effectiveness of the loss supervision, we have trained
our network without enforcing the supervision of the edge loss during
the learning process, but with the same architecture.
Table~\ref{tab:res} shows that our network performs very similarly to
V-Net \citep{Milletari2016} without edge supervision, since ours employs
similar residual blocks as V-Net in its main encoder-decoder.
In essence, the boundary stream also impacts the downstream layers of
the encoder by emphasizing edges during training.

\begin{figure}
\centering
\def\x{0.182}
\includegraphics[width=\x\linewidth,height=\x\linewidth]{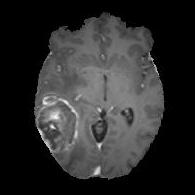}
\includegraphics[width=\x\linewidth,height=\x\linewidth]{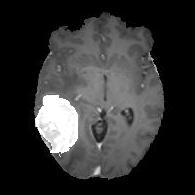}
\includegraphics[width=\x\linewidth,height=\x\linewidth]{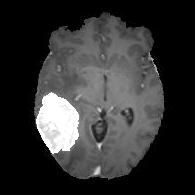}
\includegraphics[width=\x\linewidth,height=\x\linewidth]{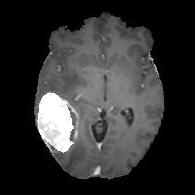}
\includegraphics[width=\x\linewidth,height=\x\linewidth]{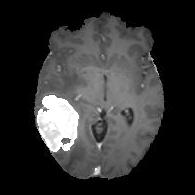}\\[3pt]
\includegraphics[width=\x\linewidth,height=\x\linewidth]{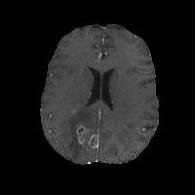}
\includegraphics[width=\x\linewidth,height=\x\linewidth]{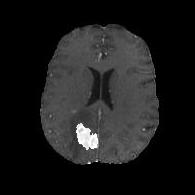}
\includegraphics[width=\x\linewidth,height=\x\linewidth]{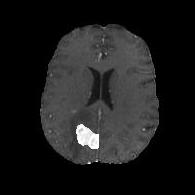}
\includegraphics[width=\x\linewidth,height=\x\linewidth]{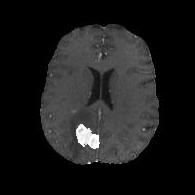}
\includegraphics[width=\x\linewidth,height=\x\linewidth]{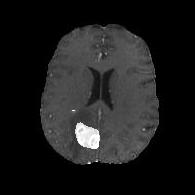}\\[3pt]
\includegraphics[width=\x\linewidth,height=\x\linewidth]{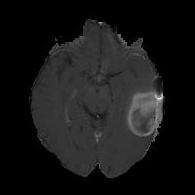}
\includegraphics[width=\x\linewidth,height=\x\linewidth]{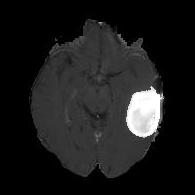}
\includegraphics[width=\x\linewidth,height=\x\linewidth]{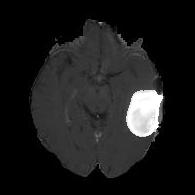}
\includegraphics[width=\x\linewidth,height=\x\linewidth]{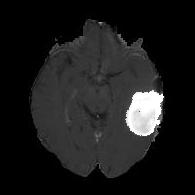}
\includegraphics[width=\x\linewidth,height=\x\linewidth]{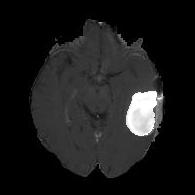}\\[3pt]
\includegraphics[width=\x\linewidth,height=\x\linewidth]{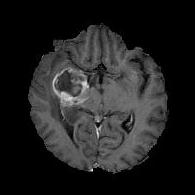}
\includegraphics[width=\x\linewidth,height=\x\linewidth]{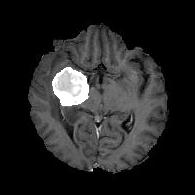}
\includegraphics[width=\x\linewidth,height=\x\linewidth]{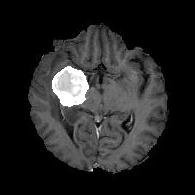}
\includegraphics[width=\x\linewidth,height=\x\linewidth]{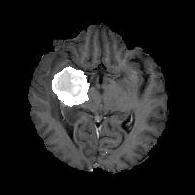}
\includegraphics[width=\x\linewidth,height=\x\linewidth]{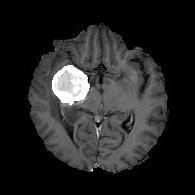}\\[3pt]
\includegraphics[width=\x\linewidth,height=\x\linewidth]{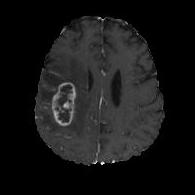}
\includegraphics[width=\x\linewidth,height=\x\linewidth]{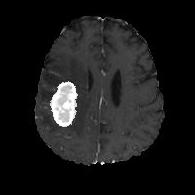}
\includegraphics[width=\x\linewidth,height=\x\linewidth]{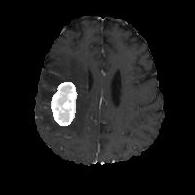}
\includegraphics[width=\x\linewidth,height=\x\linewidth]{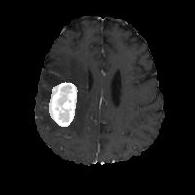}
\includegraphics[width=\x\linewidth,height=\x\linewidth]{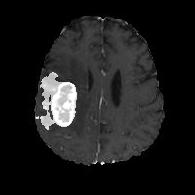}\\[3pt]
\includegraphics[width=\x\linewidth,height=\x\linewidth]{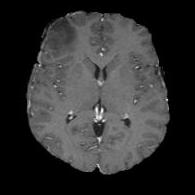}
\includegraphics[width=\x\linewidth,height=\x\linewidth]{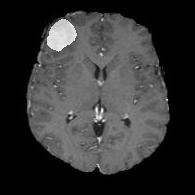}
\includegraphics[width=\x\linewidth,height=\x\linewidth]{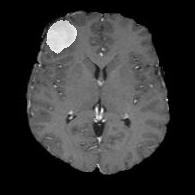}
\includegraphics[width=\x\linewidth,height=\x\linewidth]{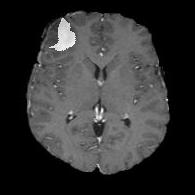}
\includegraphics[width=\x\linewidth,height=\x\linewidth]{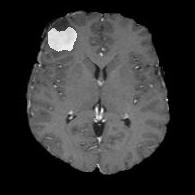}\\[3pt]
\includegraphics[width=\x\linewidth,height=\x\linewidth]{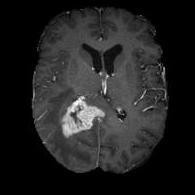}
\includegraphics[width=\x\linewidth,height=\x\linewidth]{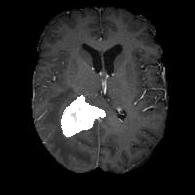}
\includegraphics[width=\x\linewidth,height=\x\linewidth]{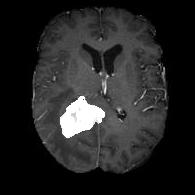}
\includegraphics[width=\x\linewidth,height=\x\linewidth]{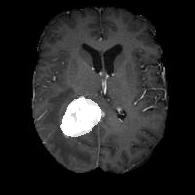}
\includegraphics[width=\x\linewidth,height=\x\linewidth]{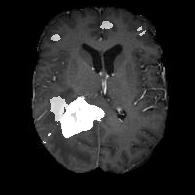}\\[3pt]
\makebox[\x\linewidth]{(a) Input image}  \makebox[\x\linewidth]{(b) Label} 
\makebox[\x\linewidth]{(c) Ours}  \makebox[\x\linewidth]{(d) V-Net}  \makebox[\x\linewidth]{(e) U-Net}
\caption[Qualitative comparison of 2D edge-aware predictions]
        {Qualitative comparison of 2D edge-aware predictions.}
\label{fig_data}
\end{figure}

\paragraph{Comparison to Competing Methods:}
We have compared the performance of our model against the most popular
deep learning-based semantic segmentation networks, U-Net
\citep{Ronneberger15} and V-Net \citep{Milletari2016}
(Figure~\ref{fig_data}). Our model outperforms both by a considerable
margin in all evaluation metrics. In particular, U-Net performs poorly
in most cases due to the high false positive of its segmentation
predictions, as well as the imprecision of its boundaries. The
powerful residual block in the V-Net architecture seems to alleviate
these issues to some extent, but V-Net also fails to produce
high-quality boundary predictions. The emphasis of learning useful
edge-related information during the training of our network appears to
effectively regularize the network such that boundary accuracy is
improved.

\section{3D Edge-Aware Encoder-Decoders}

In this section, we empirically study the models developed in Section~\ref{sec:3d-edge-aware}. 

\subsection{Dataset}

Kidney Tumor Segmentation Challenge (KiTS 2019) provides data of
multi-phase 3D CTs, voxel-wise ground truth labels, and comprehensive
clinical outcomes for 300 patients who underwent nephrectomy for
kidney tumors between 2010 to 2018 at University of Minnesota
\citep{heller2019kits19}. 210 patients were randomly selected for the
training set and the remaining 90 patients were left as a testing set.
The annotation was performed in the transverse plane with regular
subsampling of series in the longitudinal direction with roughly 50
annotated slices depicting the Kidney for each patient. The labels for
excluded slices were computed by using a contour interpolation
algorithm \citep{heller2019kits19}.

\subsection{Evaluation metrics}

We have adopted the same three evaluation metrics as outlined by KiTS
2019 challenge. Kidneys dice denote the segmentation performance when
considering both kidneys and tumors as the foreground whereas tumor
dice considers everything except the tumor as background. Composite
dice is simply the average of kidneys dice and tumor dice.

\subsection{Results}

\begin{figure}
\def\x{0.495}
\includegraphics[width=\x\linewidth,height=\x\linewidth]{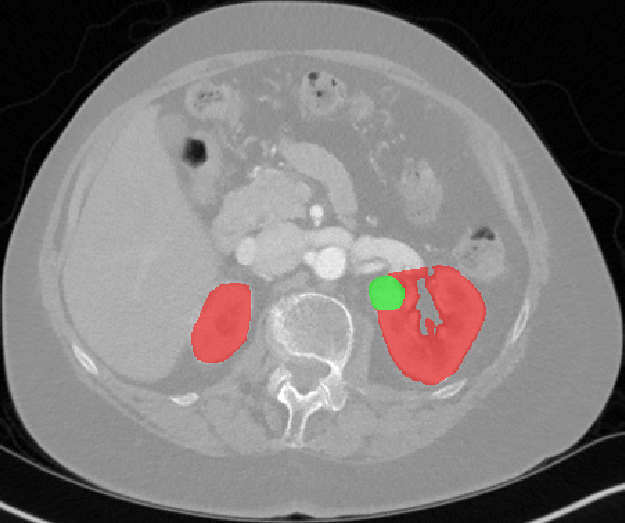}
\hfill
\includegraphics[width=\x\linewidth,height=\x\linewidth]{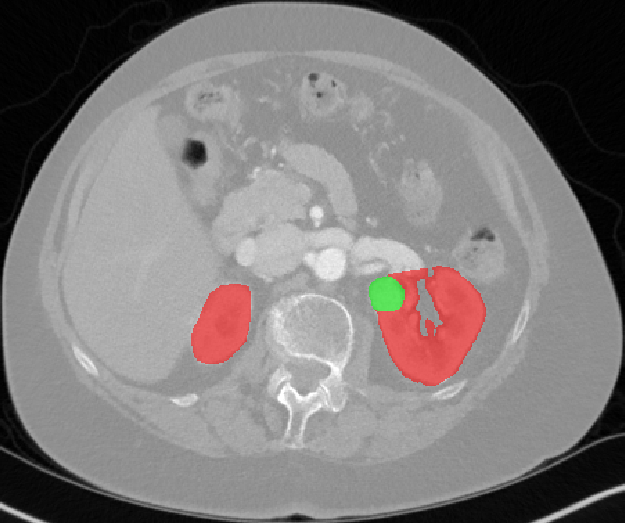}\\[4pt]
\includegraphics[width=\x\linewidth,height=\x\linewidth]{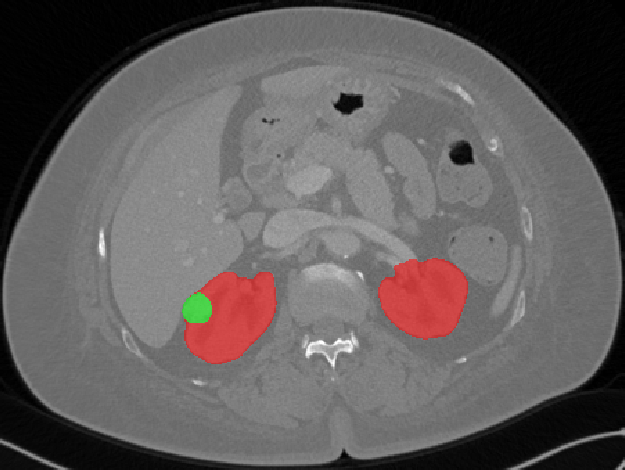}
\hfill
\includegraphics[width=\x\linewidth,height=\x\linewidth]{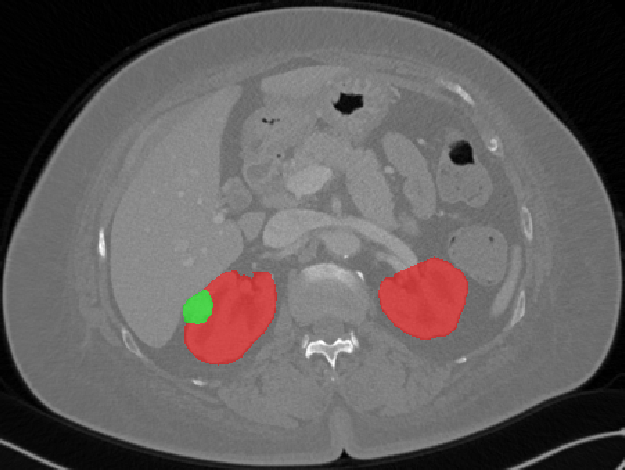}\\
\makebox[\x\linewidth]{(a) Our Predictions} \hfill \makebox[\x\linewidth]{(b) Ground truth Labels}
\caption[Qualitative comparison of volumetric edge-aware predictions]
        {Visualization of (a) our model's predictions and (b) ground
        truth labels.}
\label{fig:res_vis}
\end{figure}

Table~\ref{tab:results} represents the evaluation results of our model
on our own dataset partition. We divided the training set of KiTS 2019
dataset into our own subsets for training and validation and evaluated
the performance of a our proposed model. Figure~\ref{fig:res_vis}
illustrates the segmentation visualizations of our method and their
corresponding ground truth from two cases in the validation set of our
own partition.

\begin{table} \centering
\begin{tabular}{ccc}
\toprule
Kidneys Dice & Tumor Dice & Composite Dice \\ 
\midrule
0.96 & 0.82 & 0.89 \\
\bottomrule
\end{tabular}
\caption[Evaluation results of 3D edge-aware network on the KiTS 2019
test set] {Evaluation results of the 3D edge-aware network on the KiTS
2019 test set.}
\label{tab:results}
\end{table}

\section{Plug-and-Play Edge-Aware CNNs}

In this section, we empirically study the models developed in Section~\ref{sec:EG-CNN}.

\subsection{Implementation Details}

DALS is implemented in Tensorflow \citep{abadi2016tensorflow}. We
trained it on an NVIDIA Titan XP GPU and an Intel® Core™ i7-7700K CPU
@ 4.20GHz. All the input images were first normalized and resized to a
predefined size of $256\times256$ pixels. The size of the mini-batches
is set to 4, and the Adam optimization algorithm was used with an
initial learning rate of 0.001 that decays by a factor of 10 every 10
epochs. The entire inference time for DALS takes $1.5$ seconds. All
model performances were evaluated by using the Dice coefficient,
Hausdorff distance, and BoundF.

\begin{table} \centering
\setlength{\tabcolsep}{4pt}
\resizebox{1.0\linewidth}{!}{%
\begin{tabular}{llcccc}
\toprule
Model & Edge  & Average Dice & ET  Dice  & TC Dice & WT Dice  \\
\midrule
U-Net  & None   &  0.8305$\pm$0.0035&	0.7375$\pm$0.0021&	0.8480$\pm$0.0056& 0.9060$\pm$0.0021  \\
V-Net &None& 0.8281$\pm$0.0035&	0.7255$\pm$0.0049&	0.8570$\pm$0.0042& 0.9020$\pm$0.0014	 \\
Seg-Net &None&  0.8300$\pm$0.0033&0.7330$\pm$0.0042&0.8550$\pm$0.0049& 0.9015$\pm$0.0007\\
U-Net & EG-CNN  &  0.8406$\pm$0.0028&	0.7530$\pm$0.0113&	0.8630$\pm$0.0014& 0.9006$\pm$0.0042\\
V-Net &EG-CNN&  0.8386$\pm$0.0051&	0.7460$\pm$0.0056&	0.8605$\pm$0.0035& 0.9095$\pm$0.0063  \\
Seg-Net &EG-CNN&\textbf{0.8570$\pm$0.0007}&\textbf{0.7680$\pm$0.0113}&\textbf{0.8850$\pm$0.0070}&\textbf{0.9180$\pm$0.0028}\\
\bottomrule
\end{tabular}
}
\bigskip
\caption[Evaluation results of EG-CNN et al.~on the BraTS 2019
dataset] {Evaluation results on the BraTS 2019 dataset in terms of
overall and tumor subregions Dice scores. The Edge column determines
whether EG-CNN is utilized with the backbone architecture.}
\label{tab:result}
\end{table}

\begin{figure} \centering
\def\x{0.242}
\includegraphics[width=\x\linewidth,height=\x\linewidth]{edge_cnn/case7_img.png}
\includegraphics[width=\x\linewidth,height=\x\linewidth]{edge_cnn/case7_gt.png}
\includegraphics[width=\x\linewidth,height=\x\linewidth]{edge_cnn/case7_ours.png}
\includegraphics[width=\x\linewidth,height=\x\linewidth]{edge_cnn/case7_segnet.png}\\[3pt]
\includegraphics[width=\x\linewidth,height=\x\linewidth]{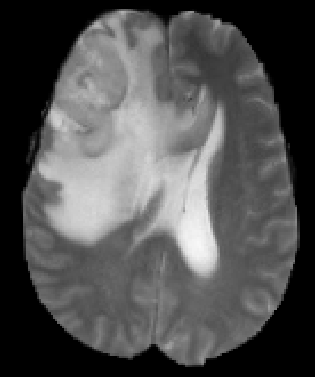}
\includegraphics[width=\x\linewidth,height=\x\linewidth]{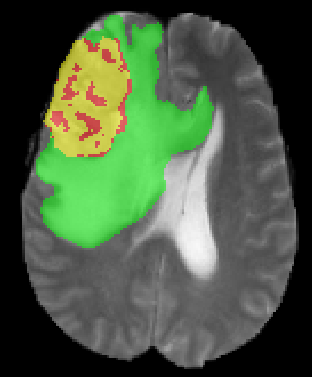}
\includegraphics[width=\x\linewidth,height=\x\linewidth]{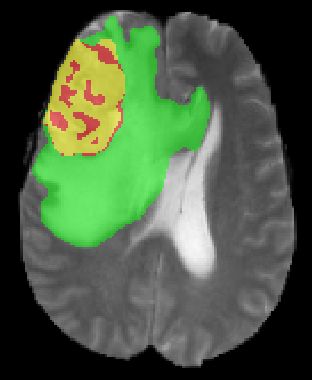}
\includegraphics[width=\x\linewidth,height=\x\linewidth]{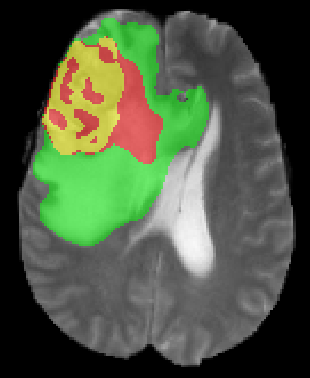}
\vspace{0.75pt}
\makebox[\x\linewidth]{(a) Input images} 
\makebox[\x\linewidth]{(b) Semantic Labels}
\makebox[\x\linewidth]{(c) Seg-Net+EG-CNN}
\makebox[\x\linewidth]{(d) Seg-Net}
\caption[Visualization of the in-plane segmentation output in BraTS] {Visualization of the in-plane segmentation outputs for tumor
subregions in the BraTS 2019 dataset. Red, green, and yellow labels
denote TC, WT, and ET subregions, respectively.}
\label{fig:brats9}
\end{figure}

\subsection{Results}

We evaluated the EG-CNN module when it is used to augment popular
medical image segmentation models: U-Net \citep{Ronneberger15}, V-Net
\citep{Milletari2016}, and Seg-Net \citep{Myronenko18}. We modified each
architecture to adopt them to the given task and to be similar to the
others for a more fair comparison. For both the U-net and V-net, we
changed the normalization to Groupnorm, to better handle a small batch
size, and adjusted the number of layers to a roughly equivalent number
between the networks. For each dataset we trained the main CNN
segmentation network with and without the EG-CNN in order to validate
the contribution of our proposed module. We estimated the accuracy of
each model in terms of Dice score for each class and of the overall
average.

\subsubsection{BraTS 2019}

Table~\ref{tab:result} reports the accuracy of the model on each of
the classes: Whole Tumor (WT), Tumor Core (TC), and Enhancing Tumor
(ET), as well as the overall overage accuracy. According to our
benchmarks, including the EG-CNN consistently increases the overall
and subregion Dice scores in all cases. In the case of brain tumor
segmentation, the EG-CNN has effectively learned highly complex and
irregular boundaries of certain subregions. Therefore, it improves
the segmentation quality around the edges, which leads to overall
better segmentation performance. Figure~\ref{fig:brats9} illustrates
how the addition of the EG-CNN to a standalone Seg-Net
\citep{Myronenko18} improves the quality of segmentation.

The quality of the predicted edges also validates the effectiveness of
our proposed edge-aware loss function, since the boundaries are crisp
and avoid the thickening effect around edges. Such a phenomenon
usually occurs when a naive loss function such as binary cross entropy
is utilized for the task of edge prediction without taking
precautions. Moreover, our model results in more fine-grained
boundaries and visually attractive edges because the learned predicted
boundaries are eventually fused with the final prediction output of
the main encoder-decoder architecture.

Since the addition of the EG-CNN module increases the number of free
parameters of the overall model, we have also experimented with larger
standalone models (by increasing their depth and/or width), but doing
so did not result in the better validation accuracy. This indicates
that our module improves the overall segmentation accuracy not due to
the model capacity increase, but due to the extra emphasis of edge
information.

\begin{table} \centering
\setlength{\tabcolsep}{4pt}
\begin{tabular}{llccc}
\toprule
Model & Edge & Kidneys Dice & Tumor Dice &  Composite Dice \\
\midrule
U-Net& None    &  0.9515$\pm$0.0049&	0.8245$\pm$0.0091&	0.8880$\pm$0.0070	  \\
V-Net&None     &  0.9370$\pm$0.0065&	0.8072$\pm$0.0072&	0.8720$\pm$0.0068	  \\
Seg-Net&None     &  0.9530$\pm$0.0028&	0.8235$\pm$0.0049&	0.8892$\pm$0.0038	  \\
U-Net&  EG-CNN   &  0.9620$\pm$0.0056&	0.8270$\pm$0.0084&	0.8945$\pm$0.0070	  \\
V-Net&EG-CNN     &  0.9483$\pm$0.0048&	0.8275$\pm$0.0087&	0.8879$\pm$0.0067	  \\
Seg-Net& EG-CNN     &  \textbf{0.9647$\pm$0.0051}&	\textbf{0.8353$\pm$0.0025}&	\textbf{0.9000$\pm$0.0038}\\
\bottomrule
\end{tabular}
\caption[Evaluation results of EG-CNN et al.~on the KiTS 2019 dataset]
        {Evaluation results of EG-CNN on the KiTS 2019 dataset for
        kidneys, tumor, and composite Dice functions. The Edge column
        determines whether EG-CNN is utilized with the backbone
        architecture.}
\label{tab:kits}
\end{table}

\begin{figure} \centering
\def\x{0.242}
\includegraphics[width=\x\linewidth,height=\x\linewidth]{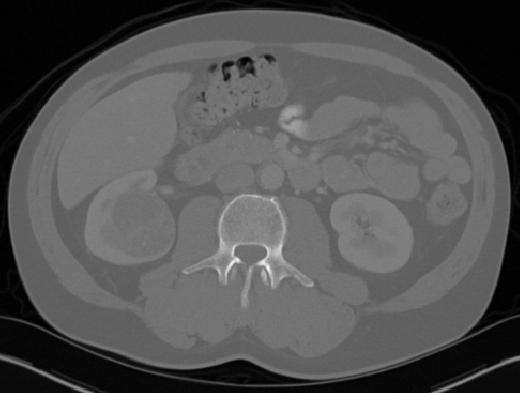}
\includegraphics[width=\x\linewidth,height=\x\linewidth]{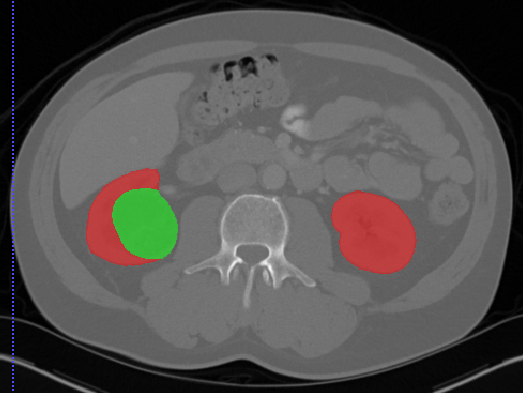}
\includegraphics[width=\x\linewidth,height=\x\linewidth]{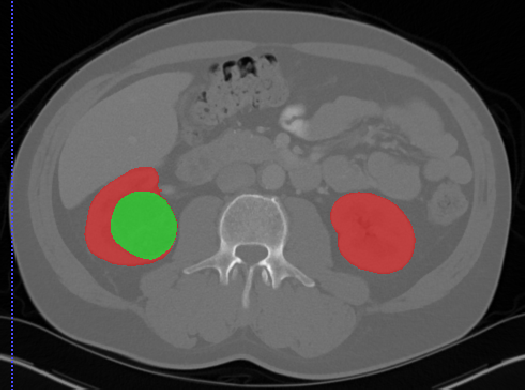}
\includegraphics[width=\x\linewidth,height=\x\linewidth]{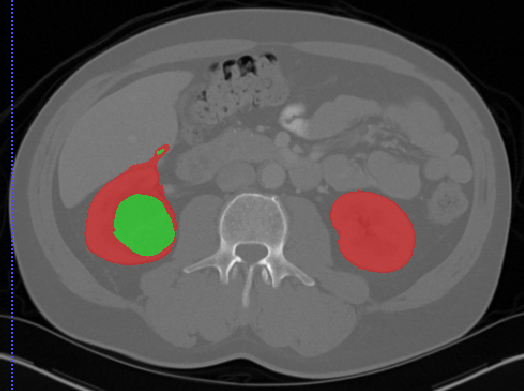}\\[3pt]
\includegraphics[width=\x\linewidth,height=\x\linewidth]{edge_cnn/kits_im_2.png}
\includegraphics[width=\x\linewidth,height=\x\linewidth]{edge_cnn/kits_gt_2.png}
\includegraphics[width=\x\linewidth,height=\x\linewidth]{edge_cnn/kits_edge_2.png}
\includegraphics[width=\x\linewidth,height=\x\linewidth]{edge_cnn/kits_base_2.png}

\makebox[\x\linewidth]{(a) Input images} 
\makebox[\x\linewidth]{(b) Semantic Labels}
\makebox[\x\linewidth]{(c) Seg-Net+EG-CNN}
\makebox[\x\linewidth]{(d) Seg-Net}
\caption[Qualitative comparison between Seg-Net and Seg-Net+EG-CNN on
KiTS] {Visualization of the segmentation performance of Seg-Net
and Seg-Net+EG-CNN on the KiTS 2019 challenge. The green and red masks
denote tumor and kidneys, respectively.}
\label{fig:kits}
\end{figure}

\subsubsection{KiTS 2019}

The achieved accuracy of the model for kidneys and kidney tumor
classes, as well as the overall accuracy are presented in
Table~\ref{tab:kits}. Similar to the results achieved on the BraTS
2019 dataset, the addition of EG-CNN has consistently improved the
segmentation performance. Visual comparisons of the output
segmentation and boundary predictions are presented in
Figure~\ref{fig:kits}. As such, the predicted edges visually conform
to the region outlines, demonstrating that the EG-CNN module and our
proposed loss functions helped to captured the details of the edges.
This has also been reflected in the final predictions of semantic
masks.

\section{Deep Active Lesion Segmentation}

In this section, we empirically study the models developed in Section~\ref{sec:dals}.

\subsection{Multiorgan Lesion Segmentation (MLS) Dataset}

\begin{table} \centering
\setlength{\tabcolsep}{4pt}
\resizebox{\linewidth}{!}{%
\begin{tabular}{llcccccccr}
\toprule
Organ        & Modality    & \# Samples  & $\text{Mean}_\text{GC}$ & $\text{Var}_\text{GC}$ & $\text{Mean}_\text{GH}$ & $\text{Var}_\text{GH}$ & Lesion Radius (pixels)\\
\midrule
Brain &    MRI      &    369     &    0.56    & 0.029  & 0.907  & 0.003 &  17.42 $\pm$ 9.516    \\
Lung  &    CT      &    87      &    0.315   & 0.002  & 0.901  & 0.004 &  15.15 $\pm$ 5.777    \\
Liver &    CT      &    112     &    0.825   & 0.072  & 0.838  & 0.002 &  20.483 $\pm$ 10.37    \\
Liver &    MRI      &    164     &    0.448   & 0.041  & 0.891  & 0.003 &   5.459  $\pm$ 2.027  \\
\bottomrule
\end{tabular}
}
\caption[MLS dataset statistics] {MLS dataset statistics. GC and GH
denote Global Contrast and Global Heterogeneity, respectively.}
\label{tab:stanford-dataset}
\end{table}

As shown in Table~\ref{tab:stanford-dataset}, the MLS dataset includes
images of highly diverse lesions in terms of size and spatial
characteristics such as contrast and homogeneity. The liver component
of the dataset consists of 112 contrast-enhanced CT images of liver
lesions (43 hemangiomas, 45 cysts, and 24 metastases) with a mean
lesion radius of 20.483 $\pm$ 10.37 pixels and 164 liver lesions from
3T gadoxetic acid enhanced MRI scans (one or more LI-RADS (LR), LR-3,
or LR-4 lesions) with a mean lesion radius of 5.459 $\pm$ 2.027
pixels. The brain component consists of 369 preoperative and
pretherapy perfusion MR images with a mean lesion radius of 17.42
$\pm$ 9.516 pixels. The lung component consists of 87 CT images with a
mean lesion radius of 15.15 $\pm$ 5.777 pixels. For each component of
the MLS dataset, we used 85\% of its images for training, 10\% for
testing, and 5\% for validation.

\subsection{Results}

\begin{table} \centering
\setlength{\tabcolsep}{4pt}
\resizebox{\linewidth}{!}{%
\begin{tabular}{ c c c c c c c}
\toprule
Dataset & Model & Dice   &   CI &  Hausdorff &   CI   &   BoundF\\ 
\midrule
\multirow{4}[7]{*}{Brain MR} 
    & U-Net &  
0.776 $\pm$ 0.214 & 
0.090 & 
2.988 $\pm$  1.238 & 
0.521 & 
0.826\\ 
\cmidrule{2-7}
    & CNN Backbone & 
0.824 $\pm$ 0.193 & 
0.078 &
2.755 $\pm$ 1.216 & 
0.490 &
0.891\\ 
\cmidrule{2-7}
    & Level-set & 
0.796 $\pm$ 0.095 & 
0.038& 
2.927 $\pm$ 0.992& 
0.400&
0.841\\ 
\cmidrule{2-7}
    & \textbf{DALS} & 
\textbf{0.888 $\pm$ 0.0755} & 
\textbf{0.030}& 
\textbf{2.322 $\pm$ 0.824}& 
\textbf{0.332}&
\textbf{0.944}\\ 
\midrule
\multirow{4}[7]{*}{Lung CT} 
    & U-Net &  
0.817 $\pm$ 0.098 & 
0.0803 & 
2.289 $\pm$ 0.650 & 
0.530 & 
0.898\\ 
\cmidrule{2-7}
    & CNN Backbone & 
0.822 $\pm$ 0.115 & 
0.0944 &
2.254 $\pm$ 0.762 & 
0.6218 &
0.900\\ 
\cmidrule{2-7}
    & Level-set & 
0.789 $\pm$ 0.078 & 
0.064& 
3.270 $\pm$ 0.553& 
0.451&
0.879\\ 
\cmidrule{2-7}
    & \textbf{DALS} & 
\textbf{0.869 $\pm$ 0.113} & 
\textbf{0.092}& 
\textbf{2.095 $\pm$ 0.623}& 
\textbf{0.508}&
\textbf{0.937}\\ 
\midrule
\multirow{4}[7]{*}{Liver MR} 
    & U-Net &  
0.769 $\pm$ 0.162 & 
0.093 & 
1.645 $\pm$  0.598 & 
0.343 & 
0.920\\ 
\cmidrule{2-7}
    & CNN Backbone & 
0.805 $\pm$ 0.193 & 
0.110 &
1.347 $\pm$ 0.671 & 
0.385 &
0.939\\ 
\cmidrule{2-7}
    & Level-set & 
0.739 $\pm$ 0.102 & 
0.056& 
2.227 $\pm$ 0.576& 
0.317&
0.954\\ 
\cmidrule{2-7}
    & \textbf{DALS} & 
\textbf{0.894 $\pm$  0.065} & 
\textbf{0.036}& 
\textbf{1.298 $\pm$ 0.434}& 
\textbf{0.239}&
\textbf{0.987}\\ 
\midrule
\multirow{4}[7]{*}{Liver CT} 
    & U-Net &  
0.698 $\pm$  0.149 & 
0.133 & 
4.422 $\pm$ 0.969 & 
0.866 & 
0.662\\ 
\cmidrule{2-7}
    & CNN Backbone & 
0.801 $\pm$ 0.178 & 
0.159 &
3.813 $\pm$  1.791 & 
1.600 &
0.697\\ 
\cmidrule{2-7}
    & Level-set & 
0.765 $\pm$ 0.039 & 
0.034& 
3.153 $\pm$ 0.825& 
0.737&
0.761 \\ 
\cmidrule{2-7}
    & \textbf{DALS} & 
\textbf{0.846 $\pm$ 0.090}& 
\textbf{0.080}& 
\textbf{3.113 $\pm$ 0.747}& 
\textbf{0.667}&
\textbf{0.773}\\ 
\bottomrule
\end{tabular}
}
\caption[Quantitative comparison of DALS and others on the MLS
dataset] {Quantitative comparison of segmentation performance of DALS
and other methods on the MLS dataset. CI denotes the confidence
interval.}
\label{tab:lidc-ltrc2}
\end{table}

\begin{figure}
\centering
\includegraphics[width=0.49\linewidth,trim={36 6 0 0},clip]{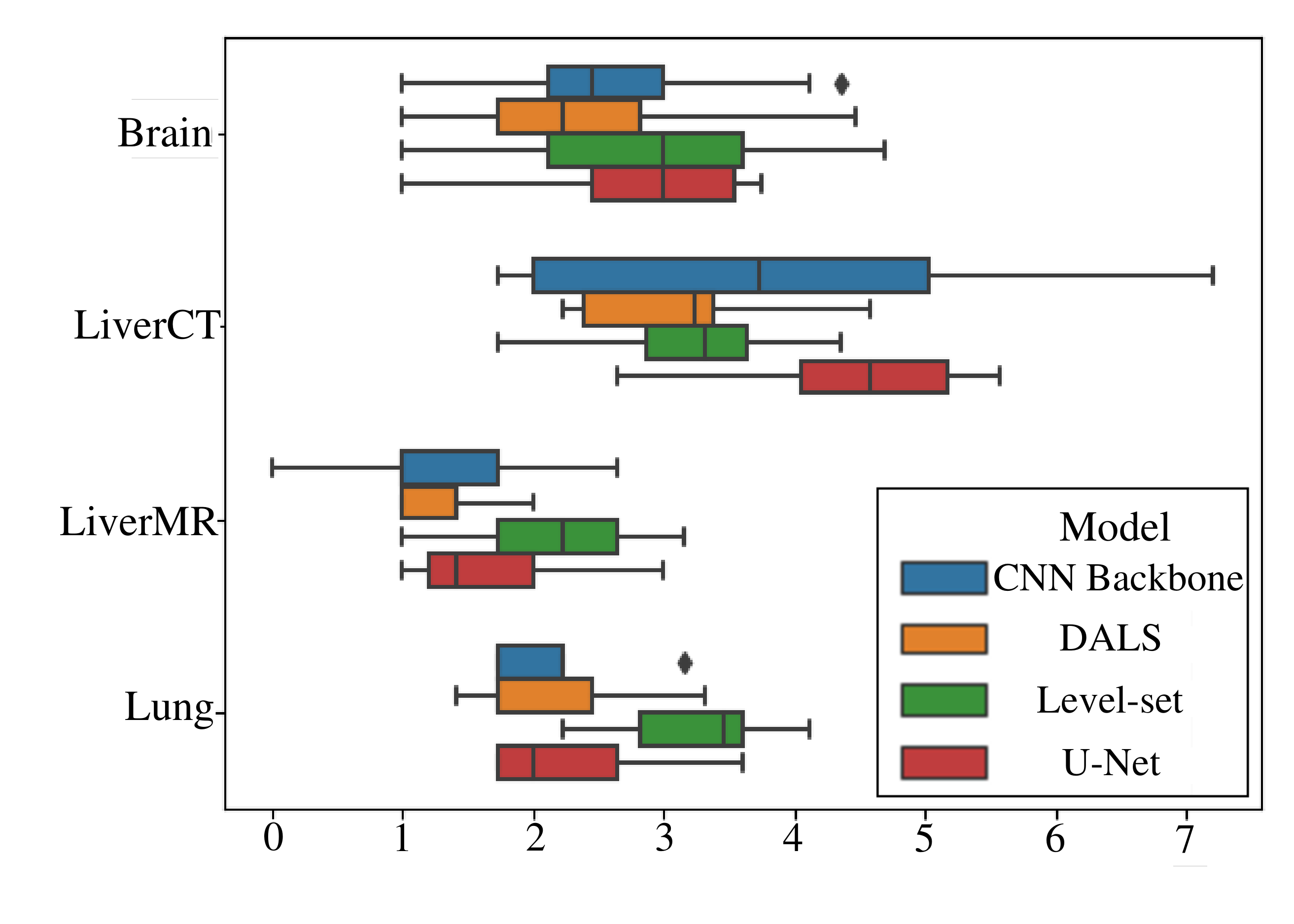}
\hfill
\includegraphics[width=0.48\linewidth,trim={36 -5 0 0},clip]{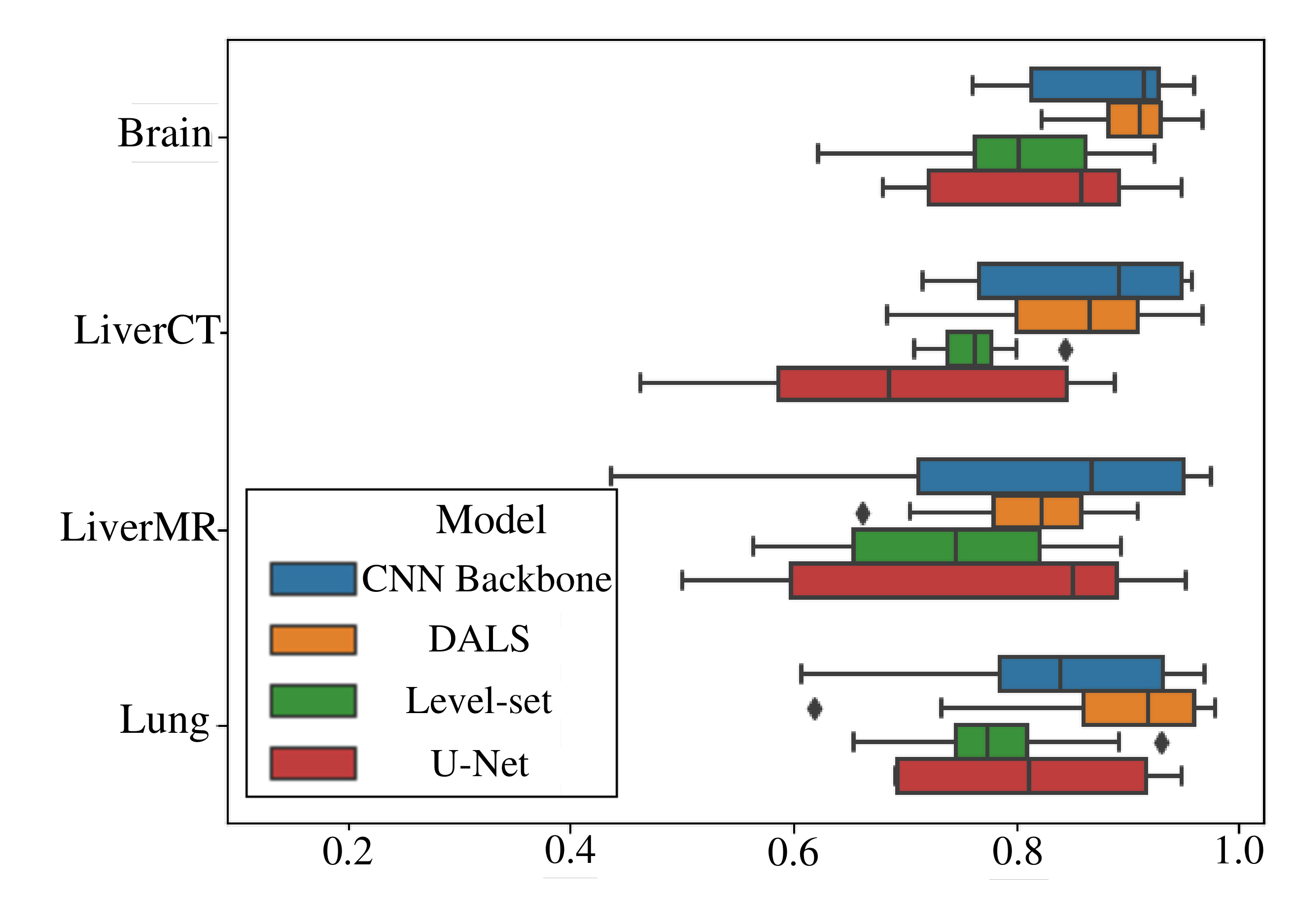}
\makebox[0.52\linewidth]{(a)} \hfill \makebox[0.40\linewidth]{(b)} \hfill
\caption[Box and whisker plots of DALS's predictions] {Box and whisker
plots of: (a) Dice score; (b) Hausdorff distance.}
\label{fig:box_plots}
\end{figure}

\subsubsection{Algorithm Comparison}

We have quantitatively compared our DALS against U-Net
\citep{Ronneberger15} and manually-initialized level-set ACM with
scalar $\lambda$ parameter constants as well as its backbone CNN. The
evaluation metrics for each organ are reported in
Table~\ref{tab:lidc-ltrc2} and box and whisker plots are shown in
Figure~\ref{fig:box_plots}. Our DALS achieves superior accuracies
under all metrics and in all datasets. Furthermore, we evaluated the
statistical significance of our method by applying a Wilcoxon paired
test on the calculated Dice results. Our DALS performed significantly
better than the U-Net ($p<0.001$), the manually-initialized ACM
($p<0.001$), and DALS's backbone CNN on its own ($p<0.005$).

\begin{figure}
\centering
\def\x{0.16}
\includegraphics[width=\x\linewidth,height=\x\linewidth,trim={000 000 000 000},clip]{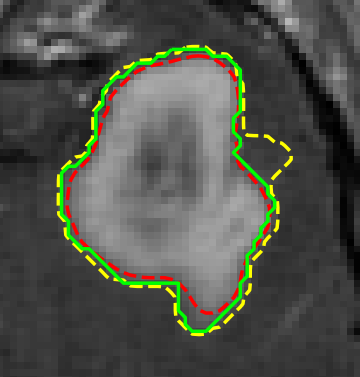}
\includegraphics[width=\x\linewidth,height=\x\linewidth,trim={150 120 150 120},clip]{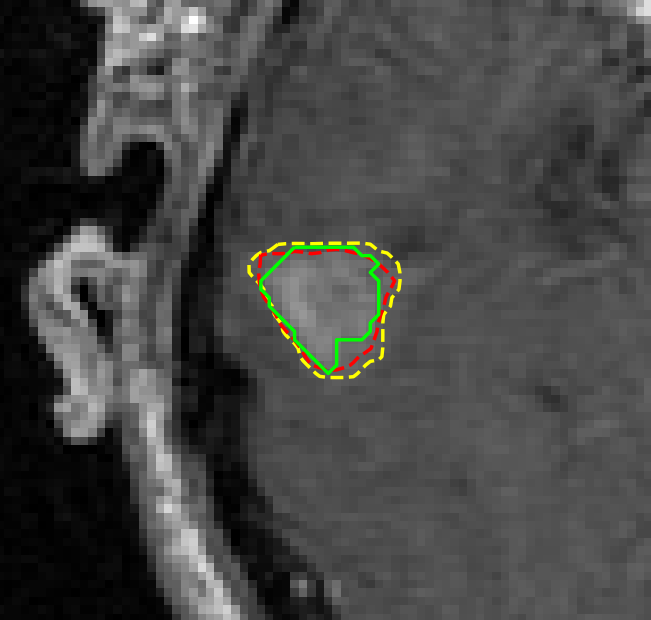} 
\includegraphics[width=\x\linewidth,height=\x\linewidth,trim={120 170 180 170},clip]{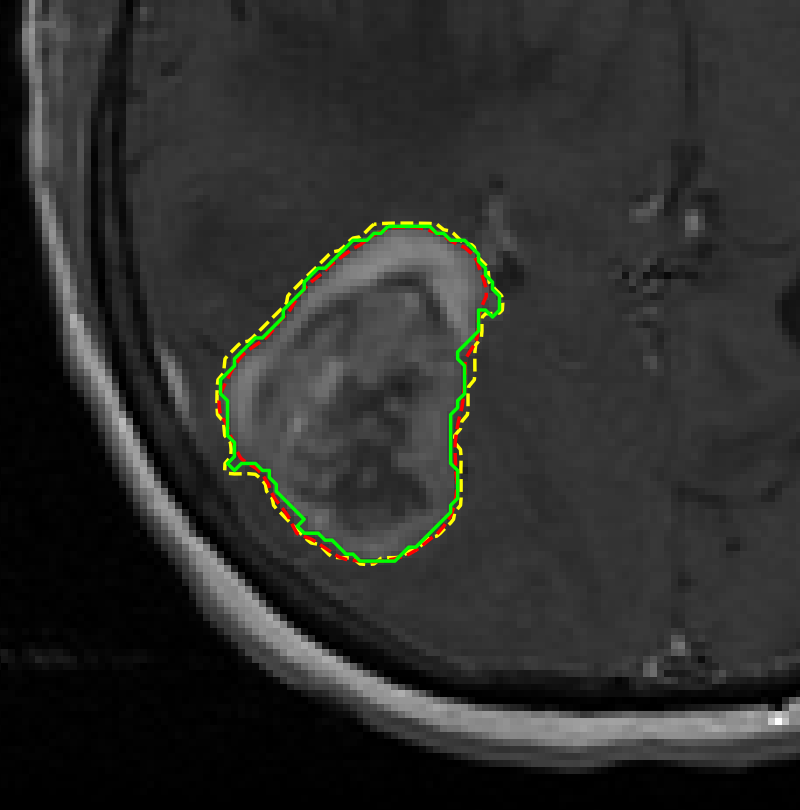} 
\includegraphics[width=\x\linewidth,height=\x\linewidth,trim={150 120 150 120},clip]{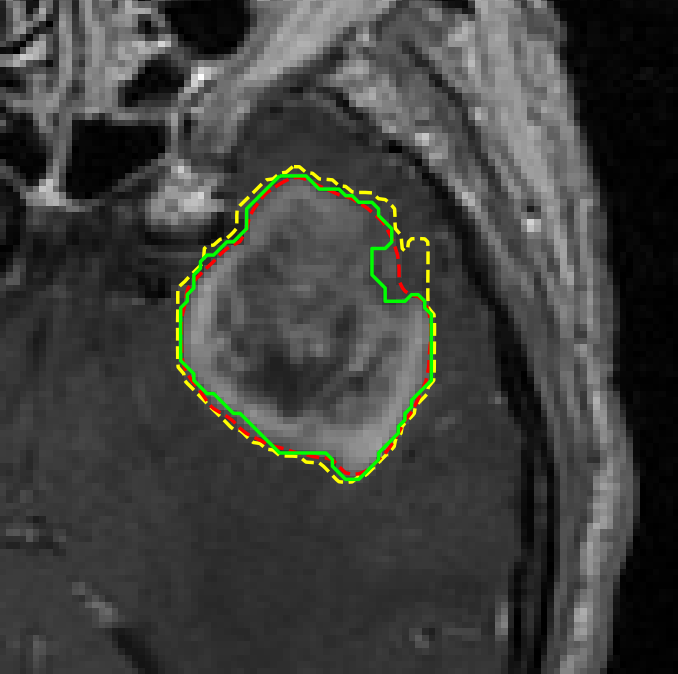} 
\includegraphics[width=\x\linewidth,height=\x\linewidth,trim={150 120 150 120},clip]{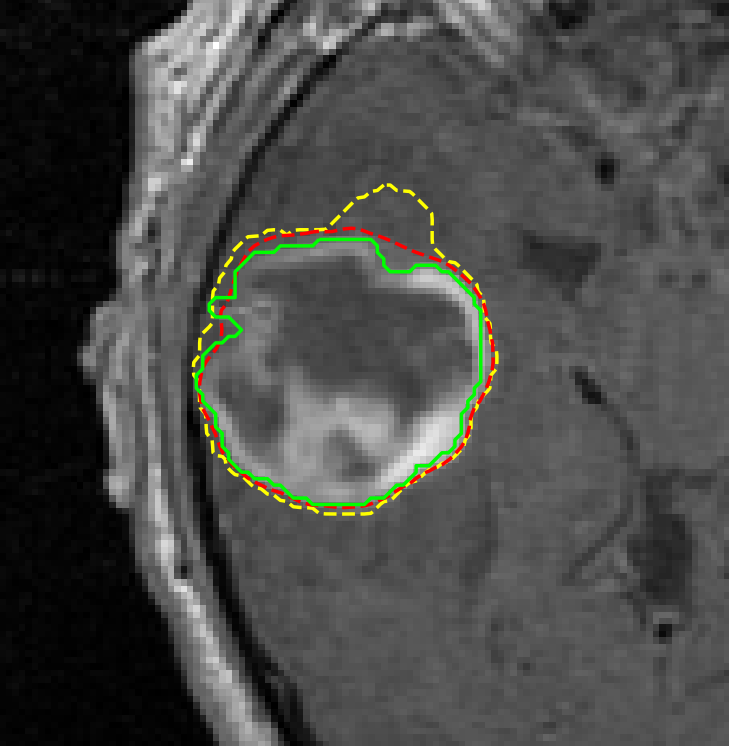}\\[3pt]
\includegraphics[width=\x\linewidth,height=\x\linewidth,trim={150 120 150 120},clip]{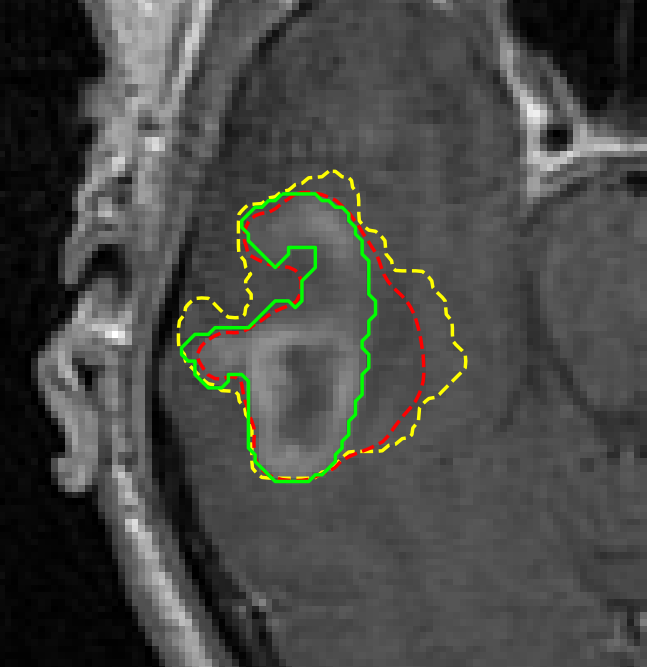} 
\includegraphics[width=\x\linewidth,height=\x\linewidth,trim={110 120 120 120},clip]{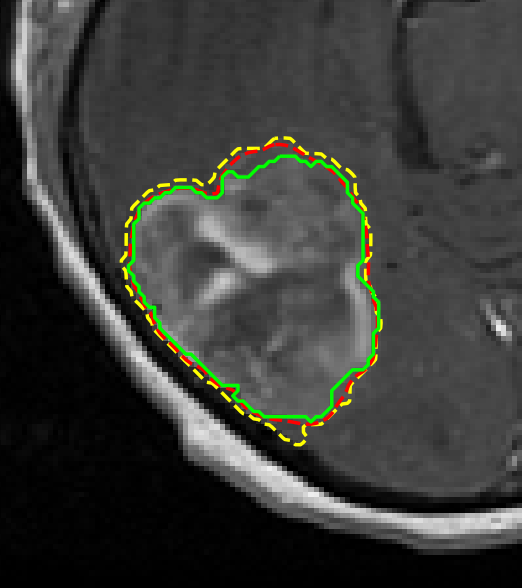} 
\includegraphics[width=\x\linewidth,height=\x\linewidth]{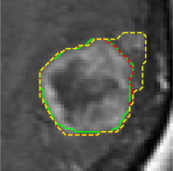} 
\includegraphics[width=\x\linewidth,height=\x\linewidth]{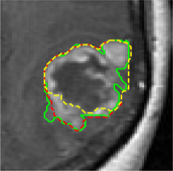} 
\includegraphics[width=\x\linewidth,height=\x\linewidth]{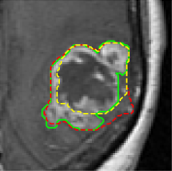}\\[3pt]
\includegraphics[width=\x\linewidth,height=\x\linewidth,trim={150 120 150 120},clip]{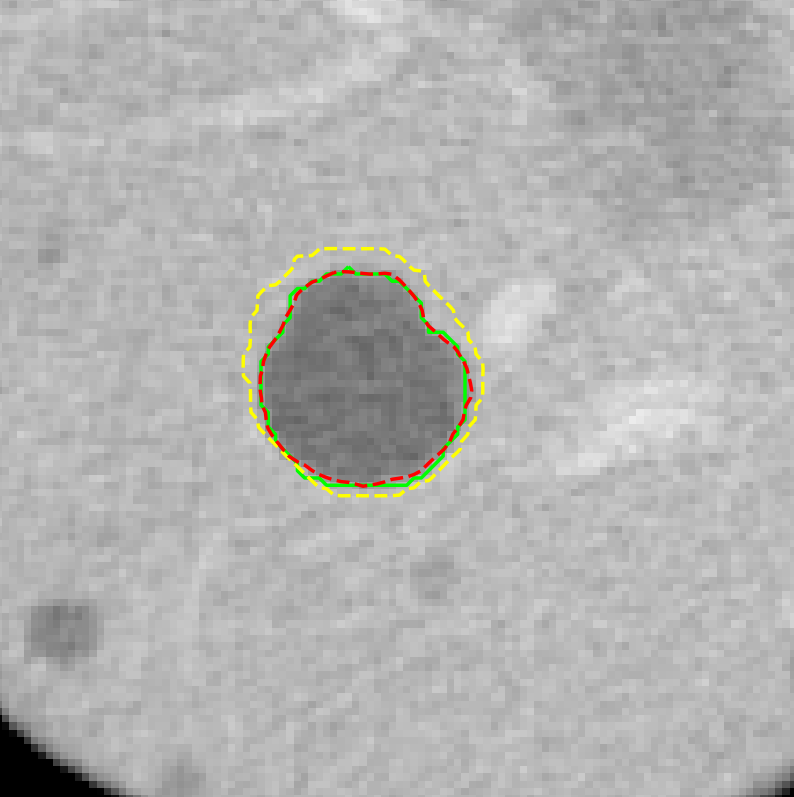} 
\includegraphics[width=\x\linewidth,height=\x\linewidth,trim={150 120 150 120},clip]{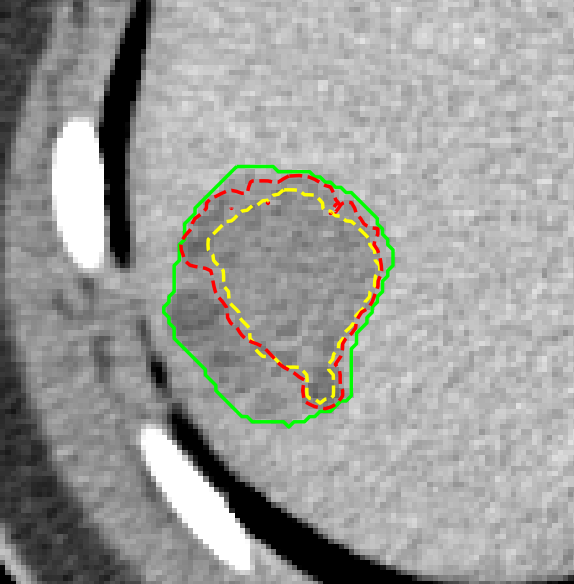} 
\includegraphics[width=\x\linewidth,height=\x\linewidth,trim={150 120 150 120},clip]{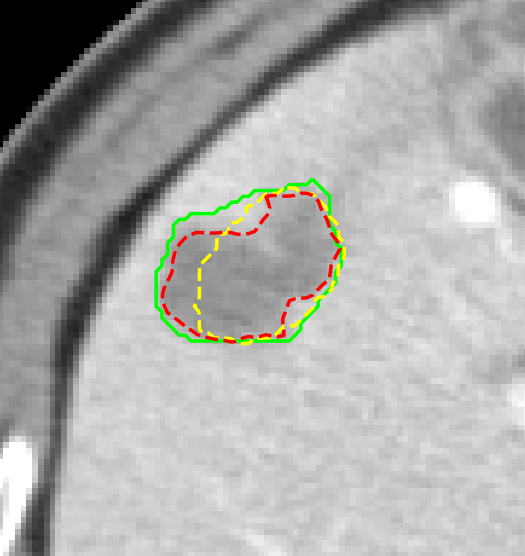} 
\includegraphics[width=\x\linewidth,height=\x\linewidth,trim={150 120 150 120},clip]{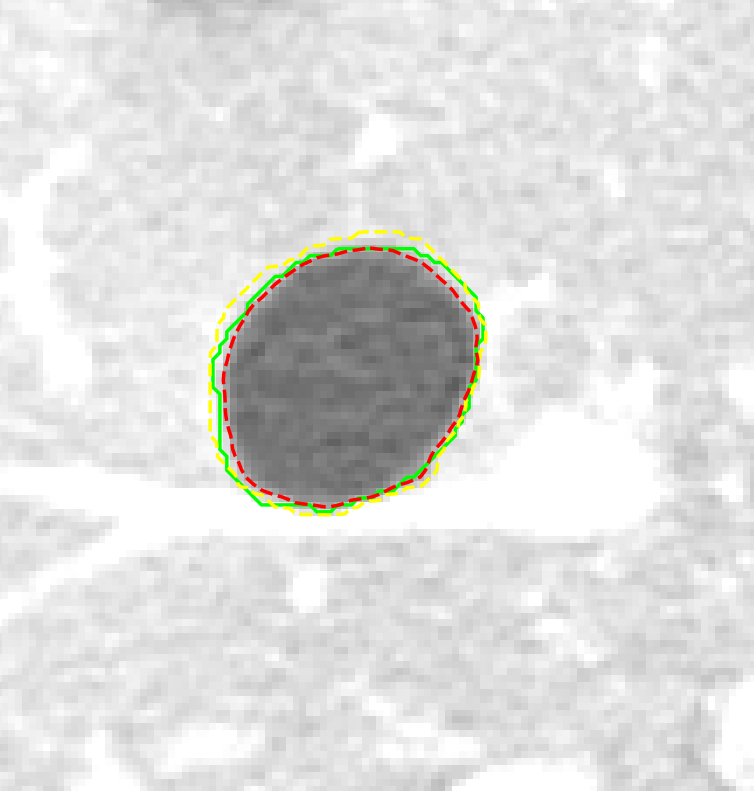} 
\includegraphics[width=\x\linewidth,height=\x\linewidth,trim={150 120 150 120},clip]{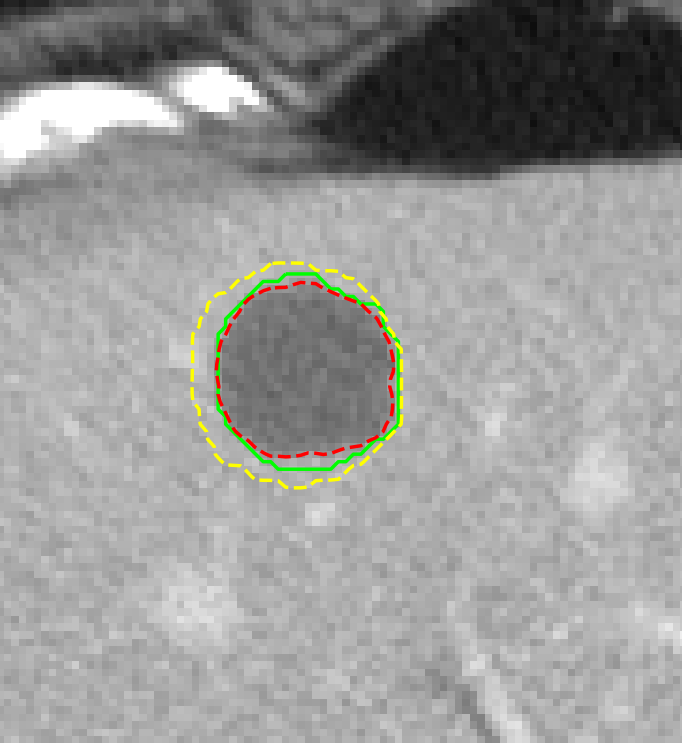}\\[3pt]
\includegraphics[width=\x\linewidth,height=\x\linewidth,trim={110 120 130 120},clip]{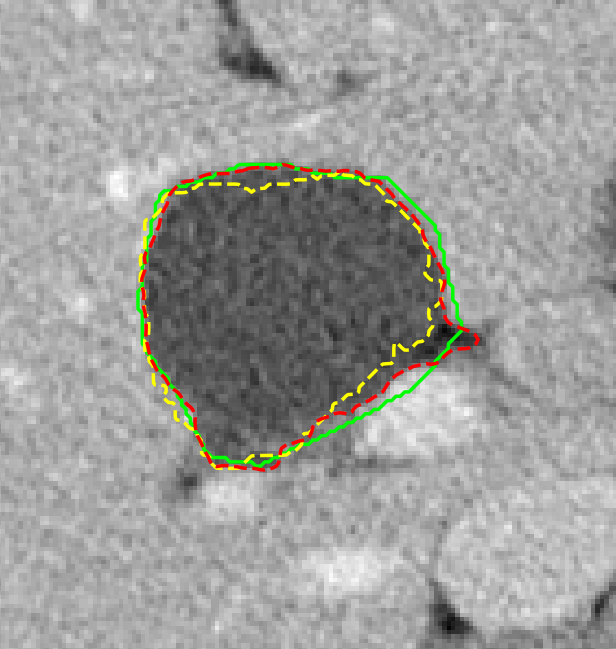} 
\includegraphics[width=\x\linewidth,height=\x\linewidth,trim={150 120 150 120},clip]{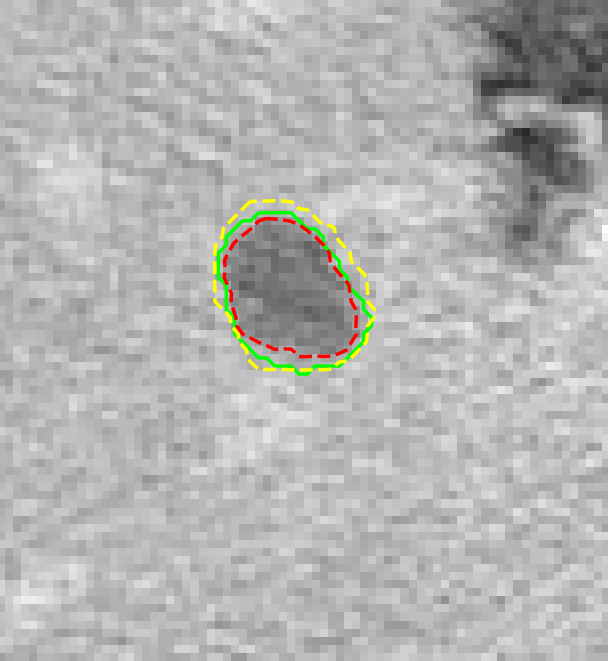} 
\includegraphics[width=\x\linewidth,height=\x\linewidth]{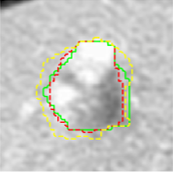} 
\includegraphics[width=\x\linewidth,height=\x\linewidth]{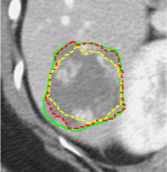} 
\includegraphics[width=\x\linewidth,height=\x\linewidth]{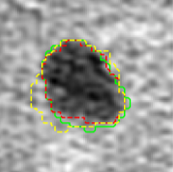}\\[3pt]
\includegraphics[width=\x\linewidth,height=\x\linewidth,trim={150 120 150 120},clip]{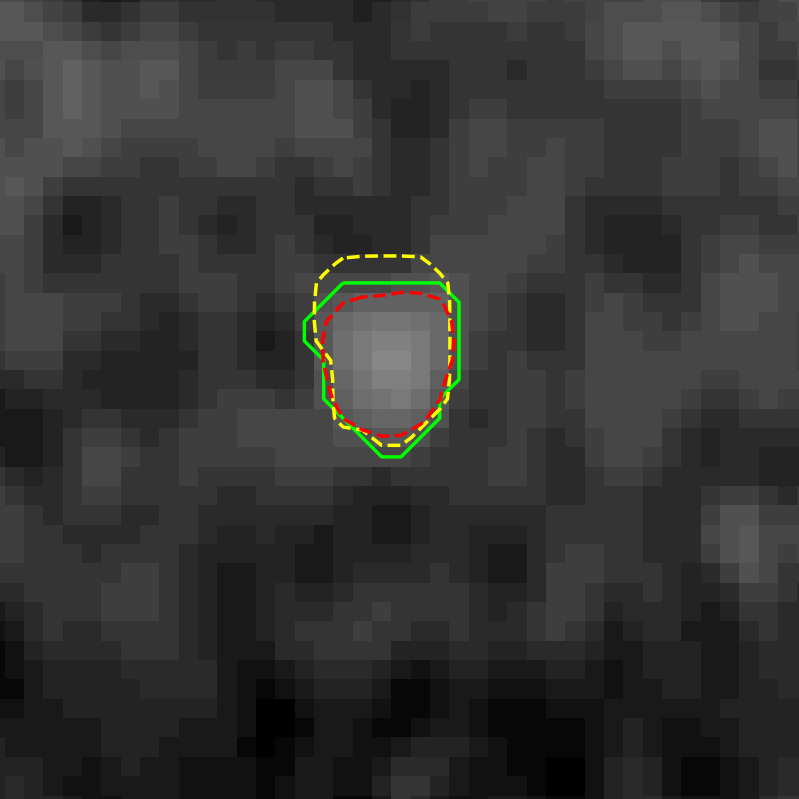} 
\includegraphics[width=\x\linewidth,height=\x\linewidth,trim={150 120 150 120},clip]{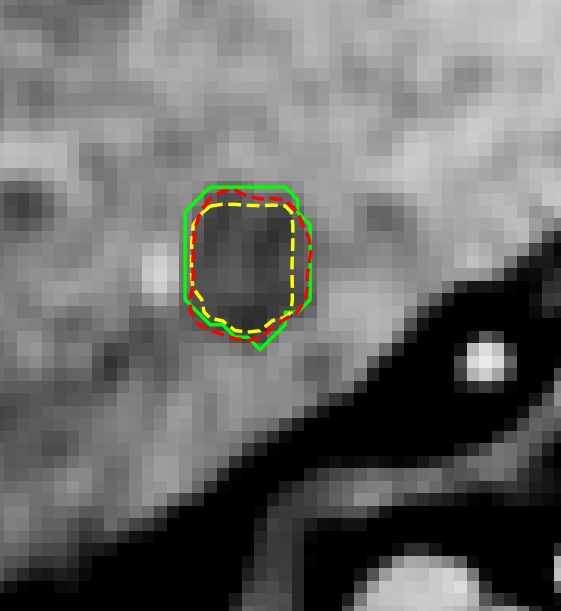} 
\includegraphics[width=\x\linewidth,height=\x\linewidth,trim={150 120 150 120},clip]{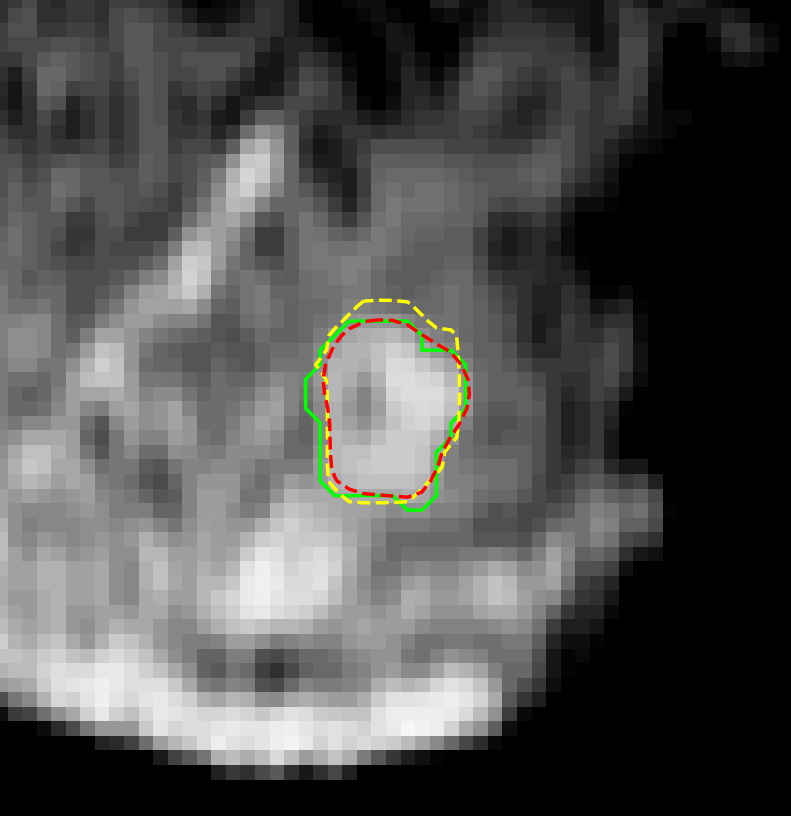} 
\includegraphics[width=\x\linewidth,height=\x\linewidth,trim={150 120 150 120},clip]{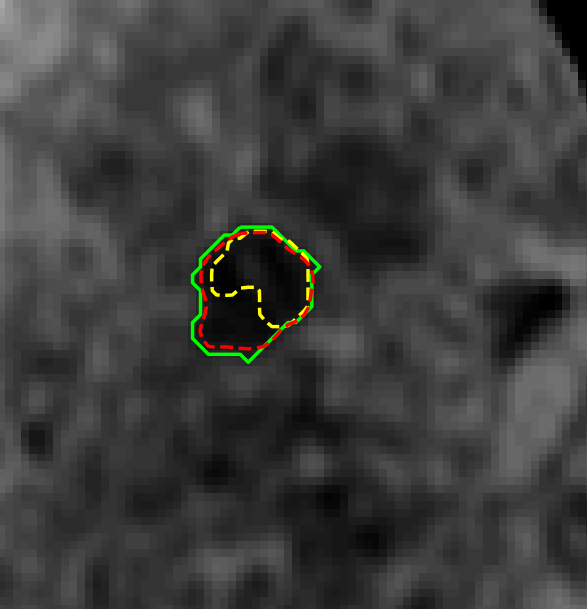} 
\includegraphics[width=\x\linewidth,height=\x\linewidth,trim={150 120 150 120},clip]{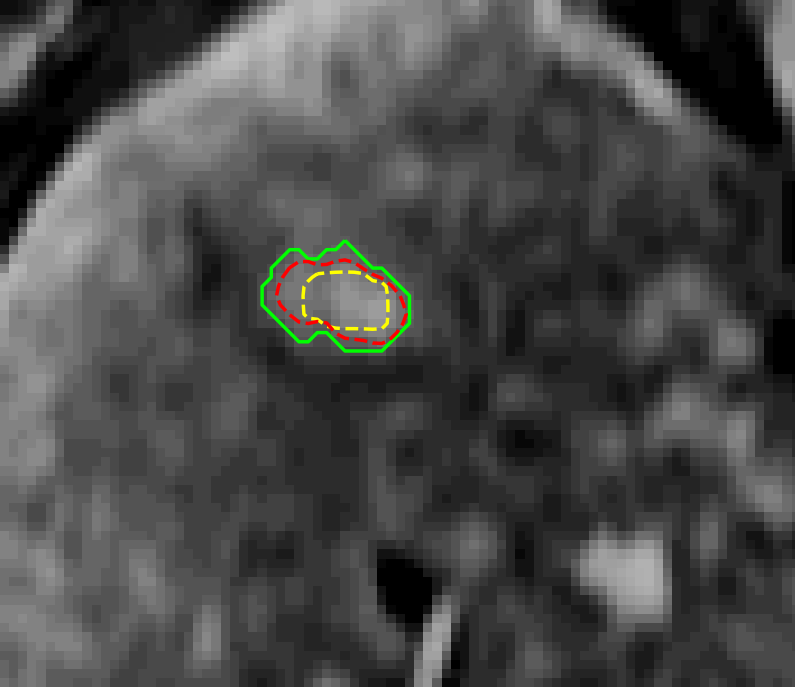}\\[3pt]
\includegraphics[width=\x\linewidth,height=\x\linewidth,trim={150 120 150 120},clip]{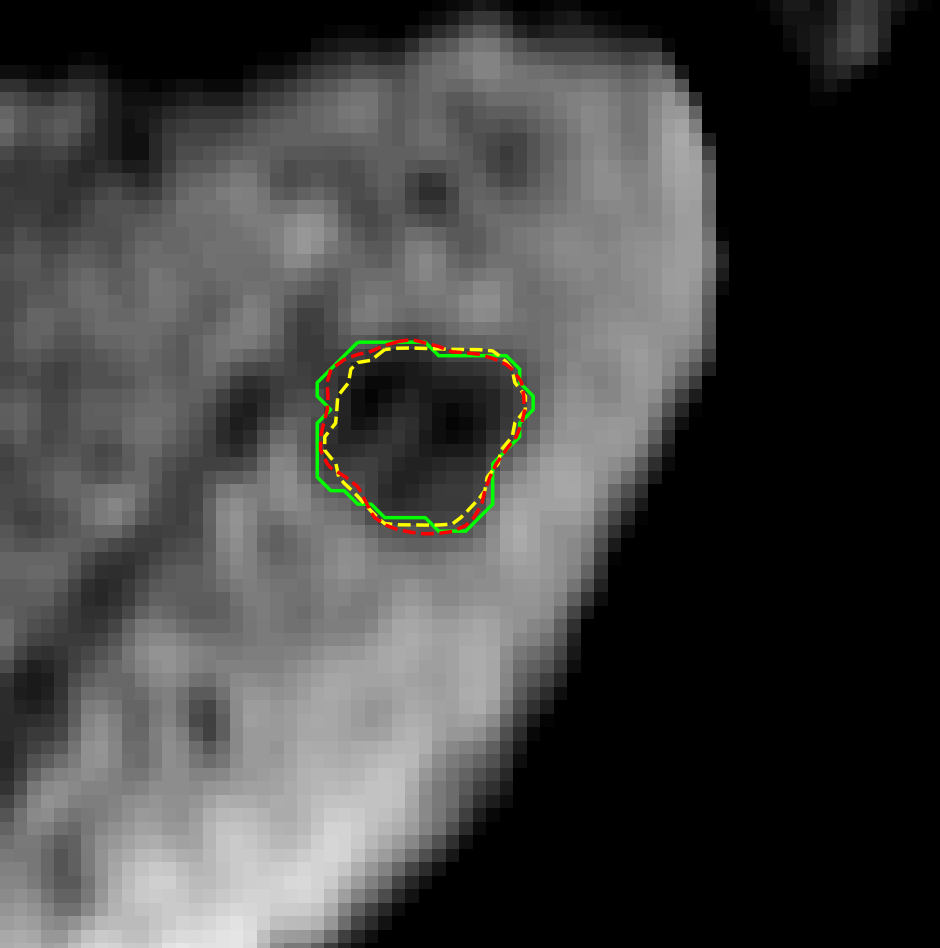} 
\includegraphics[width=\x\linewidth,height=\x\linewidth,trim={150 120 150 120},clip]{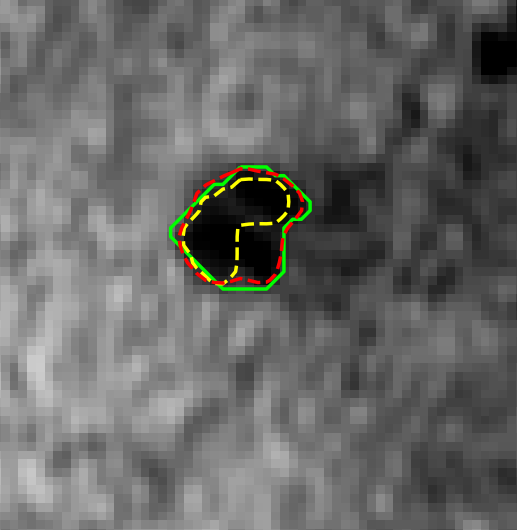} 
\includegraphics[width=\x\linewidth,height=\x\linewidth]{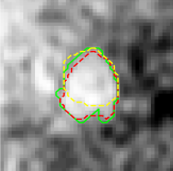} 
\includegraphics[width=\x\linewidth,height=\x\linewidth]{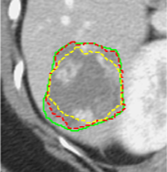} 
\includegraphics[width=\x\linewidth,height=\x\linewidth]{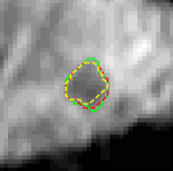}\\[3pt]
\includegraphics[width=\x\linewidth,height=\x\linewidth,trim={150 120 150 120},clip]{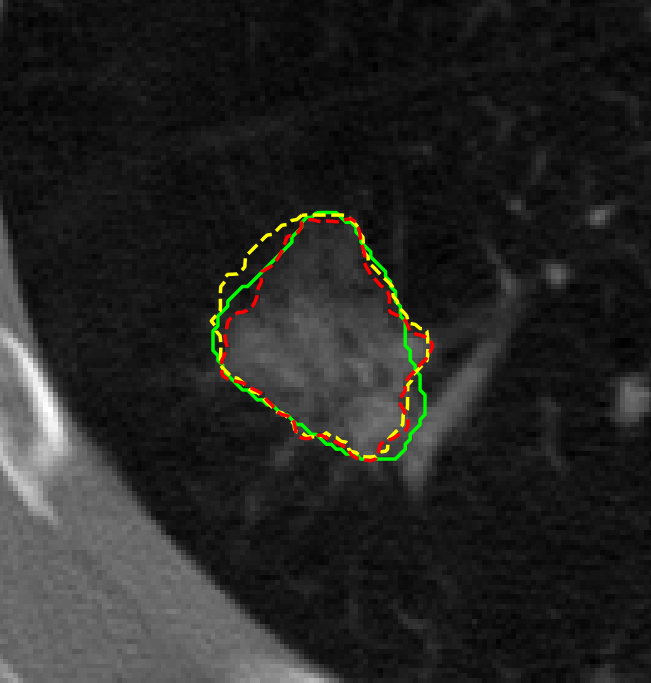} 
\includegraphics[width=\x\linewidth,height=\x\linewidth,trim={150 120 150 120},clip]{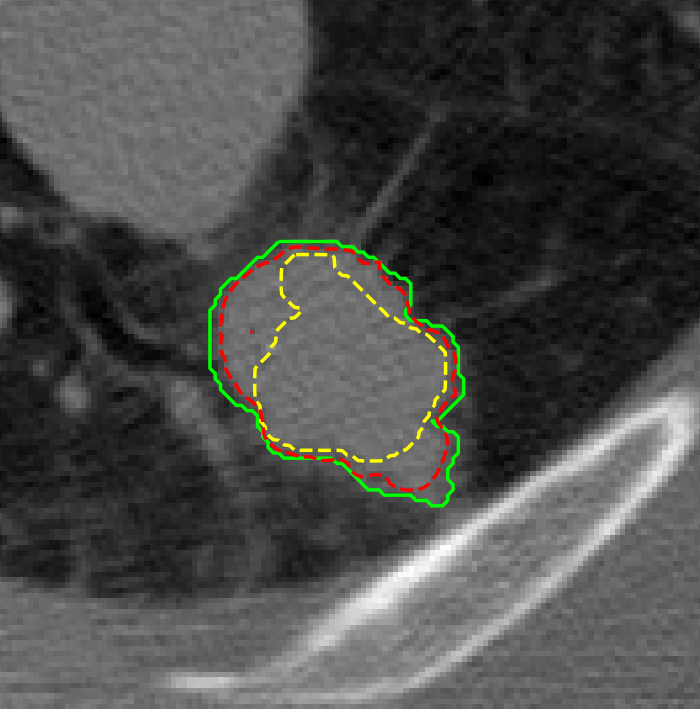} 
\includegraphics[width=\x\linewidth,height=\x\linewidth,trim={150 120 150 120},clip]{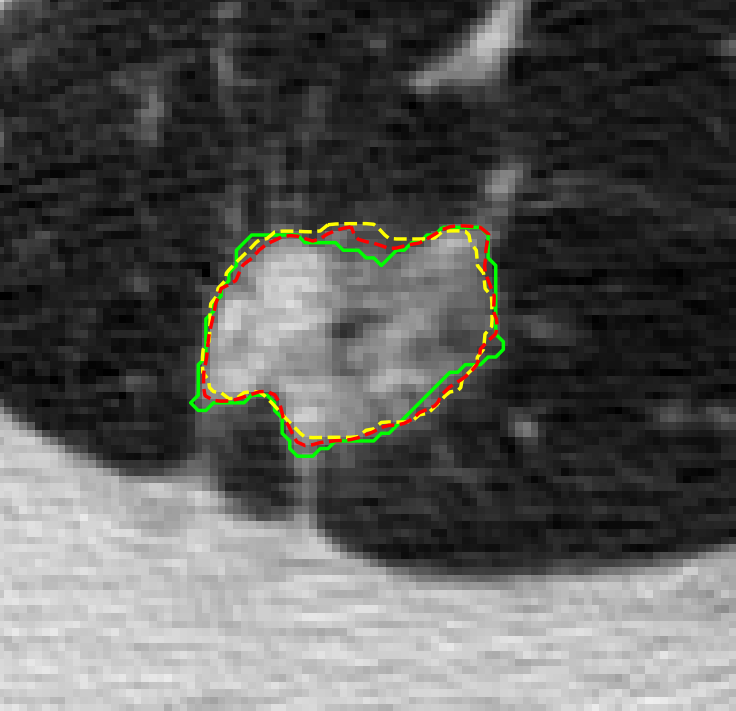} 
\includegraphics[width=\x\linewidth,height=\x\linewidth,trim={150 120 150 120},clip]{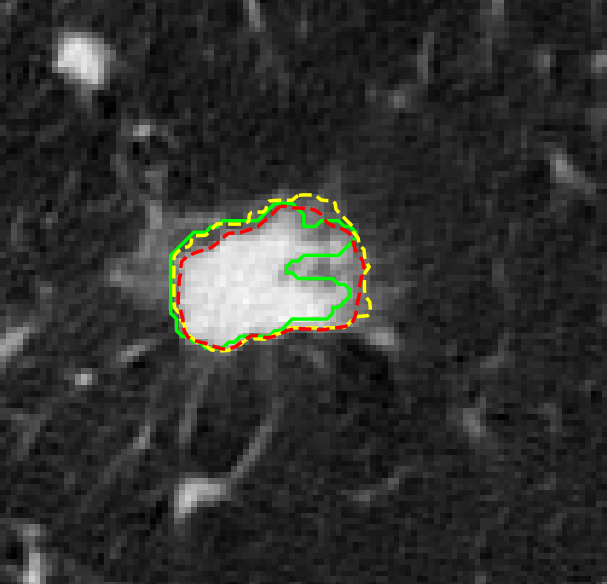} 
\includegraphics[width=\x\linewidth,height=\x\linewidth,trim={150 120 150 120},clip]{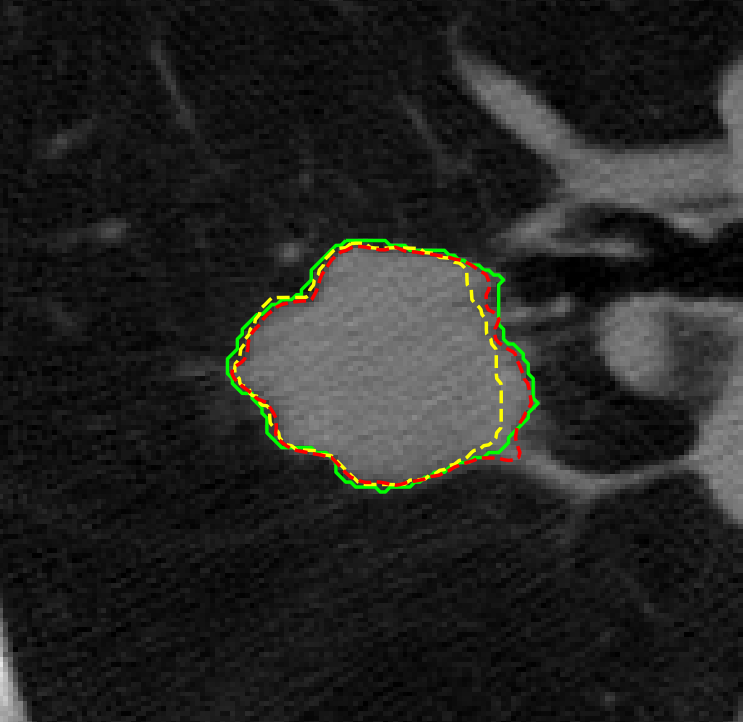}\\[3pt]
\includegraphics[width=\x\linewidth,height=\x\linewidth,trim={150 120 150 120},clip]{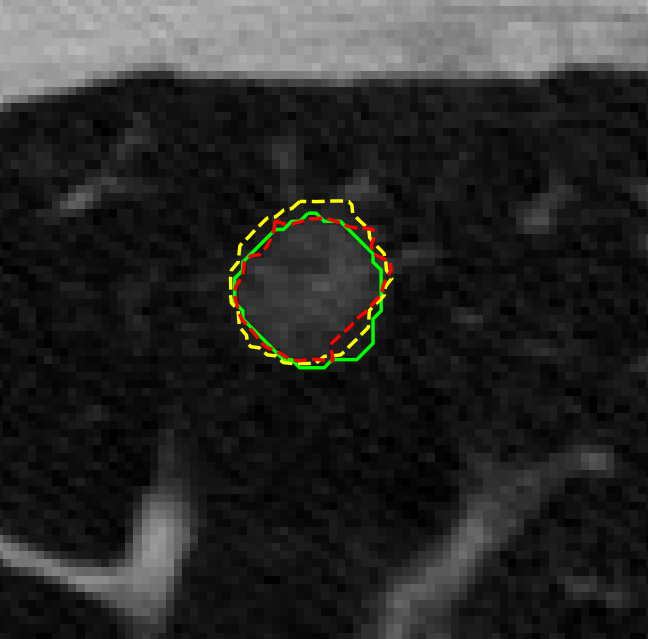} 
\includegraphics[width=\x\linewidth,height=\x\linewidth,trim={150 120 150 120},clip]{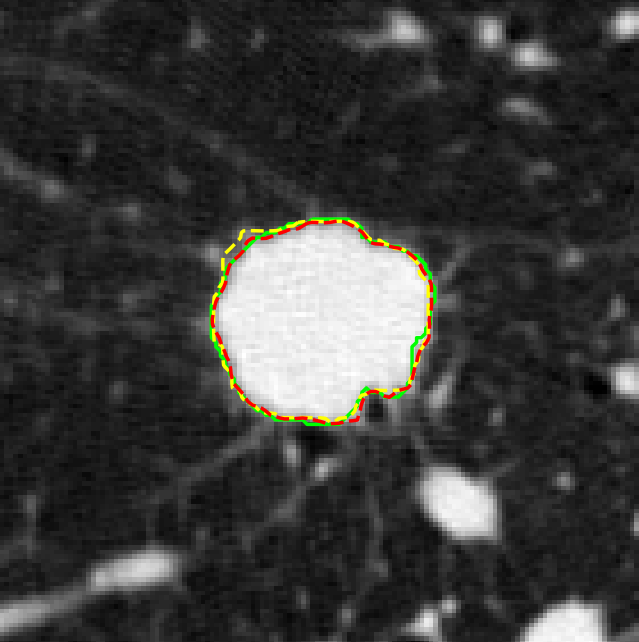} 
\includegraphics[width=\x\linewidth,height=\x\linewidth]{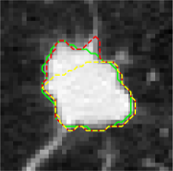} 
\includegraphics[width=\x\linewidth,height=\x\linewidth]{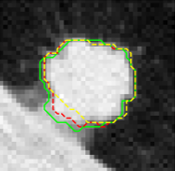} 
\includegraphics[width=\x\linewidth,height=\x\linewidth]{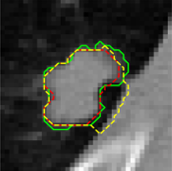}\\
\caption[Qualitative comparison of DALS and competing methods]{Comparison of the output segmentation of our DALS (red) against the U-Net (yellow) and ground truth (green).}
\label{fig:result-comp}
\end{figure}

As shown in Figure~\ref{fig:result-comp}, the DALS segmentation
contours conform appropriately to the irregular shapes of the lesion
boundaries, since the learned parameter maps, $\lambda_1(x,y)$ and
$\lambda_2(x,y)$, provide the flexibility needed to accommodate the
irregularities. In most cases, the DALS has also successfully avoided
local minima and converged onto the true lesion boundaries, thus
enhancing segmentation accuracy. DALS performs well for different
image characteristics, including low contrast lesions, heterogeneous
lesions, and noise.

\begin{figure}
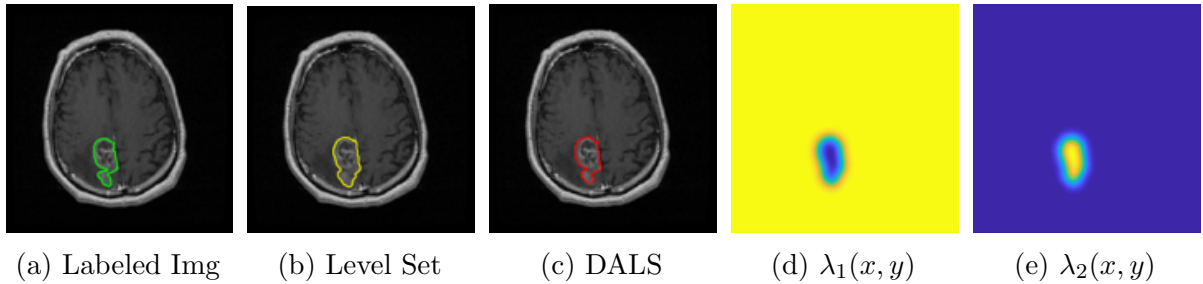

    \centering
    \subcaptionbox{Labeled Img}{\includegraphics[width=0.19\linewidth]{fig5_gt.png}} \hfill
    \subcaptionbox{Level Set}{\includegraphics[width=0.19\linewidth]{fig5_man.png}} \hfill
    \subcaptionbox{DALS}{\includegraphics[width=0.19\linewidth]{fig5_dlac.png}} \hfill
    \subcaptionbox{$\lambda_{1}(x,y)$}{\includegraphics[width=0.19\linewidth]{fig5_l1.png}} \hfill
    \subcaptionbox{$\lambda_{2}(x,y)$}{\includegraphics[width=0.19\linewidth]{fig5_l2.png}}
    \caption[Visualization of estimated parameter maps] {Visualization
    of estimated parameter maps. (a) Labeled image. (b) Level-set
    (analogous to scalar $\lambda$ parameter constants). (c) DALS
    output. (d), (e) Learned parameter maps $\lambda_{1}(x,y)$ and
    $\lambda_{2}(x,y)$.}
    \label{fig:lambda_comp}
\end{figure}

The contribution of the parameter functions was validated by comparing
the DALS against a manually initialized level-set ACM with scalar
parameters constants as well as with DALS's backbone CNN on its own.
As shown in Figure~\ref{fig:lambda_comp}, the encoder-decoder has
predicted the $\lambda_{1}(x,y)$ and $\lambda_{2}(x,y)$ feature maps
to guide the contour evolution. The learned maps serve as an attention
mechanism that provides additional degrees of freedom for the contour
to adjust itself precisely to regions of interest.

The segmentation outputs of our DALS and the manual level-set ACM in
Figure~\ref{fig:lambda_comp} demonstrate the benefits of using
parameter functions to accommodate significant boundary complexities.
Moreover, our DALS outperformed the manually-initialized ACM and its
backbone CNN in all metrics across all evaluations on every organ.

\section{Trainable Deep Active Contours}

In this section, we empirically study the models developed in Section~\ref{sec:dtac}.

\subsection{Datasets}

\subsubsection{Vaihingen}

The Vaihingen buildings dataset consists of 168 building images of
size $512\times512$
pixels.\footnote{\url{http://www2.isprs.org/commissions/comm3/wg4/2d-sem-label-vaihingen.html}}
The labels for each image are generated by using a semi-automated
approach. We used 100 images for training and 68 for testing,
following the same data partition as in \citep{marcos2018learning}. In
this dataset, almost all the images contain multiple instances of
buildings, some of which are located at the image edges.

\subsubsection{Bing Huts}

The Bing Huts dataset consists of 605 images of size $64\times64$
pixels.\footnote{\url{https://www.openstreetmap.org/\#map=4/38.00/-95.80}}
We followed the same data partition used in
\citep{marcos2018learning}, employing 335 images for training and 270
images for testing. This dataset is especially challenging due the low
spatial resolution and contrast of the images.

\subsection{Evaluation Metrics}

To evaluate our model's performance, we utilized four different
metrics---Dice, mean Intersection over Union (mIoU), Boundary F
(BoundF) \citep{cheng2019darnet}, and Weighted Coverage (WCov)
\citep{w_cov}.

The Dice (F1) score of an image given the ground truth mask $G$ and
the prediction $Y$ is
\begin{equation}
\text{Dice}(G,Y) = \frac{2 \sum_{i=1}^{N}G_{i}Y_{i} }{\sum_{i=1}^{N} G_{i}^2 + \sum_{i=1}^{N}Y_{i}^2 + \epsilon},
\label{eq:dice2}
\end{equation}
where $G_{i}$ and $Y_{i}$ denote a pixels in $G$ and $Y$, and $N$ is the number of pixels in the image.

Similarly, the IoU score measures the overlap of two objects by
calculating the ratio of intersection over union according to
\begin{equation}
\text{IoU}(G, Y) = \frac{|G \cap Y|}{|G \cup Y|}.
\label{eq:wcov}
\end{equation} 

BoundF computes the average of Dice scores over 1 to 5 pixels around
the boundaries of the groundtruth.

In WConv, the maximum overlap output is selected and the IoU between
the ground truth and best output is calculated. IoUs for all instances
are summed up and weighted by the area of the ground truth instance.
Assuming that $S_{G} = \{r_{1}^{S_{G}},\dots,r_{|S_{G}|}^{S_{G}}\}$ is
a set of ground truth regions and $S_{Y} =
\{r_{1}^{S_{Y}},\dots,r_{|S_{Y}|}^{S_{Y}}\}$ is a set of prediction
regions for single image, and $|r_{j}^{S_{G}}|$ is the number of
pixels in $r_{j}^{S_{G}}$, the weighted coverage can be expressed as
\begin{equation}
\text{WCov}(S_{G}, S_{Y}) =
\frac{1}{N}\sum_{j=1}^{|S_{G}|}{|r_{j}^{S_{G}}|}
\max\limits_{k=1...|S_{Y}|} \text{IoU}(r_{j}^{S_{G}}, r_{k}^{S_{Y}}).
\label{eq:wcov2}
\end{equation} 

\subsection{Ablation Studies}
\label{sec:ablation}

\subsubsection{Single Instance Segmentation}

\begin{table}[t!] \centering
\setlength{\tabcolsep}{4pt}
\resizebox{\columnwidth}{!}{%
\begin{tabular}{ll cccc cccc}
\toprule
\multicolumn{2}{c}{Method}  & \multicolumn{4}{c}{Vaihingen} & \multicolumn{4}{c}{Bing Huts} \\
   \cmidrule(lr){1-2} \cmidrule(lr){3-6} \cmidrule(lr){7-10}
{Approach}    &   Backbone & Dice  &   mIoU &   WCov  &  BoundF   &   Dice  &   mIoU &   WCov  &  BoundF  \\
\midrule
{FCN}    &  UNet  & 87.40 &	78.60 &	81.80 &	40.20 & 77.20 &	64.90&	75.70&	41.27  \\
{FCN}  & ResNet  & 84.20	& 75.60&	77.50&	38.30 & 79.90 &	68.40 &	76.14&	39.19 \\
{FCN}  & Mask R-CNN  & 86.00	&76.36 & 81.55 &	36.80 & 77.51& 65.03 &	76.02 &	65.25 \\
{FCN}  &  Ours & 79.30	& 66.50 &	68.60 &	68.0 & 80.23 &	66.98 &	77.15 &   40.19     \\
{FCN} &  DSAC  & -- &	81.00 &	81.40 &64.60	 & -- &69.80	&73.60	& 30.30	  \\
{FCN} &  DarNet  & -- &87.20	 &86.80 &	76.80 & -- &74.50	&77.50	& 37.70	  \\
{DSAC } &  DSAC  & -- &71.10	 &70.70	 &	36.40 & -- &38.70	&44.60	& 37.10	  \\
{DSAC } &  DarNet  & -- &60.30	 &61.10	 &	24.30 & -- &57.20	&63.00	& 15.90	  \\
{DarNet } &  DarNet  & 93.66 &88.20	 &88.10	 &	75.90 & 85.21 &75.20	&77.00	& 38.00	  \\
{DTAC, \small Const $\lambda$}  & Ours &  91.18 &	83.79 &  82.70 &	73.21 & 84.53 &	73.02 & 74.21 & 48.25	 \\
{DTAC}  & Ours&  \textbf{94.26}  &	\textbf{89.16}	& \textbf{90.54}   & \textbf{78.12} & \textbf{89.12} &	\textbf{80.39}   & \textbf{81.05}    &	\textbf{53.50}  \\
\bottomrule
\end{tabular}
}
\caption[Single instance quantitative comparison of DTAC and others]
        {Model Evaluations of DTAC and others: Single Instance
        Segmentation.}
\label{tab:datasets-perf}
\end{table}

Although most of the images in the Vaihingen dataset depict multiple
instances of buildings, the DarNet and DSAC models can deal only with
a single building instance at a time. For a fair comparison against
these approaches, we report single instance segmentation results in
the exact same manner as \citep{marcos2018learning} and
\citep{cheng2019darnet}. As reported in Table~\ref{tab:datasets-perf},
our DTAC outperforms both DarNet and DSAC in all metrics on both the
Vaihingen and Bing Huts datasets.

\begin{figure}
\centering
\includegraphics[width=0.135\linewidth,height=0.135\linewidth]{fig4_DTAC/128_gt_v}
\hfill
\includegraphics[width=0.135\linewidth,height=0.135\linewidth]{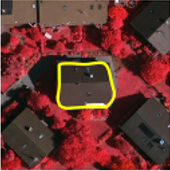}
\hfill
\includegraphics[width=0.135\linewidth,height=0.135\linewidth]{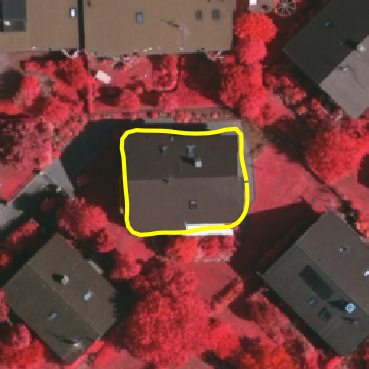}
\hfill
\includegraphics[width=0.135\linewidth,height=0.135\linewidth]{fig4_DTAC/128_dcac_v.jpg}
\hfill
\includegraphics[width=0.135\linewidth,height=0.135\linewidth,trim={80
35 20 20},clip]{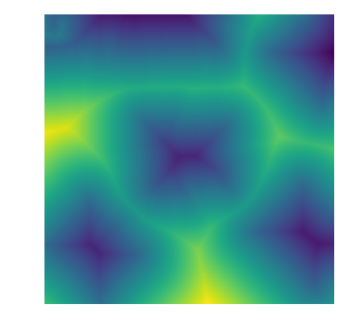} \hfill
\includegraphics[width=0.135\linewidth,height=0.135\linewidth,trim={80
35 20 20},clip]{fig4_DTAC/128_l1_v} \hfill
\includegraphics[width=0.135\linewidth,height=0.135\linewidth,trim={80
35 20 20},clip]{fig4_DTAC/128_l2_v}\\[4pt]
\includegraphics[width=0.135\linewidth,height=0.135\linewidth]{fig4_DTAC/049_gt_v}
\hfill
\includegraphics[width=0.135\linewidth,height=0.135\linewidth]{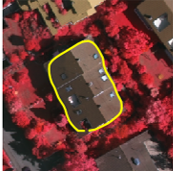}
\hfill
\includegraphics[width=0.135\linewidth,height=0.135\linewidth]{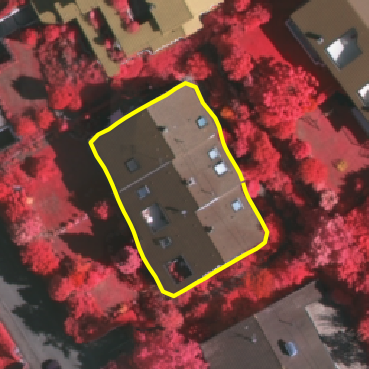}
\hfill
\includegraphics[width=0.135\linewidth,height=0.135\linewidth]{fig4_DTAC/049_dcac_v.jpg}
\hfill
\includegraphics[width=0.135\linewidth,height=0.135\linewidth,trim={80
35 20 20},clip]{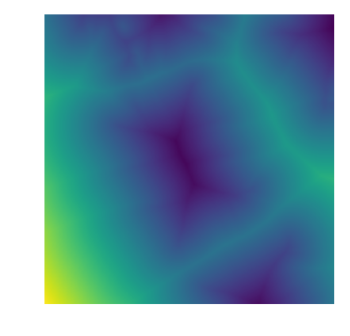} \hfill
\includegraphics[width=0.135\linewidth,height=0.135\linewidth,trim={80
35 20 20},clip]{fig4_DTAC/049_l1_v} \hfill
\includegraphics[width=0.135\linewidth,height=0.135\linewidth,trim={80
35 20 20},clip]{fig4_DTAC/049_l2_v}\\[4pt]
\includegraphics[width=0.135\linewidth,height=0.135\linewidth]{fig4_DTAC/163_gt_v}
\hfill
\includegraphics[width=0.135\linewidth,height=0.135\linewidth]{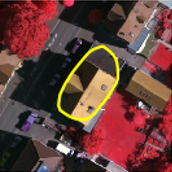}
\hfill
\includegraphics[width=0.135\linewidth,height=0.135\linewidth]{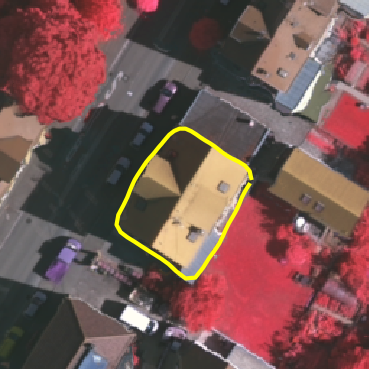}
\hfill
\includegraphics[width=0.135\linewidth,height=0.135\linewidth]{fig4_DTAC/163_dcac_v.jpg}
\hfill
\includegraphics[width=0.135\linewidth,height=0.135\linewidth,trim={80
35 20 20},clip]{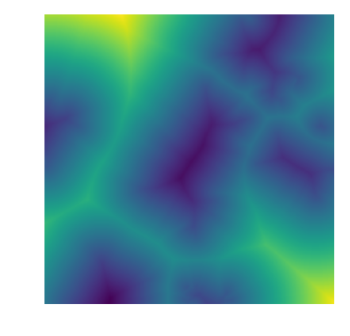} \hfill
\includegraphics[width=0.135\linewidth,height=0.135\linewidth,trim={80
35 20 20},clip]{fig4_DTAC/163_l1_v} \hfill
\includegraphics[width=0.135\linewidth,height=0.135\linewidth,trim={80
35 20 20},clip]{fig4_DTAC/163_l2_v}\\[4pt]
\includegraphics[width=0.135\linewidth,height=0.135\linewidth]{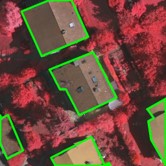}
\hfill
\includegraphics[width=0.135\linewidth,height=0.135\linewidth]{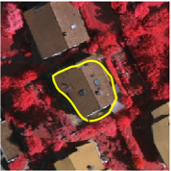}
\hfill
\includegraphics[width=0.135\linewidth,height=0.135\linewidth]{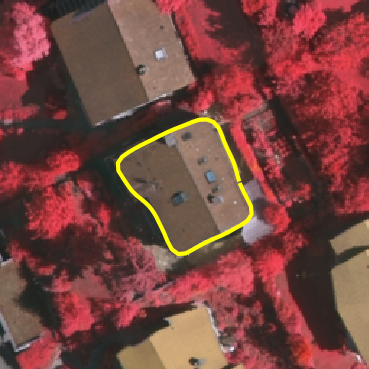}
\hfill
\includegraphics[width=0.135\linewidth,height=0.135\linewidth]{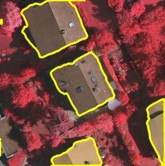}
\hfill
\includegraphics[width=0.135\linewidth,height=0.135\linewidth,trim={80
35 20 20},clip]{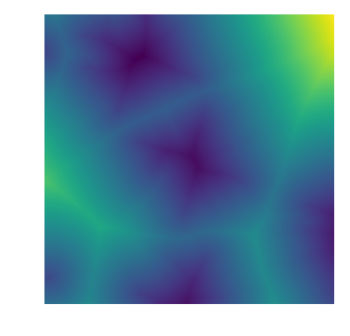} \hfill
\includegraphics[width=0.135\linewidth,height=0.135\linewidth,trim={80
35 20 20},clip]{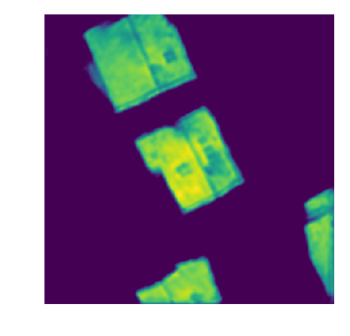} \hfill
\includegraphics[width=0.135\linewidth,height=0.135\linewidth,trim={80
35 20 20},clip]{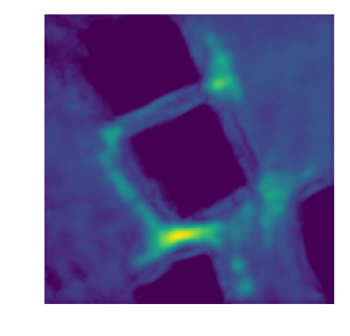}\\[4pt]
\hfill
\includegraphics[width=0.135\linewidth,height=0.135\linewidth]{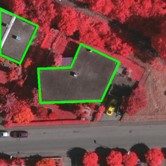}
\hfill
\includegraphics[width=0.135\linewidth,height=0.135\linewidth]{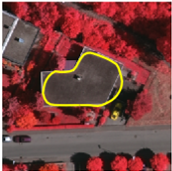}
\hfill
\includegraphics[width=0.135\linewidth,height=0.135\linewidth]{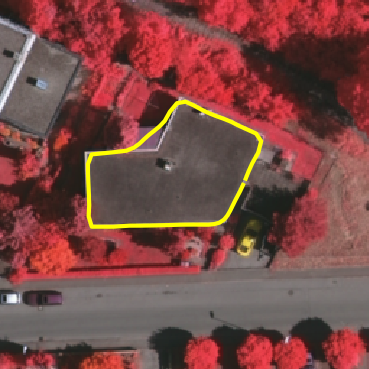}
\hfill
\includegraphics[width=0.135\linewidth,height=0.135\linewidth]{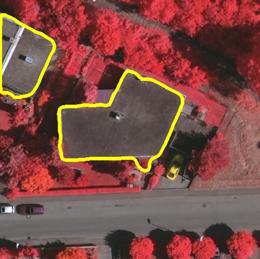}
\hfill
\includegraphics[width=0.135\linewidth,height=0.135\linewidth]{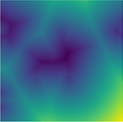} \hfill
\includegraphics[width=0.135\linewidth,height=0.135\linewidth]{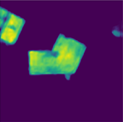} \hfill
\includegraphics[width=0.135\linewidth,height=0.135\linewidth]{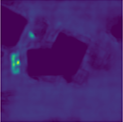}\\[4pt]
\includegraphics[width=0.135\linewidth,height=0.135\linewidth]{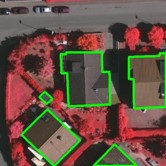}
\hfill
\includegraphics[width=0.135\linewidth,height=0.135\linewidth]{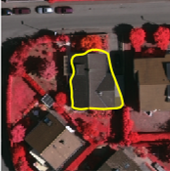}
\hfill
\includegraphics[width=0.135\linewidth,height=0.135\linewidth]{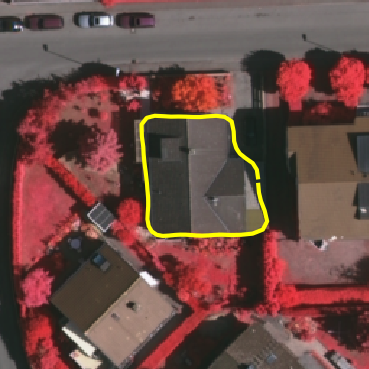}
\hfill
\includegraphics[width=0.135\linewidth,height=0.135\linewidth]{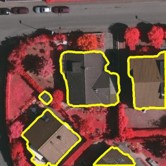}
\hfill
\includegraphics[width=0.135\linewidth,height=0.135\linewidth]{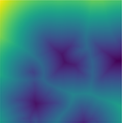} \hfill
\includegraphics[width=0.135\linewidth,height=0.135\linewidth]{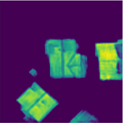} \hfill
\includegraphics[width=0.135\linewidth,height=0.135\linewidth]{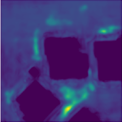}\\[4pt]
\includegraphics[width=0.135\linewidth,height=0.135\linewidth]{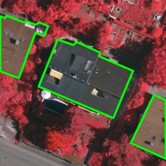}
\hfill
\includegraphics[width=0.135\linewidth,height=0.135\linewidth]{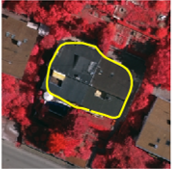}
\hfill
\includegraphics[width=0.135\linewidth,height=0.135\linewidth]{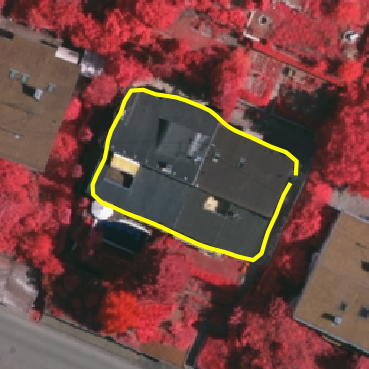}
\hfill
\includegraphics[width=0.135\linewidth,height=0.135\linewidth]{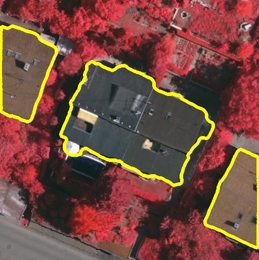}
\hfill
\includegraphics[width=0.135\linewidth,height=0.135\linewidth]{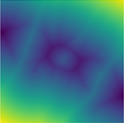} \hfill
\includegraphics[width=0.135\linewidth,height=0.135\linewidth]{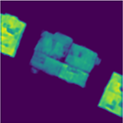} \hfill
\includegraphics[width=0.135\linewidth,height=0.135\linewidth]{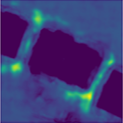}\\[4pt]
\begin{subfigure}{\linewidth}
  \centering
  \subcaptionbox{\centering Labeled Image}{\includegraphics[width=0.135\linewidth,height=0.135\linewidth]{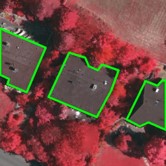}}
\hfill
\subcaptionbox{\centering DSAC}{\includegraphics[width=0.135\linewidth,height=0.135\linewidth]{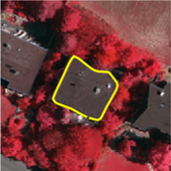}}
\hfill
\subcaptionbox{\centering DarNet}{\includegraphics[width=0.135\linewidth,height=0.135\linewidth]{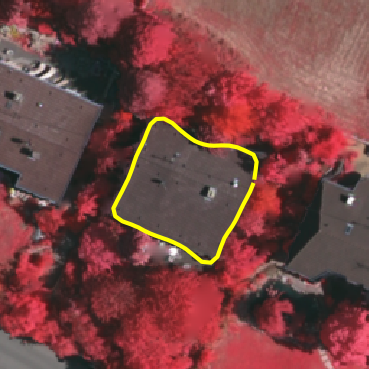}}
\hfill
\subcaptionbox{\centering Our DTAC}{\includegraphics[width=0.135\linewidth,height=0.135\linewidth]{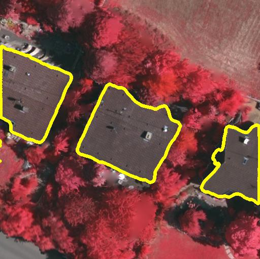}}
\hfill
\subcaptionbox{\centering Initialization}{\includegraphics[width=0.135\linewidth,height=0.135\linewidth]{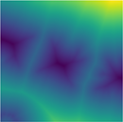}}
\hfill
\subcaptionbox{\centering $\lambda_1(x,y)$}{\includegraphics[width=0.135\linewidth,height=0.135\linewidth]{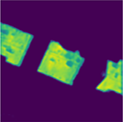}}
\hfill
\subcaptionbox{\centering $\lambda_2(x,y)$}{\includegraphics[width=0.135\linewidth,height=0.135\linewidth]{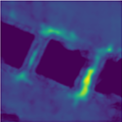}}\\[4pt]
\end{subfigure}
\caption[Qualitative comparison of DTAC and others on the
Vaihingen dataset] {Comparative visualization of the labeled image,
the output of DSAC, the output of DarNet, and the output of our DTAC,
for the Vaihingen dataset: (a) Image with label (green), (b) DSAC
output, (c) DarNet output, (d) our DTAC output, (e) DTAC learned
initialization map, (f) $\lambda_{1}(x,y)$ and (g) $\lambda_{2}(x,y)$
for the DTAC.}
\label{fig:final_comparison}
\end{figure}

\begin{figure}
\centering
\includegraphics[width=0.135\linewidth,height=0.135\linewidth]{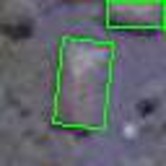}
\hfill
\includegraphics[width=0.135\linewidth,height=0.135\linewidth]{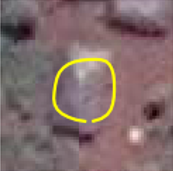}
\hfill
\includegraphics[width=0.135\linewidth,height=0.135\linewidth]{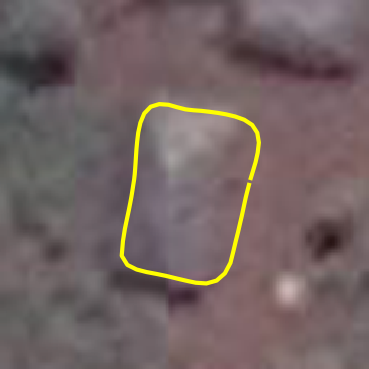}
\hfill
\includegraphics[width=0.135\linewidth,height=0.135\linewidth]{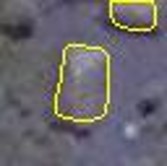}
\hfill
\includegraphics[width=0.135\linewidth,height=0.135\linewidth]{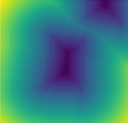}
\hfill
\includegraphics[width=0.135\linewidth,height=0.135\linewidth]{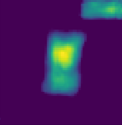}
\hfill
\includegraphics[width=0.135\linewidth,height=0.135\linewidth]{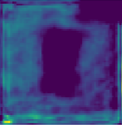}\\[4pt]
\includegraphics[width=0.135\linewidth,height=0.135\linewidth]{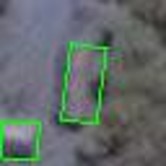}
\hfill
\includegraphics[width=0.135\linewidth,height=0.135\linewidth]{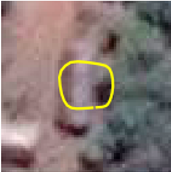}
\hfill
\includegraphics[width=0.135\linewidth,height=0.135\linewidth]{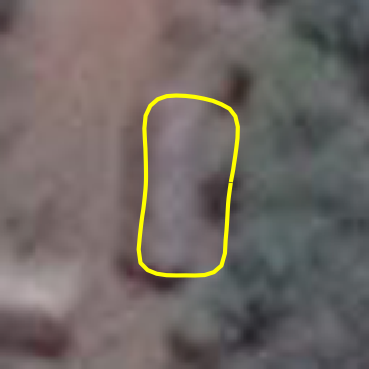}
\hfill
\includegraphics[width=0.135\linewidth,height=0.135\linewidth]{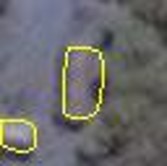}
\hfill
\includegraphics[width=0.135\linewidth,height=0.135\linewidth]{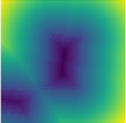}
\hfill
\includegraphics[width=0.135\linewidth,height=0.135\linewidth]{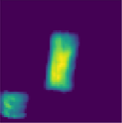}
\hfill
\includegraphics[width=0.135\linewidth,height=0.135\linewidth]{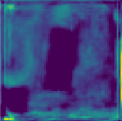}\\[4pt]
\begin{subfigure}{\linewidth}
  \centering
  \subcaptionbox{\centering Labeled Image}{\includegraphics[width=0.135\linewidth,height=0.135\linewidth]{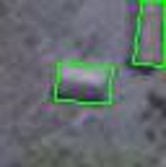}}
\hfill
\subcaptionbox{\centering DSAC}{\includegraphics[width=0.135\linewidth,height=0.135\linewidth]{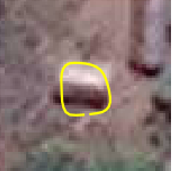}}
\hfill
\subcaptionbox{\centering DarNet}{\includegraphics[width=0.135\linewidth,height=0.135\linewidth]{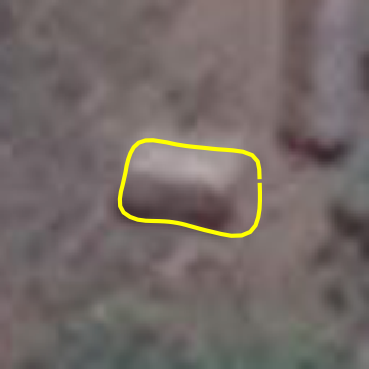}}
\hfill
\subcaptionbox{\centering Our DTAC}{\includegraphics[width=0.135\linewidth,height=0.135\linewidth]{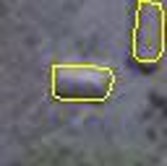}}
\hfill
\subcaptionbox{\centering Initialization}{\includegraphics[width=0.135\linewidth,height=0.135\linewidth]{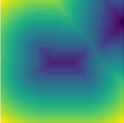}}
\hfill
\subcaptionbox{\centering $\lambda_1(x,y)$}{\includegraphics[width=0.135\linewidth,height=0.135\linewidth]{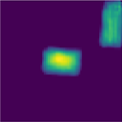}}
\hfill
\subcaptionbox{\centering $\lambda_2(x,y)$}{\includegraphics[width=0.135\linewidth,height=0.135\linewidth]{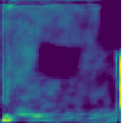}}\\[4pt]
\end{subfigure}
\caption[Qualitative comparison of DTAC and others on the
Bing dataset] {Comparative visualization of the labeled image, the
output of DSAC, the output of DarNet, and the output of our DTAC, for
the Bing Huts dataset: (a) Image with label (green), (b) DSAC output,
(c) DarNet output, (d) our DTAC output, (e) DTAC learned
initialization map, (f) $\lambda_{1}(x,y)$ and (g) $\lambda_{2}(x,y)$ for the
DTAC.}
\label{fig:final_comparison2}
\end{figure}

As shown in Figures~\ref{fig:final_comparison}, with the Vaihingen
dataset, both the DarNet and DSAC models struggle to cope with the
topological changes of the buildings and fail to appropriately capture
sharp edges, while our framework readily handles these challenges in
most cases. With the Bing Huts dataset, as shown in Figure~\ref{fig:final_comparison2}, both the DarNet and DSAC models
are able to localize the buildings, but they mainly over-segment the
buildings in many cases. This may be due to their inability to
distinguish the building from the surrounding soil because of the low
contrast and small size of the image. Comparing the segmentation
output of DSAC (Figure~\ref{fig:final_comparison2}b), DarNet
(Figure~\ref{fig:final_comparison2}c), and DTAC
(Figure~\ref{fig:final_comparison2}d), our DTAC model performs well in
a low contrast dataset, producing more accurate boundaries than the
earlier models.

\begin{table} \centering
\setlength{\tabcolsep}{4pt}
\resizebox{\columnwidth}{!}{
\begin{tabular}{ll cccc cccc}
\toprule
\multicolumn{2}{c}{Method}  & \multicolumn{4}{c}{Vaihingen} & \multicolumn{4}{c}{Bing Huts} \\
\cmidrule(lr){1-2} \cmidrule(lr){3-6} \cmidrule(lr){7-10}
Approach    &   Backbone & Dice  &   mIoU &   WCov  &  BoundF   &   Dice  &   mIoU &   WCov  &  BoundF  \\
\midrule
{FCN}     & UNet  &81.00&	69.10&	72.40&	34.20& 71.58&	58.70&	65.70&	40.60  \\ 
{FCN}     & ResNet  & 80.10 & 67.80   &	70.50&	32.50 & 74.20 &	61.80&	66.59&	39.48 \\ 
{FCN}     & Mask R-CNN  & 82.00 & 72.20   &	73.50 &	29.80 & 76.12 & 63.40 & 0.7051 & 0.7041 \\ 
{FCN}     & Ours & 89.30	& 81.00 &	82.70 &	49.80 & 75.23 &	60.31 & 72.41	 & 41.12	  \\
{DTAC, \small Const $\lambda$}  & Ours&  90.80	&83.30	&83.90	&47.20	  & 81.19 & 68.34 & 75.29 & 44.61 \\
{DTAC} & Ours&   \textbf{95.20} & \textbf{91.10} &	\textbf{91.71} &	\textbf{69.02} &  \textbf{83.24} &	\textbf{71.30}   & \textbf{78.45}    &	\textbf{48.49} \\
\bottomrule
\end{tabular}
}
\caption[Multiple instance quantitative comparison of DTAC and others]
        {Model Evaluations of DTAC and others: Multiple Instance
        Segmentation.}
\label{tab:datasets-perf2}
\end{table}

\subsubsection{Multiple Instance Segmentation}

We now compare the performance of DTAC against popular models such as
Mask R-CNN for multiple instance segmentation of all buildings in the
Vaihingen and Bing Huts datasets. Our extensive benchmarks confirm
that our DTAC model comfortably outperforms Mask R-CNN and other
method by a wide margin as reported in Table~\ref{tab:datasets-perf2}.
Although Mask R-CNN seems to be able to fairly localize the building
instances, the fine-grained details of boundaries are lost, as is
attested by the BoundF metric. The performance of other CNN-based
approaches follow the same trend in our benchmarks.

\subsubsection{Local and Fixed Weighted Parameters}

\begin{figure}
\subcaptionbox{\centering Labeled Image}{\includegraphics[width=0.19\linewidth,height=0.19\linewidth]{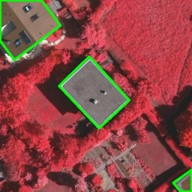}}\hfill
\subcaptionbox{\centering DTAC, constant
$\lambda$s}{\includegraphics[width=0.19\linewidth,height=0.19\linewidth]{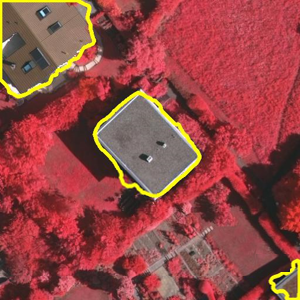}}\hfill
\subcaptionbox{\centering DTAC}{\includegraphics[width=0.19\linewidth,height=0.19\linewidth]{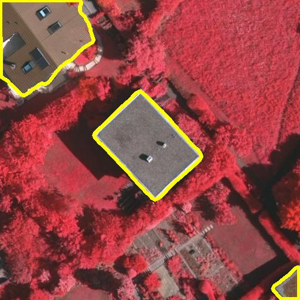}}\hfill
\subcaptionbox{\centering $\lambda_1(x,y)$}{\includegraphics[width=0.19\linewidth,height=0.19\linewidth,trim={80
35 20 20},clip]{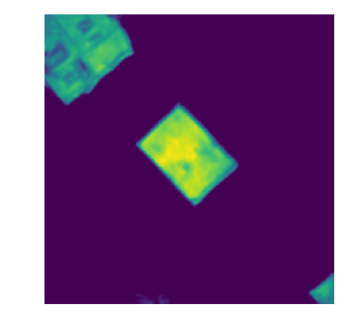}}\hfill
\subcaptionbox{\centering $\lambda_2(x,y)$}{\includegraphics[width=0.19\linewidth,height=0.19\linewidth,trim={80
35 20 20},clip]{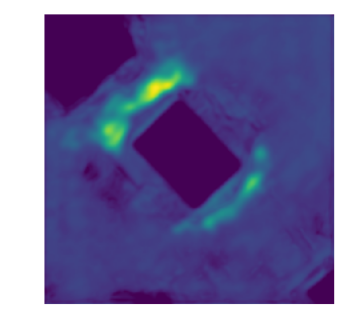}}
\caption[Visualization of DTAC's learned feature maps] {Visualization
of DTAC's learned feature maps. (a) Labeled image. (b) DTAC output with
constant weighted parameters. (c) DTAC output. (d),(e) Learned parameter
maps $\lambda_{1}(x,y)$ and $\lambda_{2}(x,y)$.}
\label{fig:const-lambda}
\end{figure}

To validate the contribution of the local weighted parameters in the
level-set ACM, we also trained our DTAC on both the Vaihingen and Bing
Huts datasets by allowing just a single trainable scalar parameter,
constant over the entire image, for both $\lambda_1$ and $\lambda_2$.
As presented in Table~\ref{tab:datasets-perf}, for both the Vaihingen
and Bing Huts datasets, this ``constant-$\lambda$'' formulation (the
Chan-Vese formulation \citep{chan2001active,lankton2008localizing})
still outperforms the baseline CNN in most evaluation metrics for both
single-instance and multiple-instance buildings, thus establishing the
effectiveness of the end-to-end training of our DTAC. Nevertheless,
the DTAC with its full $\lambda_1(x,y)$ and $\lambda_2(x,y)$ maps
outperforms this constant-$\lambda$ version by a wide margin in all
experiments and metrics. A key metric of interest in this comparison
is the BoundF score, which elucidates that our local formulation
captures the details of the boundaries more effectively by locally
adjusting the inward and outward forces on the contour.
Figure~\ref{fig:const-lambda} shows that our DTAC has perfectly
delineated the boundaries of the building instances. However, the DTAC
hobbled by the constant-$\lambda$ formulation has over-segmented these
instances.

\subsection{Number of Iterations}

\begin{figure}
\centering
\subcaptionbox{}{\includegraphics[width=0.42\linewidth,height=0.32\linewidth,trim={0 0 0 40},clip]{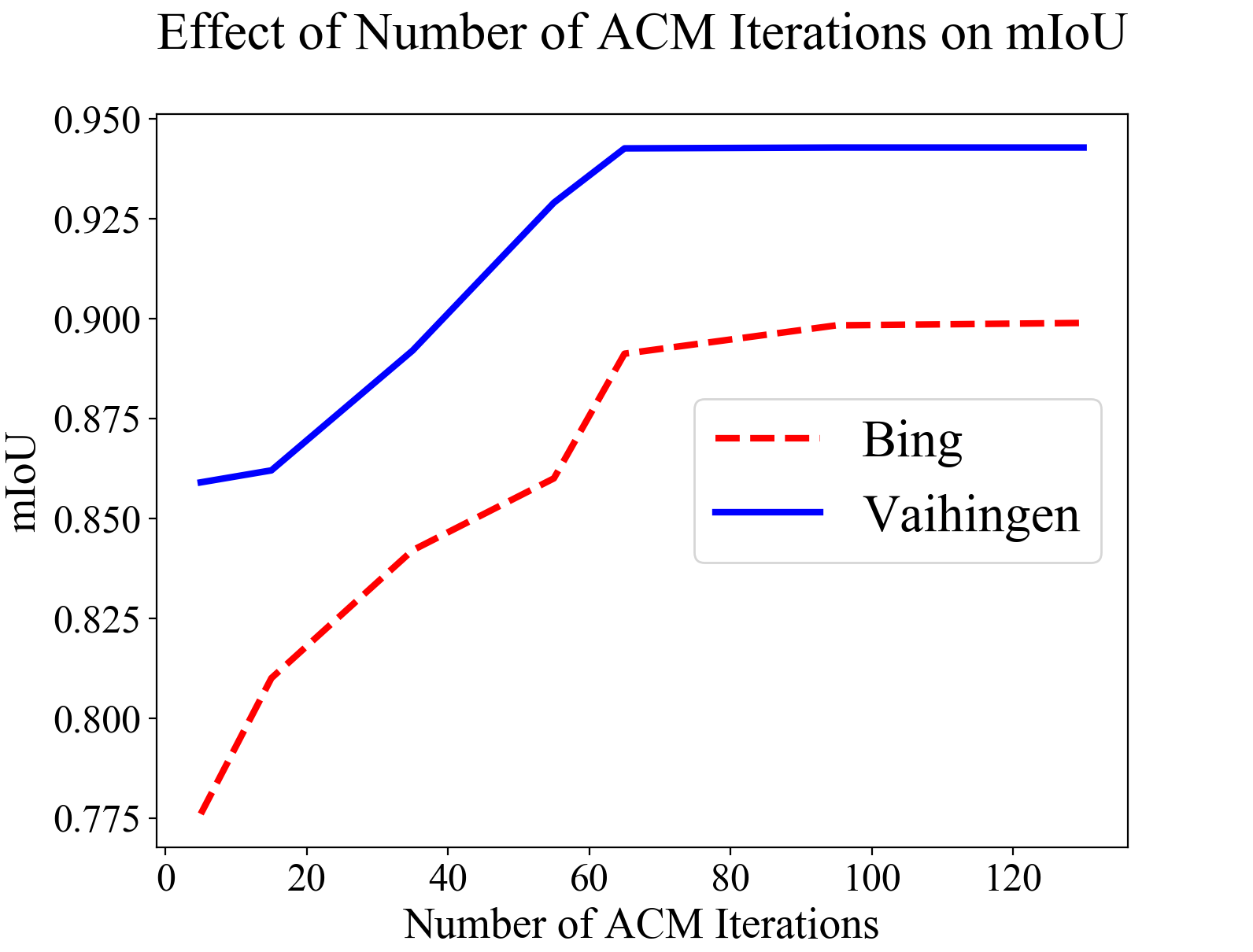}} 
\qquad
\subcaptionbox{}{\includegraphics[width=0.42\linewidth,height=0.32\linewidth,trim={0 0 0 40},clip]{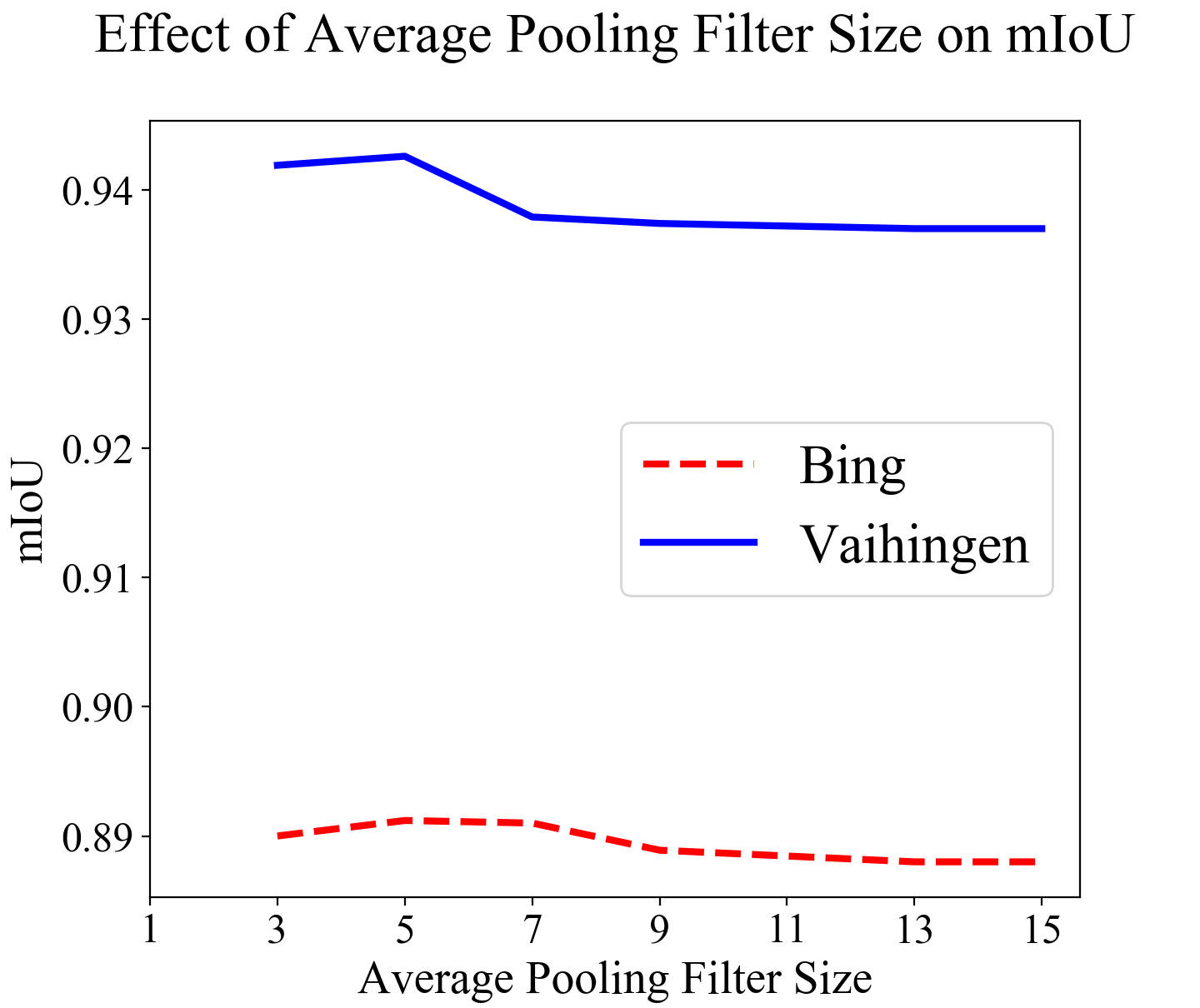}}
\caption[Effects of the number of ACM iterations and filter size
on overall accuracy] {The effects of (a) varying the number of ACM
iterations on mIoU and (b) varying the average pooling filter size on
mIoU.}
\label{fig:exp_plots}
\end{figure}

The direct learning of an initialization map as well as its efficient
end-to-end implementation have enabled the DTAC to require a
significantly lower number of iterations to converge with a better
chance of avoiding the undesirable local minima. As illustrated in
Figure~\ref{fig:exp_plots}a, we have extensively investigated the
effect of the number of iterations on the overall mIoU for both
Vaihingen and Bing datasets, and our results show that DTAC exhibits a
robust performance after a certain threshold. Therefore, we have
chosen a fixed number $N=60$ iterations for optimal performance, which
runs in less than one second in TensorFlow.

\subsection{Average Pooling Filter Size}

The average pooling filter size is an important hyper-parameter in the
extraction of localized image statistics. As illustrated in
Figure~\ref{fig:exp_plots}b, we have investigated the effect of the
average pooling filter size on the overall mIoU for both Vaihingen and
Bing datasets. Our experiments indicate that filter values that are
too small are sub-optimal while excessively large values defeat the
benefits of the localized formulation. Consequently, we set a filter
size of $f=5$ for the DTAC.

\section{Few-Shot Semantic Segmentation}

In this section, we empirically study the models developed in Chapter~\ref{cha:few}.

\subsection{Datasets} 


To evaluate SegAVA, we used the PASCAL VOC 2012
\citep{pascal-voc-2007} dataset, which consists of 20 categories that
are divided equally into 4 partitions, with 5 categories in each
partition. We trained our model on 3 partitions and evaluated on the
remaining partition with cross-validation.


\subsection{Evaluation Metrics} 

To evaluate SegAVA's performance, we utilized the mean Intersection
over Union (mIoU) metric to measure the intersection over the union
for each foreground class and took the average result over all the
classes. The IoU score measures the overlap of two objects by
calculating the ratio of intersection over union according to
\begin{equation}
\text{IoU}(G, Y) = \frac{|G \cap Y|}{|G \cup Y|},
\label{eq:wcov3}
\end{equation} 
where $G$ denotes the ground truth mask and $Y$ denotes the prediction
mask.


\subsection{Implementation Details}
\label{sec:impl}

We have implemented SegAVA in
Pytorch.\footnote{\href{https://pytorch.org/}{https://pytorch.org/}}
Like \citep{wang2019panet}, we used a VGG16 network
\citep{simonyan2014very} pretrained on ImageNet
\citep{russakovsky2015imagenet} as our feature extractor in the image
space, and resized the input images to $417\times417$ pixels and
randomly augmented them using horizontal flipping. All the training
and testing was performed on an Nvidia Titan RTX GPU, and an Intel®
Core™ i7-7700K CPU @ 4.20GHz.

We trained our model for 150,000 iterations with a batch size of 2 and
a stochastic gradient descent algorithm with an initial learning rate
of 0.001 and a momentum value of 0.09 and weight decay of 0.0005. The
learning rate was reduced by a factor of 10 in every 10,000
iterations. To determine the value of hyper-parameter $\gamma$ in
(\ref{eq:PANet_6}), we performed a grid search from minimum to maximum
values of $0.5$ and $4$ for $\gamma$, stepping by $0.5$, and our
experiments confirmed that using $\gamma=0.5$ provides the optimal
balance between the two loss terms.

\subsection{Evaluation}

\begin{table} \centering
\setlength{\tabcolsep}{4pt}
\resizebox{\linewidth}{!}{%
\begin{tabular}{l ccccc ccccc r}
\toprule
\multirow{2}{*}{Method} & \multicolumn{5}{c}{1-shot} & \multicolumn{5}{c}{5-shot} &  \\
\cmidrule(lr){2-6} \cmidrule(lr){7-11}
& part-1 & part-2 & part-3 & part-4 & Mean & part-1 & part-2 & part-3 & part-4 & Mean & $\Delta$ \\ 
\midrule
OSLSM \citep{shaban2017one} &  33.6 &55.3 &40.9& 33.5& 40.8& 35.9& 58.1& 42.7& 39.1& 43.9& 3.1    \\
co-FCN \citep{rakelly2018conditional}&  36.7& 50.6& 44.9& 32.4& 41.1& 37.5& 50.0& 44.1& 33.9& 41.4& 0.3    \\
SG-One \citep{zhang1810sg}&  40.2 & 58.4 &48.4& 38.4& 46.3& 41.9& 58.6& 48.6& 39.4& 47.1 &0.8    \\
AMP \citep{AMP} &  41.9 & 50.2 & 46.7 & 34.7 & 43.4& 41.8 &55.5& 50.3& 39.9& 46.9 &3.5     \\
Meta-Seg \citep{meta_seg} & 42.2& 59.6& 48.1 &44.4& 48.6& 43.1 &62.5& 49.9& 45.3 &50.2& 1.6  \\
MDL \citep{multi_scale} &  39.7 &58.3& 46.7 &36.3 &45.3 &40.6& 58.5& 47.7& 36.6& 45.9& 0.6   \\
PANet-init \citep{wang2019panet} & 30.8 &40.7& 38.3& 31.4& 35.3& 41.6& 52.7& 51.68& 40.8& 46.7& 11.4  \\
$\text{OS}_\text{Adv}$ \citep{YANG2020225} &  46.9 &59.2& 49.3& 43.4& 49.7& 47.2& 58.8& 48.8& \textbf{47.4}& 50.6& 0.9   \\
Feat Weight \citepalias{feature_weights}
&  \textbf{47.0}& 59.6 &52.6 &\textbf{48.3}& \textbf{51.9} & 50.9 &62.9& 56.5 &50.1 &55.1& 3.2     \\
PANet \citep{wang2019panet} & 42.3 &58.0& 51.1& 41.2& 48.1& 51.8& 64.6& 59.8& 46.5& 55.7& 7.6  \\
\textbf{SegAVA} &  44.1& \textbf{59.8}& \textbf{52.9}& 45.6& 50.6& \textbf{51.9}&\textbf{65.1} &\textbf{60.2} &47.2 & \textbf{56.1}&5.5    \\
\bottomrule
\end{tabular}
}
\caption[Quantitative comparison of SegAVA and others on the PASCAL-5i
dataset] {Results from SegAVA for 1-way, 1-shot and 1-way, 5-shot
segmentation on the PASCAL-5i dataset using mean IoU as the measure of
accuracy. $\Delta$ represents the difference between the 1-shot and
5-shot means.}
\label{PANet_table}
\end{table}

\subsubsection{1-Way, 1-Shot and 5-Shot Segmentation}

\begin{figure}
\begin{subfigure}{\linewidth}
\centering
\includegraphics[width=\linewidth]{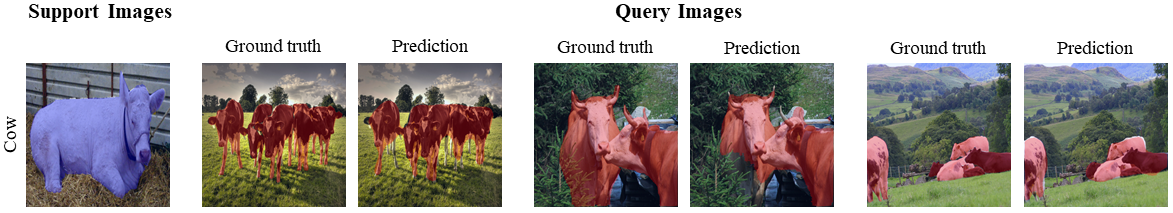}  
\label{fig:sub-first}
\end{subfigure}
\begin{subfigure}{\linewidth}
\centering
\includegraphics[width=\linewidth]{segava/fig3_2.png}  
\label{fig:sub-first1}
\end{subfigure}
\begin{subfigure}{\linewidth}
\centering
\includegraphics[width=\linewidth]{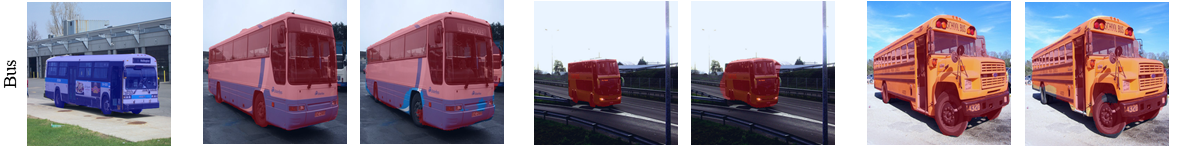}  
\label{fig:sub-first2}
\end{subfigure}
\begin{subfigure}{\linewidth}
\centering
\includegraphics[width=\linewidth]{segava/fig3_4.png}  
\label{fig:sub-first3}
\end{subfigure}
\begin{subfigure}{\linewidth}
\centering
\includegraphics[width=\linewidth]{segava/neurips_fig3_last.png}  
\label{fig:sub-first4}
\end{subfigure}
\caption[SegAVA's predictions in 1-way, 1-shot task on the PASCAL-5i
dataset] {Example results from evaluating SegAVA in 1-way, 1-shot
segmentation on the PASCAL-5i dataset.}
\label{fig:results}
\end{figure}

As detailed in Table~\ref{PANet_table}, our experiments for the tasks
of 1-way 1-shot and 1-way 5-shot semantic segmentation demonstrates
competitive performance on the PASCAL-5i dataset. For 1-way, 5-shot
segmentation, our model achieves a new state-of-the art performance
and consistently outperforms competing approaches such as PANet
\citep{wang2019panet} and $\text{OS}_\text{Adv}$ \citep{YANG2020225},
except for part-4. For 1-way, 1-shot segmentation, we have achieved
state-of-the-art results on part-2 and part-3 while also being
competitive to Feat Weight \citep{feature_weights} with respect to the
overall mean. Our qualitative results (Figure~\ref{fig:results}) show
that single or multiple instances belonging to the same class have
been appropriately segmented.

\begin{table} \centering
\setlength{\tabcolsep}{4pt}
\begin{tabular}{l cc cc}
\toprule
\multirow{2}{*}{Annotations} & \multicolumn{2}{c}{SegAVA} & \multicolumn{2}{c}{PANet} \\
\cmidrule(lr){2-3}
\cmidrule(lr){4-5}
& 1-shot & 5-shot & 1-shot & 5-shot\\ 
\midrule
Dense & 50.6 &56.1 & 48.1&55.7    \\
Scribble     &46.9  &55.3 & 44.8& 54.6   \\
Bounding box     & 47.2 & 53.5& 45.1& 52.8  \\
\bottomrule
\end{tabular}
\caption[Quantitative comparison of SegAVA and PANet] {Comparison
between SegAVA and PANet in semi-supervised segmentation. Results are
expressed in mean-IoU.}
\label{PANet_table_3_weak}
\end{table}

\subsection{Semi-Supervised Segmentation}

\begin{figure}
\centering
\includegraphics[width=0.9\linewidth]{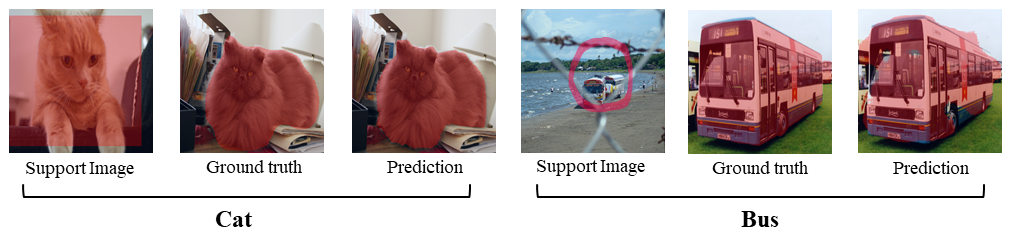}
\caption[Predictions of semi-supervised segmentation with SegAVA]
        {Example results from SegAVA on 1-way, 1-shot segmentation
        using both bounding boxes and scribble annotations.}
\label{fig:segava_semi}
\end{figure}

We have also validated the effectiveness of our SegAVA by using
bounding box and scribble annotations. As reported in
Table~\ref{PANet_table_3_weak}, our model generalizes well when using
these weaker types of annotations, and is still able to extract the
important features of the support set and localize and segment the
objects of interest in the query images. For the task of 1-way, 5-shot
segmentation, the performance of our model using scribble annotations
is surprisingly close to when dense level masks are made available.
Our model outperforms PANet in all tasks using both types of weaker
annotations. Qualitative results for semi-supervised segmentation are
shown Figure~\ref{fig:segava_semi}.

\begin{figure}
\centering
\includegraphics[width=0.9\linewidth]{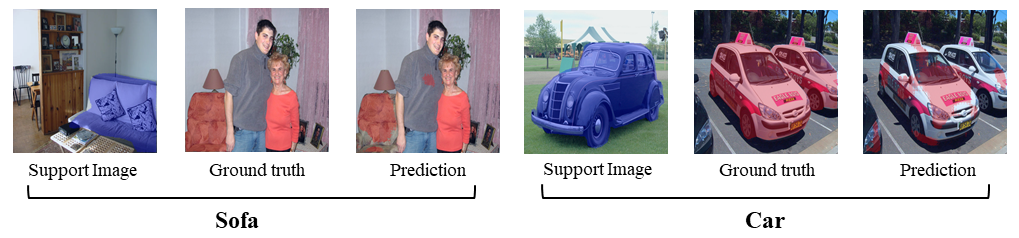}
\caption[Failure cases in SegAVA's predictions] {Example of failure
cases from evaluating SegAVA in 1-way, 1-shot segmentation on the
PASCAL-5i dataset.}
\label{fig:segava_failure}
\end{figure}
\subsection{Failure Cases}
\label{sec:fail}

Figure~\ref{fig:segava_failure} shows example failure cases of our
model. First, in some instances, our model is unable to fully
delineate the segmentation masks and may additionally produce
undesired patches. This can be resolved by incorporating
post-processing methods. Second, the model is unable recognize some
cases, which may be due to the extracted features in the image or
latent space being insufficient for certain classes.

\subsection{Active Contour Assisted Few-Shot Segmentation}

\begin{figure}
\def\x{0.24}
\includegraphics[width=\x\linewidth,height=\x\linewidth]{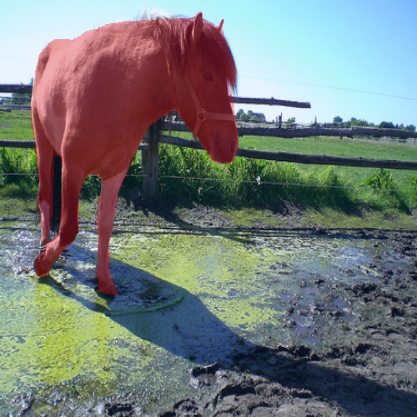}
\hfill
\includegraphics[width=\x\linewidth,height=\x\linewidth]{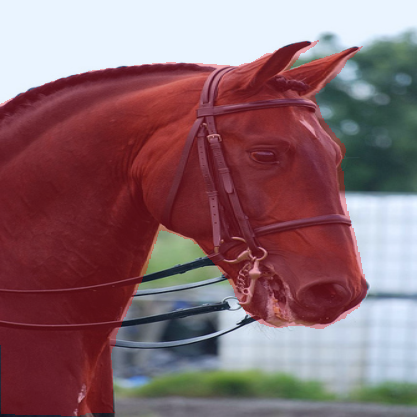}
\hfill
\includegraphics[width=\x\linewidth,height=\x\linewidth]{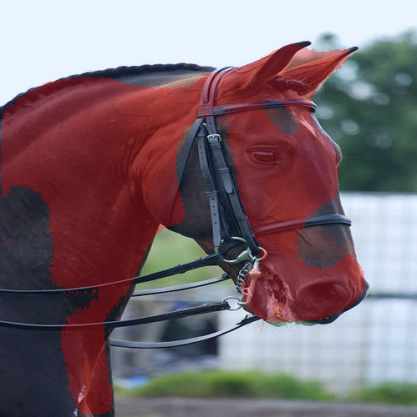}
\hfill
\includegraphics[width=\x\linewidth,height=\x\linewidth]{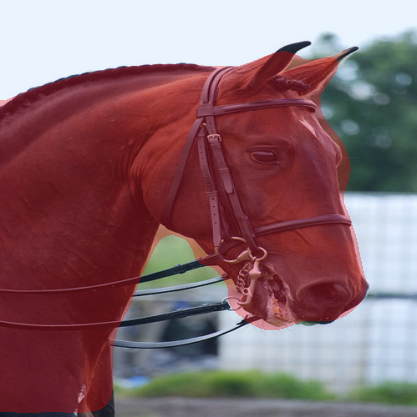}
\\[5pt]
\includegraphics[width=\x\linewidth,height=\x\linewidth]{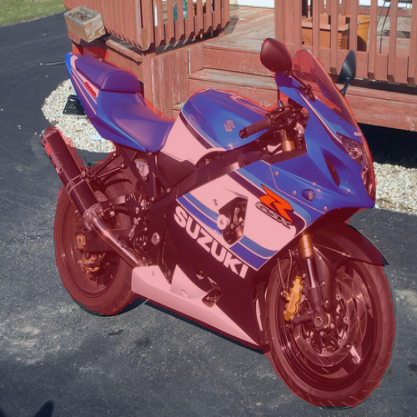}
\hfill
\includegraphics[width=\x\linewidth,height=\x\linewidth]{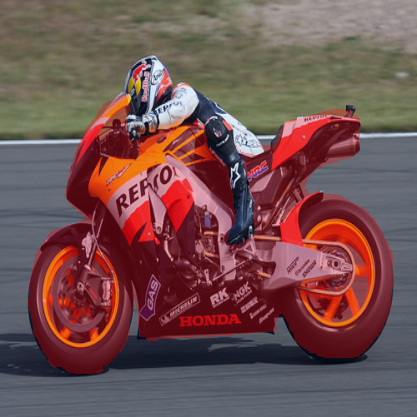}
\hfill
\includegraphics[width=\x\linewidth,height=\x\linewidth]{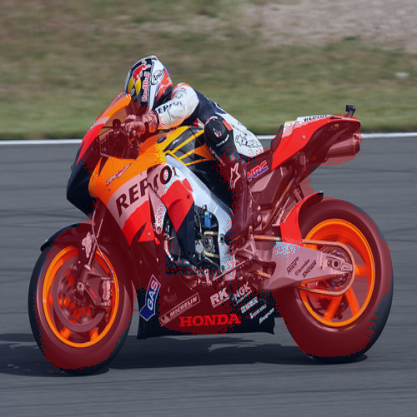}
\hfill
\includegraphics[width=\x\linewidth,height=\x\linewidth]{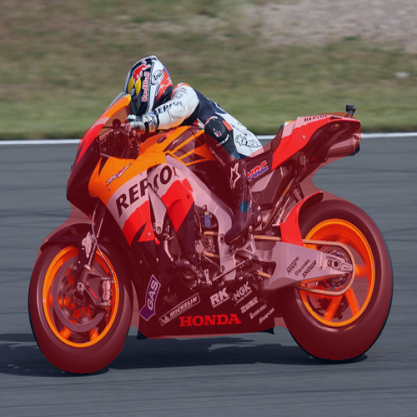}
\\[5pt]
\includegraphics[width=\x\linewidth,height=\x\linewidth]{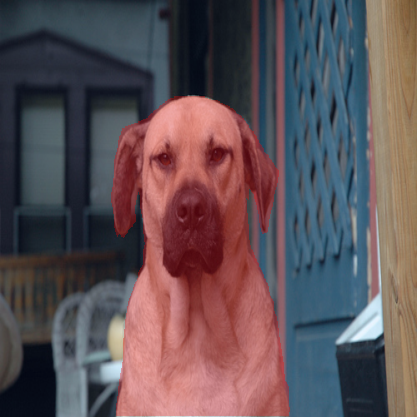}
\hfill
\includegraphics[width=\x\linewidth,height=\x\linewidth]{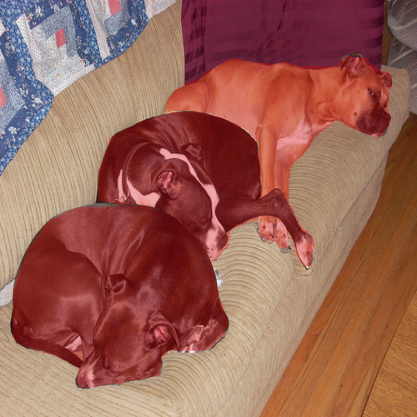}
\hfill
\includegraphics[width=\x\linewidth,height=\x\linewidth]{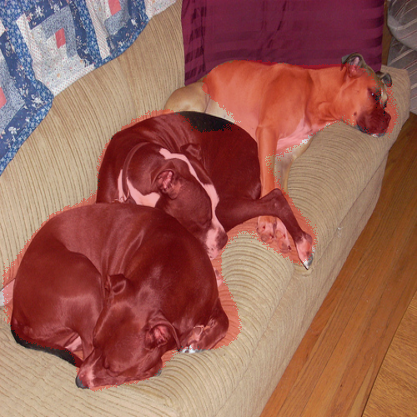}
\hfill
\includegraphics[width=\x\linewidth,height=\x\linewidth]{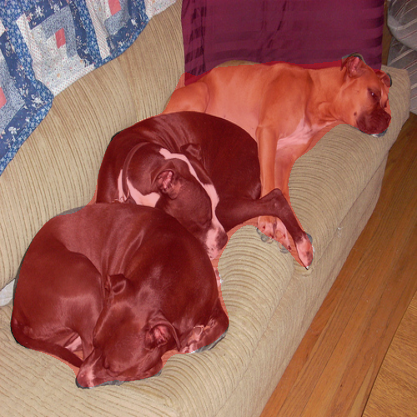}
\makebox[\x\linewidth]{(a)} \hfill \makebox[\x\linewidth]{(b)} \hfill
\makebox[\x\linewidth]{(c)} \hfill \makebox[\x\linewidth]{(d)}
\caption[Qualitative comparison of SegAVA and SegAVA+DALS]
        {Qualitative comparison of SegAVA and SegAVA+DALS. (a) Support
        Image. (b) Groundtruth. (c) SegAVA. (d) Combined. Qualitative
        comparison of SegAVA and SegAVA+DALS for 1-way, 1-shot and
        1-way, 5-shot segmentation tasks on the PASCAL-5i dataset.
        Combined denotes a model that consists of SegAVA as the
        backbone and DALS as a post-processor.}
\label{combined_compare}
\end{figure}

\begin{table} \centering
\setlength{\tabcolsep}{4pt}
\begin{tabular}{lcc}
\toprule
Model & 1-shot & 5-shot \\
\midrule
SegAVA & 50.6& 56.1\\
Combined  & \textbf{52.0} & \textbf{57.3}\\
\bottomrule
\end{tabular}
\medskip
\caption[Quantitative comparison of SegAVA and SegAVA+DALS]
        {Quantitative comparison of SegAVA and SegAVA+DALS for 1-way,
        1-shot and 1-way, 5-shot segmentation tasks on the PASCAL-5i
        dataset. Combined denotes a model that consists of SegAVA as
        the backbone and DALS as a post-processor.}
\label{res_combined}
\end{table}

We further leveraged our proposed DALS framework along with SegAVA to
delineate and refine the initial segmented boundaries. Quantitative
comparisons, as presented in Table~\ref{res_combined}, demonstrate the
benefits of leveraging DALS as post-processor with SegAVA. For 1-way,
1-shot and 1-way, 5-shot segmentation tasks, the combined framework
has improved the overall mIoU by 1.4\% and 1.2\%, respectively.

Figure~\ref{combined_compare} shows the qualitative comparison between
SegAVA and the combined framework with DALS. Evidently, DALS
delineates the mis-segmented regions and improves the segmentation
accuracy around the edges.

\chapter{Conclusions and Future Work}
\label{cha:conclusions}

\section{Conclusions}

This thesis has contributed several novel, fully automatic image
segmentation pipelines that can robustly produce precise object and
region boundaries and semantic segmentation masks, and learn to do so
from limited amounts of training data both in supervised and
semi-supervised, few-shot learning settings.


Motivated by the shortcomings of CNNs in tasks requiring segmentation
predictions with precise boundaries, we first proposed an end-to-end,
edge-aware model for semantic segmentation. Our network explicitly
accounts for object edge information by using a dedicated shape stream
that processes the feature maps at the boundary level and fuses the
multiscale contextual information of the boundaries with the encoder
output of the regular segmentation stream. Additionally, edge-aware
loss functions emphasize the learning of edge information during
training by tuning the weights of the downstream encoder and
regularizing the network to prioritize boundaries. We validated the
effectiveness of our approach on the task of brain tumor segmentation
using the BraTS 2018 dataset. The results indicate that our network
produces more accurate segmentation outputs with fine-grained
boundaries in comparison to the popular segmentation networks U-Net
and V-Net.

Second, we built upon the notion of edge-aware networks and proposed
an end-to-end volumetric (3D) edge-aware framework for the reliable
and automated segmentation of kidneys and kidney tumors. Our network
consists of a an encoder-decoder architecture equipped with a boundary
stream that processes the edge information separately and is
supervised by edge-aware losses. We have validated the effectiveness
of our approach by training and testing our model on 2019 MICCAI KiTS
Kidney Tumor Segmentation Challenge dataset. Our method achieved
dice scores of $0.9742$ and $0.8103$ for kidney and tumor repetitively
and an overall composite dice score of $0.8923$ and ranks 9th overall
in terms of composite dice among the 100 participants of this
challenge.

Third, we introduced the EG-CNN, a plug-and-play module for
boundary-aware CNN segmentation, which can be paired with an existing
encoder-decoder architecture to improve the segmentation accuracy. Our
EG-CNN does not require any additional annotation effort since edge
information can be extracted from the ground truth segmentation masks.
Supervised by edge-aware and consistency loss functions, the EG-CNN
learns to emphasize the edge representations by leveraging the feature
maps of intermediate resolutions in the encoder of the main stream and
feeding them into a series of edge-gated layers. We evaluated the
EG-CNN against three popular 3D segmentation architectures, U-Net,
V-Net, and Seg-Net, in the tasks of brain and kidney tumor
segmentation on the BraTS19 and KiTS19 datasets. Our results indicate
that the addition of the proposed EG-CNN consistently improves the
segmentation accuracy in all the benchmarks.

We then turned out attention taking advantage of the automated,
data-driven nature of CNNs and the precision and versatility of ACMs
to devise a powerful image segmentation pipeline.

We first presented Deep Active Lesion Segmentation (DALS), a novel
framework that combines the capabilities of the CNN and the level-set
ACM to yield a robust, fully automatic medical image segmentation
method that produces more accurate and detailed boundaries compared to
competing state-of-the-art methods. The DALS framework includes an
encoder-decoder that feeds a level-set ACM with per-pixel parameter
functions. We evaluated our framework in the challenging task of
lesion segmentation with a new dataset, MLS, which includes a variety
of images of lesions of various sizes and textures in different organs
acquired through multiple imaging modalities. Our results affirm the
effectiveness our DALS framework.

Second, we introduced a novel image segmentation framework, called
DTAC, which is an end-to-end trainable unification of ACMs and CNNs.
For this purpose, we proposed a new locally-penalized Eulerian energy
model that includes pixel-wise learnable parameters that can adjust
the contour to precisely capture and delineate the boundaries of
objects of interest in the image. We tackled the problem of building
instance segmentation on two challenging datasets, Vaihingen and Bing
Huts, as test cases and our model significantly outperforms the
current state-of-the-art method, DarNet. Furthermore, unlike DarNet
and DSAC, which rely on the manual initialization of ACM contours, our
model is fully automatic, as its backbone CNN learns initialization
maps as well as weighted parameters that can guide the contour to
avoid suboptimal solutions. Moreover, DarNet and DSAC are limited to
segmenting a single building at a time, whereas our DTAC can segment
multiple buildings simultaneously. In view of the level of success
that DTAC has achieved in this application and the fact that it
features a general Eulerian ACM formulation, it seems readily
applicable to other segmentation tasks in various domains, wherever
purely CNN filter-based approaches can benefit from the versatility
and precision of ACMs in delineating object boundaries in images.

Finally, we addressed semi-supervised learning of image segmentation
models, particularly the challenging problem of few-shot segmentation
by feature alignment in the image and latent spaces of support and
query samples. Our SegAVA model leverages a latent stream as well as
an encoder-decoder stream to extract the most essential discriminative
semantic embeddings and learns similarities in both spaces. The latent
stream consists of two variational autoencoders, conditioned on the
support and query sets, that jointly learn to generate the input
images and discriminatively identify the most common class-specific
representations using a Wasserstein-2 metric. These embedding are then
decoded to the image space and concatenated with common representation
that are found by comparing support and query extracted features using
our fully convolutional decoder. We trained and tested SegAVA using
the PASCAL-5i dataset, demonstrating new state-of-the-art performance
in 1-shot and 5-shot scenarios. We also validated the SegAVA model in
a semi-supervised setting where only bounding boxes are provided, and
our results demonstrate the sustained effectiveness of our approach.

\section{Future Work}

The research presented in this thesis can be further developed along
the following avenues:
\begin{enumerate}
\item 
Edge-aware networks can be further studied by adding modules that
assist the boundary stream better to delineate and find the
representations that correspond to the edges. In particular, a module
such as the proposed differentiable level-set ACM can be added to each
stage of the boundary stream to further refine the edges. Since the
ACM part is capable of back-propagating error, the entire framework
can be trained in an end-to-end manner.

Additionally, notion of fusing the output of the boundary stream and
main stream can be further studied and developed. In this work, we
presented a simple yet effective scheme for our fusion module, with an
emphasis on reducing the number of free parameters. However, future
efforts can include fusing the output of each stage of the boundary
and main streams in multiple resolutions to ensure the capture of
fine-grained details.

Finally, the end-to-end trainable active contours can be extended to
volumetric applications in order to interact with 3D encoder-decoders,
which are pervasive in medical image analysis. The efficiency of the
proposed framework is a noteworthy factor in such integration as
previous 3D ACM methods were prohibitively expensive computationally
and as a result have not been extensively utilized in medical imaging
to date.

\item
End-to-end trainable deep active contours were leveraged in a
supervised learning setting. However, given the fact that ACMs usually
do not require training labels, an intriguing research direction is to
utilize the proposed framework in semi-supervised or unsupervised
setting. Different levels of supervision can be employed in
initializing the contour. For instance, starting from a bounding box
that identifies a rough estimate of the target region, the trainable
ACM can be leveraged to fully delineate the segmentation mask.

Another important research direction for trainable deep active
contours is interactive segmentation and active learning where a user
may provide additional input and correct the mistakes of the
segmentation model as necessary, from which the model can learn.

\item
A promising direction for our SegAVA framework is to study the synergy
between it and other segmentation techniques in which the user can
interactively provide additional supervision, whether in the form of
mouse clicks or bounding boxes/scribbles, to improve the accuracy of
the segmentation. Additionally, our end-to-end trainable active
contours is a promising candidate for integration with SegAVA.

Another important direction is to study additional modules that can be
utilized along with SegAVA to provide the means for cross-domain,
few-shot segmentation. This may be realized by leveraging additional
transformation layers to accommodate domain shift and various feature
distributions.

Finally, applications such as medical image analysis, where
establishing dense, pixel-wise annotated datasets is very costly, can
benefit from our SegAVA model.

\end{enumerate}

\appendix

\graphicspath{ {cmbbe/figures/} }

\chapter{Fast and Automatic Segmentation of Pulmonary Lobes
from Chest {CT} Using a Progressive Dense V-Network}
\label{cha:appendix}

The material in this appendix was published as \citep{imran2019fast},
which is an expanded version of the publication
\citep{Imran2018automatic}.

\section{Abstract}

Automatic, reliable lobe segmentation is crucial to the diagnosis, assessment, and quantification of pulmonary diseases. Existing pulmonary lobe segmentation techniques are  prohibitively slow, undesirably rely on prior (airway/vessel) segmentation, and/or require user interactions for optimal results. We introduce a reliable, fast, and fully automated lung lobe segmentation method based on a Progressive Dense V-Network (PDV-Net). The proposed method can segment lung lobes in one forward pass of the network, with an average runtime of 2 seconds using a single Nvidia Titan XP GPU. An extensive robustness analysis of our method demonstrates reliable lobe segmentation of both healthy and pathological lungs in CT images acquired by scanners from different vendors, across various CT scan protocols and acquisition parameters.

\section{Introduction}

\begin{figure}
\centering
\includegraphics[width=0.7\linewidth]{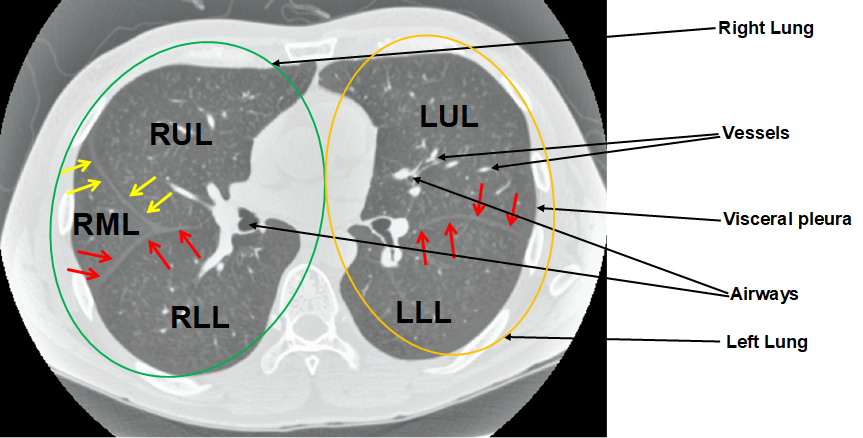} 
\caption[An axial lung CT slice with visible fissures] {An axial lung CT slice with visible fissures. The left upper lobe (LUL) and left lower lobe (LLL) are defined by a major fissure (indicated by red arrows); the right upper lobe (RUL), right middle lobe (RML), and right lower lobe (RLL) are defined by a major fissure (indicated by red arrows) and a minor fissure (indicated by yellow arrows).} 
\label{fig:axial-fissures}
\end{figure}

Human lungs are divided into five lobes. The inner membrane of the lung (visceral pleura) folds towards the center of the lung and creates double layer fissures that define the five lobes. The lobar boundaries are made of two major (oblique) fissures and a minor (horizontal) fissure. As shown in Figure~\ref{fig:axial-fissures}, the left lung has two lobes separated by a major fissure---the upper (superior) lobe and the lower (inferior) lobe. Along with upper and lower lobes, the right lung has a middle lobe; a major fissure separates the upper lobe from the middle lobe and a minor fissure separates the lower lobe from the middle lobe. Each of the five lobes is functionally independent, with its own bronchial and vascular systems.

\begin{figure}
\centering
\subcaptionbox{}{
\includegraphics[height=0.25\linewidth, width=0.3\linewidth]{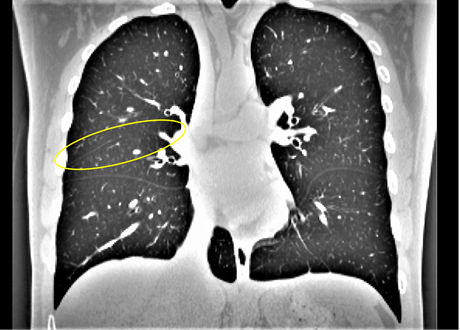} 
\label{fig:coronal-fissures}}
\qquad
\subcaptionbox{}{
\includegraphics[height = 0.25\linewidth, width=0.3\linewidth]{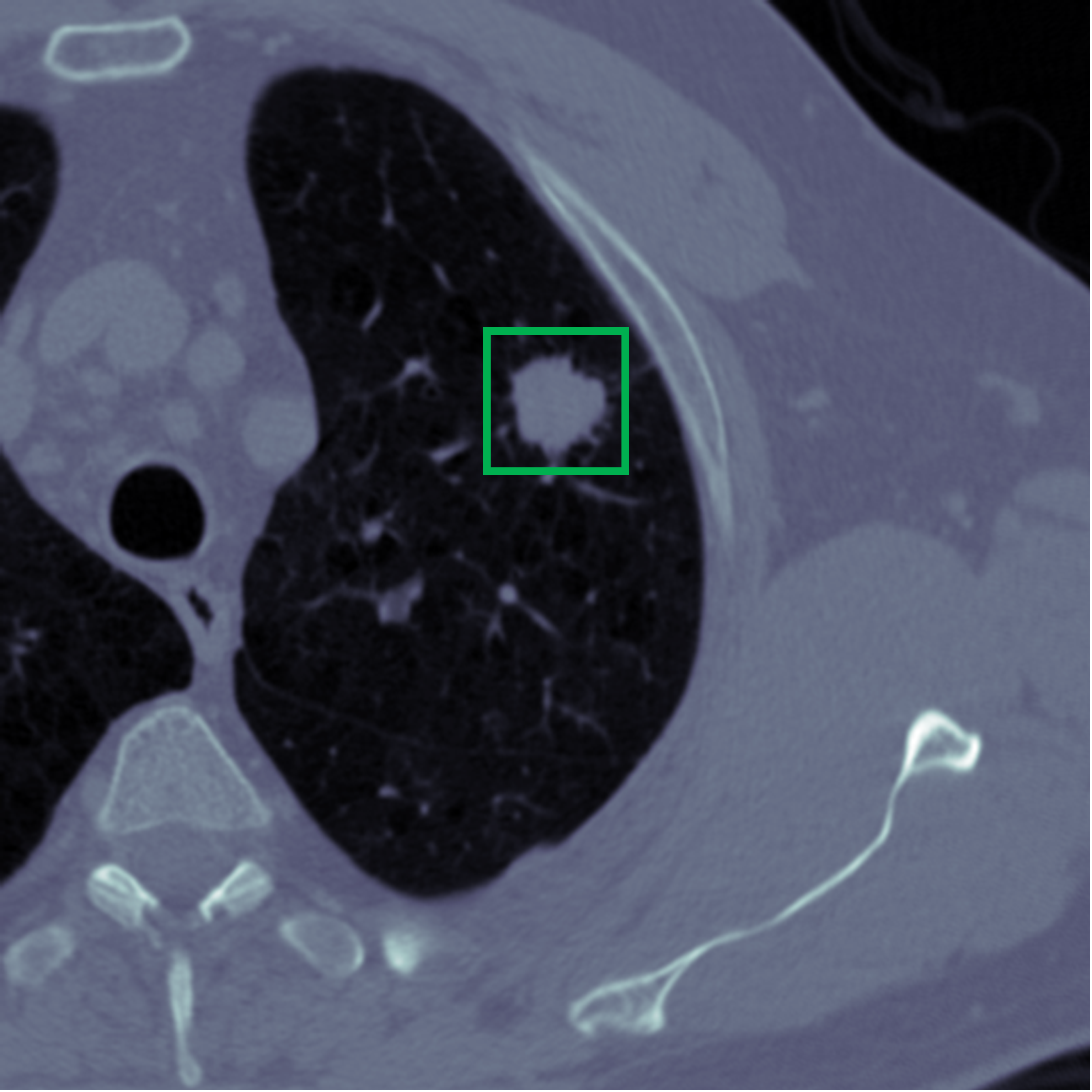} 
\label{fig:axial-nodule}}
\qquad
\subcaptionbox{}{
\includegraphics[height=0.25\linewidth, width=0.3\linewidth]{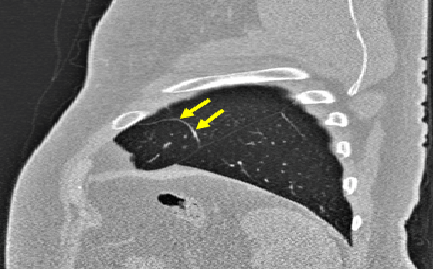} 
\label{fig:sagittal-accessory}}
\qquad
\subcaptionbox{}{
\includegraphics[height=0.25\linewidth, width=0.3\linewidth]{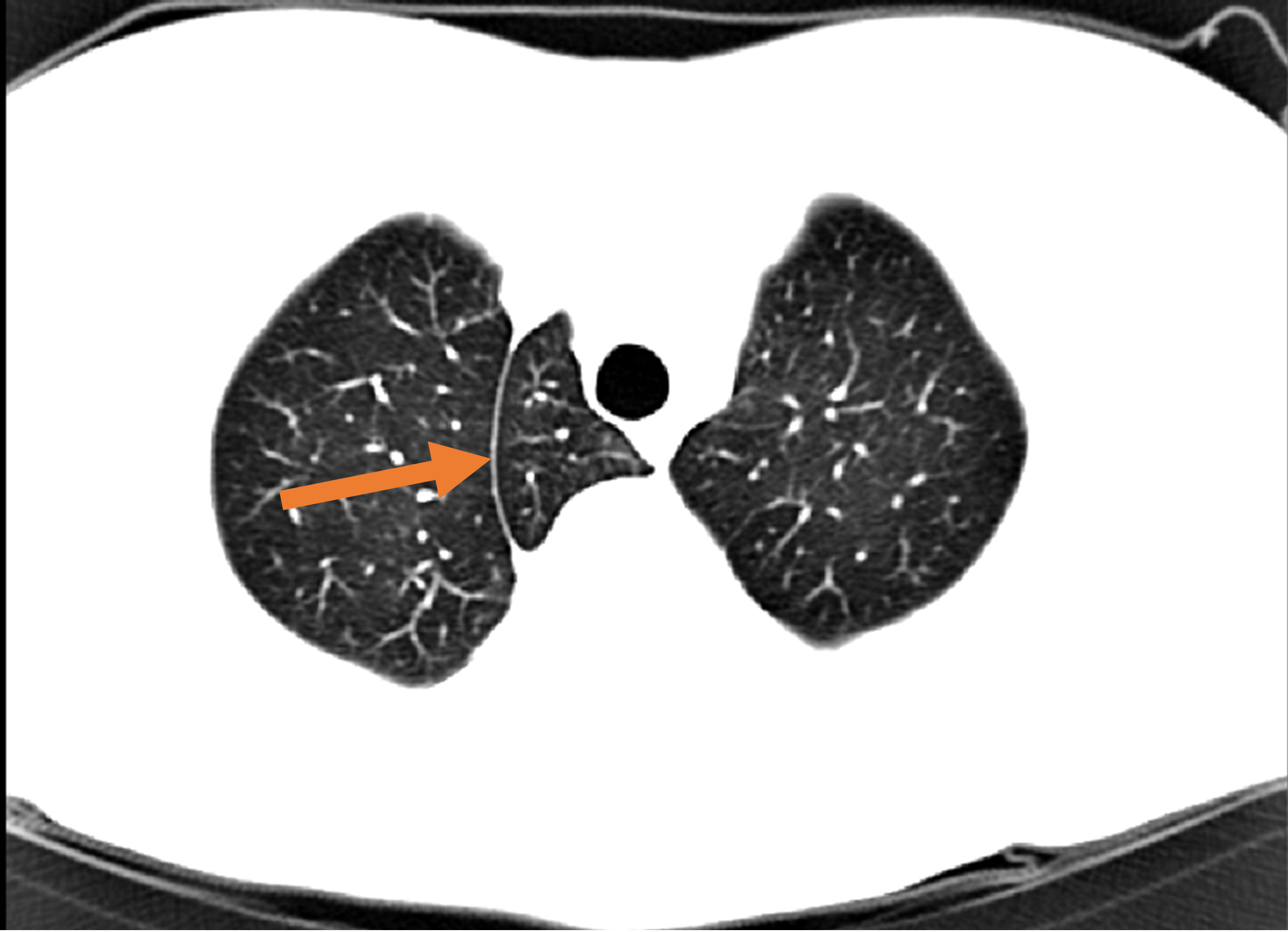} 
\label{fig:axial-azygos}}
\caption[Visualization of different fissure types] {(a) A coronal slice where the major fissures are complete and visible, but the minor fissure (circled) is incomplete. (b) Nodule shown in the bounding box. (An example nodule report: 5mm nodule found in the left upper lobe).  (c) Accessory fissure (arrows) in a left lung sagittal slice, which looks similar in shape to a minor fissure. (d) Azygos fissure (arrow) in an axial slice creates an extra lobe (azygos lobe) in the right lung} 
\label{fig:fissure-cases}
\end{figure}

 Automatic segmentation of the lung lobes is important for both clinical and technical purposes. From the clinical perspective, automatic lung lobe segmentation can help radiologists review chest CT scans more efficiently. This is because radiologists often report their pulmonary findings by indicating the affected lung lobe, whose identification requires them to navigate through the nearby slices and search for fissure lines, which are often visually indistinct. Automatic lung lobe segmentation can eliminate the need for such a tedious and time-consuming process. From the technical perspective, accurate lung lobe segmentation can assist several subsequent clinical tasks, including nodule malignancy prediction (cancers mostly occur in the left or right upper lobes), automatic lobe-aware report generation for each nodule (see Figure~\ref{fig:fissure-cases}(a)), and assessment and quantification of chronic obstructive pulmonary diseases (COPD) and interstitial lung diseases (ILD), by narrowing down the search space to the lung lobes most-likely to be affected.

 However, identifying fissures poses a challenge for both human and machine perception. First, fissures are most often incomplete, not extending to the lobar boundaries. This is shown in Figure~\ref{fig:fissure-cases}(a) where the horizontal fissure is incomplete, unlike the oblique fissures. Several studies in the literature have confirmed the incompleteness of fissures as a very common phenomenon. After reviewing 100 fixed and inflated lung specimens, \citet{Raasch1982} found incomplete right major fissures in 70\% of the cases, left major in 46\% of the cases, and 94\% across the minor fissures. Moreover, the studies of \citet{Gulsun2006} and \citet{Aziz2004} also showed more than 50\% incompleteness in pulmonary fissures. Second, the visual characteristics of lobar boundaries change in the presence of pathologies.
The changes could also be related to their thicknesses, locations, and shapes. Third, there also exist other fissures in the lungs that can be misinterpreted as the major and minor fissures that separate the lobes. Examples include accessory fissures (see the sagittal slice in Figure~\ref{fig:fissure-cases}(c)) and azygos fissures (see the axial slice in Figure~\ref{fig:fissure-cases}(d)). 

To address the need for accurate and robust lobe segmentation, we have pursued a fully automatic and reliable deep learning solution based on a Progressive Dense V-Network (PDV-Net) \citep{Imran2018automatic}. Our PDV-Net model inputs an entire CT volume and generates accurate segmentation of the lung lobes in about 2 seconds in only a single forward pass of the network, eliminating the need for any user interaction or any prior segmentation of the lungs, vessels, or airways, which are common assumptions in the design of existing models. Extensive robustness analyses demonstrate that our proposed method performs reliably for CT scans acquired using various imaging protocols from both healthy and pathological patients.

\section{Related Work}

There have been several efforts to segment lung lobes using semi-automatic and automatic techniques. We categorize these approaches into two groups: \emph{reliant approaches}, which rely on a prior segmentation or anatomical information, and \emph{non-reliant approaches}, which do not rely on such prior segmentations.

\subsection{Reliant Approaches} 

\subsubsection{Prior-Based Segmentation} 

Reliant approaches require as input a segmentation mask of lungs or lobes (different modalities), airways and vessels, or fissure initialization. A good example of the latter is the work by \citet{Doel2012}, in which lobe segmentation is performed based on an initialization via fissure detection. In another example of fissure initialization, \citet{iwano2013pulmonary} proposed semi-automatic and automatic lobe segmentation methods based on region-growing. The semi-automatic approach requires major and minor fissure initialization, whereas for the automatic approach, recognition of lobar bronchi and localization of fissures are performed prior to the final lobar segmentation. On average, the semi-automatic approach takes approximately 80 seconds and the automatic approach takes approximately 44 seconds per case.

A number of works depend on prior segmentation of airways, vessels, or fissures. The work by \citet{Bragman2017} is a good representative, wherein the suggested method relies on the prior segmentation of airways and vessels. Specifically, a population model of fissure priors was constructed and combined with patient-specific anatomical information for non-parametric surface fitting. Despite the promising results, the model lacks robustness and its reliance on prior knowledge limited the study. In recent work, \citet{Giuliani2018} proposed an approach to segment lobes from an approximate segmentation based on the airway tree. The final lobe segmentation was generated by combining the approximate segmentation with all the lung structures (airways, vessels, lungs, and fissures) segmentation using a multilevel graph cut algorithm. This segmentation method is highly reliant on the quality of the prior airway and vessel segmentations, as well as anatomical knowledge. \citet{Lassen2013} proposed a watershed-based lobe segmentation method by combining anatomical information from lungs, fissures, vessels, and bronchi. Despite reporting improved segmentation in the presence of incomplete fissures, the failure of individual prior segmentations limited the performance of the overall segmentation. Based on this work, \citet{lassen2017fast} proposed an interactive lobe segmentation method to interactively correct lobe segmentation error through user inputs. However, this improvement was obtained at the price of prolonged segmentation sessions.  \citet{lim2016fully} performed quantification of emphysema in 66 patients with moderate to severe emphysema who had undergone CT for lung volume reduction planning. They used lobar segmentation from four different prototypes for inter-software variability in lobe-wise emphysema quantification. Although the lobe segmentation performance is not reported, it is dependent on prior airway and vessel segmentation. 

Other works also rely on prior lung or lobe segmentation masks. For example, \citet{Bauer2018} segmented the lung lobes in the expiration phase based on a prior lobe segmentation mask obtained from a CT image acquired in the inspiration phase. An automated lung and lobe segmentation pipeline was proposed by \citet{blaffert2010completely}, in which a lung model mesh based on watershed segmentation is adapted to lobar segmentation. Final lobe regions are obtained by adjusting based on overlaid lungs in a post-processing step. However, the authors do not report a quantitative evaluation of lobar segmentation. The model takes 20 seconds to perform lobar segmentation in each CT scan.

\subsubsection{Atlas-Based Segmentation} 

Another variation of reliant segmentation is registration using mutual information with a previously segmented atlas. The performance of final lobe segmentation is greatly dependent on the performance of the segmentation algorithm used in creating a reference atlas. Among atlas-based approaches for lobe segmentation, \citet{Ross2010} employed the thin-plate spline and a maximum a posteriori estimation method using a manually-defined atlas as a reference. Fissure points were selected based on the atlas and the final lobe segmentation was generated after a post-processing step. Although this method did not rely on any prior airway and vessel segmentation, the execution time was long. Moreover, the creation of the atlas is very cumbersome and prone to poor results in pathological lung cases. By contrast, \citet{Pu2009} performed lobe segmentation by fitting an implicit function to fissures without reliance on prior airway or vessel segmentation. Although they achieved good accuracy for healthy lungs, the performance of their method degraded in the case of lungs with abnormal orientations. Unlike the other atlas-based segmentations, \citet{vanRikxoort2010} made use of multiple atlases for lobe segmentation. Their method showed promise albeit at the expense of slow execution.

\subsection{Non-Reliant Segmentation} 

Recently, a few convolutional neural-network-based lobe segmentation techniques have been proposed \citep{George2017, Ferreira2018, Wang2018}. The segmentation method of \citet{George2017} employs a 2D fully convolutional network followed by a 3D random walker algorithm. This approach does not rely on a prior segmentation of airways or vessels nor on any pre-computed atlases; however, it cannot generate lobe segmentation in a single pass, nor in an end-to-end manner. Furthermore, the 3D random walker algorithm relies on a number of heuristics for the initialization of seeds and weights. \citet{Ferreira2018} proposed a lobe segmentation model based on a fully regularized V-Net model with deep supervision and carefully chosen regularization. Although the performance looks impressive, the model was trained with few examples, so it lacks generalizability and may not be effective for varying CT scan cases. A 3D Dense Net-based lobe segmentation method was proposed by \citet{Wang2018}. Although they reported good accuracy for pathological lungs, their lobe segmentation method relies on prior lung segmentation and assumes the presence of five lobes, which might not always be the case (e.g., \citep{LOLA2011}).     

Our work \citep{Imran2018automatic} mitigates the aforementioned limitations---namely, reliance on prior masks, slow runtime, and lack of robustness---through an end-to-end learning network. Without relying on any prior airway/vessel segmentation or anatomical knowledge or atlases, our method performs lobe segmentation in a single pass of the network. Owing to the full utilization of the 3D context in our model, the resulting lobe segmentation is smooth and nearly noise-free, which eliminates the need for any subsequent post-processing to fill holes or remove noisy patches from outside the lung area.
Our method shows promise for the potential clinical use in quantification of pulmonary diseases and automatic generation of radiological reports.

\section{Materials and Methods}

\subsection{Progressive Dense V-Net}

Combining ideas from dense V-Networks \citep{Gibson2018} and progressive holistically-nested networks \citep{harrison2017}, we propose a new architecture---the Progressive Dense V-Network (PDV-Net), an end-to-end solution for organ segmentation in 3D volumetric data. 

\begin{figure}
\centering
\includegraphics[width=\linewidth]{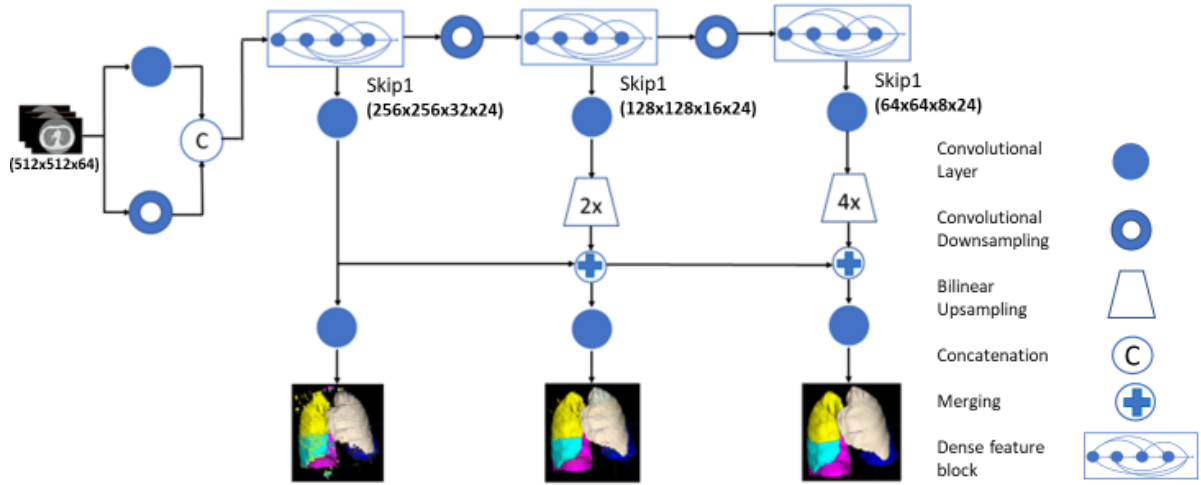}
\caption[PDV-Net model for lobe segmentation] {PDV-Net model for the segmentation of lung lobes. Segmentation outputs in different pathways are progressively improved for the final result.}
\label{fig:pdv-net}
\end{figure}

As shown in Figure~\ref{fig:pdv-net}, the input to the network is first down-sampled  and concatenated with a strided $5\times 5\times 5$ convolution of the input with 24 kernels. The concatenation result is then passed to 3 dense feature blocks, each consisting of 5, 10, and 10 densely-wired convolution layers respectively. The growth rates of the dense blocks are set to 4, 8, and 16 respectively.  All the convolutional layers in a dense block have a kernel size of $3\times 3\times 3$ and are followed by batch normalization and parametric rectified linear units (PReLU). The outputs of the dense feature blocks are consecutively  utilized in low and high resolution passes via convolutional down-sampling and skip connections. This enables the generation of feature maps at three different resolutions. The outputs of the skip connections of the second and third dense feature blocks are further up-sampled in order to be consistent with the size of the output in the first skip connection. The feature maps from skip1 are passed to a convolutional layer followed by a softmax, which outputs the probability maps. In the second pathway, the feature maps from skip1 and skip2 are merged and the output probability maps are produced by a convolutional layer followed by softmax. Similarly, we obtain the final segmentation resulting from the merged feature maps resulted from the skip2 and skip3 connections. Unlike the dense V-Net, the PDV-Net generates the final output by progressively improving the outputs from the previous pathways. 

The PDV-Net is trained using a subset $S$ of a volumetric medical image dataset $(\mathcal{D})$. The training set $(S)$ contains 3D CT scan images and their corresponding ground truth labels. So, $S = (\mathcal{X}_n, \mathcal{Y}_n)$, for $n = 1,\dots, N$, where the input volumes $\mathcal{X}_n^{(m)} = {x_i^{(n)}; i = 1,\dots |\mathcal{X}|_n}$, and the corresponding ground truth labeled volumes $\mathcal{Y}_n^{(m)} = {y_i^{(n)}}$ $i = 1,\dots |\mathcal{Y}|_n, y_l^{(n)} \in \{0\dots L\}$. Here, $|S|$ is the total number of training examples passed to the network and $L$ is the  number of labels provided in the ground truth data through per-voxel labeling $(l)$. To train the PDV-Net, we use a Dice loss function \citep{Milletari2016} at each level of the progressive network, which directly maximizes the similarity between the predicted values and the ground truth over all voxels. This loss properly handles the class imbalance problem prevalent in lung lobe segmentation: lung lobes have different sizes and background regions can be large. We employ  a multi-class Dice for the segmentation task:
\begin{equation}
\label{eq:loss}
d = \sum_{l=1}^{L}\frac{\sum_{j=1}^{Z}p_j^{l}g_j^{l}}{\sum_{j=1}^{Z}(p_j^{l})^2+\sum_{j=1}^{Z}(g_j^{l})^2},
\end{equation}
where \(Z\) is the total number of voxels, \(L\) is the number of classes, \(p_j^{l}\) denotes the predicted probabilities for each class, and \(g_j^l\) denotes the corresponding ground truth for each class. 

\subsection{Data} 

\begin{figure}
\def\x{0.24}

\includegraphics[width=0.32\linewidth,height=0.32\linewidth]{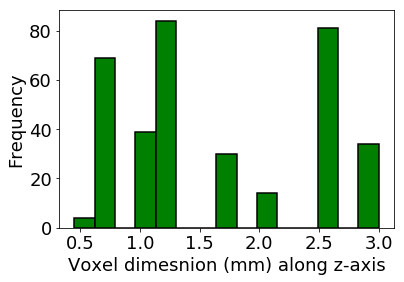}
\hfill
\includegraphics[width=0.32\linewidth,height=0.32\linewidth]{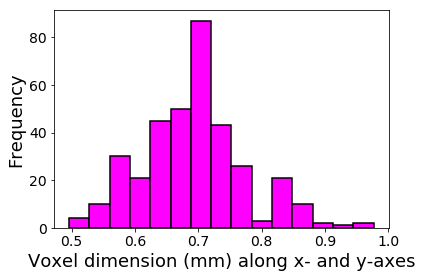}
\hfill
\includegraphics[width=0.32\linewidth,height=0.32\linewidth]{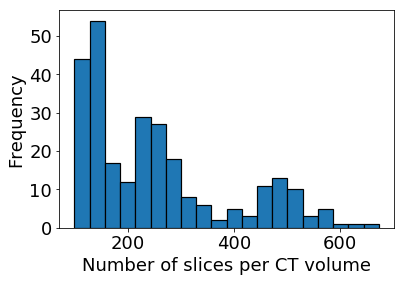}
 
\caption[Histograms of number of slices and voxel dimensions] {Histograms (from left to right) of the number of slices per volume; voxel dimensions along the $x$ and $y$ axes; and voxel dimensions along the $z$ axis of lung CT scans in the entire LIDC dataset.}
\label{fig:slices-voxels}
\end{figure}

We used 3 public datasets to evaluate our model: 
\begin{enumerate}
\item We selected a subset of chest CT volumes (354 cases) from the LIDC dataset \citep{Armato2011} for annotation. To ensure variation in the data, the CT scans were selected such that both challenging and visible fissures are well-represented in the dataset. The lobe segmentation ground truth masks were generated in a semi-automatic fashion by multiple human annotators using the chest imaging platform feature of 3D Slicer. To mitigate bias in the ground truth, the generated masks were later refined and validated by an expert radiologist. The dataset was partitioned into 270 training and 84 test cases. 10\% of the training set was utilized as the validation set to select values for the hyper-parameters. The CT scans used in the experiment have a variable number of slices with each CT volume containing 100 to 672 slices of size $512\times 512$ pixels. Figure~\ref{fig:slices-voxels} shows the histograms of the number of slices per volume, and of the voxel dimensions which vary between 0.49--0.98\,mm, 0.49--0.98\,mm, and 0.45--3.00\,mm along the $x$, $y$, and $z$ axes, respectively. Therefore, the selected CT scans used for pulmonary lobe segmentation not only exhibit varying shapes of fissures and lobes, but also show a variable number of slices and voxel sizes. 

\item We selected 154 CTs from the LTRC database \citep{karwoski2008processing}. The LTRC dataset includes lobe masks for pathological cases that have clear evidence of COPD or ILD diseases, including emphysema and fibrosis. The LTRC cases allow us to measure the robustness of our model against pathologies in the lungs.

\item We used 55 cases of the Lobe and Lung Analysis (LOLA11) challenge \citep{LOLA2011} and submitted our results to the challenge organizers for evaluation.
\end{enumerate}

\section{Experiments}

\subsection{Baselines for Comparison} 

For our baseline comparison, we used a U-Net architecture \citep{Ronneberger15} and a dense V-Net. The former is used in the most recent published article for lung lobe segmentation \citep{George2017} and the latter is a strong baseline for comparison, which we are the first to employ for lung lobe segmentation.

\subsection{Implementation Details}

For our PDV-Net and the dense V-Net, the training volumes were first normalized, followed by rescaling to $512\times512\times64$, using one NVIDIA Titan XP GPU. Due to the large memory footprint of the model, the gradient check-pointing method \citep{Bulatov2018} was used for memory-efficient back-propagation. Additionally, batch-wise spatial dropout \citep{Gibson2018} is incorporated for regularization purposes. The training was performed on a 64-bit Intel Xeon E5-2697 v4 2.30\,GHz CPU system with 256\,GB of RAM. We used the Adam optimizer \citep{Kingma2014} with a learning rate of 0.01 and a weight decay of $10^{-7}$.

For the 2D U-Net model, the implemented architecture is symmetric and consists of four contracting and expanding layers, starting with 16 features in the first layer and doubling the number of features in each step. Each contracting layer consists of two $3\times3$ convolutions and a ReLU activation followed by a $2\times2$ max-pooling layer. The expansion path consists of an up-convolution with feature concatenation from the respective contracting layer, and two $3\times3$ convolutions. In addition, all the ReLU layers are preceded by a batch-normalization layer. To improve the training process, we also used a generalized Dice score as the loss function, such that the contribution of each class in the image to the gradients is balanced. We trained the network with axial slices from all the training volumes, each sized $512\times 512$ pixels and normalized to have values between 0 and 1. To avoid over-fitting to the background class, we used only the axial slices, wherein at least one lung lobe is present. We used the Adam optimizer with a learning rate of $5\times 10^{-5}$ and batches of 10 images.

\section{Results and Discussion}

\subsection{LIDC Results}

\begin{figure}
\centering
 \resizebox{\linewidth}{!}{%
  \begin{tabular}{cccccc}
& {\large Raw Slice} & {\large GT} & {\large U-Net} & {\large DV-Net} & {\large PDV-Net}\\
\noalign{\smallskip}

    \rotatebox{45}{\large Axial} &
	\includegraphics[width=0.25\linewidth, trim={0.25cm 2cm 0.25cm 2.5cm},clip]{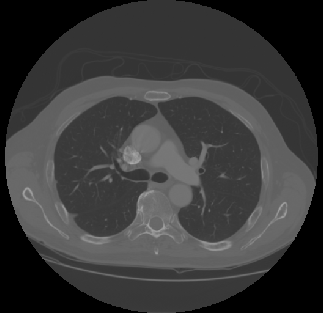} &
	\includegraphics[width=0.25\linewidth, trim={0.25cm 2cm 0.25cm 2.5cm},clip]{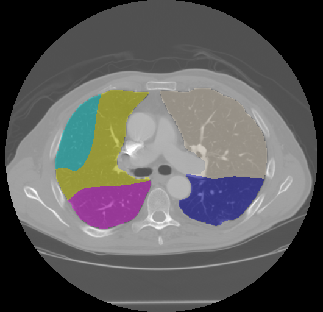} &
	\includegraphics[width=0.25\linewidth, trim={0.25cm 2.5cm 0.25cm 3cm},clip]{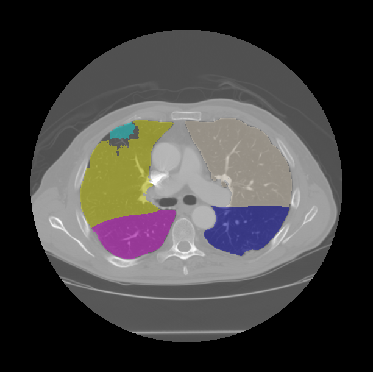} &
	\includegraphics[width=0.25\linewidth, trim={0.25cm 2cm 0.25cm 2.5cm},clip]{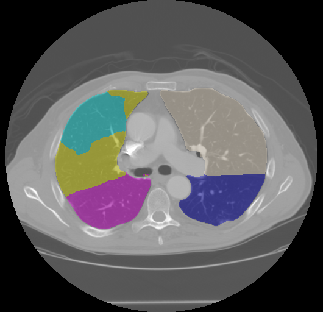} &
    \includegraphics[width=0.25\linewidth, trim={0.25cm 2cm 0.25cm 2.5cm},clip]{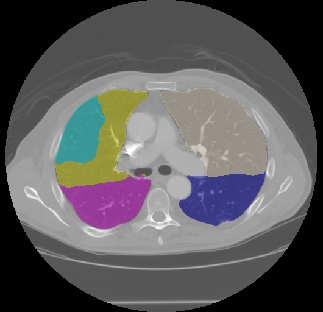} 
    \smallskip\\
        
    \rotatebox{45}{\large Coronal} &
    \includegraphics[width=0.25\linewidth, trim={0.25cm 2.5cm 0.25cm 1.5cm},clip]{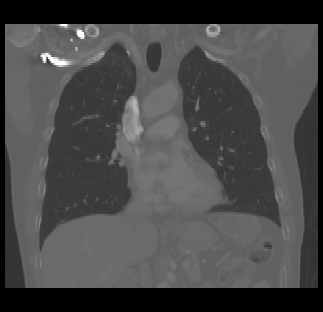} &
	\includegraphics[width=0.25\linewidth, trim={0.25cm 2.5cm 0.25cm 1.5cm},clip]{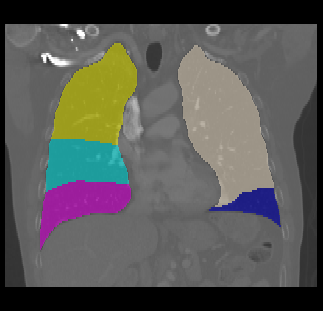} &
	\includegraphics[width=0.25\linewidth, trim={0.25cm 3.6cm 0.25cm 2.5cm},clip]{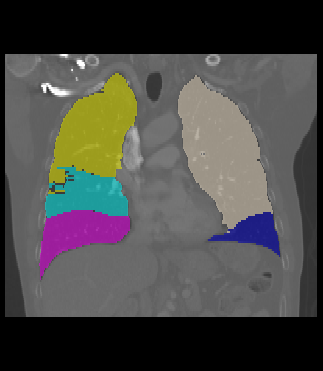} &
	\includegraphics[width=0.25\linewidth, trim={0.25cm 2.5cm 0.25cm 1.5cm},clip]{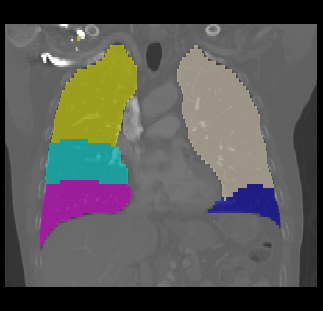} &
    \includegraphics[width=0.25\linewidth, trim={0.25cm 2.5cm 0.25cm 1.5cm},clip]{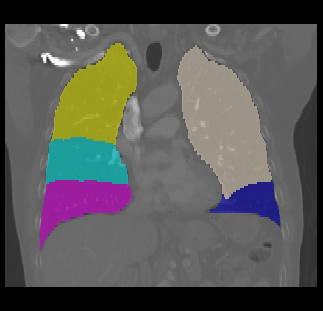} 
    \smallskip\\
    
    \rotatebox{45}{\large Sagittal} &
    \includegraphics[width=0.25\linewidth, trim={12.0cm 1.5cm 11.0cm 3.0cm},clip]{990ct_s179.png} &
	\includegraphics[width=0.25\linewidth, trim={12.0cm 1.5cm 11.0cm 3.0cm},clip]{990gt_s179.png} &
    \includegraphics[width=0.25\linewidth, trim={12.0cm 1.5cm 11.0cm 3.0cm},clip]{990u_s179.png} &
    \includegraphics[width=0.25\linewidth, trim={12.0cm 1.5cm 11.0cm 3.0cm},clip]{990dv_s179.png} &
    \includegraphics[width=0.25\linewidth, trim={12.0cm 1.5cm 11.0cm 3.0cm},clip]{990pdv_s179.png}
    \smallskip\\
    
    &
    \rotatebox{45}{\large 3D} &
	\includegraphics[width=0.25\linewidth, trim={16.0cm 5.0cm 11.0cm 5.0cm},clip]{46g_3d.png} &
     \includegraphics[width=0.25\linewidth, trim={16.0cm 5.0cm 11.0cm 5.0cm},clip]{46u_3d.png} &
    \includegraphics[width=0.25\linewidth, trim={16.0cm 5.0cm 11.0cm 5.0cm},clip]{46dv_3d.png} &
    \includegraphics[width=0.25\linewidth, trim={16.0cm 5.0cm 11.0cm 5.0cm},clip]{46pdv_3d.png} 
    
 \end{tabular} }
  \caption[Qualitative comparison of PDV-Net's outputs] {Qualitative comparison of PDV-Net's superior performance, both in slice and volume level, against DV-Net and U-Net. Note how noisy patches and rough boundaries are removed from the final segmentation generated by the PDV-Net. Color coding: almond:~LUL, blue:~LLL, yellow:~RUL, cyan:~RML, pink:~RLL.}
  \label{fig:vis-lidc}
  
\end{figure}

Table~\ref{tab:lidc-ltrc} shows the calculated overall and lobe-wise Dice scores and standard deviations for each of the models. Our PDV-Net model, with an overall score of $0.939 \pm 0.020$, significantly outperformed the 2D model and yielded consistently larger Dice scores for each of the lung lobes against both the DV-Net and U-Net. Moreover, the lower standard deviation for each lobe indicates that our progressive model is more robust. Figure~\ref{fig:vis-lidc} provides a qualitative comparison between the three models, showing that our PDV-Net model captures lung fissures better than the 2D U-Net and DV-Net. The superiority of our PDV-Net model is evident both in slice (axial, coronal, sagittal) and 3D views.

\begin{table} \centering
\setlength{\tabcolsep}{4pt}
\resizebox{\linewidth}{!}{%
\begin{tabular}{ l l c c c c c c}
\toprule
Dataset & Model & RUL & RML & RLL & LUL & LLL & Overall\\ 
\noalign{\smallskip}
\midrule
\noalign{\smallskip}
\multirow{3}{*}{LIDC(84)} 
    & 2D U-Net &  
0.908 $\pm$ 0.049 & 
0.844 $\pm$ 0.076 & 
0.940 $\pm$ 0.054 & 
0.959 $\pm$ 0.042 & 
0.949 $\pm$ 0.056 & 
0.920 $\pm$ 0.043\\
\cmidrule{2-8}
    & 3D DV-Net & 
0.929 $\pm$ 0.036 & 
0.873 $\pm$ 0.058 &
0.951 $\pm$ 0.018 & 
0.958 $\pm$ 0.020 &
0.949 $\pm$ 0.041 & 
0.932 $\pm$ 0.023 \\ 
\cmidrule{2-8}
    & 3D PDV-Net & 
0.937 $\pm$ 0.031 & 
0.882 $\pm$ 0.057& 
0.956 $\pm$ 0.017& 
0.966 $\pm$ 0.014&
0.966 $\pm$ 0.037& 
0.939 $\pm$ 0.020\\ 
\midrule
\multirow{3}{*}{LTRC(154)} 
    & 2D U-Net &  
0.914 $\pm$ 0.039 & 
0.866 $\pm$ 0.054 & 
0.952 $\pm$ 0.023 & 
0.961 $\pm$ 0.023 & 
0.954 $\pm$ 0.021 & 
0.929 $\pm$ 0.025\\
\cmidrule{2-8}
  & 3D DV-Net & 
0.949 $\pm$ 0.013 & 
0.901 $\pm$ 0.021 & 
0.959 $\pm$ 0.009 & 
0.961 $\pm$ 0.007 &
0.958 $\pm$ 0.012 &
0.946 $\pm$ 0.008\\ 
\cmidrule{2-8}
  & 3D PDV-Net & 
0.952 $\pm$ 0.011& 
0.908 $\pm$ 0.020& 
0.961 $\pm$ 0.008& 
0.966 $\pm$ 0.006&
0.960 $\pm$ 0.010&
0.950 $\pm$ 0.007\\ 
\bottomrule
\end{tabular}}
\caption[Quantitative comparison of PDV-Net and others on LIDC and LTRC] {Performance comparison of the proposed 3D progressive dense V-net with the 2D U-net and 3D dense V-net models in segmenting 84 LIDC and 154 LTRC cases. Mean Dice score and standard deviation for each lobe have been reported.}
\label{tab:lidc-ltrc}
\end{table}
\setlength{\tabcolsep}{1.4pt}

\begin{figure}
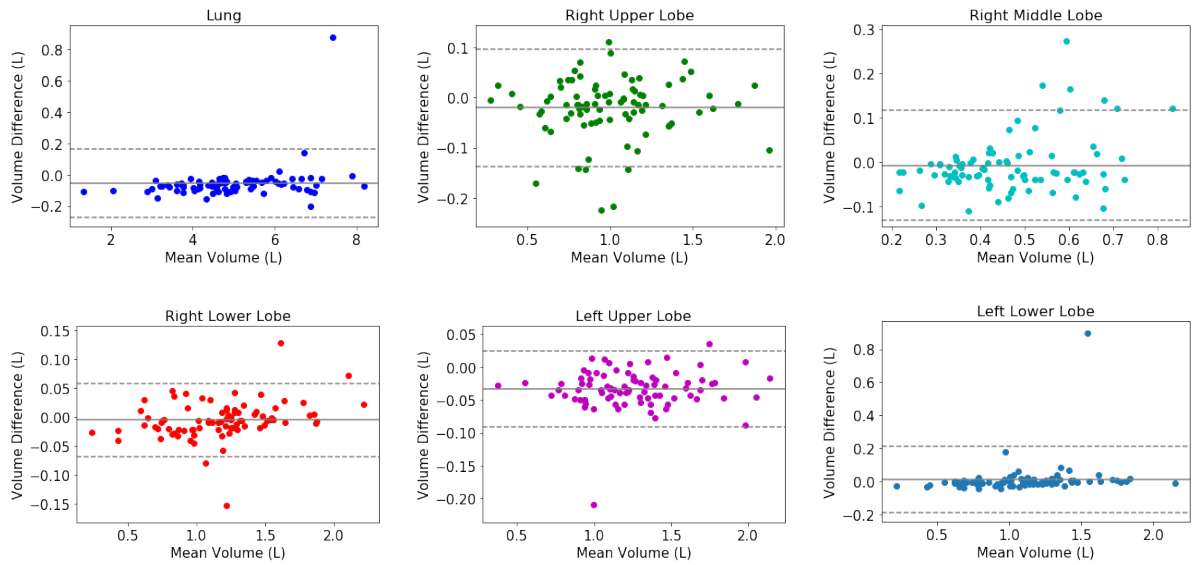

\centering
\includegraphics[width=0.32\linewidth]{bla-alt_lung.png}
\hfill
\includegraphics[width=0.32\linewidth]{bla-alt_rul.png}
\hfill
\includegraphics[width=0.32\linewidth]{bla-alt_rml.png} 
\\[10pt]
\includegraphics[width=0.32\linewidth]{bla-alt_rll.png}
\hfill
\includegraphics[width=0.32\linewidth]{bla-alt_lul.png}
\hfill
\includegraphics[width=0.32\linewidth]{bla-alt_lll.png}
\caption[Bland-Altman plots of PDV-Net's outputs] {Bland-Altman plots show the agreement between 3D PDV-Net and ground truth.}
\label{fig:bland-altman}
\end{figure}

We further used Bland-Altman plots to measure the agreement between our PDV-Net and ground truth segmentations of the 84 LIDC cases (Figure~\ref{fig:bland-altman}). Good agreement was observed between our segmentation model and ground truth in every plot (Lung and LLL being the two best agreements). Pearson correlation showed that all six volume sets in ground truth are strongly correlated with the corresponding six volume sets in the PDV-Net segmentation, with $p < 0.001$. 

\subsection{LTRC Results} 

Table~\ref{tab:lidc-ltrc} shows that the 3D progressive dense V-Net achieves an average Dice score of $0.950 \pm 0.007$, significantly improving the dense V-Net ($0.946 \pm 0.008$). Once again, the progressive dense V-Net model outperformed the 2D U-Net model with an average Dice score of $0.929 \pm 0.025$. Individual lobes were segmented better by our 3D progressive dense V-Net model than by the 3D dense V-Net and the 2D U-Net models (Table \ref{tab:lidc-ltrc}). Note that the LTRC dataset includes many pathological cases where the fissure lines are either invisible, distorted, or absent in the presence of pathologies such as emphysema, fibrosis, etc. As a result, lobe segmentation becomes more challenging. Nevertheless, our model performed well in segmenting lobes in pathological cases from the LTRC dataset. Moreover, our model outperformed the model of \citet{George2017} in segmenting the LTRC cases both in Dice score (0.941 $\pm$ 0.255) and inference speed (4-8 minutes per case). 

\subsection{LOLA11 Results} Our segmentation results for the LOLA11 cases  were evaluated by the organizers of LOLA11. To be consistent with our previous analyses, the Jaccard scores computed by the organizers were converted to Dice scores. The results are shown in Table~\ref{tab:lola-score}. Our method achieved an overall Dice score of 0.934, which is very competitive to the state-of-the-art reliant method \citep{Bragman2017} with a Dice score of 0.938, while outperforming the methods of \citet{Giuliani2018} and \citet{vanRikxoort2010}. 

\begin{table} \centering
\setlength{\tabcolsep}{4pt}
\begin{tabular}{ l c c c c }
\toprule
Lobe & Mean $\pm$ SD & $Q_1$ & Median & $Q_3$\\ 
\midrule
RUL & 0.9518 $\pm$ 0.1750 & 0.9371 & 0.9688 & 0.9881 \\
RML & 0.8621 $\pm$ 0.4149 & 0.8107 & 0.9284 & 0.9663 \\ 
RLL & 0.9581 $\pm$ 0.1993 & 0.9621 & 0.9829 & 0.9881 \\
LUL & 0.9551 $\pm$ 0.2160 & 0.9644 & 0.9834 & 0.9924 \\
LLL & 0.9342 $\pm$ 0.3733 & 0.9546 & 0.9805 & 0.9902 \\
\midrule
Overall & 0.9345 \\
\citep{Giuliani2018} & 0.9282 \\
\citep{Bragman2017} & 0.9384 \\
\citep{vanRikxoort2010} & 0.9195 \\
\noalign{\smallskip}
\bottomrule
\end{tabular}
\caption[Performance evaluation of our 3D PDV-Net on LOLA dataset] {Performance evaluation of our 3D PDV-Net model on 55 LOLA cases, showing lobe-wise mean Dice scores, standard deviations, median
scores, first quartiles, and third quartiles. Jaccard score to Dice score conversion: $\text{Dice} = 2\times \text{Jaccard}/(1+\text{Jaccard})$.}
\label{tab:lola-score}
\end{table}
\setlength{\tabcolsep}{1.4pt}

\begin{figure}
\def\x{0.24}

\includegraphics[width=\x\linewidth,height=\x\linewidth]{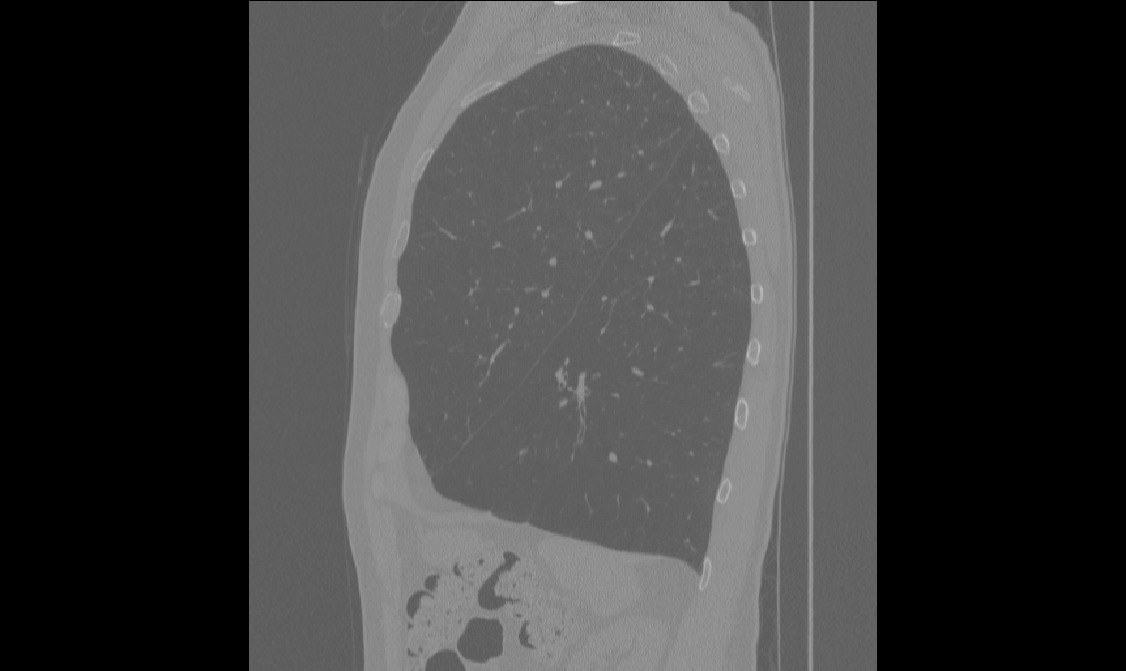}
\hfill
\includegraphics[width=\x\linewidth,height=\x\linewidth]{lola-8_left346.png}
\hfill
\includegraphics[width=\x\linewidth,height=\x\linewidth]{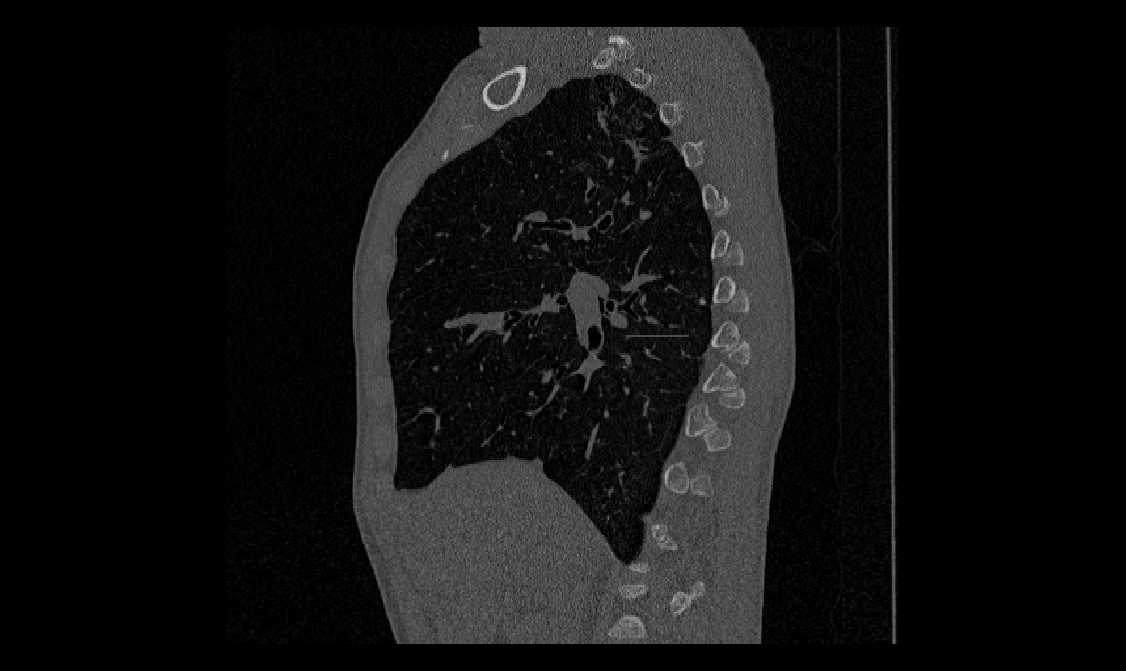}
\hfill
\includegraphics[width=\x\linewidth,height=\x\linewidth]{lola-6_right217.png}

\vspace{0.75pt}

\includegraphics[width=\x\linewidth,height=\x\linewidth]{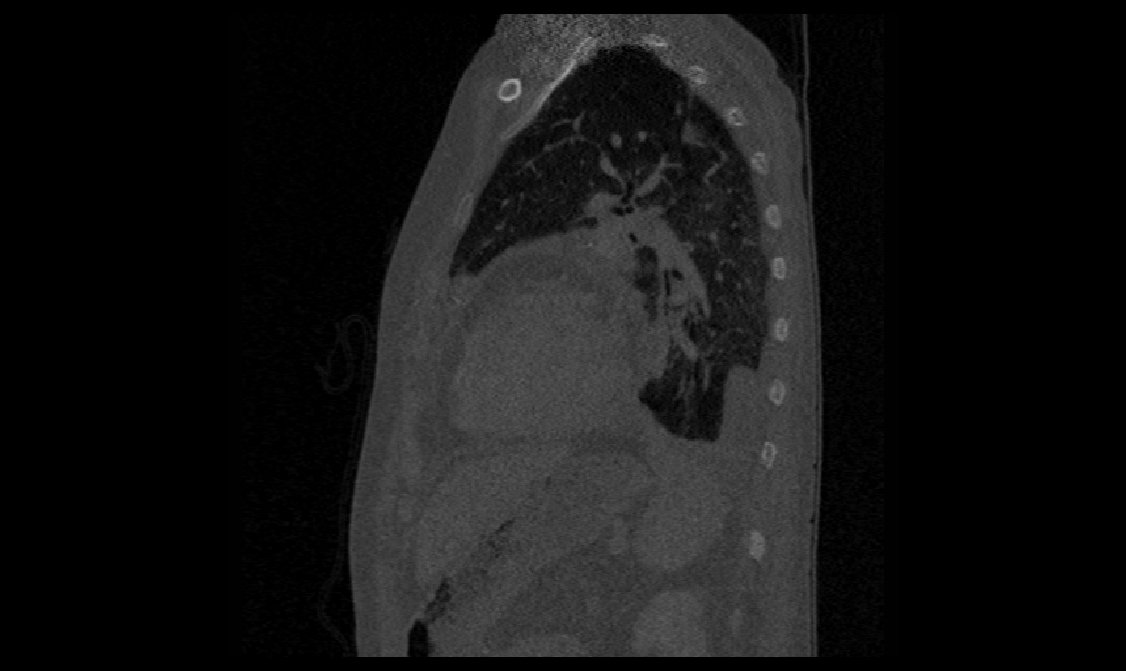}
\hfill
\includegraphics[width=\x\linewidth,height=\x\linewidth]{lola-21_left332.png}
\hfill
\includegraphics[width=\x\linewidth,height=\x\linewidth]{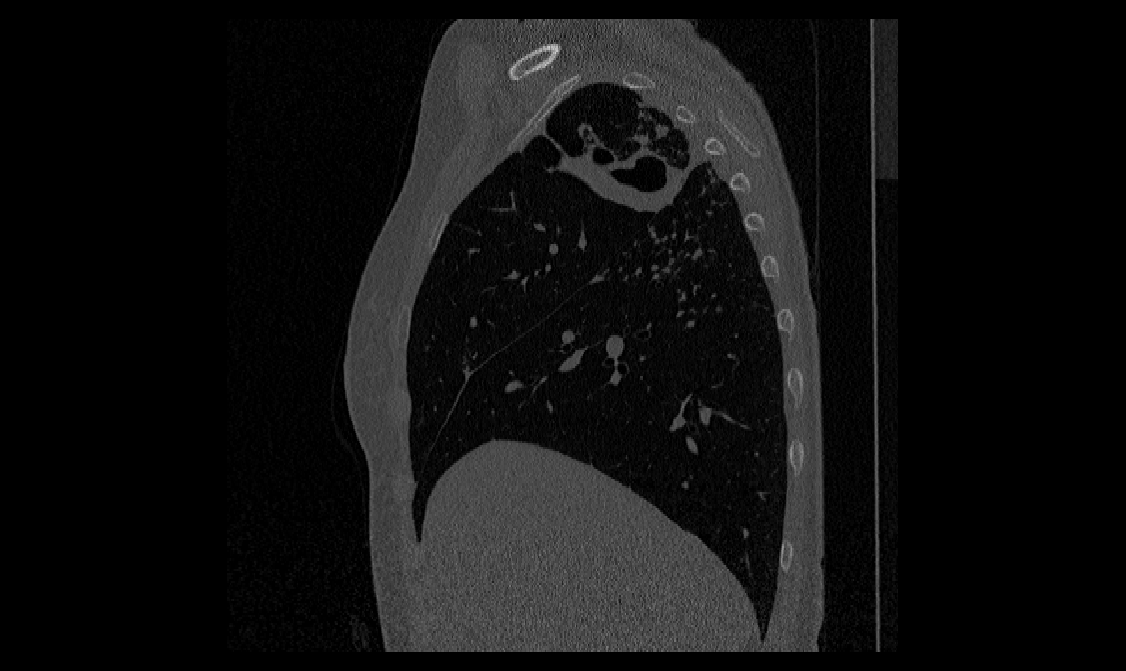}
\hfill
\includegraphics[width=\x\linewidth,height=\x\linewidth]{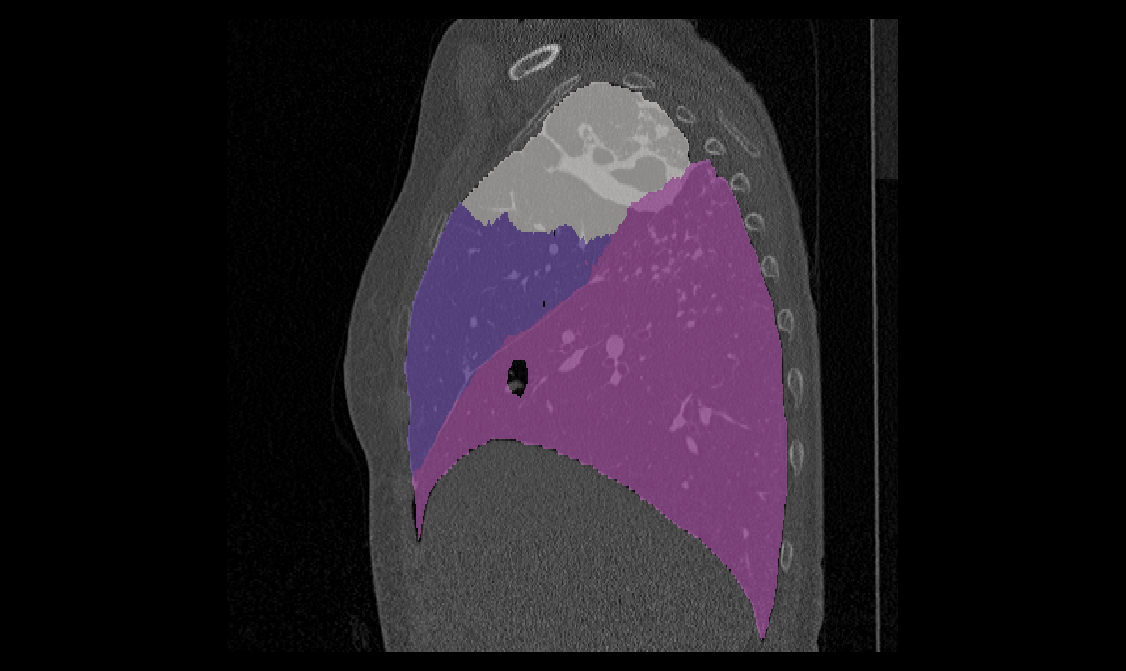}

\makebox[\x\linewidth]{(a)} \hfill \makebox[\x\linewidth]{(b)} \hfill
\makebox[\x\linewidth]{(c)} \hfill \makebox[\x\linewidth]{(d)}

\caption[Visualization of LOLA11 segmentation outputs] {(a) Input image. Outputs of : (b) Segmentation. (c) Input image. (d) Segmentation. Sagittal plane visualization of LOLA11 segmentation by our 3D PDV-Net: good cases (upper row) and failure cases (bottom row).}
\label{fig:lola-cases}
\end{figure}

Figure~\ref{fig:lola-cases} shows the segmentation results for the LOLA11 cases.  For the left lung in Case~8, the LUL and LLL Dice scores were 0.9940 and 0.9926, respectively. For the right lung in Case~6, the scores are as follows: RUL: 0.9580, RML: 0.9480, and RLL: 0.9869. Again, for the left lung of Case 21, the segmentation Dice scores were relatively low. For the left lung in Case~21, the LUL score was 0.8170 and the LLL score was 0.3035. For the right lung in Case~55, although the right lower lobe was segmented with a high Dice score of 0.9818, because of the invisibility of the horizontal fissure, the RUL and RML had low segmentation Dice scores of 0.6827 and 0.7499, respectively.

\subsection{Robustness Analysis} 

We further investigated the robustness of our model by grouping the 84 LIDC cases in three ways. For the first grouping, the Dice scores were put in three different Z-spacing buckets: $\hbox{Z-spacing}\le 1$, $1<\hbox{Z-spacing}<2$, and $\hbox{Z-spacing}\ge 2$. In the second grouping, the Dice scores were put in four manufacturer buckets: GE, Philips, Siemens, and Toshiba. In the third grouping, the Dice scores were grouped according to the reconstruction kernel into 3 buckets: soft, lung, and bone. A one-way ANOVA analysis confirmed that there were no significant differences ($p\hbox{-value}<0.05$) between the average Dice scores of the buckets within each grouping, suggesting that our model is robust against the choice of reconstruction kernel, size of reconstruction interval, and different CT scanner vendors. Moreover, nodule volume in each of the 84 cases does not affect the lobe segmentation performance. There is no correlation between nodule volume and lobe segmentation accuracy, as indicated by the Pearson correlation ($p\hbox{-value}<0.05$). 

\begin{figure}
\centering
\includegraphics[width=\linewidth]{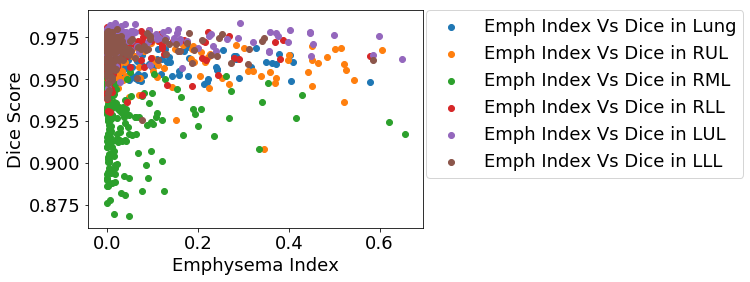}
\caption[Plots of lobe-wise and overall segmentation accuracy] {Plots of lobe-wise and overall segmentation accuracy (Dice scores) of our model versus the emphysema indices of the LTRC test cases reveal insignificant correlation.}
\label{fig:emph-dice}
\end{figure}

We also studied how the segmentation correlation is affected by lung pathologies, by analyzing the correlation between Dice scores and the emphysema index; i.e., the proportion of the lungs affected by emphysema (in the range 0--1). For the LTRC cases, we associated lobe-wise emphysema indices by calculating the proportion of emphysema voxels (voxels marked as emphysema in the LTRC ground truth) in each of the lobes, as well as overall emphysema indices for both lungs. Figure~\ref{fig:emph-dice} shows plots of the per-lobe and overall emphysema indexes versus segmentation performance. The small Pearson correlation ($p\hbox{-value}<0.05$) reveals that the lobe segmentation accuracy is uncorrelated with the emphysema index, confirming the robustness of our model in segmenting lobes in pathological cases.

\subsection{Speed Analysis} 

Our 3D PDV-Net model takes approximately 2 seconds to segment lung lobes from one CT scan using a single Nvidia Titan XP GPU, which is six times faster than the 2D U-Net model. To our knowledge from the lung lobe segmentation models reported in literature, ours is by far the fastest model. Note that no prior published research considered a 3D convolutional model for lung lobe segmentation.

\section{Conclusions}

Reliable and automatic lung lobe segmentation is a challenging task, especially in the presence of pathologies and incomplete fissures. We introduced a new 3D CNN-based segmentation technique, namely, Progressive Dense V-Networks (PDV-Nets), and applied it to the automatic, fast, and reliable segmentation of lung lobes from chest CT scans. We evaluated our method using three test datasets---84 cases from LIDC, 154 cases from LTRC, and 55 cases from LOLA11. Our results demonstrated that our model outperforms, or at worst performs comparably to, the state-of-the-art while running at an average speed of 2 seconds per case, without requiring any prior segmentation. Furthermore, we demonstrated the robustness of our method against varying configurations of CT reconstruction, choice of CT imaging device vendor, and the presence of lung pathologies.

\bibliographystyle{UCLAthesis}
\bibliography{thesis}

\end{document}